\documentclass[
	12pt,				
	openright,			
	oneside,			
	a4paper,			
	french,				
	spanish,            
	brazil,             
	english				
	]{abntex2}
\usepackage{lmodern}			
\usepackage[T1]{fontenc}		
\usepackage[utf8]{inputenc}	
\usepackage{lastpage}			
\usepackage{indentfirst}		
\usepackage{color}			    
\usepackage{graphicx}			
\usepackage{subcaption}
\usepackage{microtype} 		    
\usepackage[final]{pdfpages}	
\usepackage{eucal}              
\usepackage{rotating}           
\usepackage{lipsum}				
\usepackage{amsmath,amssymb}
\usepackage[brazilian,hyperpageref]{backref}	 
\usepackage[alf,abnt-emphasize=bf]{abntex2cite}
\usepackage{amsthm}
\theoremstyle{definition}
\newtheorem{definition}{Definition}[section]

\renewcommand*{\backrefalt}[4]{
	\ifcase #1 %
	\or
	\else
	\fi}%

\titulo{Vector Field Neural Networks}
\autor{Daniel Santana Nogueira Vieira}
\local{Rio de Janeiro}
\data{2018}
\orientador{Jo\~ao Antonio Recio da Paix\~ao}
\coorientador{}
\instituicao{
  Universidade Federal do Rio de Janeiro
  \break
   Instituto de Matemática
  \break
  INSTITUTO TÉRCIO PACITTI DE APLICAÇÕES E PESQUISAS COMPUTACIONAIS
  \break
  PROGRAMA DE PÓS-GRADUAÇÃO EM INFORMÁTICA}
\tipotrabalho{Dissertação (Mestrado)}
\preambulo{Dissertação de Mestrado submetida ao Corpo Docente do Departamento de Ciência da Computação do Instituto de Matemática, e Instituto Tércio Pacitti de Aplicações e Pesquisas Computacionais da Universidade Federal do Rio de Janeiro, como parte dos requisitos necessários para obtenção do título de Mestre em Informática.}
\definecolor{blue}{RGB}{41,5,195}
\makeatletter
\hypersetup{
		pdftitle       = {\@title}, 
		pdfauthor      = {\@author},
    	pdfsubject     = {\imprimirpreambulo},
	    pdfcreator     = {LaTeX with abnTeX2},
		pdfkeywords    = {abnt}{latex}{abntex}{abntex2}{trabalho acadêmico}, 
		colorlinks     = false,       		
    	linkcolor      = blue,          	
    	citecolor      = blue,        		
    	filecolor      = magenta,    		
		urlcolor       = blue,
		bookmarksdepth = 4
}
\makeatother
\makeindex
\begin{document}
\selectlanguage{english}
\frenchspacing 
\imprimircapa
\setcounter{page}{0}
\imprimirfolhaderosto*

\begin{fichacatalografica}
    \includepdf{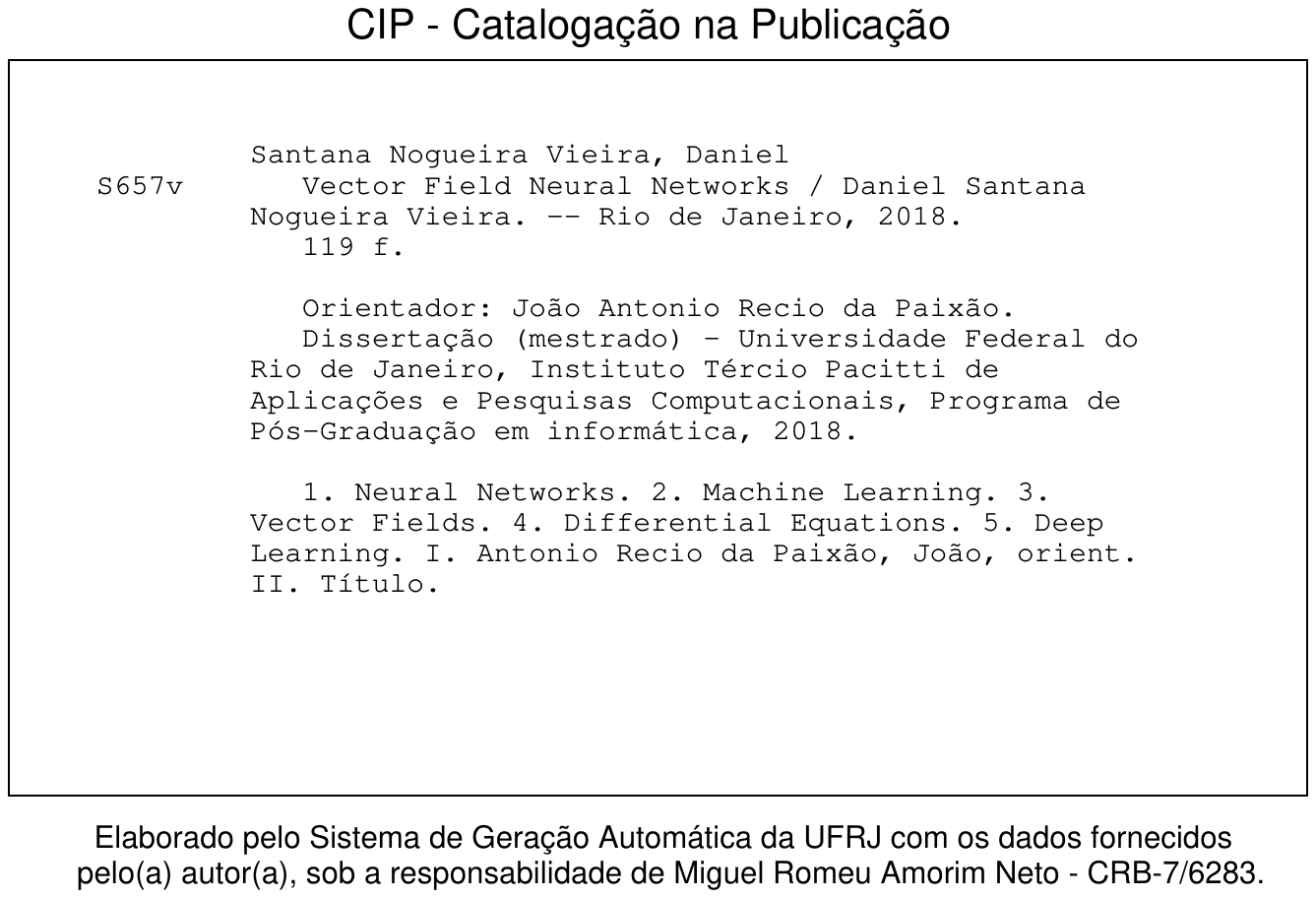}
\end{fichacatalografica}
\thispagestyle{empty}

%
\begin{folhadeaprovacao}
  \begin{center}
    {\ABNTEXchapterfont\mdseries\MakeTextUppercase\imprimirautor}
    \vspace*{\fill}\vspace*{\fill}
    \begin{center}
      \ABNTEXchapterfont\bfseries\MakeTextUppercase\imprimirtitulo
    \end{center}
    \vspace*{\fill}
    \hspace{.45\textwidth}
    \begin{minipage}{.5\textwidth}
        \imprimirpreambulo
    \end{minipage}%
    \vspace*{\fill}
   \end{center}
   
   Aprovado em: \imprimirlocal, 20 de Dezembro de 2018.
   
   \assinatura{\imprimirorientador \:(Orientador), D.Sc. , UFRJ } 
   \assinatura{Jo\~ao Carlos Pereira da Silva \:(Membro, D.Sc, UFRJ)}
   \assinatura{Marcello Goulart Teixeira \:(Membro, D.Sc, UFRJ}

   \begin{center}
    \vspace*{0.5cm}
    {\imprimirlocal}
    \par
    {\imprimirdata}
    \vspace*{1cm}
  \end{center} 
\end{folhadeaprovacao}
\begin{agradecimentos}
The author acknowledges the Coordena\c{c}\~ao de Aperfei\c{c}oamento de Pessoal de N\'ivel Superior (CAPES) for funding.
\end{agradecimentos}
\begin{epigrafe}
    \vspace*{\fill}
	\begin{flushright}
		\textit{``People want what they can not have. They want certainty.`` \\
		}
	\end{flushright}
\end{epigrafe}
\begin{resumo}[Abstract]
\begin{otherlanguage*}{english}

This work begins by establishing a mathematical formalization between different geometrical interpretations of Neural Networks, providing a first contribution.  From this starting point, a new interpretation is explored, using the idea of implicit vector fields moving data as particles in a flow. A new architecture, Vector Fields Neural Networks(VFNN), is proposed based on this interpretation, with the vector field becoming explicit. A specific implementation of the VFNN using Euler's method to solve ordinary differential equations (ODEs) and gaussian vector fields is tested. The first experiments present visual results remarking important features of the new architecture and providing another contribution with the geometrically interpretable regularization of model parameters. Then, the new architecture is evaluated for different hyperparameters and inputs, with the objective of evaluating the influence on model performance, computational time, and complexity. The VFNN model is compared against the known basic models  Naive Bayes, Feed Forward Neural Networks, and Support Vector Machines(SVM), showing comparable, or better, results for different datasets. Finally, the conclusion provides many new questions and ideas for improvement of the model that can be used to increase model performance.

 \vfill
\end{otherlanguage*}
\end{resumo}

\begin{resumo}

Este trabalho se inicia estabelecendo a formalização matemática entre diferentes intereptações geométricas de Redes Neurais, apresentando uma primeira contribuição. A partir disto, uma nova interpretação é explorada utilizando a ideia de campos vetoriais implícitos movendo dados como partículas em um escoamento. Uma nova arquitetura, Vector Field Neural Networks (VFNN), é proposta baseada nesta interpretação, utilizando campos vetoriais explícitos. Uma implementação específica do modelo VFNN é testado utilizando o método de Euler para solução de equações diferenciais ordinárias (EDOs) e campos vetoriais gaussianos. Os primeiros experimentos apresentam resultados visuais marcando importantes características da nova arquitetura e apresentando uma nova contribuição através de uma regularização geometricamente interpretável de parâmetros do modelo. Em seguida, a nova arquitetura é avaliada para diferentes hiper-parâmetros e entradas, com o objetivo de avaliar a influencia dos mesmos na performance do modelo, tempo computacional e complexidade. O modelo VFNN é então comparado com os conhecidos modelos básicos de Naive Bayes, Redes Neurais Feed Forward e Support Vector Machines(SVM), demonstrando resultados comparáveis ou superiores em diferentes datasets. Por fim, a conclusão apresenta diversas questões e ideias para melhora do modelo que podem ser utilizadas para aumentar a sua performance.

 \vfill
 \textbf{Palavras-chave}: Campos vetoriais. Redes Neurais. Equa\c{c}\~oes Diferenciais. Interpretação Geométrica. Regularização. Fun\c{c}\~oes de Kernel. Apredizagem Profunda.
\end{resumo}

\pdfbookmark[0]{\listfigurename}{lof}
\listoffigures*
\cleardoublepage
\pdfbookmark[0]{\listtablename}{lot}
\listoftables*
\cleardoublepage
\begin{siglas}
    \item[ACC] Accuracy
    \item[AUCROC] Area Under the Receiver Operator Characteristic Curve
    \item[FFNN] Feed Foward Vector Field Neural Networks
    \item[GAN] Generative Adversarial Networks
    \item[LSTM] Long Short-Term Memory Networks
    \item[MSE] Mean Squared Error
    \item[NAND] NOT-AND
    \item[NB] Naive Bayes
    \item[ODE] Ordinary Differential Equation
    \item[RBF] Radial Basis Function
    \item[RBFNN] Radial Basis Function Neural Networks
    \item[RelU] Rectified Linear Units
    \item[RNN] Recurrent Neural Networks
    \item[ROC] Receiver Operator Characteristic
    \item[SVM] Support Vector Machines
    \item[UCI] University of California Irvine
    \item[VC] Vapnik–Chervonenkis
    \item[VFNN] Vector Field Neural Networks

\end{siglas}
\pdfbookmark[0]{\contentsname}{toc}
\tableofcontents*
\cleardoublepage
\textual
\chapter{Introduction}
\label{chapterIntroduction}

\section{Motivation}

Machine Learning is currently a topic of large interest in both media \cite{brynjolfsson}, industry\cite{columbus}, and academia \cite{LIU20151636}\cite{Butler2018}\cite{degregory}. It has shown remarkable results in a series of important problems, such as medicine \cite{degregory}, \cite{LIU20151636}, chemistry, engineering \cite{Butler2018}, and even professional games like Go \cite{silver2016mastering}. One of the most important research themes in Machine Learning is the problem of classifying objects in classes (e.g facial recognition systems and medical imaging diagnosis).

\begin{figure}[!ht]
    \centering
    \caption{\vspace{-0.1cm} Elements of a Neural Network. }
    \begin{subfigure}[b]{0.45\textwidth}
        \includegraphics[width=\textwidth]{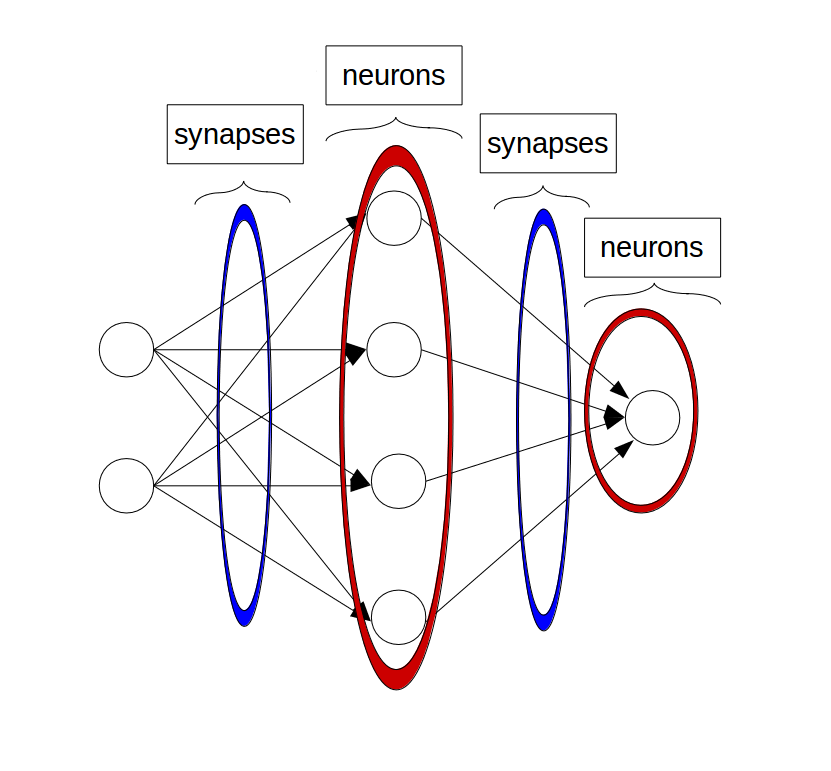}
        \label{fig:archmeanintro}
    \end{subfigure}
    \vspace{-0.25cm}
    \caption*{\vspace{-0.1cm} Source: Author. }
    \label{fig:archmeaningintro}
\end{figure}

Among many techniques that can be used to solve classification problems, one is Neural Networks. Neural Networks inherit its name from the biological inspiration for its creation. The model specifies that a series of nodes, called neurons, are connected by synapses (hence, a Neural Network) between themselves (Figure \ref{fig:archmeaningintro}). It has gained significant attention in the last years after it broke records on image classification problems \cite{bennenson_records} and proved to be useful in other areas such as  natural language processing \cite{goodfellow2016deep}. 

However, there are important problems concerning the use of Neural Networks. Neural Networks\cite{goyal2016interpreting} and several other models, like Support Vector Machines \cite{barakat2006rule}, lack result interpretability. Thus, they are commonly known as Black Box models \cite{freitas2010importance}, and it is not possible, or rather difficult, to provide a reason for a human comprehend on why a decision was made. This is specially important for businesses involving legal implications for acceptance and denial (e.g. credit approval, healthcare insurance, etc.), raising issues about privacy, ethics and law \cite{eu-679-2016}. Nonetheless, even though this is a growing concern, it will not be addressed by this dissertation.

A different issue is the interpretation of the model itself. That is, how can we approach, understand, teach, and visualize the model? Is it possible to associate it with familiar mathematical structures and concepts? This is important to provide a \emph{mindset} to the researchers and people who work with Machine Learning. It helps to better understand and explore the model if we can approach it through different representations. We will detail known geometrical interpretations of Neural Networks and as a first contribution, we will mathematically formalize the connections between them.

\begin{figure}[!ht]
    \centering
    \caption{\vspace{-0.1cm}A classification problem with two classes (orange and blue). On the first image the original dataset is shown. The second plot shows the decision boundary for each class after training with the use of colouring. The last plot illustrates the function value for each class in the domain. }
    \begin{subfigure}[b]{0.7\textwidth}
        \includegraphics[width=\textwidth]{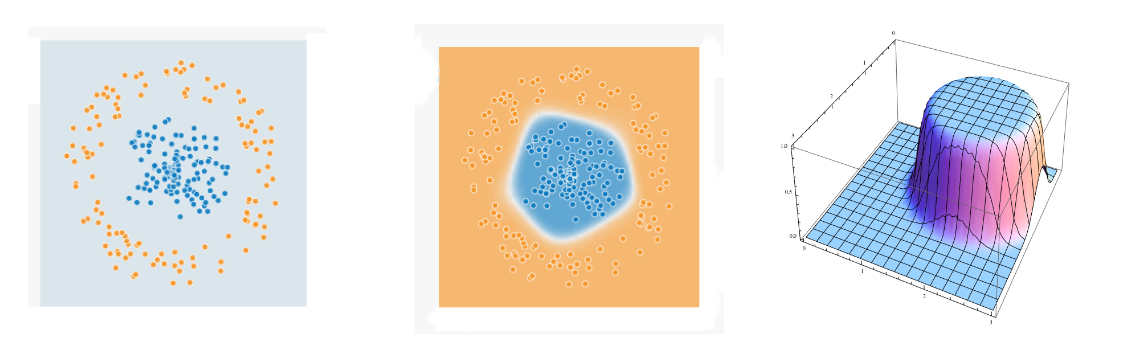}
        \label{fig:netfuc_intro}
    \end{subfigure}
    \vspace{-0.25cm}
    \caption*{\vspace{-0.1cm} Source: \cite{tensorplay} and \cite{stackplay}. }
    \label{fig:netasfunc_intro}
\end{figure}

The classical approach is to address the problem of comprehending Neural Networks behavior, in a geometrical sense, by understanding them as approximation of functions \cite{hornik1991approximation} (Figure \ref{fig:netasfunc_intro}). This approach makes use of our extensive knowledge of functions, one of the fundamental objects in mathematics, and their behavior.

\begin{figure}[!ht]
    \centering
    \caption{\vspace{-0.1cm} Two ways to visualize the same classification problem. }
    \begin{subfigure}[b]{0.4\textwidth}
        \caption{The decision line is shown on the original space, as the result of all layers combined.}
        \includegraphics[width=\textwidth]{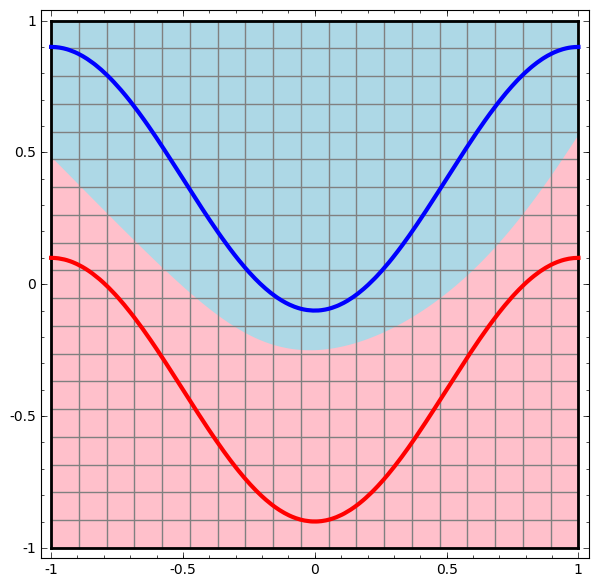}
        \label{fig:colah2_1_intro}
    \end{subfigure}
    \begin{subfigure}[b]{0.4\textwidth}
        \caption{The linear classifier is applied after the transformation, evident by the straight line.}
        \includegraphics[width=\textwidth]{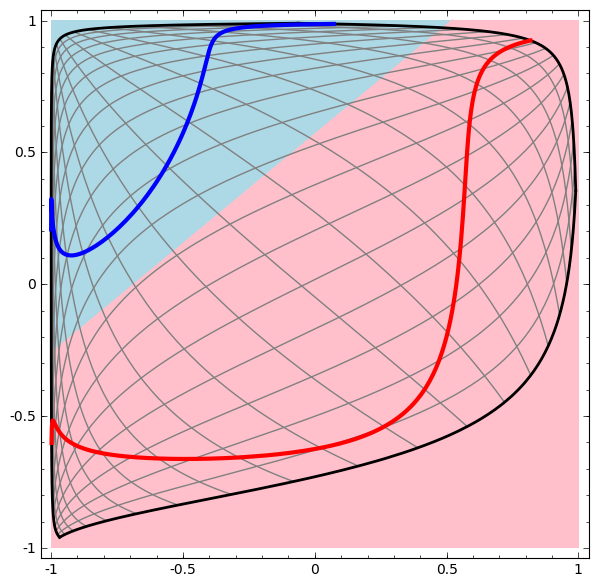}
        \label{fig:colah2_2_intro}
    \end{subfigure}
    \vspace{-0.25cm}
    \caption*{\vspace{-0.1cm} Source: \cite{colah}. }\label{fig:colah2_intro}
\end{figure}

Another possible interpretation considers the Neural Networks as transformations of original data space \cite{colah}. The objective is to smoothly change the dataset to a linearly separable configuration (Figure \ref{fig:colah2_intro}).

Finally, the approach we are going to develop, explain, and focus in this dissertation was suggested by Olah \cite{colah} at the end of his work. It interprets Neural Networks through vector fields, and tries to explore the benefits of using this widely established mathematical tool.

\begin{figure}[!ht]
    \centering
    \caption{The first plot presents the input data, the second plot shows the vector field, and finally the last plot shows the transformed data.}
    \begin{subfigure}[b]{0.28\linewidth}
         \includegraphics[width=\linewidth]{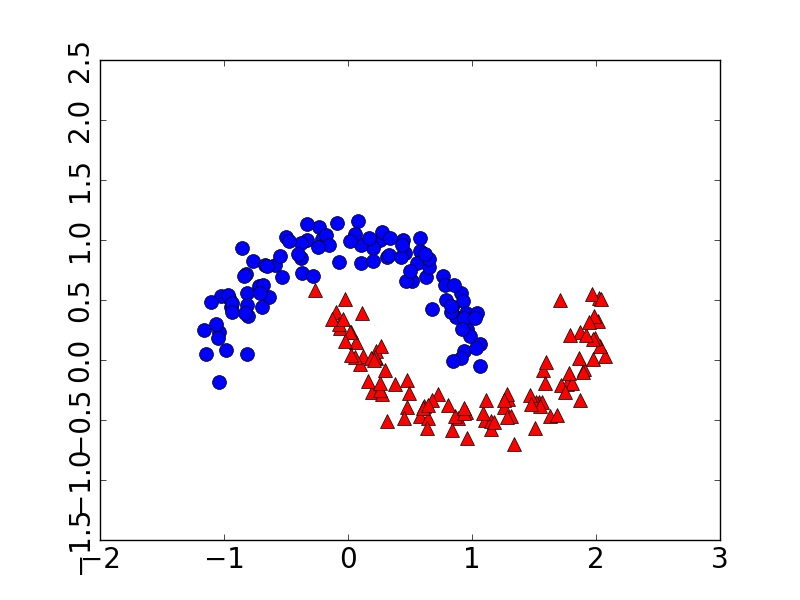}
         \label{fig:original_intro}
    \end{subfigure}
    \begin{subfigure}[b]{0.3\linewidth}
         \includegraphics[width=\linewidth]{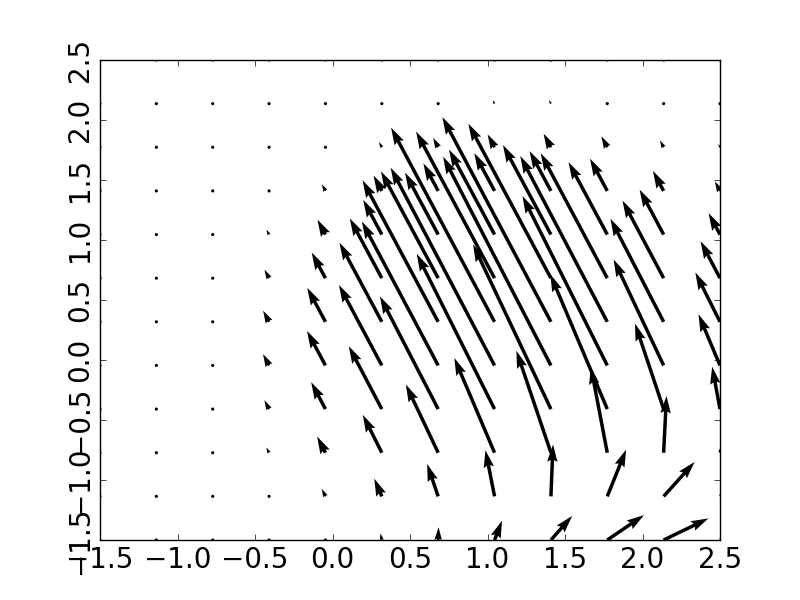}
         \label{fig:field_intro}
    \end{subfigure}
     ~ 
    \begin{subfigure}[b]{0.28\linewidth}
         \includegraphics[width=\linewidth]{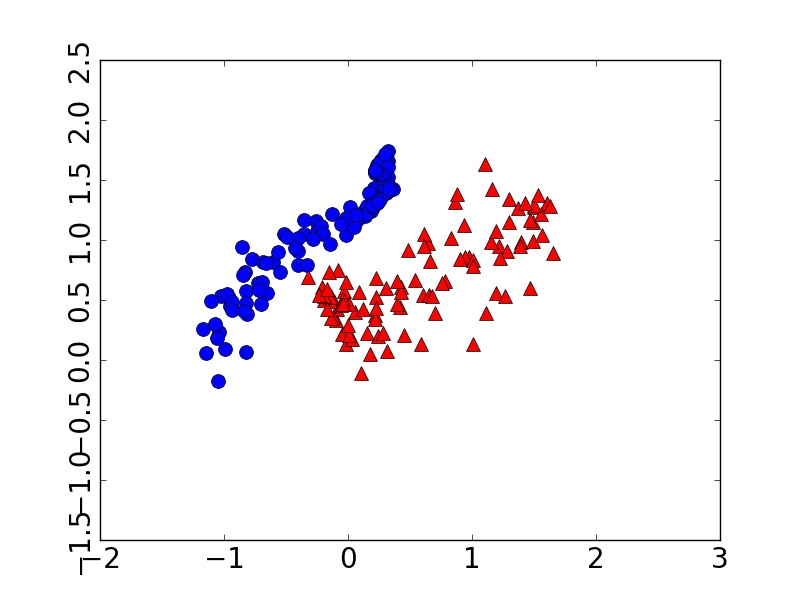}
         \label{fig:transformed_intro}
     \end{subfigure}
     \vspace{-0.25cm}
     \caption*{Source: Author.}
     \label{intro_intro}
 \end{figure}

The model becomes equivalent to a vector field being implicitly applied to the data. That is, there is an implicit vector field, that we are not aware, that is responsible for its classification properties. Every Neural Network becomes equivalent to a movement inside an implicit vector field (Figure \ref{intro_intro}). This property of being implicit is problematic, since it makes harder to adopt the benefits of the vector field interpretation.

Moreover, the increasing difficulty in improving Neural Networks models has also raised problems as it has been hard to consistently improve and develop models. Typically, when dealing with an architecture (a structure that defines how neurons are arranged, or placed, in relation to each other) the researchers use a series of heuristics which have shown promising results. However, most of the time they do not have an explanation or good intuition why it helped. Also, when designing new architectures the effect is greater, with many proposed architectures lacking even a mathematical or physical analogy, let alone proofs, to justify its creation. All of this poses a risk as it can mask potential problems and make it difficult to have a consistent approach to improve and develop Neural Networks  when dealing with new problems.

Taking the last two problems into account, as well as the vector field interpretation, a novel architecture is proposed. It uses explicit vector fields as activation functions and interprets the problem of classification as a problem of creating a flow that leads data to a good configuration. The explicit nature of the vector field eases the exploration of benefits and addresses the problem of consistent improvement and creation of architectures. A solution to move points in a flow can be found by solving ordinary differential equations (ODEs).

The new architecture will also contribute by having a direct geometrical (and statistical) meaning to the regularization of model parameters. This property will be explored in a visual result which displays the utility of choosing meaningful parameters. 

Using  the vector fields representation it will be shown that the new architecture is comparable in performance to known and widely used techniques such as Naive Bayes, Support Vector Machines, and Feed Forward Neural Networks. Further tests evaluate hyperparameters influence over performance. Finally, its adoption provides a physically  and mathematically rich \emph{mindset} and theoretical foundation for further exploring and increasing model performance.

\section{Related Work}

The idea of using vector fields, or ODEs, together with Neural Networks is not new. Previous works used theses concepts separetly for visualization, creation of architectures, improvement of performance, and so on. Here, a brief summary of those works will be provided.

A relevant amount of recent work has appeared regarding approaches to Neural Networks systems with ODEs \cite{DBLP:journals/corr/abs-1710-09513}\cite{DBLP:journals/corr/abs-1710-10121}\cite{DBLP:journals/corr/HaberR17}\cite{E2017}\cite{DBLP:journals/corr/HeZRS15}. Numerous benefits can be achieved, as explained by Weinan:

\begin{citacao}
For example, one can use adaptive time step size, which corresponds to choosing the layers adaptively. One can use high order or even implicit discretization, and these do not yet have an analog in deep neural networks. One can also use advanced numerical methods for training, such as the multi-grid method or the parallel shooting method.

The vast majority of the applied mathematics community is familiar with differential equations. \cite{E2017}
\end{citacao}

Works by others have further developed the approach  by  proposing architectures that could reduce computing time while maintaining similar accuracy to state-of-the-art models \cite{chang2017multi}\cite{chang2017reversible}.  


Vector fields have also been recently used to analyze the optimization problem in Generative Adversarial Networks (GANs) \cite{mescheder2017numerics}, achieving remarkable results for visualization and comprehension of GANs limitations and how to extend them.

\section{Problem Definition and Objective}
As proposed in the motivation, we have two issues to address: to present geometrical interpretations of Neural Networks; and to create a new architecture that provides a clear path for improvement. It must be clear that our aim is not to provide interpretability to the results themselves, but, to provide a physically, geometrically and mathematically inspired interpretation, which can be used for improving models and results.

First, it is important to successfully explain the most used interpretations of Neural Networks, and the benefits and problems they have which need to be addressed. Then, to present the new interpretation that we would like to work upon, and how it compares to others.

Now, we should develop a new architecture that is mathematically sound with the new interpretation, evaluate how hyperparameters and randomness influences performance, and test it in different problems against known models.

\begin{itemize}
    \item  Interpretation of Neural Networks:
   \begin{itemize}
     \item  explain known interpretations, and
     \item  explore the new interpretation.
   \end{itemize}
   
    \item  Create a new Neural Network architecture:
   \begin{itemize}
     \item  create the mathematical model and the architecture itself,
     \item  test how hyperparameters and randomness influence the performance, complexity and time, and
     \item  compare it against other models to verify its feasibility of learning.
   \end{itemize}   
\end{itemize}

\section{Organization}

The dissertation has been structured to provide a path for the reader and  it is composed by the following chapters: Basic Concepts, Neural Networks Interpretations, Vector Field Neural Networks, Design of Experiments, Results and Discussions, and Conclusion.

Chapter 2 (Basic Concepts) contains the basic information needed to be understood by the reader in order to fully understand the contents of the present work. It summarizes the concepts of vector fields, Machine Learning and neural networks, exemplifying use cases, providing definitions, and mathematical and physical reasoning.

Chapter 3 (Interpretability and Visualization) goes through 3 different types of interpretations for the Neural Networks model as a way to provide further insight into the model. It starts with a classic approach by functions, then an interpretation based on a transformation that moves data, and finally one extended from it and based on fluid motion.

Chapter 4 (Vector Field Neural Networks) uses the aforementioned fluid motion interpretation to create a new model architecture of Neural Network. The Neural Network updating scheme is developed, and preliminary results are discussed to prove its feasibility. The architecture and results developed in this chapter are published\cite{vieira}.

Chapter 5 (Design of Experiments) presents the computational experiment which will be used to evaluate the architecture proposed in chapter 4 thoroughly. The metrics, datasets, and hyperparameters, will be briefly explained, and their choices justified.

Chapter 6 (Results and Discussion) contains the results from the computational experiments described in chapter 5 where they will be shown and analyzed. 

Chapter 7 (Conclusion) summarizes the history of this text. Where did we started? What path did we took? What did the results told us? Were the objectives reached? And what can we do to further expand the research area based on what was exposed?

Appendices A, B, C, D, E, F and G contain the full extension of the results, summarized at chapter 6 for readability's sake.

\chapter{Basic Concepts}
\label{chapterBasic}

Some basic concepts need to be defined in order to allow the reader to fully understand the objective proposed. Vector fields and differential equations will be reviewed briefly, providing some examples, interpretations  and basic properties that will be important later on. Similarly, Machine Learning and Neural Networks concepts will be given an introduction for those who are not familiar with it.

\section{Vector Fields}

\begin{figure}[!ht]
    \centering
    \caption{\vspace{-0.1cm} A vector field visualized with two different techniques.}
    \begin{subfigure}[b]{0.35\textwidth}
        \includegraphics[width=\textwidth]{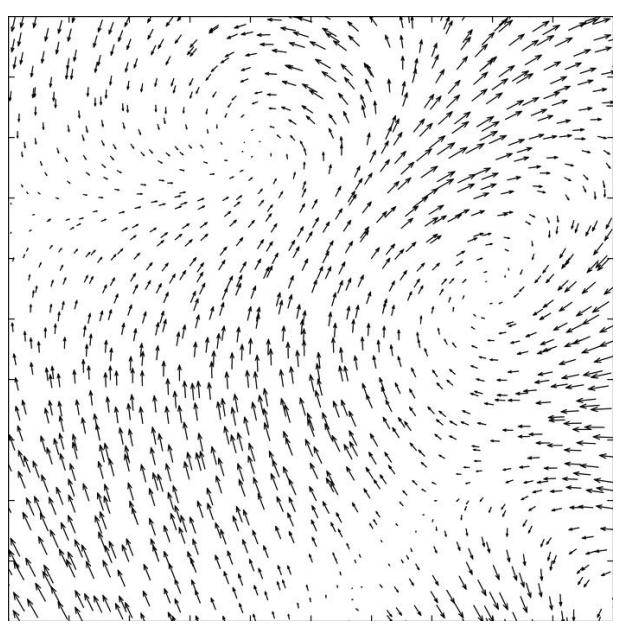}
        \label{fig:flow_jic}
    \end{subfigure}
    \begin{subfigure}[b]{0.35\textwidth}
        \includegraphics[width=\textwidth]{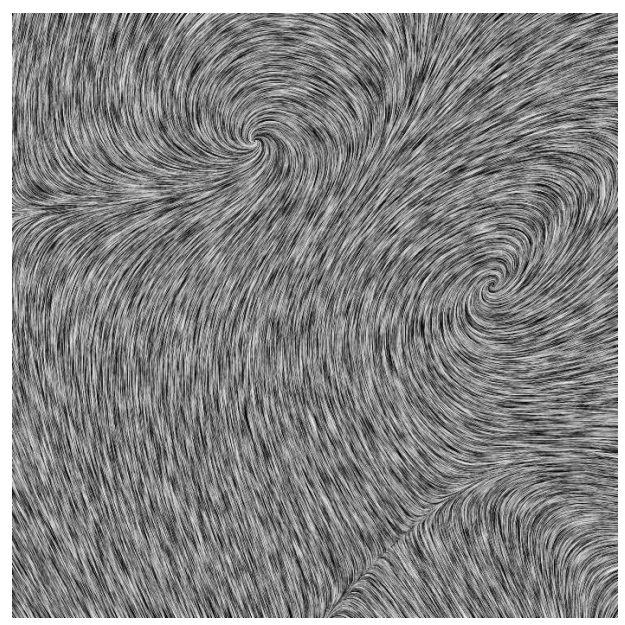}
        \label{fig:flow_lit}
    \end{subfigure}
    \vspace{-0.25cm}
    \caption*{\vspace{-0.1cm} Source: \cite{vecfield2d}.}
    \label{fig:flows}
\end{figure}

Vector fields (Figure \ref{fig:flows}) are fundamental objects in the scientific mindset. It can be used to describe everything from velocity, motion,change, heat transfer, astrophysics and solid mechanics\cite{marsden2011vector}. The very equations that govern fluid dynamics, known as Navier Stokes equations\cite{fox1985introduction}, can be written in a vector field fashion since vector fields are useful to display function variations, i.e. representing derivatives. Now we can use this tool to develop a new interpretation of neural networks, and how they are learning. But first, one needs to understand the basics of what is a vector field, its properties, and the objects associated to it. 

\begin{figure}[!ht]
    \centering
    \caption{\vspace{-0.1cm} Velocity vector fields in two different flow situations.}
    \begin{subfigure}[b]{0.4\textwidth}
        \caption{A vector field describing circular flow in a bathtub.}
        \includegraphics[width=\textwidth]{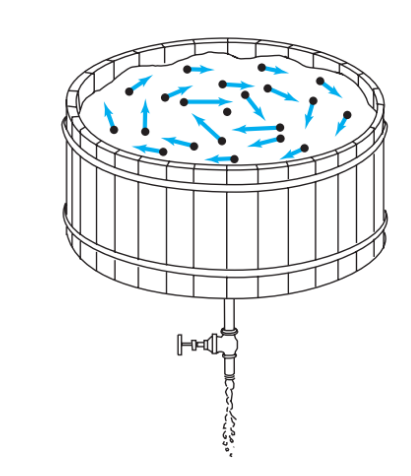}
        \label{fig:tub}
    \end{subfigure}
    ~ 
    \begin{subfigure}[b]{0.4\textwidth}
        \caption{A vector field describing the velocity of flow in a pipe.}
        \includegraphics[width=\textwidth]{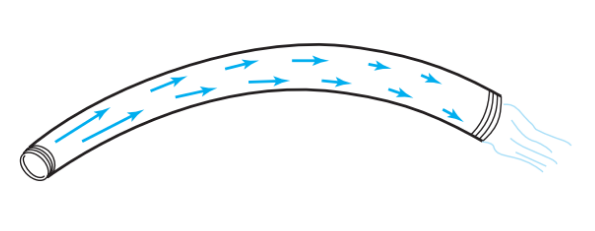}
        \label{fig:tube}
    \end{subfigure}
    \vspace{-0.25cm}
    \caption*{\vspace{-0.1cm} Source: \cite{marsden2011vector}.}\label{fig:tubes}
\end{figure}

Consider the problem of water flowing on Figure \ref{fig:tubes}. Here we depict the water flowing in a bathtub and a tube, along with a series of arrows that vary in length and direction. In this case, what is shown is the speed of an infinitesimal body of water in that point in space. The vector field is a collection of such arrows, spanning all over the flow domain, which tell us the velocity of a particle of fluid in that location. It provides us with a way to describe the velocity for all points in the domain.


\begin{definition}
A \textbf{vector field} in $\mathbb{R}^{Machine Learning}$ is a map $\mathbf{K}$: $X$ $\subset \mathbb{R}^{Machine Learning}\rightarrow \mathbb{R}^{n}$ that assigns to each point $x \in X$ a vector $\mathbf{K}(x)$. 
\end{definition}


\begin{figure}[!ht]
    \centering
    \caption{\vspace{-0.1cm} A vector field in three dimensions, each point $x$ in space has a designated arrow $\mathbf{K}(x)$ ($F(X)$ in the plot below).}
    \begin{subfigure}[b]{0.6\textwidth}
        \includegraphics[width=\textwidth]{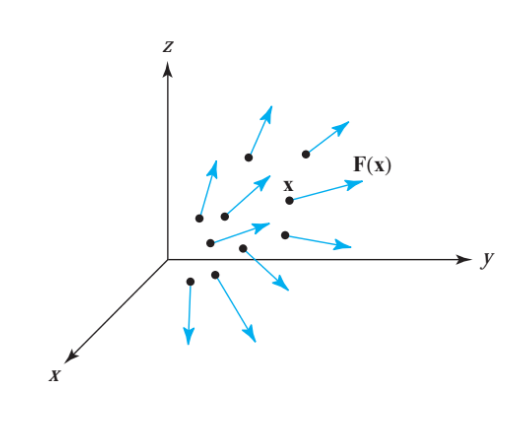}
        \label{fig:vec_def}
    \end{subfigure}
    \vspace{-0.25cm}
    \caption*{\vspace{-0.1cm} Source: \cite{marsden2011vector}.}\label{fig:vecdef}
\end{figure}

In Figure \ref{fig:vecdef},  the vector field $\mathbf{K}$ $\subset \mathbb{R}^{3}$ has three component scalar fields ( maps which lead to a real number ) $K_1, K_2, K_3$:
\begin{center}
$\mathbf{K}(x,y,z) = (K_1(x, y, z), K_2(x, y, z), K_3(x, y, z)).$
\end{center}

In general, we have  $n$ scalar fields composing a vector field on $\mathbb{R}^{n}$. 

\begin{figure}[!ht]
    \centering
    \caption{\vspace{-0.1cm} Here, in both cases, it is easy to see the drawing of the flowlines from the velocity vector field of the flow. }
    \begin{subfigure}[b]{0.4\textwidth}
        \includegraphics[width=\textwidth]{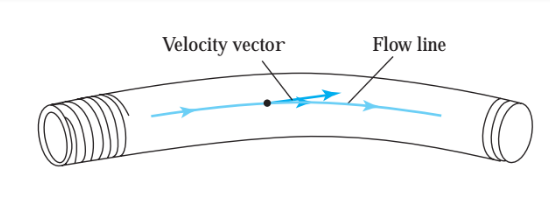}
        \label{fig:streamdef}
    \end{subfigure}
    ~ 
    \begin{subfigure}[b]{0.4\textwidth}
        \includegraphics[width=\textwidth]{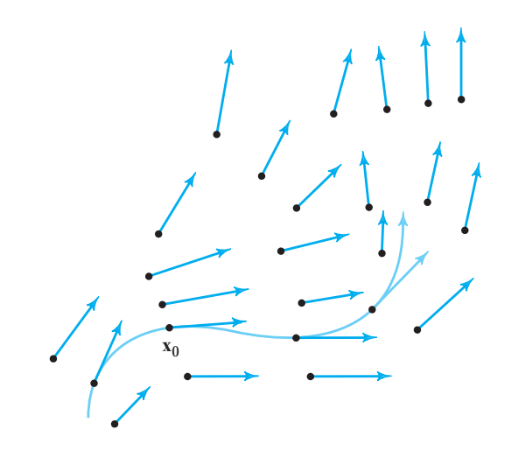}
        \label{fig:steamedo}
    \end{subfigure}
    \vspace{-0.25cm}
    \caption*{\vspace{-0.1cm} Source: \cite{marsden2011vector}.}\label{fig:streams}
\end{figure}

Now we can direct our attention towards an important concept that rises from the use of vector fields: streamlines, defined as the integral curve of the vector field. Geometrically, these lines are tangent in each point of the vector field (Figure \ref{fig:streams}). The streamlines are equal to the curve of trajectory for each fluid particle, representing the movement of the fluid.

Formally, a streamline $c(t)$ of a given vector field $\mathbf{K}$ can be defined as follows:

\begin{definition}
If $\mathbf{K}$ is a vector field, a \textbf{streamline} is a path $c(t)$ such that:
\begin{center}
    $c'(t) = \mathbf{K}(c(t)),\qquad t\ \in\ \mathbb{R},\ t\geqslant0.$
\end{center}
\end{definition}

Following this definition, the vector field $\mathbf{K}$ represents the derivative of path $c(t)$, thus, the path can be found by integrating over the field. In a physical context, the vector field could represent the velocity of a car and the path $c(t)$, as the name implies, would represent the circuit driven by the car.

Now that we have briefly explained the basics of vector fields, it is time to redirect our attention towards the connection between vector fields and differential equations. Remember that the streamlines have been defined to have their derivative equal to the vector field. Look back to Figure \ref{fig:streams}. Imagine now that we are integrating the differential equation that represents the position of a particle for a given time and initial state. The curve $c(t)$ has the very solution for this problem. Therefore, it is possible to view the streamlines as a way to find the solution to a given differential equation problem. When a specific solution is found, we use the initial conditions given by the particle position and start to move it along the vector field.

 Thus, since the analytical solution is not always possible or, rather, unfeasible, having an approximated way to calculate streamlines is important. The easiest method for doing such a thing is known as Euler's method \cite{Butcher1987}. The method is based on the assumption that, for small enough intervals, the linear approximation holds true. The path is created by taking small steps in the direction provided by the vector field.

It is useful to give a definition to the unique solution of the streamline $c(t)$. Let $\mathbf{K}$, a vector field on $\mathbb{R}^{n}$ be a smooth function $\mathbf{K}: \mathbb{R}^{n} \to \mathbb{R}^{n}$. Consider the corresponding ordinary differential equation (ODE):

\begin{equation} \label{eq:field_v1}
\frac{\partial{\phi(x,\ t)}}{\partial{t}} = \mathbf{K}(\phi(x,\ t)),
\end{equation}
\begin{equation} \label{field_v2}
\phi(x,\ 0) = x,
\end{equation}

where $x$ states the initial position, which is used to find the unique solution $\phi(x, t)$, that gives the position of $x$ after time $t$. The curve $\phi(X, t)$ which solves the ODE is called a \textbf{flow} of the vector field $\mathbf{K}$.

Since,

\begin{equation*} \label{eq:field_v3}
\frac{\partial{\phi(x,\ t)}}{\partial{t}} = \lim_{h\rightarrow 0} = \frac{\phi(x,\ t+h)- \phi(x,\ t)}{h} = \mathbf{K}(\phi(x,\ t)), \\
\end{equation*}
\begin{equation*} \label{eq:field_v4}
\phi(x,\ t + h) - \phi(x,\ t) = h\mathbf{K}(\phi(x,\ t)),
\end{equation*}

then, we can use Euler's method \cite{Butcher1987} to approximate $\phi(x , t)$ by $\phi_t$, the solution of the ODE, with $\phi_i \approx \phi(x, ih)$ for the discretization, $h$ being the step size, $i$ the current step, and $N$ the maximum number of steps:

\begin{equation} \label{euler_it}
\phi_{i+1} = \phi_i + h\mathbf{K}(\phi_i),  \qquad 0\leqslant i\leqslant N\ \in\ \mathbb{N}.
\end{equation}

Considering Euler's method,  it is known that for $h \to 0$, the flow $\phi(X, t)$ is calculated exactly.

One of the main problems of Euler's method is the need for small steps. The linear approach is only successful in producing a correct solution when the step is small compared to the function derivative which is being integrated. There are other problems, such as the method stability \cite{Butcher1987} that rises from adopting such a simple approximation. However, it still is a powerful tool for the right problems and provides the basis for more complicated methods.




\section{Machine Learning}

Machine Learning is a set of computer learning methods and heuristics which can be used when certain conditions are met. Informally speaking, these conditions can be summarized as following:  a pattern exists, we can not pin it down mathematically, and there is data available \cite{abu2012learning}. 

The first one is trivial, and defines the problem to be solved. The second condition has to do with the fact that, if we already have a mathematical formulation that solves that problem, specially if it is a mathematical modeling with good theoretical basis, then, it does not make sense to apply Machine Learning. After all, the problem is already solved by a solution that is more robust  and accountable than any Machine Learning model would be. For instance, there is no point in trying to predict the uniform motion of a point with Machine Learning, if we already have a perfect fine physics formula for that. Finally, the last condition lies at the heart of every Machine Learning work, and differentiates it from other artificial intelligence approaches. Every decision taken by a Machine Learning model is based on the data provided to it. The entire concept of Machine Learning lies on the premise that we are not able to produce the correct rules that an artificial intelligence system would need to follow to perform a certain task. As such, we pass down this task to the heuristics and methods aforementioned, and with the use of data, try to find a solution.

\begin{figure}[!ht]
    \centering
    \caption{\vspace{-0.1cm} Workflow of a typical Machine Learning problem.}
    \begin{subfigure}[b]{0.8\textwidth}
        \includegraphics[width=\textwidth]{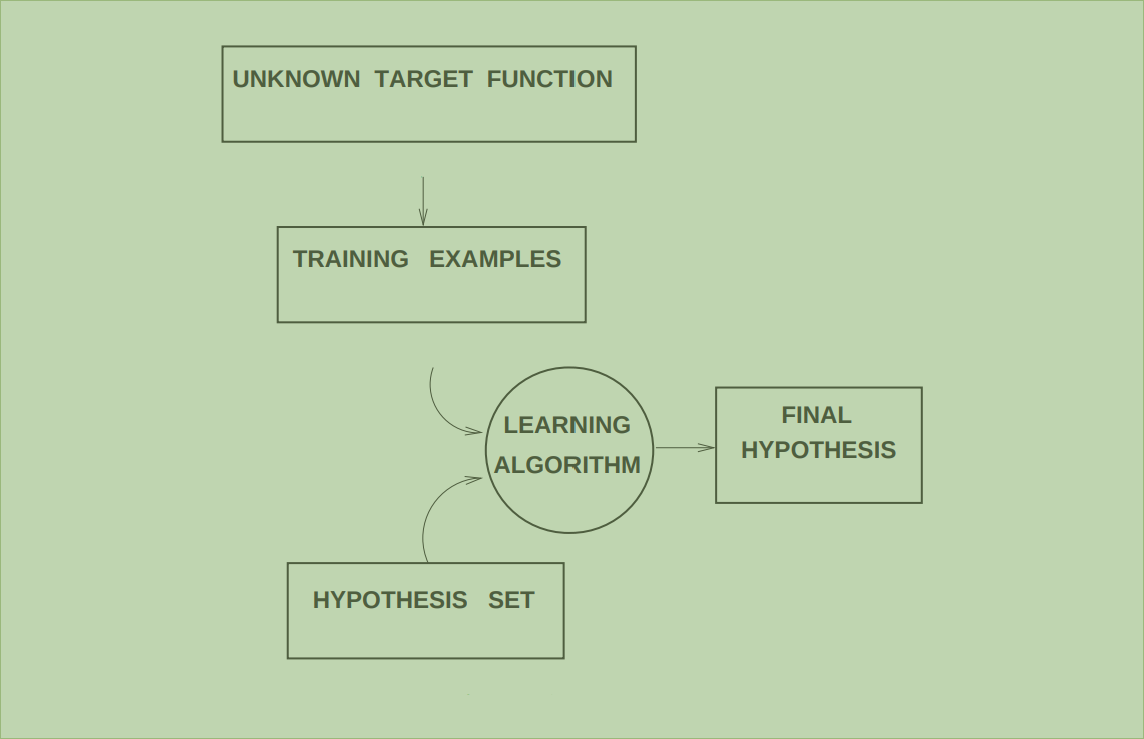}
        \label{fig:lflow}
    \end{subfigure}
    \vspace{-0.25cm}
    \caption*{\vspace{-0.1cm} Source: Adapted from \cite{abu2012learning}.}
    \label{fig:learnflow}
\end{figure}

To better define Machine Learning, it is important that we define its workflow. Figure \ref{fig:learnflow} shows what a typical Machine Learning structure looks like. We have some unknown function, a pattern, as we stated before, that we wish to learn. This pattern is able to provide a series of examples that we can use to approximate it, these are our training examples. Then, we need to chose a learning algorithm, which will be responsible of picking a final hypothesis from the hypothesis set.

This process requires a good and clear understanding of what the hypothesis set and the learning algorithm mean. The hypothesis set is the place from where we draw our hypothesis that aims to approximate the unknown function, the true pattern. This hypothesis set, then, depends on what form we want our model to have. For instance, it is possible to pick a hypothesis set that contains all the hyperplanes for a given space. Therefore, when we choose the hypothesis set, we restrict our final answer to one of the members of the set (a hyperplane, in our example). However, we have not spoken about how to find this said hyperplane. All that we have is the idea that we need to chose a fitting hyperplane. That is where the learning algorithm comes into play. The learning algorithm will make the final choice on which, among the entire hypothesis set (the family of all hyperplanes in this example), will be used to provide the answer. It is important not to mix the two of them together, as different hypothesis set can use the same learning algorithm and different algorithms may, in some cases, be used for a hypothesis set. For instance, the gradient descent is a key part of a Neural Networks implementation, however, it is not part of the hypothesis set, it is the learning algorithm which is typically used to find one of the possible outcomes. Now that we have a good understanding of the general process and the difference and importance between the learning algorithm and the hypothesis set we can further explore the Machine Learning structure.

Although not completely agreed upon, it is customary to break Machine Learning tasks into three different types: supervised learning, unsupervised learning, and reinforcement learning \cite{abu2012learning}. Supervised learning, in which this dissertation is focused, are those problems where we have both input data and output data. So, to put it another way, the domain is known, as well as the image of the function which we are aiming to learn. The learning process occurs by looking at the difference between the answer provided by our model, and the expected answer, adjusting the model to achieve the correct output. Examples of supervised learning can be seem on many different areas, a famous one being the recognition of animals, objects and people on pictures.

\begin{figure}[!ht]
    \centering
    \caption{\vspace{-0.1cm} Example of classification and regression problem when dealing with temperature.}
    \begin{subfigure}[b]{0.8\textwidth}
        \includegraphics[width=\textwidth]{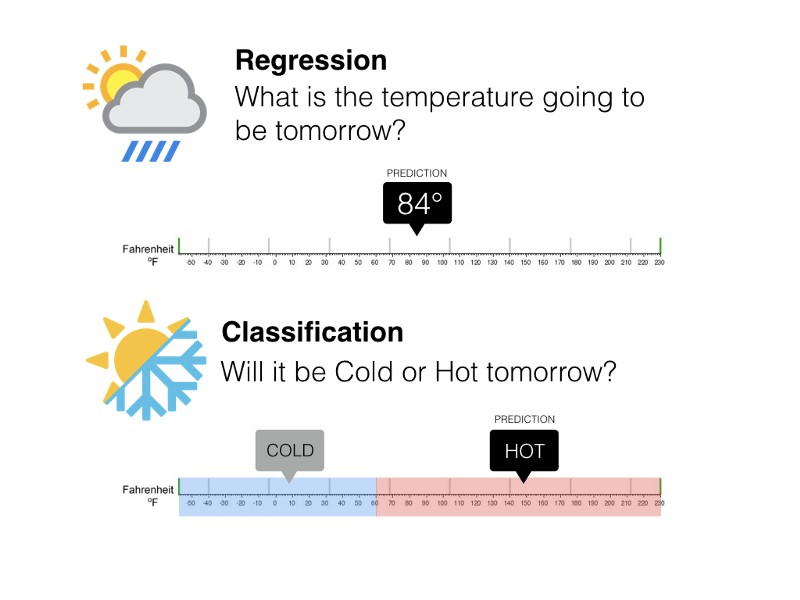}
        \label{fig:classreg}
    \end{subfigure}
    \vspace{-0.25cm}
    \caption*{\vspace{-0.1cm} Source: \cite{alireza}.}
    \label{fig:classregression}
\end{figure}

Our desire in this work is to focus on the specific problem of supervised learning. When we take a closer look, it is also possible to break down supervised learning into two different parts: regression and classification. Figure \ref{fig:classregression} shows an illustration of a regression and classification problem. The regression problem deals with a function that has a codomain that is continuous, such as temperature. Meanwhile, the classification problem, as the name indicates, uses a function to split the data between previously selected discrete classes (such as cold and hot).

\begin{figure}[!ht]
    \centering
    \caption{\vspace{-0.1cm} Supervised problem of classification with boundary line between patients healthy and sick and regression with tendency line of survivability in years.}
    \begin{subfigure}[b]{0.8\textwidth}
        \includegraphics[width=\textwidth]{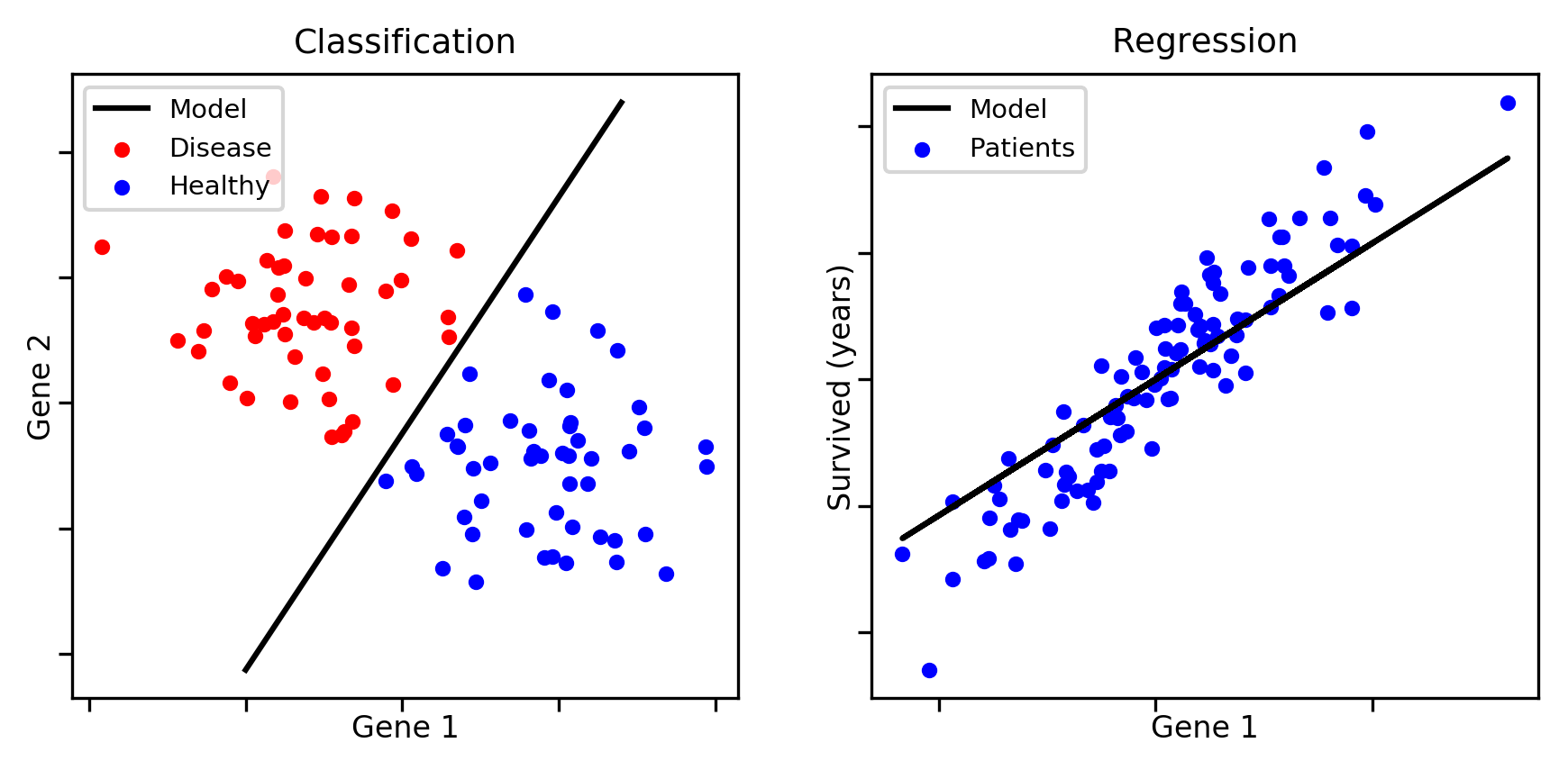}
        \label{fig:classreg2}
    \end{subfigure}
    \vspace{-0.25cm}
    \caption*{\vspace{-0.1cm} Source: \cite{freecodecamp}.}
    \label{fig:classregression2}
\end{figure}

Visually speaking, Figure \ref{fig:classregression2} helps us see this difference from another point of view. At the classification problem in the first image, we are interested in finding the boundary line which separates the classes of interest, and the model represents this line. However, in the regression problem the aim is to find the function (a line in the example) which predicts the continuous function value for the domain.

\begin{figure}[!ht]
    \centering
    \caption{\vspace{-0.1cm} Visualizing underfitting, proper fitting and overfitting in a regression problem using polynomial fit of different degrees.}
    \begin{subfigure}[b]{1.0\textwidth}
        \includegraphics[width=\textwidth]{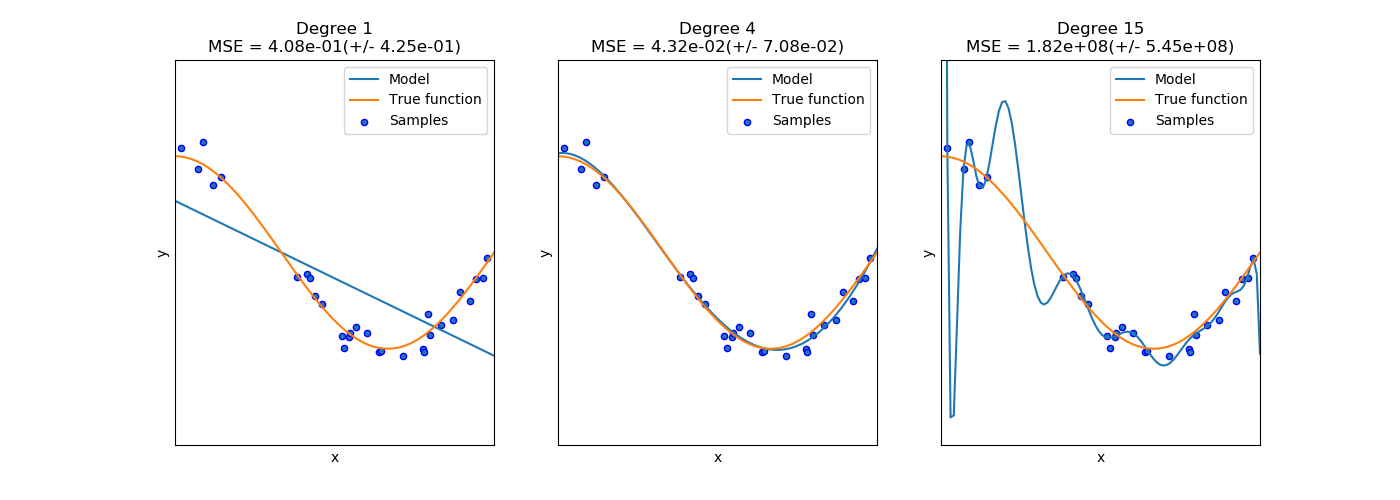}
        \label{fig:underoverr}
    \end{subfigure}
    \vspace{-0.25cm}
    \caption*{\vspace{-0.1cm} Source: \cite{scikitunder}.}
    \label{fig:underoverregression}
\end{figure}

In both scenarios of supervised learning, one has to be careful about the fitting of the data. Overfitting is a known problem of some Machine Learning hypothesis set, and happens when a model is too complex (informally, too many parameters) and is able to "memorize" the data. However, it is also possible for the model to underfit the data. Although not common, a model may be too simple to represent the function, and by that, miss the correct output. Figure \ref{fig:underoverregression}  provides a clear example of underfitting and overfitting happening in a regression scenario. In the first image, the degree of the polynomial is small, the corresponding straight line is insufficient to capture the data pattern and we have a high mean squared error. The second image shows a model with a good choice of for the degree of the polynomial. The model is capable of predicting the pattern, and the mean squared error is low. At last, the third image shows an overfitted model, where the degree of the polynomial is so high that the model is able to fit exactly the data points, which informally means "memorizing" the sample, thus, ignoring the underlying pattern. Nevertheless, when measured out-of-sample (against data not used in the training), the error is high.

\begin{figure}[!ht]
    \centering
    \caption{\vspace{-0.1cm} Visualizing underfitting, proper fitting and overfitting in a classification problem.}
    \begin{subfigure}[b]{0.8\textwidth}
        \includegraphics[width=\textwidth]{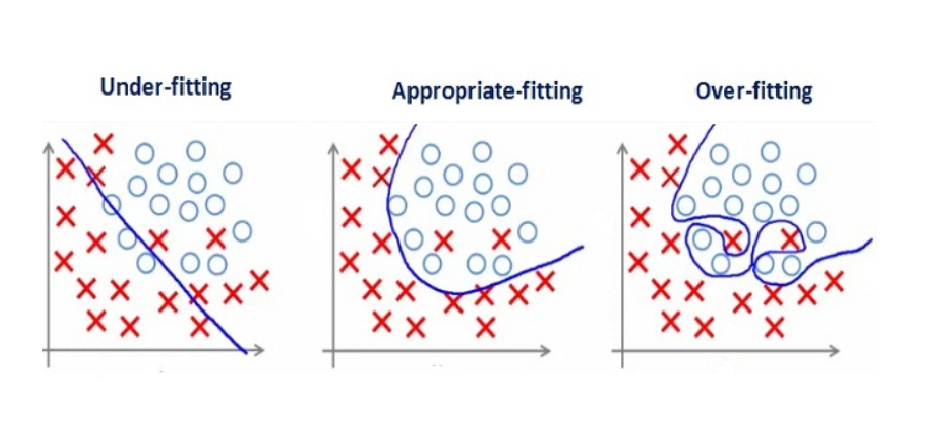}
        \label{fig:underoverc}
    \end{subfigure}
    \vspace{-0.25cm}
    \caption*{\vspace{-0.1cm} Source: Adapted from \cite{anup}.}
    \label{fig:underoverclassification}
\end{figure}

Similarly, Figure \ref{fig:underoverclassification} shows the fitting problem in a classification problem with a class displayed as blue circles, and the other class as red crosses. In this case, the boundary line represents the model, and the underfitting case is represented by a boundary that can not separate the two classes in a satisfactory fashion. An appropriate fitting is shown next, where the boundary line separates classes without being influenced by outliers, errors, and noise. Finally, the third image is an overfit model which fails to discover the correct pattern which separates classes as it tries to get every point in the training set correct.

As a way to deal with the problems of fitting, it is common pratice to break the dataset into 3 different parts: training, validation and test \cite{abu2012learning}.

Many methods and heuristics can be used to perform Machine Learning, and the researcher's task is to be able to pick the one which correctly addresses the problem without suffering from one or more of the issues explained here. Up next, we are going to explore Neural Networks, one among the many possible choices, and explain how it works, what it does, and some of its advantages and disadvantages.

\section{Neural Networks}

Machine Learning research enjoyed  a recent growth in popularity and importance, partly\cite{michaelnielsen} from the results achieved by a particular hypothesis set, or, rather, a set of similar hypothesis sets: the Neural Networks and Deep Learning models. Together with massive amounts of available data, these models were responsible for new records in a series of learning problems inside computer vision area \cite{bennenson_records} which followed its use in different areas such as  natural language processing, weather prediction, house pricing, etc. \cite{goodfellow2016deep}.

Before we can explain Neural Networks, or even Deep Learning, models, we need to establish the concept of a Perceptron. The Perceptron is a simple function that was developed by Frank Rosenblatt \cite{rosenblatt1961principles} in the 50s and 60s, based on the works of Warren McCulloch and Walter Pitts \cite{mcculloch1943logical}.

\begin{figure}[!ht]
    \centering
    \caption{\vspace{-0.1cm} The Perceptron. }
    \begin{subfigure}[b]{0.45\textwidth}
        \includegraphics[width=\textwidth]{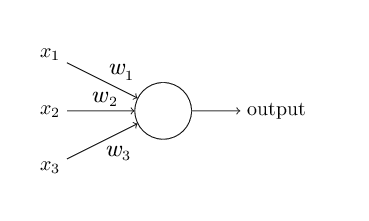}
        \label{fig:perceptron}
    \end{subfigure}
    \vspace{-0.25cm}
    \caption*{\vspace{-0.1cm} Source: \cite{michaelnielsen}. }\label{fig:perceptron_def}
\end{figure}

Take a look at Figure \ref{fig:perceptron_def}, the Perceptron receives a series of binary signal inputs, and, after assigning a weight to each input, computes the binary output signal. If the weighted sum $\sum_{j} w_j x_j$ is less than a threshold it assumes a value of zero, or else it assumes a value of one.

However, for notation simplicity, it is useful to redefine the way the output is calculated. Instead of using a threshold, we transfer it to the left side of the equation and add a constant, known as bias $b$, to the weighted sum. Also, it is useful to write the sum as the inner product between weights and inputs. Now, the equation is the following:
\[
Output=
\begin{cases}
    0,& \text{if } w^T x + b\leq 0,\\
    1,&  \text{if } w^T x + b> 0,
\end{cases}
\]
where $w$ is the vector of weights and $w^T x$ is the inner product that calculates the weighted sum of inputs.
 
As mentioned by Nielsen \cite{michaelnielsen} the bias indicates the likeability of a neuron being activated. If the bias is high, it is easier for the activation to occur - also, the opposite happens when the bias is negative.

\begin{figure}[!ht]
    \centering
    \caption{\vspace{-0.1cm} A Perceptron implementation of the NAND logic operator. }
    \begin{subfigure}[b]{0.45\textwidth}
        \includegraphics[width=\textwidth]{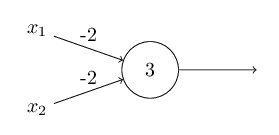}
        \label{fig:perceptron_nan}
    \end{subfigure}
    \vspace{-0.25cm}
    \caption*{\vspace{-0.1cm} Source: \cite{michaelnielsen}. }\label{fig:perceptron_nand}
\end{figure}

\begin{figure}[!ht]
    \centering
    \caption{\vspace{-0.1cm} The logic operator for bitwise sum and the equivalent Perceptron implementation. }
    \begin{subfigure}[b]{0.45\textwidth}
        \includegraphics[width=\textwidth]{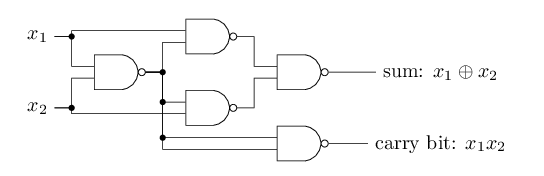}
        \label{fig:logic_sum}
    \end{subfigure}
    \begin{subfigure}[b]{0.45\textwidth}
        \includegraphics[width=\textwidth]{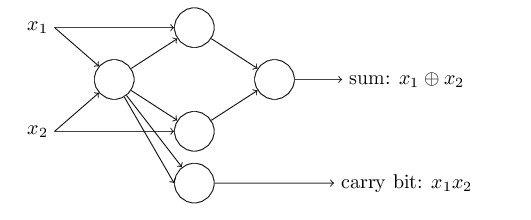}
        \label{fig:perceptron_sum}
    \end{subfigure}
    \vspace{-0.25cm}
    \caption*{\vspace{-0.1cm} Source: \cite{michaelnielsen}. }
    \label{fig:bitwise}
\end{figure}

These Perceptrons can be connected in a series of ways to represent different logic functions. In fact, Figure \ref{fig:perceptron_nand} shows a Perceptron with weights that models the NAND logical function. This enable the Perceptrons to, theoretically, represent any type of existent logical function \cite{michaelnielsen} - as the NAND operation includes all of the basic logic operations inside of it. The analogy with logic circuits can be seen as a possible interpretation of Neural Networks. It is possible, for instance, to create the logical operator that makes the bitwise sum of two bits, as shown by Figure \ref{fig:bitwise}.

\begin{table}
\begin{center}
\caption{NAND operator logic table.}
\begin{tabular}{ |c|c|c|c|c| } 
\hline
$x_1$ & $x_2$ & $ w \cdot x + b$ & $ Output$ \\
\hline 
0& 0 & 3 & 1 \\ 
0& 1 & 1 & 1 \\ 
1& 0 & 1 & 1 \\ 
1& 1 & -2 & 0  \\
\hline
\end{tabular}
\label{nandtable}
\caption*{ Source: Author.}
\end{center}
\end{table}

\begin{figure}[!ht]
    \centering
    \caption{\vspace{-0.1cm} Elements of a Neural Network. }
    \begin{subfigure}[b]{0.45\textwidth}
        \includegraphics[width=\textwidth]{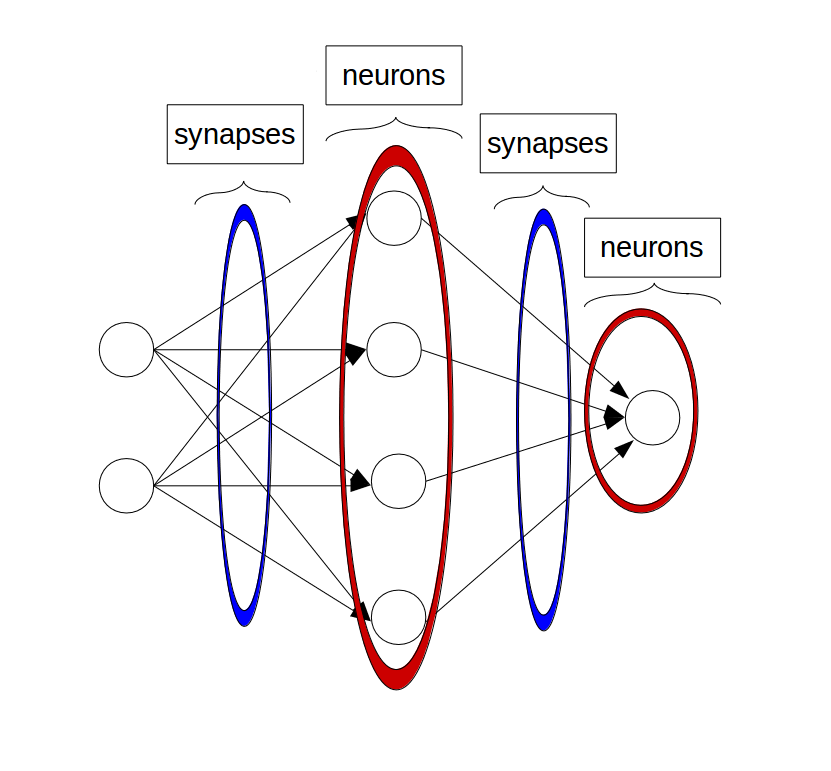}
        \label{fig:archmean}
    \end{subfigure}
    \vspace{-0.25cm}
    \caption*{\vspace{-0.1cm} Source: Author. }
    \label{fig:archmeaning}
\end{figure}

The idea of a Neural Network comes from, literally, creating a network of slightly different Perceptrons. Inspired by the biological analogy of the connection between neurons in the human brain, each Perceptron is called a neuron and the weights  connecting inputs and neurons are called synapses (Figure \ref{fig:archmeaning}).

\begin{figure}[!ht]
    \centering
    \caption{\vspace{-0.1cm} The sigmoid and ReLU activation functions that can be used in Neural Network Perceptrons. }
    \begin{subfigure}[b]{0.45\textwidth}
        \includegraphics[width=\textwidth]{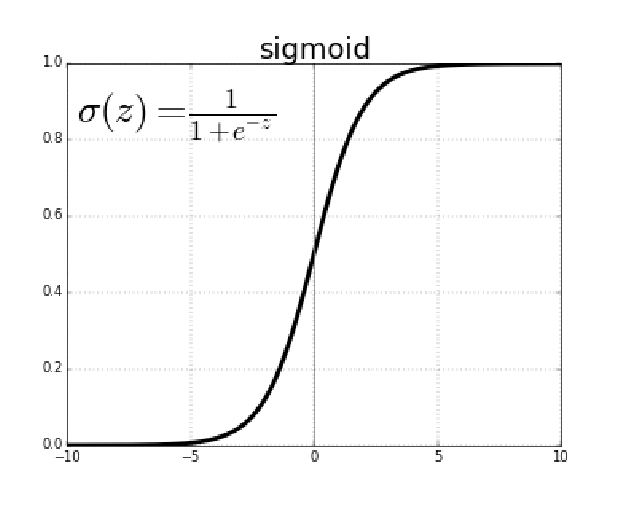}
        \label{fig:sig}
    \end{subfigure}
    \begin{subfigure}[b]{0.45\textwidth}
        \includegraphics[width=\textwidth]{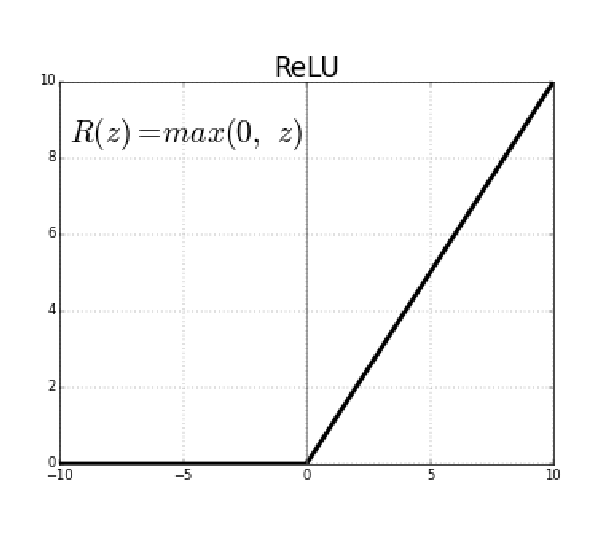}
        \label{fig:relu}
    \end{subfigure}
    \vspace{-0.25cm}
    \caption*{\vspace{-0.1cm} Source: \cite{towardsdata}. }
    \label{fig:sigrelu}
\end{figure}

Nevertheless, the original Perceptron is not what is used today in the creation of Neural Networks, because it suffer from some numerical and optimization issues during the learning process ( for instance,  the sum of weighted inputs does not have a maximum or minimum value). Because of that, a series of activation functions were adopted by Neural Networks. These functions act upon the weighted sum and usually have some nice properties, like a minimum and maximum value, continuity, etc. The family of sigmoid functions (Figure \ref{fig:sigrelu}) was used extensively, however the problem of vanishing gradient\cite{michaelnielsen} has lead to the use of different functions known as rectified linear unit function (ReLU).

\begin{figure}[!ht]
    \centering
    \caption{\vspace{-0.1cm} A feedfoward Neural Network with one hidden layer. }
    \begin{subfigure}[b]{0.7\textwidth}
        \includegraphics[width=\textwidth]{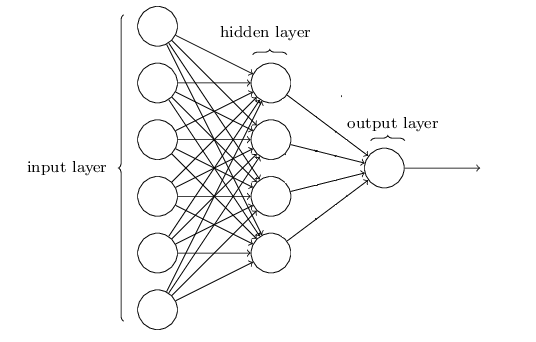}
        \label{fig:neuralnet}
    \end{subfigure}
    \vspace{-0.25cm}
    \caption*{\vspace{-0.1cm} Source: Adapted from \cite{michaelnielsen}. }
    \label{fig:neuralnetwork}
\end{figure}

The simplest Neural Network is the feedfoward Network ( Figure \ref{fig:neuralnetwork}). The name comes from the fact that the network feeds its input in a straight line from left to right. Each neuron outputs its result to every neuron in the next layer. (Newer archictectures, such as RNN \cite{kaparthy_rnn} and LSTM \cite{colah_lstm},  do not follow this principle and are a bit more complicated to visualize and understand.)

In Figure \ref{fig:neuralnetwork}, it is possible to see three different layers, with each column of circles representing a layer. The first one is the input layer, the last one is the output layer, and the layer in-between is called a hidden layer.

\begin{figure}[!ht]
    \centering
    \caption{\vspace{-0.1cm} A feedfoward Neural Network with one hidden layer of 4 neurons. }
    \begin{subfigure}[b]{0.7\textwidth}
        \includegraphics[width=\textwidth]{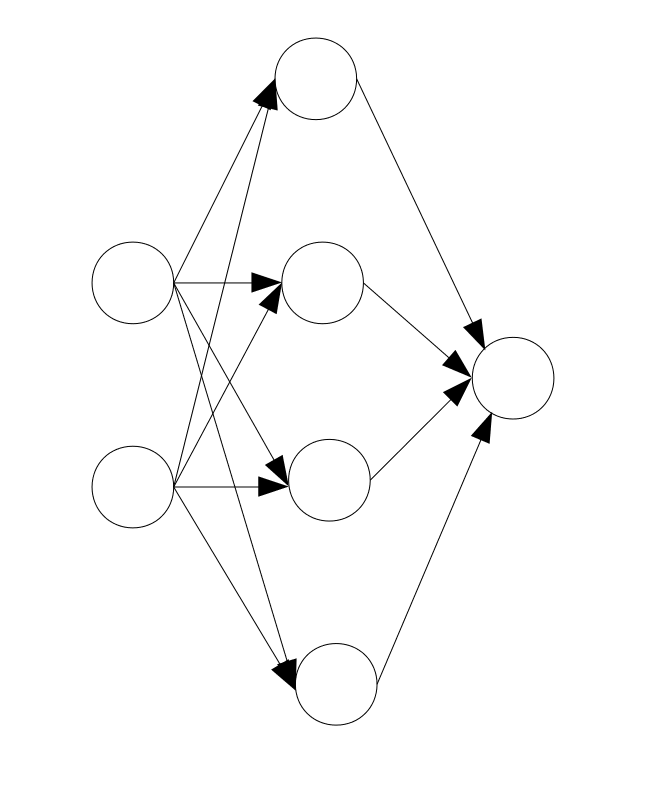}
        \label{fig:shallowcomp}
    \end{subfigure}
    \vspace{-0.25cm}
    \caption*{\vspace{-0.1cm} Source: Author. }
    \label{fig:shallowcomplexity}
\end{figure}

\begin{figure}[!ht]
    \centering
    \caption{\vspace{-0.1cm} A feedfoward Deep Learning netowork with 3 hidden layers of 2 neurons. }
    \begin{subfigure}[b]{0.7\textwidth}
        \includegraphics[width=\textwidth]{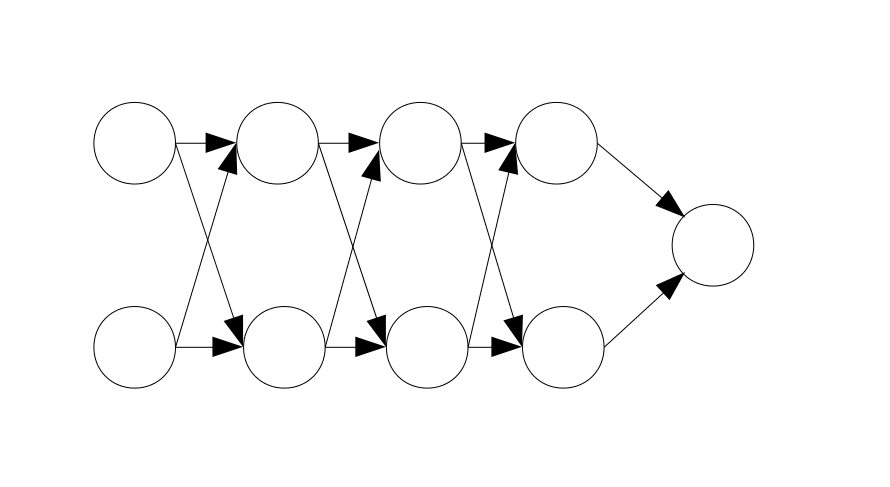}
        \label{fig:deepcomp}
    \end{subfigure}
    \vspace{-0.25cm}
    \caption*{\vspace{-0.1cm} Source: Author. }
    \label{fig:deepcomplexity}
\end{figure}

It is known that Neural Networks with only one hidden layer and large enough number of neurons can approximate any function \cite{Cybenko1989}\cite{hornik1991approximation}\cite{HORNIK1989359}. However, in practical terms it is computationally expensive to indefinitely increase the number of neurons in a single layer. It was the idea of increasing the number of hidden layers and the larger computational capacity of today's computers that allowed Deep Learning models to exist and be able to successfully learn their parameters. A Deep Learning model is simply a network with a multitude of hidden layers. The large number of possible combinations between hidden layers have shown remarkable results, increasing model complexity while maintaining feasibility \cite{michaelnielsen}. Figure \ref{fig:shallowcomplexity}  and Figure \ref{fig:deepcomplexity} show two different networks: the first one is a shallow Neural Network, with one hidden layer with a large depth (4 neurons in the hidden layer), meanwhile the second one is a Deep Learning network with 4 smaller hidden layers (2 neurons in hidden layer).


\begin{figure}[!ht]
    \centering
    \caption{\vspace{-0.1cm} Feedfoward Neural Network with appropriate activation function for learning. }
    \begin{subfigure}[b]{0.7\textwidth}
        \includegraphics[width=\textwidth]{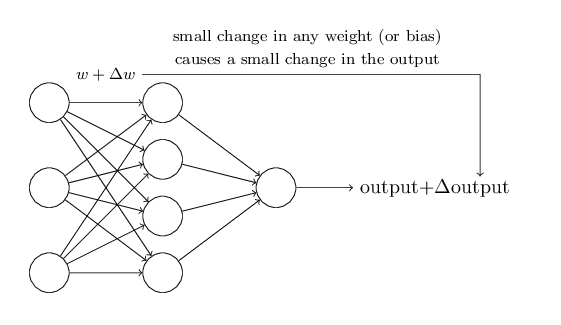}
        \label{fig:perc_change}
    \end{subfigure}
    \vspace{-0.25cm}
    \caption*{\vspace{-0.1cm} Source: \cite{michaelnielsen}. }
    \label{fig:perceptron_change}
\end{figure}

But can we learn the parameters? The learning problem is, in itself, a rather difficult process of optimization. We are looking for a set of variables that approximates our result to the correct output. The process of learning differs from the type of problem at hand, however, for the sake of simplicity, let us consider the supervised problem. In general, we desire the function to have a certain characteristic: that small changes in the parameters cause small changes in the output. This condition makes it possible to change our parameters just a little bit each time, and get closer to the solution (Figure \ref{fig:perceptron_change}). This explains the need for a smooth activation function, previous mentioned.

A different issue is to determine the function which is going to be optimized. Usually, a cost function $C$ (also known as  error function) is used after the output from the network is calculated. This cost function is responsible to evaluate the difference between the ground truth $y$ and the prediction $\tilde{y}$. From this evaluation, it is possible to finally optimize our parameters (weights and bias) and start learning. The general optimization problem can be stated as follows:

\begin{equation}\label{opt_basic}
\min_{\zeta} \quad  C_{\zeta}(y, \ \tilde{y}_{\zeta})
\end{equation}
where $\zeta$ are the parameters which define our model.

The optimization problem can be solved in a variety of ways, but the simplest and most common one is through gradient descent. Since all functions are defined by the user, it is easy to calculate the influence over each parameter. However, it is only possible to feasibly calculate the gradient because of a technique known as backpropagation \cite{michaelnielsen}\cite{goodfellow2016deep}, a combination of dynamic programming and the chain rule, where the calculation of one hidden layer is used to fasten the calculation of the gradient of previous layers. 

Overall, the Neural Networks models have shown good results in a series of different problems, using backpropagation and increased number of hidden layers to create feasible models to deal with hard problems. Yet, there are some problems that still need to be addressed. First, because of its nature, the result from a network model can not be easily understood by humans. That means that sometimes we may have an answer for a problem, but no way to tell why that is the answer. This is specially troublesome in some areas like medicine, banking, and any business that requires explanation to clients. Beyond that, there is also a growing concern that Deep Learning methods have good accuracy because they have so much parameters that it is possible for them to overfit the training data and achieve an apparent good solution. This could prove to be problematic when dealing with any factors that vary slightly from the expected. Lastly, and what will be the focus of our next chapter and motivation for designing a new model, is the problem of creating and using models. There is a myriad of networks architectures, optimization routines, parameters initialization techniques, and activation functions being used now in research and industry. There is a growing concern with how the researcher and business can pick between them, create new ones, communicate them, and have some proof, or at least stronger intuition about why they should help solve the problem. The lack of interpretation for model parameters is also a similar problem, which we intend to help mitigate. In the next chapter we are going to present three different interpretations of Neural Networks, and how they can help to tackle these problems.

\chapter{Neural Networks Interpretations}
\label{chapterinterpretability}

When dealing with a new problem or understanding something for the first time as a student, a powerful tool one can have is the ability to visualize geometrically and physically to comprehend how a tool, concept, algorithm, etc. works. Also, it is important to have completely different interpretation, visualizations, and representations of the same object. For example, a matrix is the fundamental linear algebra object, and can also be represented as geometrical transformations, color maps, graphs or images. This is no different when dealing with research problems: by having multiple ways to interpret phenomena or models, the user can get new insights into how and why something happens or works. For this reason, following last chapter's conclusion, we are going to present three different interpretations of Neural Networks. How they help us gain insight on the model, what we can extract from them, what problems they may have, and how we can create new models.


The first one is the classical approach, understanding Neural Networks as functions, and is widely used for a variety of objectives. The second one considers the transformations in the space of the data. Instead of a function adapting or changing to fit the dataset, the dataset  moves spatially to achieve a better configuration  before classification \cite{colah}. At the end of his work Olah \cite{colah} suggests exploring another interpretation, which is further developed in this dissertation, and based on the idea of vector fields. Instead of thinking about transforming the space, we are going to move the data along the direction of a vector field, similar to a fluid flowing from one place to another. The idea is to create a flow that can move the dataset to a better configuration before the classification.

\section{Classical Interpretation}
The classic way to look at a Neural Network is to view them as functions. This is an obvious way to interpret and analyze Neural Networks, given that the function is explicit in the formulation used in the model. From this interpretation, we look at the problem of learning as the task of fitting a specific function that is able to take our input data and produce the output we want. The entire Neural Network is, then, a composition of weighted sums of non-linear functions.

Formally speaking, we want to find the function $\mathcal{F}(X)$ which takes the original data $X$ and returns the image $Y$, the correct label/class:

\begin{equation*} \label{eq:inter_f}
\mathcal{F}:\ X \subset \mathbb{R}^{n}  \rightarrow  Y \subset \mathbb{R}^{m},
\end{equation*}
where $n$ is the data dimension and $m$ is the label dimension. A formal mathematical definition will help understand the next interpretations, as well as the connection between them.

One advantage that we get from visualizing, interpreting and thinking about Neural Networks as functions is familiarity with the mathematical object. People from computer science, mathematics, statistics, engineering, etc. are extremely familiar with functions. They know important properties of functions, such as: what it means to be  a continuous and smooth function, function interpolation, extrapolation and fitting of functions, etc. These are all common terms.

\begin{figure}[!ht]
    \centering
    \caption{\vspace{-0.1cm} A classification problem with two classes (orange and blue). On the first image the original dataset is shown. The second plot shows the decision boundary for each class after training with the use of colouring. The last plot illustrates the function value for each class in the domain. }
    \begin{subfigure}[b]{0.7\textwidth}
        \includegraphics[width=\textwidth]{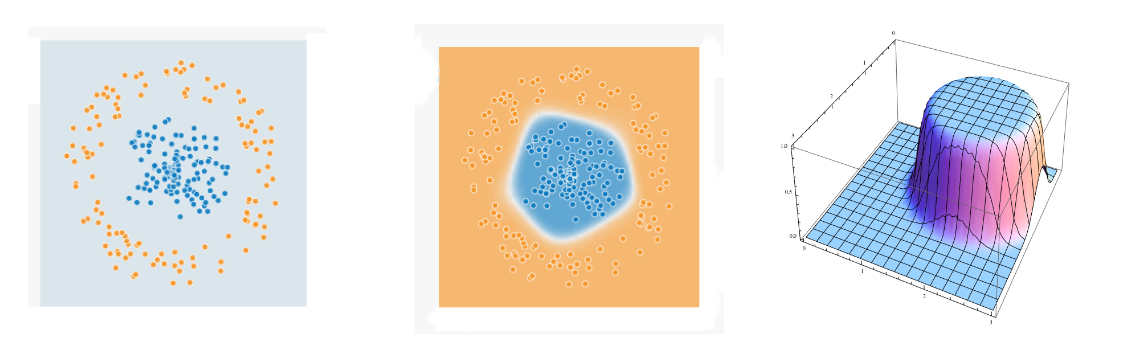}
        \label{fig:netfuc}
    \end{subfigure}
    \vspace{-0.25cm}
    \caption*{\vspace{-0.1cm} Source: \cite{tensorplay} and \cite{stackplay}. }
    \label{fig:netasfunc}
\end{figure}

For instance, take a simple problem of binary classification. Imagine now that our task is to be able to calculate the correct image of a certain function, provided a domain space $X$ which is the dataset. Let's say that we set one class with a value of zero, and the other with  one. Now, take a look at Figure \ref{fig:netasfunc}. We can see the original domain $X$ as the blue and orange dots, colouring each class. The function $\mathcal{F}(X)$, on the third plot, returns $Y$ depending on the input data, thus, separating the classes. Another visualization of functions which we can imagine is a level set of $\mathcal{F}(X) = 0.5$, a concept familiar to first year calculus students, that servers as the boundary between classes (see Figure \ref{fig:netasfunc}).

\begin{figure}[!ht]
    \centering
    \caption{\vspace{-0.1cm} The sigmoid function used by the network neuron and the step function it imitates. }
    \begin{subfigure}[b]{0.4\textwidth}
        \caption{Sigmoid function.}
        \includegraphics[width=\textwidth]{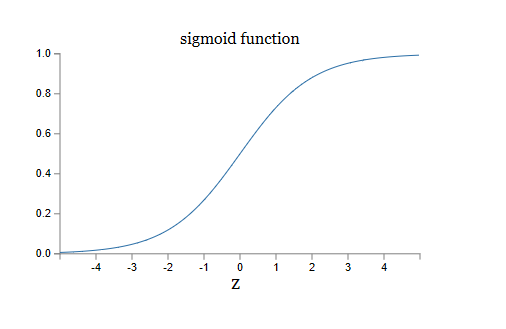}
        \label{fig:sigmoid}
    \end{subfigure}
    \begin{subfigure}[b]{0.4\textwidth}
        \caption{Step function.}
        \includegraphics[width=\textwidth]{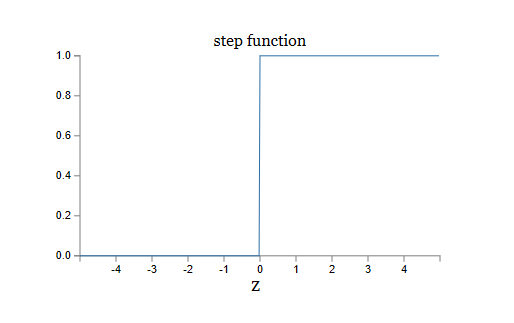}
        \label{fig:step}
    \end{subfigure}
    \vspace{-0.25cm}
    \caption*{\vspace{-0.1cm} Source: \cite{michaelnielsen}. }\label{fig:sig_step}
\end{figure}

This interpretation is  also useful to understand, for example, the universality of Neural Networks. The universality theorem states that it is possible to fit any function, within a small error, with the use of a single hidden layer, provided that it has enough neurons. The mathematical formal proof  is quite complex\cite{Cybenko1989}\cite{HORNIK1989359}\cite{hornik1991approximation} but as \cite{michaelnielsen} has shown, one can have a good intuition of the universality by visualizing how a series of step functions (which can be approximated by the sigmoid, Figure \ref{fig:sig_step}) can shape any curve on 2 dimensions.

\begin{figure}[!ht]
    \centering
    \caption{\vspace{-0.1cm} A generic function is approximated by a neural network with sigmoid activation functions. The function is represented in blue and the approximation in orange.}
    \begin{subfigure}[b]{0.4\textwidth}
        \includegraphics[width=\textwidth]{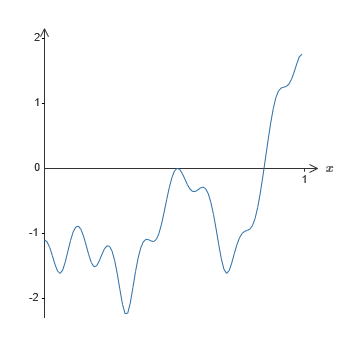}
        \label{fig:nielsen_f}
    \end{subfigure}
    \begin{subfigure}[b]{0.4\textwidth}
        \includegraphics[width=\textwidth]{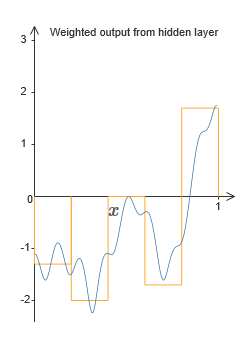}
        \label{fig:nielsen_appro}
    \end{subfigure}
    \vspace{-0.25cm}
    \caption*{\vspace{-0.1cm} Source: \cite{michaelnielsen}.}
\label{fig:universality}
\end{figure}

Figure \ref{fig:universality} shows  how a combination of a series of step functions can create the shape desired. Once we have small enough rectangles it resembles the idea of Riemann's infinitesimal approach towards integration, something that again every calculus student is exposed to (emphasizing once again the importance of familiarity). If we are able to create, theoretically, as many steps as we want, then it is possible to fit any curve within some error margin, just like the integration process by summation approximates the analytical solution.

Furthermore, how to choose activation functions is also an important, but difficult, task. The problem of vanishing gradient \cite{michaelnielsen}, and why the use of ReLU's as activation functions is important can be understood easier by taking the function viewpoint. Additionally, the smoothness requirement and need for non-linear functions can also be easier understood.

This exemplifies why a mathematically and visual based interpretation of Neural Networks is fundamental. Nonetheless, this interpretation can be confusing for humans sometimes. It is hard to tell what a decision made by a function of a million parameters is actually doing, even when we know precisely how each parameter is calculated and its mathematical influence on the final result. Moreover, even functions with few parameters are hard to be geometrically interpreted, and to imagine what change they will cause in the dataset. Therefore, the function approach does not make  use of the original space configuration, and how it can influence the model choice and result. Thus, other approaches started to emerge, using a different toolbox and point of view in an attempt to provide better answers to problems like vorticity and knots on the original data space.

There are also other known Neural Networks interpretations like the previously mentioned circuit logic interpretation in Chapter 2 \cite{michaelnielsen} and an interpretation based on functional programming\cite{colahfunc}. Nonetheless, in this dissertation our focus will be on geometrical interpretations.

\section{Space Transformation Interpretation}

To tackle some of the difficulties that rose as Neural Networks and Deep Learning models were being used, specially those concerned with the data space, a different approach cemented on the idea of space transformations was created. Olah \cite{colah} explored much of this interpretation, providing good examples and reasoning to its usefulness, besides viewing the problem that we are trying to solve from a different angle.

The general architecture of Neural Networks allows us to break the function $\mathcal{F}(X)$ in two separate parts: a transformation of space $Z(X)$, followed by a classifier $g(Z(X))$. Consider Equation \ref{eq:inter_rel1}. The original data space $X$ is transformed by the map $Z(X)$, moving the data to the transformed space $\tilde{X}$. In the transformed space, the separation happens with the use of a classifier defined by $g(\tilde{X})$.

\begin{equation} \label{eq:inter_rel1}
\mathcal{F}(X) = g(Z(X)).
\end{equation}

\begin{figure}[!ht]
    \centering
    \caption{\vspace{-0.1cm} A simple binary classification problem. One class is in blue and the other in red. }
    \begin{subfigure}[b]{0.4\textwidth}
        \includegraphics[width=\textwidth]{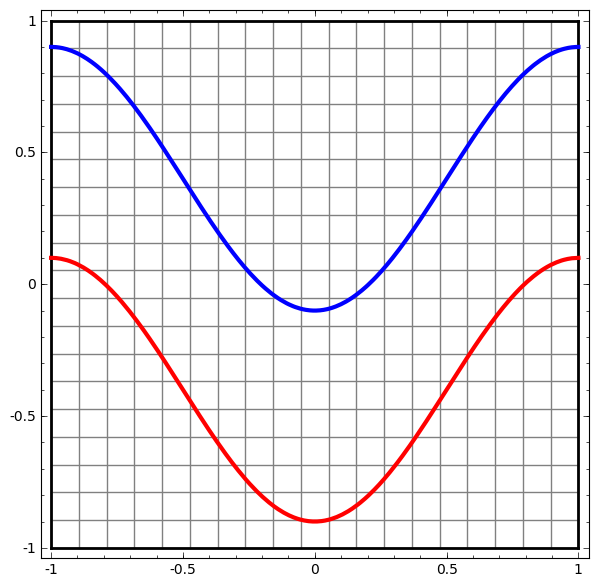}
        \caption{Training data used to learn the underlying pattern.}
        \label{fig:colah1_1}
    \end{subfigure}
    \begin{subfigure}[b]{0.4\textwidth}
        \caption{The decision line is shown on the original space, as the result of all layers combined.}
        \includegraphics[width=\textwidth]{Figures/colah_datanl.png}
        \label{fig:colah1_2}
    \end{subfigure}
    \vspace{-0.25cm}
    \caption*{\vspace{-0.1cm} Source: \cite{colah}. }\label{fig:colah1}
\end{figure}

Informally speaking, it is not hard to train the final classifier $g(\tilde{X})$, if the transformed space $\tilde{X}$ is linearly separable. Therefore, our problem is no longer to fit a function, but to enable a space transformation that, when executed over the original space, makes it possible to separate the transformed space easily (for example, linearly).

Just as we did before, it is possible to create a formal mathematical definition for the transformation and classification process:

\begin{equation*} \label{eq:inter_t1}
Z:\ X  \rightarrow  \tilde{X}, \qquad \tilde{X} \subset \mathbb{R}^{n},
\end{equation*}
\begin{equation*} \label{eq:inter_t2}
g:\ \tilde{X} \rightarrow  Y, \qquad Y \subset \mathbb{R}^{m}.
\end{equation*}

To provide some visual insight, take a look at Figure \ref{fig:colah1} where it is possible to see two different curves (red and blue) representing classes. Now, if we use a Neural Network it is possible to create a threshold between classes. This is the idea of functions being fitted, as we are looking at the space in its original configuration.

\begin{figure}[!ht]
    \centering
    \caption{\vspace{-0.1cm} Two ways to visualize the same classification problem. }
    \begin{subfigure}[b]{0.4\textwidth}
        \caption{The decision line is shown on the original space, as the result of all layers combined.}
        \includegraphics[width=\textwidth]{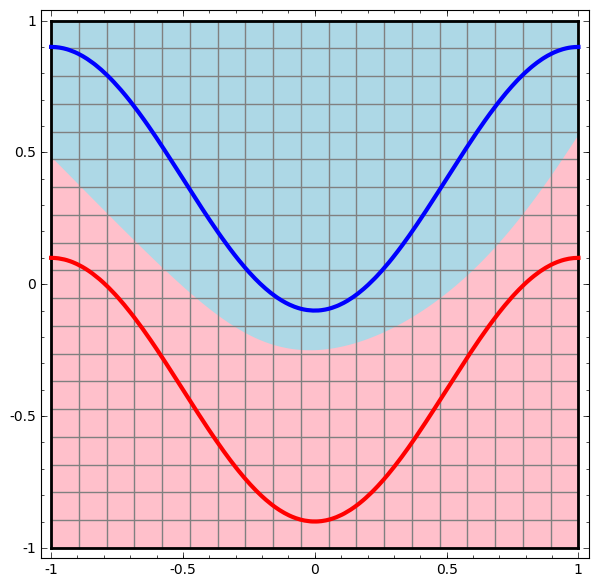}
        \label{fig:colah2_1}
    \end{subfigure}
    \begin{subfigure}[b]{0.4\textwidth}
        \caption{The linear classifier is applied after the transformation, evident by the straight line.}
        \includegraphics[width=\textwidth]{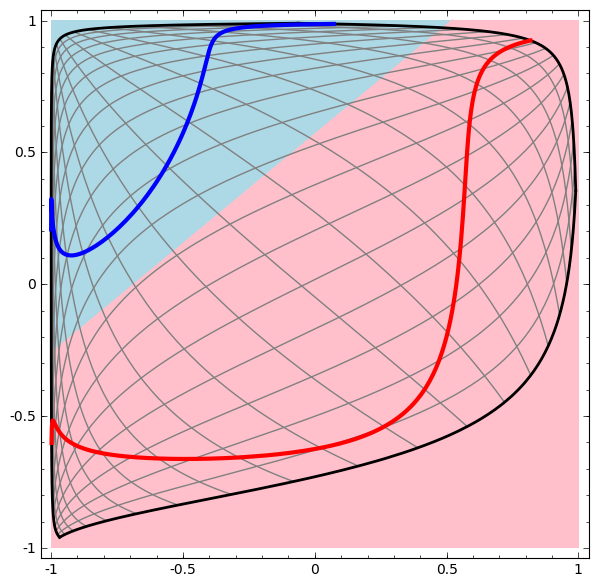}
        \label{fig:colah2_2}
    \end{subfigure}
    \vspace{-0.25cm}
    \caption*{\vspace{-0.1cm} Source: \cite{colah}. }\label{fig:colah2}
\end{figure}

However, looking at Figure \ref{fig:colah2} it is possible to see two different representations. We are looking at both  the threshold obtained by the neural network in the original space $X$, and the transformed data $\tilde{X}$ with its threshold $y = 0.5$. It is easier to pass a line between both classes in the second case. This is  one of the ideas behind this interpretation. The problem is divided between the learning task, which comprises of transforming the original space into some other space, and the final classification, made by a linear model such as a logistic regression. Informally, the final linear classification is not where the real learning lies, since the transformation beforehand is what made it possible.

\begin{figure}[!ht]
    \centering
    \caption{\vspace{-0.1cm} A different binary problem where one class (blue) encircles the other (red). In the second image it is possible to see the model shrinking the data because it lacks dimensionality. }
    \begin{subfigure}[b]{0.4\textwidth}
        \includegraphics[width=\textwidth]{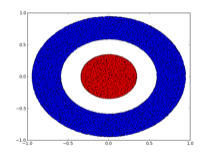}
        \label{fig:colah3_1}
    \end{subfigure}
    \begin{subfigure}[b]{0.4\textwidth}
        \includegraphics[width=\textwidth]{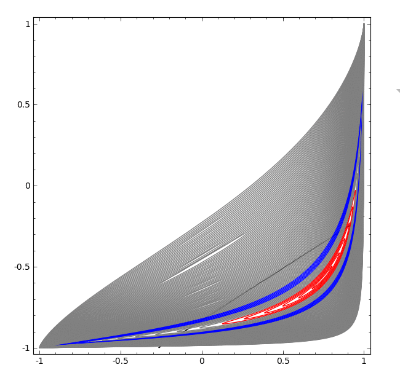}
        \label{fig:colah3_2}
    \end{subfigure}
    \vspace{-0.25cm}
    \caption*{\vspace{-0.1cm} Source: \cite{colah}.}
    \label{fig:colah3}
\end{figure}

This type of interpretation and problem redefinition makes it possible to see how model, and data dimensionality, can affect the model that we are going to use in a problem. For instance, Figure \ref{fig:colah3} shows two circles with different classes. It is simple enough to see that, if, for example, our transformation of this space is continuous from $\mathbb{R}^2$ to $\mathbb{R}^2$ this problem can not be solved. The result when trying to do so, even when using multiple hidden layer, is also shown at Figure \ref{fig:colah3}. The final classification can achieve accuracy values close to 80\%, but by looking at the transformed space we can see that it is not a good solution and the model has failed to truly learn.

\begin{figure}[!ht]
    \centering
    \caption{\vspace{-0.1cm} With three dimensions it is easy to pass a hyperplane between classes. }
    \begin{subfigure}[b]{0.4\textwidth}
        \includegraphics[width=\textwidth]{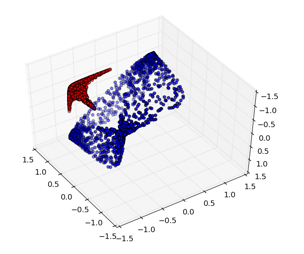}
        \label{fig:colah4_1}
    \end{subfigure}
    \vspace{-0.25cm}
    \caption*{\vspace{-0.1cm}Source: \cite{colah}. }\label{fig:colah4}
\end{figure}

However, when we change the network to allow a map from $\mathbb{R}^2$ to $\mathbb{R}^3$ to exist (check Figure \ref{fig:colah4}), the network creates a transformation in space where we can clearly see that it is easy to separate classes linearly.

\begin{figure}[!ht]
    \centering
    \caption{\vspace{-0.1cm} Another problem where two classes are entangled. The second image shows that the model is still capable of separating them linearly. }
    \begin{subfigure}[b]{0.4\textwidth}
        \includegraphics[width=\textwidth]{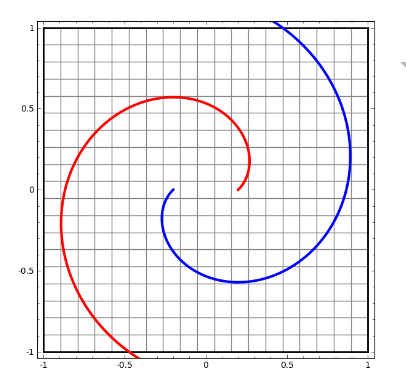}
        \label{fig:colah5_1}
    \end{subfigure}
    \begin{subfigure}[b]{0.4\textwidth}
        \includegraphics[width=\textwidth]{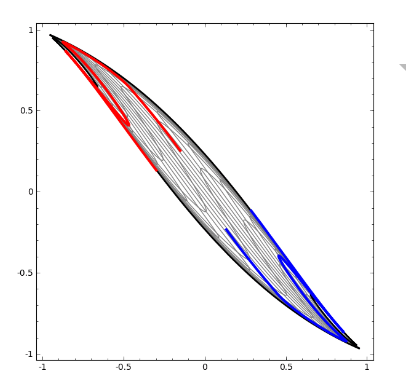}
        \label{fig:colah5_2}
    \end{subfigure}
    \vspace{-0.25cm}
    \caption*{\vspace{-0.1cm} Source: \cite{colah}. }
    \label{fig:colah5}
\end{figure}

\begin{figure}[!ht]
    \centering
    \caption{\vspace{-0.1cm} A different problem with classes more entangled. The second image shows that the same model does not perform well, as it is unable to bring a linear separation in space between classes. }
    \begin{subfigure}[b]{0.4\textwidth}
        \includegraphics[width=\textwidth]{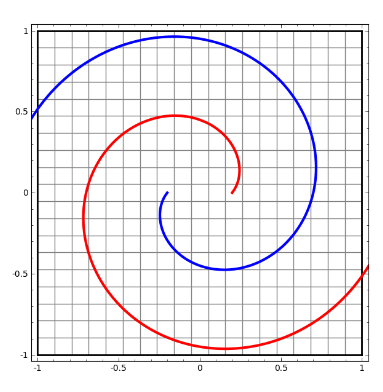}
        \label{fig:colah6_1}
    \end{subfigure}
    \begin{subfigure}[b]{0.4\textwidth}
        \includegraphics[width=\textwidth]{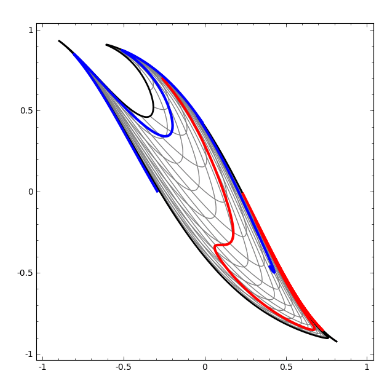}
        \label{fig:colah6_2}
    \end{subfigure}
    \vspace{-0.25cm}
    \caption*{\vspace{-0.1cm} Source: \cite{colah}. }\label{fig:colah6}
\end{figure}

A different aspect that one can examine is the variation caused by increasing data and model complexity. In Figure \ref{fig:colah5} and \ref{fig:colah6} it is possible to see how the same network performs poorly when data is more entangled, although it is not completely clear the reason why. Given that the model had the power necessary to do larger transformations the second case would also present a solution in two dimensions.

Still, this interpretation, even while providing insight into part of the issues that rose earlier does not solve all of our problems. For instance, it does not provide a clear solution to the problem of entanglement in data. Also, we can not make use of the vector fields and ODE's solving methods as easier as the next approach allow us to.

\section{Vector Field Interpretation}

The present work explores a third interpretation of Neural Networks inspired by the previous space transformation interpretation, thinking of vector fields as activation functions \cite{colah}. The interpretation is based on vector fields and the movement of points as a flow is developed.

As Equation \ref{eq:intro_v2} describes it, the transformed space $\tilde{X}$ and $Z(X)$ are reinterpreted. The $Z(X)$ transformation is viewed as a flow $\phi(X,\ t)$ which has a initial condition $X$, and is exposed to a vector field for time $t$. 

\begin{equation} \label{eq:intro_v2}
Z(X) = \phi(X,\ t) = \tilde{X},
\end{equation}
where $t$ is the elapsed time during which the particles were exposed to the vector field $\mathbf{K}(\phi(X, t))$. This means that, given initial condition $X$, integrating over $\mathbf{K}(\phi(X,\ t))$ gives us the same result produced by $Z(X)$, the transformed space $\tilde{X}$.

\begin{figure}[!ht]
    \centering
    \caption{Pipeline connecting the transformation and vector field interpretations.}
    \begin{subfigure}[b]{1.\linewidth}
         \includegraphics[width=\linewidth]{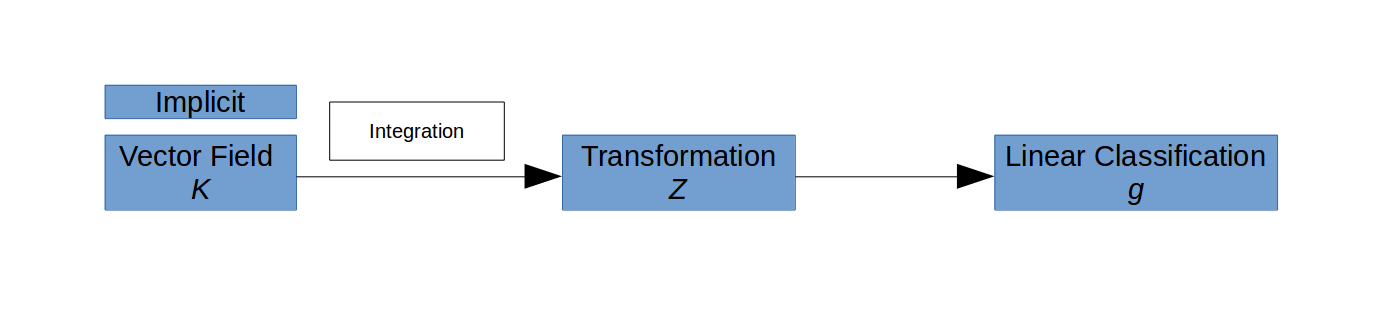}
         \label{fig:vec_pip}
    \end{subfigure}
     \vspace{-0.25cm}
     \caption*{Source: Author.}
     \label{vector_pipe}
 \end{figure}

Our interest lies in the process of moving the data, as such, we maintain the separation between the final classification layer and the displacement of data to linearly separable configuration.Figure \ref{vector_pipe} displays the pipeline which is followed. Behind the process of transformation there is an implicit field which is being integrated over time.

The main point is to consider the Neural Network model solving a problem of moving points in some useful manner, a flow $\phi(X, t)$ (defined in Chapter 2) moving a set of positions $X$. For a specific $t$ , $\phi(X,\ t)$ represents the set of particles' position after time $t$. This movement can be interpreted as an underlying implicit vector field $\mathbf{K}(\phi(X, t))$, which is produced by the series of activation functions present on neurons and leaves that at a different place $\tilde{X}$ \footnote{This approach is specially useful when we consider the data not to change its current dimension as it passes through the hidden layers. By maintaining the dimension we can keep the analogy with the flow of a fluid, although this is still possible when the dimensions changes in the hidden layers if we create dummy dimensions at $X$.}. Thus, we can define the problem from Equations \ref{eq:field_v1} and \ref{field_v2}:

\begin{equation*} \label{vec_v1}
\frac{\partial{\phi(X,\ t)}}{\partial{t}} = \mathbf{K}(\phi(X,\ t)),
\end{equation*}
\begin{equation*} \label{vec_v2}
\phi(X,\ 0) = X.
\end{equation*}

Concluding , Equation \ref{eq:intro_v3} presents the relationship between all interpretations.

\begin{equation} \label{eq:intro_v3}
\mathcal{F}(X) = g(Z(X)) = g(\phi(X,\ t)),
\end{equation}

\begin{figure}[!ht]
    \centering
    \caption{The first plot presents the dataset in the original space $X$ and the second plot shows the dataset in the transformed space $\tilde{X}$. Finally, the third and fourth plot displays transformed domain $\tilde{X}$ and the underlying vector field $\mathbf{K}(\phi(X, t))$.}
    \begin{subfigure}[b]{0.35\linewidth}
         \includegraphics[width=\linewidth]{Figures/original.jpg}
         \label{fig:vec_original}
    \end{subfigure}
     ~ 
    \begin{subfigure}[b]{0.35\linewidth}
         \includegraphics[width=\linewidth]{Figures/transformed.jpg}
         \label{fig:vec_transformed}
     \end{subfigure}
    
    \begin{subfigure}[b]{0.35\linewidth}
         \includegraphics[width=\linewidth]{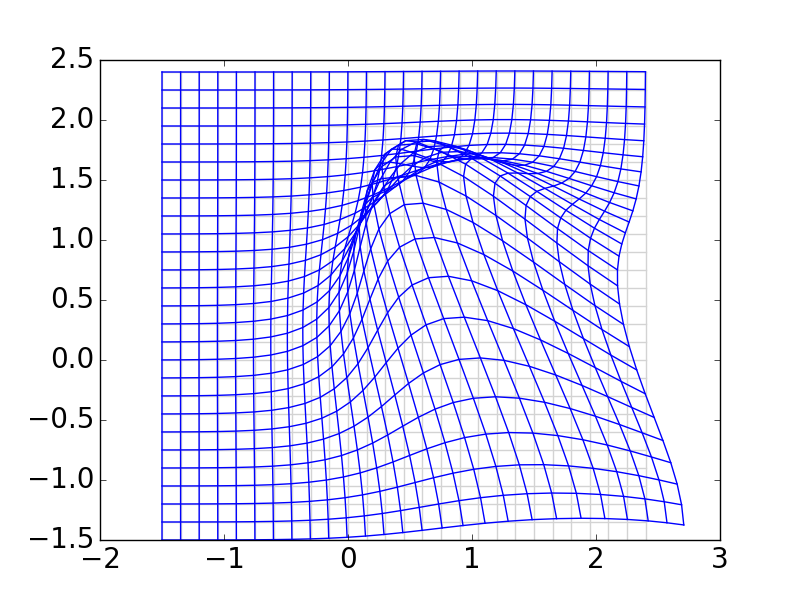}
         \label{fig:vec_mesh}
    \end{subfigure}
    \begin{subfigure}[b]{0.35\linewidth}
         \includegraphics[width=\linewidth]{Figures/vecfield.png}
         \label{fig:vec_vec}
    \end{subfigure} 
    
     \vspace{-0.25cm}
     \caption*{Source: Author.}
     \label{vector_interpretation}
 \end{figure}

It is possible in low dimension to visualize the data $X$ being displaced to the final configuration $\tilde{X}$ and the corresponding vector field $\mathbf{K}(\phi(X, t))$ applied (Figure \ref{vector_interpretation}).

Another benefit from this interpretation is that, as vector fields are widely know from engineering and physics, we can borrow interesting tools for visualization, interpretation of parameters, and also for the process of model creation. Maybe it is useful to stagnate part of the data in the flow, or to concentrate them somewhere, all of this is capable with certain types of vector fields\cite{fox1985introduction}\cite{marsden2011vector}.

Besides, by looking again at the flow analogy, it is interesting to note that by flowing along this hypothetical river, we can imagine the data to be taking a series of small steps at each time. These steps could be represented by each hidden layer, and as such we would have a clear meaning for the increase in hidden layers. Furthermore, it also brings into debate that the vector field equivalent of each hidden layer, created by the combination of all neurons in a hidden layer, should not change much between hidden layers, just as the vector field in a flow does not changes suddenly to completely different direction.

Finally, by choosing to interpret the Neural Networks like this, one can think of a particular  way to choose the vector field. Instead of learning a series of different functions that can be represented by some field it may be possible to just learn the field directly, given that we can define how to create a field from scratch. This may lead to a reduction in complexity of the model and may help fight the problem of overfitting. It also enables us to directly insert specific characteristics to the vector field depending on the necessity. Therefore, this new interpretation shows to be promising in providing new research ideas for the further development of Neural Networks and Deep Learning models.

\chapter{Vector Field Neural Networks}
\label{chaptervecfield}

In the last chapter we have spent sometime exploring the existing interpretations of Neural Networks and a new approach proposed in this work. Thus, one might beg the question: if it is indeed possible to interpret a Neural Network as an implicit vector field, why not learn the explicit vector field which would represent the change of the position of the dataset to become linearly separable.


\begin{figure}[!ht]
    \centering
    \caption{Pipeline with explicit vector field and numerical integration.}
    \begin{subfigure}[b]{1.\linewidth}
         \includegraphics[width=\linewidth]{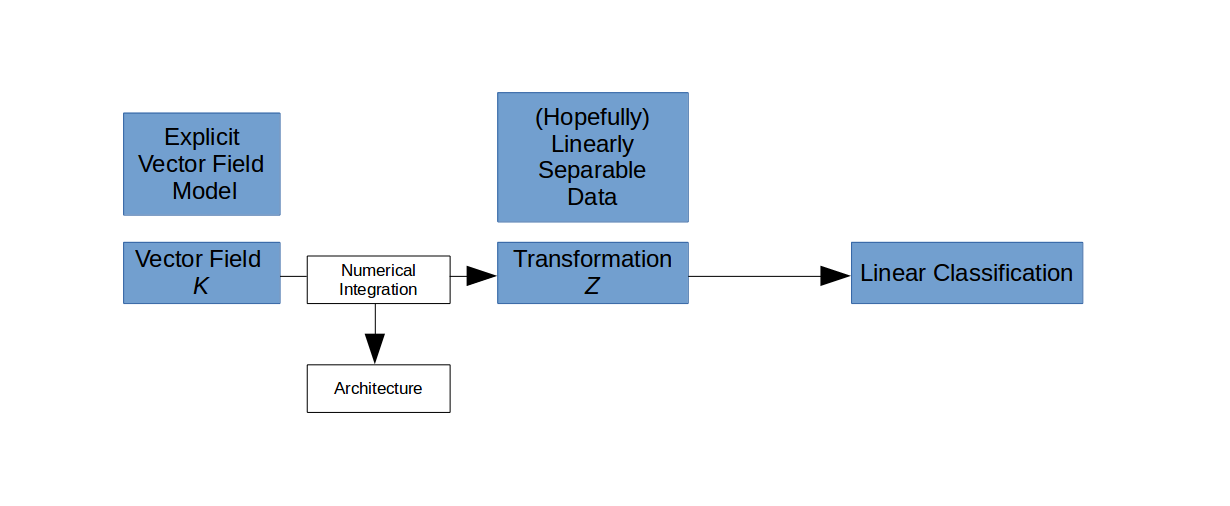}
         \label{fig:cap4_pip}
    \end{subfigure}
     \vspace{-0.25cm}
     \caption*{Source: Author.}
     \label{cap4_pipe}
 \end{figure}
 
Figure \ref{cap4_pipe} shows the modified pipeline with the idea of an explicit vector field. There is a choice of vector field model and numerical integration process enabling us to solve the ODE. This choice will define the architecture to find the solution which takes the data to the optimum linearly separable configuration for classification.

\begin{figure}[!ht]
     \centering     
     \caption{Different tasks that need to be addressed to find the problem solution.}
     \begin{subfigure}[b]{0.8\linewidth}
         \includegraphics[width=\linewidth]{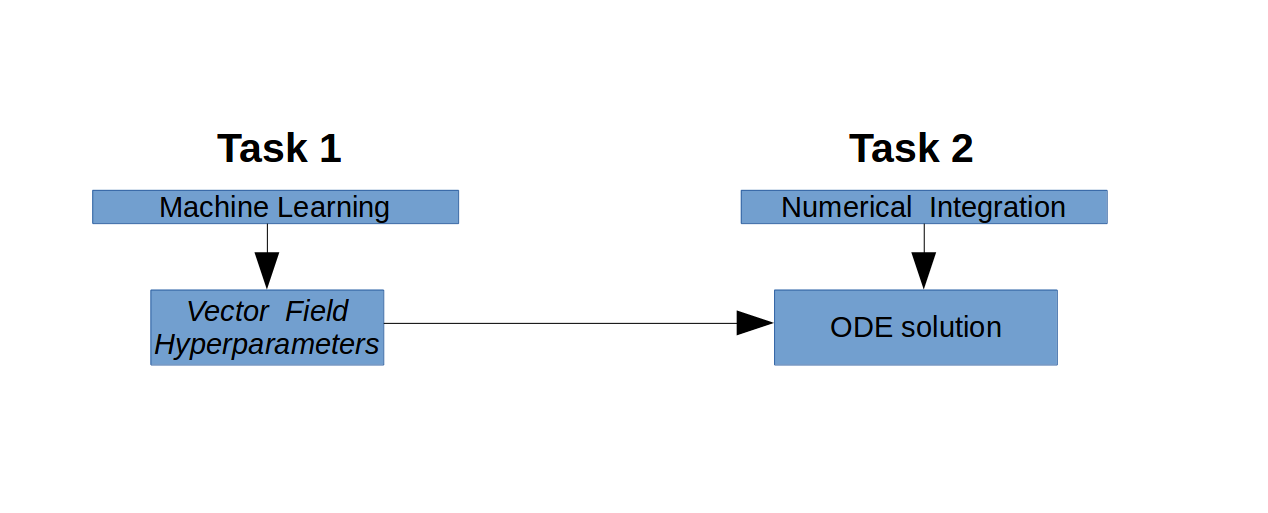}
         \label{fig:cap4_task}
     \end{subfigure}
     \vspace{-0.25cm}
     \caption*{Source: Author.}
     \label{cap4_tasks}
 \end{figure}

Here we need to make an important distinction in the problem (Figure \ref{cap4_tasks}). There are two different tasks: one is the learning process involved in finding the hyperparameters $\theta$ which define the vector field; the second is the problem of finding the transformation given the vector field, solved by the (numerical) integration of the vector field.


Now we can begin the endeavor of developing a new architecture inspired by the idea of an explicit vector field and moving of data along the vector field. First, let's state what purpose this architecture should fulfill and what we are trying to achieve. What is desired is an architecture capable learning a specific field, that can separate data in its correct classes. We are aiming to learn a vector field, and the parameter of interest that should be optimized are those that define this field. Note, however, that we have not yet spoke about the nature of this field. As one familiar with vector fields may already know, there is a myriad of different families of vector fields (vector field model) with different properties \cite{fox1985introduction}\cite{marsden2011vector}. For the moment we need to define our methodology regardless of integration technique and vector field model applied. The choice of integration method and vector field model, along with its properties should be one of the considerations that the user should take into account when trying to solve a specific problem. 


Let $\mathbf{K}$, a vector field on $\mathbb{R}^{n}$ be a smooth function $\mathbf{K}: \mathbb{R}^{n} \to \mathbb{R}^{n}$. Consider again the ordinary differential equation that was established in Chapter 2 (Equations \ref{eq:field_v1} and \ref{field_v2}):

\begin{equation*} \label{cap4_v1}
\frac{\partial{\phi(X,\ t)}}{\partial{t}} = \mathbf{K}_\theta(\phi(X,\ t)),
\end{equation*}
\begin{equation*} \label{cap4_v2}
\phi(X,\ 0) = X,
\end{equation*}

where the family of vector fields $\mathbf{K}_\theta(\phi(X, t))$ is defined by parameters $\theta$ (defined by the choice of vector field).

We propose to find the best vector field, with respect to an error function, in this family that transforms every point $X$ in the input space, in a point  $\tilde{X}_\theta$ in the transformed space such that points in distinct classes would be linearly separable. Intuitively, the vector field's streamlines represents the desired movement to enable classification.

At this moment we have defined the methodology at a high level. At implementation level, however, it is necessary to make certain choices, namely:  the numerical integration method which defines the architecture, the vector field model, and the optimization method.

\section{Architecture}

The first choice is the numerical integration method. It is possible to do something similar to what's been proposed in the present work using different ODE solving methods. We use the simplest method, Euler's method, as an integration technique to solve the ODE with the iteration process defined by Equation \ref{euler_it}, described in Chapter 2:

\begin{equation*} \label{cap4_it}
\phi_{i+1} = \phi_i + h\mathbf{K}_{\theta}(\phi_i,\ t)  \qquad 0\leqslant i\leqslant N.
\end{equation*}

\begin{sidewaysfigure}[!ht]
     \centering
     \caption{The first plot presents the dataset in the original space $X$. The second plot shows the architecure with Euler's method. The third plot displays the dataset in the transformed space $\tilde{X}$. The fourth and fifth plots exhibit the underlying vector field $\mathbf{K}(\phi(X, t))$ and the transformed domain $\tilde{X}$.}
     
     \begin{subfigure}[b]{0.22\linewidth}
         \includegraphics[width=\linewidth]{Figures/original.jpg}
         \label{fig:original}
     \end{subfigure}
     ~ 
     \begin{subfigure}[b]{0.48\linewidth}
         \includegraphics[width=\linewidth]{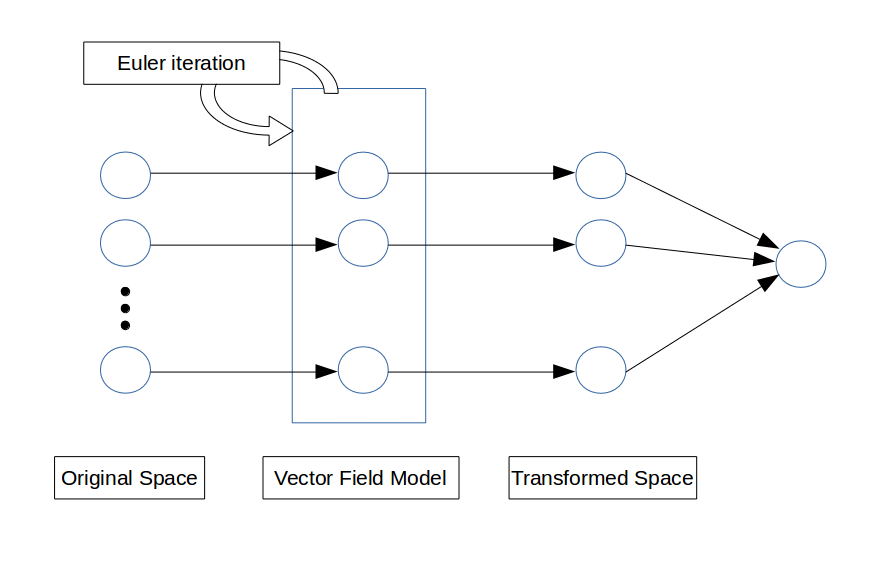}
         \label{fig:architecture}
     \end{subfigure}
         \begin{subfigure}[b]{0.22\linewidth}
         \includegraphics[width=\linewidth]{Figures/transformed.jpg}
         \label{fig:transformed}
     \end{subfigure}
     \begin{subfigure}[b]{0.25\linewidth}
         \includegraphics[width=\linewidth]{Figures/vecfield.png}
         \label{fig:field}
     \end{subfigure}
         \begin{subfigure}[b]{0.25\linewidth}
         \includegraphics[width=\linewidth]{Figures/meshgrid.png}
         \label{fig:meshgrid}
     \end{subfigure}
     \vspace{-0.25cm}
     \caption*{Source: Author.}
     \label{intro}
 \end{sidewaysfigure}

At Figure \ref{intro} is possible to visualize the input data being transformed by the vector field layer of the architecture. As previously expected the data moves to a configuration  in which it is easier to be separated linearly. The optimized vector field found through the training process can also be seen and it is possible to see the shift in space at the last image. Note that the architecture's last layer is a linear separator, which can be implemented using any known linear classifier. This provides a simplistic working example of our proposal.

The first decision of the general pipeline has been made. We have numerical integration method and, as a consequence, the architecture. However, in order to test the architecture, the need for a defined vector field model arises.


\section{Vector Field Model}

In this work  we chose a combination of gaussians and associated vectors to generate stationary fields (fields that do not change with time). The gaussian is a well known function extensively used in math, physics, statistics and engineering.

Vector field $\mathbf{K}_{\theta}(\phi(X,\ t))$ is presented in Equations \ref{kernel1} and \ref{kernel2}, where $V = \{V_1, V_2, \ldots, V_S\}$ are the vectors where $V_i \in  \mathbb{R}^{n}$ and $M=\{\mu_1, \mu_2, \ldots \mu_S \}$ are the centers of the gaussians  where $\mu_i \in  \mathbb{R}^{n}$, $\theta=V \cup M$, and $S$ is the number of gaussians,

\begin{equation} \label{kernel1}
G(\phi(X,\ t),\ \mu) = e^{-||\phi(X,\ t) - \mu||^2},       \qquad \mu \in \mathbb{R}^{n},
\end{equation}
\begin{equation}\label{kernel2}
 \mathbf{K}_{\theta}(\phi(X,\ t)) = \sum_{i =1}^{S}V_i G(\phi(X,\ t),\ \mu_i), \qquad  V_i \in \mathbb{R}^{n}.
\end{equation}

The gaussian presents some nice properties: being smooth, defined for the entire $\mathbb{R}^{n}$, and having parameters which have known intuitively and mathematical meaning (i.e. mean, $\mu$, and variance, $V$). Furthermore, if we do not solve the ODE, $Z(X)$ needs to be an analytically defined. It is important to remember that the solution for gaussian vector field can not be analytically calculated, reason why the use of numerical methods. Thus, by explicitly learning the underlying vector field and numerically finding $\phi(X, t)$ it is possible create patterns that $Z(X)$ can not,  illustrating the importance of this approach.

Please note that in a statistical sense the centers represent the distribution mean and the variance can be controlled through the $V_i$ parameters, which can be thought of as weight vectors that provide a direction to the vector field (see Figure \ref{intro} for an example). Additionally, by choosing a fixed family of vector fields it is now possible to calculate how many free parameters the model has, as a function of the number of gaussians and the dataset dimension. The number of free parameters can be calculated as following:

\begin{equation}\label{freeparameter}
p= 2Sn+n+ 1.
\end{equation}
where $p$ is the number of parameters. This will be useful later on, when we analyze the model time and complexity on different settings.

Now that we have defined the integration methods as well as the vector field model the last thing that we need to do is to solve the learning problem.


\section{Optimization}

There is a choice of linear classifier, cost function and optimization technique. Just like the two previous sections this is also a choice of the user. All but one of the choices made in this section are general choices of any Neural Network model. Many optimization techniques are available and in this dissertation we choose the gradient descent, a method which uses the derivative of the Cost function and a hyperparameter called learning rate ($\eta$) as step size to reach the minimum. Furthermore, the final layer is a logistic regression (Equation \ref{logistic}) that acts upon the transformed points $\tilde{X}_\theta$:

\begin{equation}\label{logistic}
\hat{y}_{\zeta} = \frac{1}{1+ e^{-(w^T \tilde{X}_{\theta} +b)}} ,\qquad w \in \mathbb{R}^{n}\ and\ b,\ \hat{y}_{\zeta}\in\mathbb{R}\ st.\ 0<\hat{y}_{\zeta}<1,
\end{equation}
where $w$ are the weights, $b$ the bias and $\zeta$ is the collection of model parameters $\{w,b,\theta\}$. 

The cost function is a binary cross entropy (Equation \ref{cross}) which will receive the predicted value $\hat{y}_{\zeta}$ from the logistic regression and compare it against the ground truth $y$.

\begin{equation}\label{cross}
C_{\zeta}(y, \hat{y}_{\zeta}) = \sum_{k=1}^{a}\big[ y\,ln(\hat{y}_{\zeta}) + (1-y)\,ln(1-\hat{y}_{\zeta})\big],
\end{equation}
where $a$ is the mini-batch training size. The mini-batch size is a common hyperparameter of Neural Networks, where the gradient descent uses only a portion of the available training data in each step. When the entire training set is used, it is common to state the training as full-batch.

Another common practice when using backpropagation is to include a penalization of the model parameters. This is typically done to prevent overfitting, and can be made in a variety of ways \cite{michaelnielsen}. The regularization, as it is called, increases the Cost function on purpose, taking into consideration the parameter which the user wants to reduce.

The important choice in this section is the use of a L2 regularization process for vectors $V_i$ in the optimization. The regularization of $V_i
$ has a direct geometrical (and statistical) meaning and will be later shown (\ref{fig:reg}) at the preliminary results that it performs desirable effects at the movement of $X$. 


Thus, the regularized optimization problem can be summarized in Equation \ref{cost_function}:

\begin{equation}\label{cost_function}
\min_{\zeta} \quad O_{\zeta, \lambda} =  C_{\zeta}(y, \hat{y}_{\zeta}) + \lambda\sum_{i=1}^{S}\|V_i\|^2,
\end{equation}
where the first term is the Cost function and the second term is the regularization with $\lambda$ being regularization hyperparameter.

\section{Preliminary Visual Results}

From this moment forward, in this work, let's consider the case when only one step is taken ($N=1$) in the integration process, and the step size $h$ is also equal to one. This reduces the problem to that of a shallow network with a simple optimization problem. Rightfully, someone might ask: why is it important to consider this case, when all of the work until now was promoted using the idea of multiple steps? The truth is that it is easier to proceed this way. To take smaller steps towards the final objective gives us the chance to deal with expected problems, as well as unexpected, in a more controllable version of our problem. Therefore, if something happens that is not expected when moving to a Deep Learning optimization, there is a standing ground on which we can take a look and search for clues. For instance, the accuracy of the shallow version serves as a benchmark to assert Deep Learning performance.


Finally, the gradient of $ O_{\zeta, \lambda}$ with respect to $\mu_{j,i}$ and $V_{j,i}$ are presented in Equations \ref{grad1} and \ref{grad2}:

\begin{equation}\label{grad1}
\frac{\partial{O_{\zeta, \lambda}}}{\partial{\mu_{i}}} = 2(\hat{y}_{\zeta} - y)G(X , \mu_i)(X - \mu_{i})(w^TV_i)
\end{equation}
\begin{equation}\label{grad2}
\frac{\partial{O_{\zeta, \lambda}}}{\partial{V_{i}}} = (\hat{y}_{\zeta} - y)G(X, \mu_i)w + \eta\lambda V_{i}.
\end{equation}

\begin{figure}[!ht]
     \centering     
     \caption{Specific architecture after setting integration method, number of steps, size of step and vector field model. Each $G$ represents one of $S$ gaussians composing the vector field}
     \begin{subfigure}[b]{0.8\linewidth}
         \includegraphics[width=\linewidth]{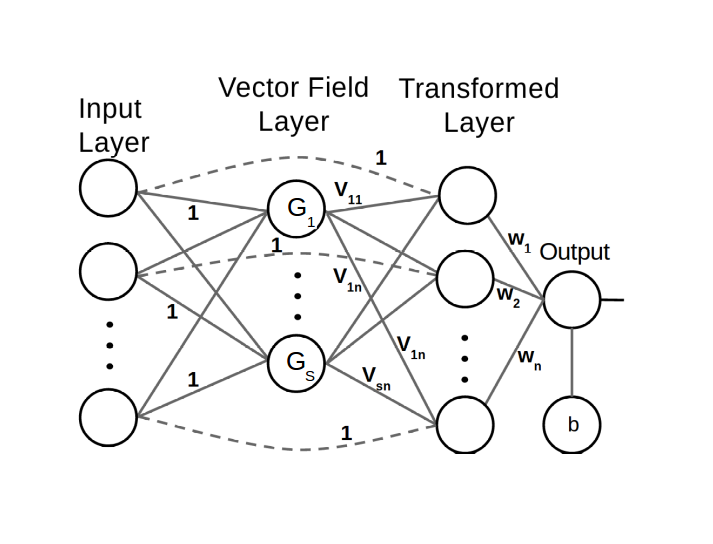}
         \label{fig:cap4_archimp}
     \end{subfigure}
     \vspace{-0.25cm}
     \caption*{Source: Author.}
     \label{cap4_archimple}
 \end{figure}

After taking a series of decisions with respect to the learning model we have the option to be more specific about the general architecture shown at Figure \ref{intro}. Considering the vector field model, our vector field $K$ is composed of a series of gaussians $G_i$, from one to $S$, which act on the original data space $X$. Since we only take one step with Euler method ($N=1$, $h=1$), it is also possible to take away the iteration process and simply feed the original dataset into the transformed space plus the output from the vector field model. The logistic regression is used as a final linear classifier and, as such, the last layer has the parameters $w$ and $b$ (Figure \ref{cap4_archimple}).

\begin{figure}[!ht]
    \centering
    \caption{\vspace{-0.1cm} Sin, Moons and Circle datasets, respectively.}\label{fig:datasets}
    \begin{subfigure}[b]{0.31\textwidth}
        \includegraphics[width=\textwidth]{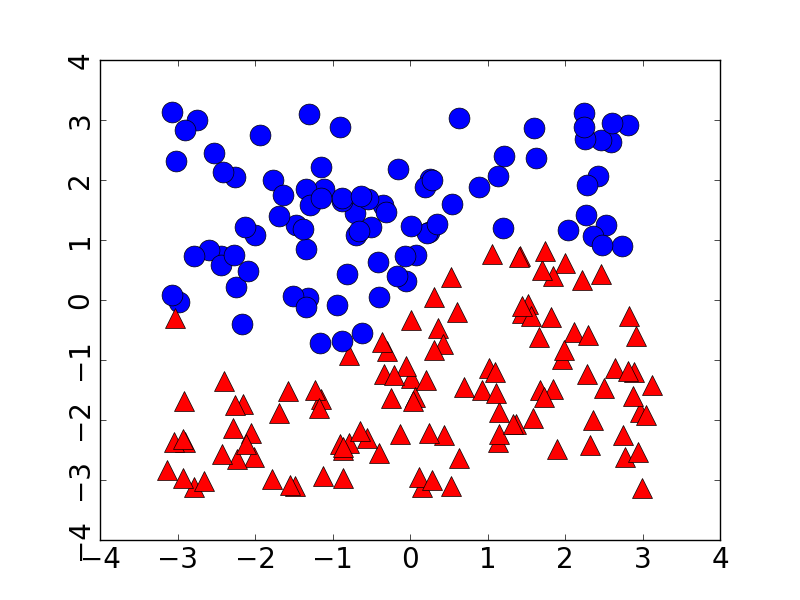}
        \label{fig:sin}
    \end{subfigure}
    ~ 
    \begin{subfigure}[b]{0.31\textwidth}
        \includegraphics[width=\textwidth]{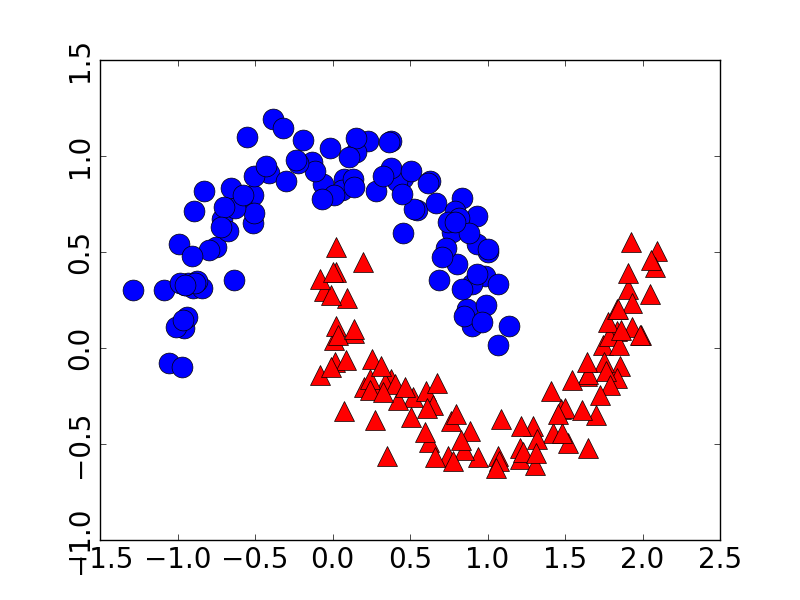}
        \label{fig:moons}
    \end{subfigure}
    ~ 
    \begin{subfigure}[b]{0.31\textwidth}
        \includegraphics[width=\textwidth]{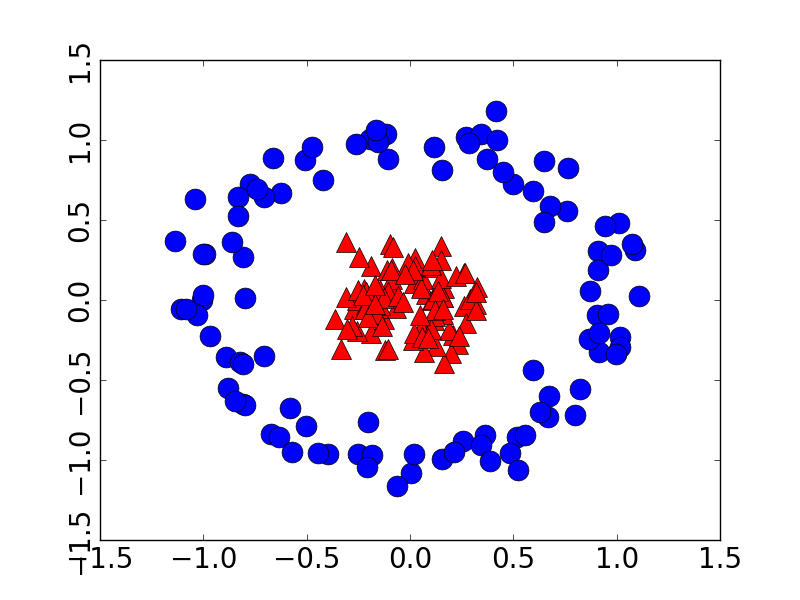}
        \label{fig:circles}
    \end{subfigure}
    \vspace{-0.25cm}
    \caption*{Source: Author.}
\end{figure}

Following with our step-by-step process, now it is possible to define a small experiment in two dimensions to showcase the architecture performance. Two \textit{scikit-learn} Machine Learning datasets, moons and circle, from \textit{scikit-learn} \cite{scikit-learn} and a sin dataset (created by the author) were used. Figure \ref{fig:datasets} shows the datasets.

\begin{figure}[!htb]
     \centering
     \caption{\vspace{-0.1cm}Cost vs. Epochs. \textit{Circles} dataset with $\theta = \{0.03, 0.3, 3.0\}$ respectively.}\label{fig:cost}
     \begin{subfigure}[b]{0.31\textwidth}
         \includegraphics[width=\textwidth]{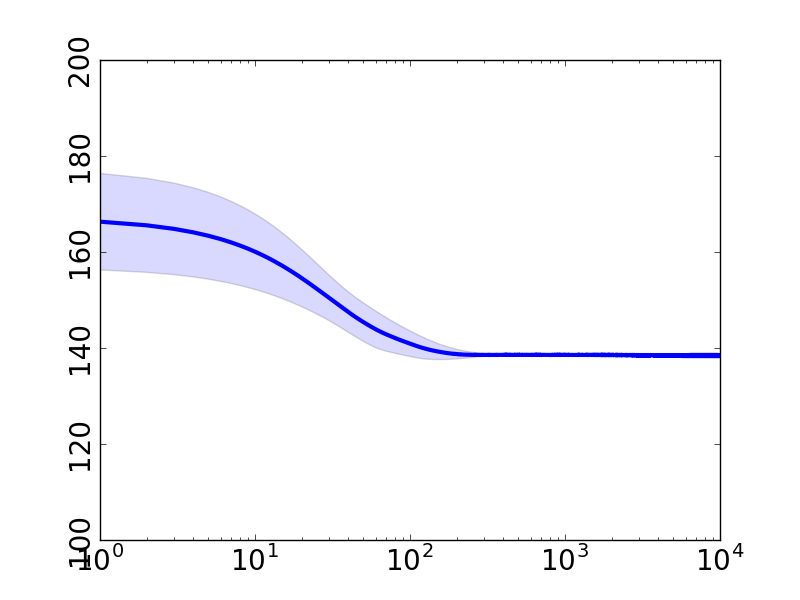}
         \label{fig:cost003}
     \end{subfigure}
     ~ 
     \begin{subfigure}[b]{0.31\textwidth}
         \includegraphics[width=\textwidth]{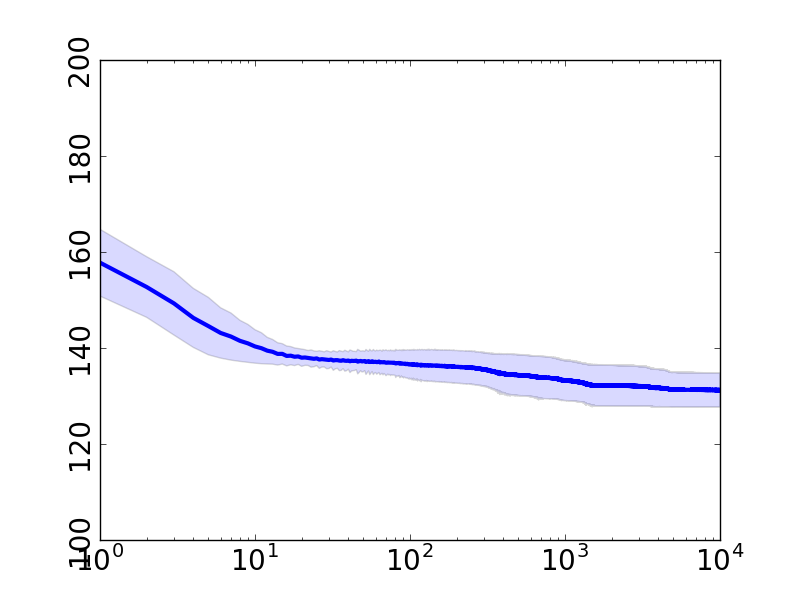}
         \label{fig:cost030}
     \end{subfigure}
     ~ 
     \begin{subfigure}[b]{0.31\textwidth}
         \includegraphics[width=\textwidth]{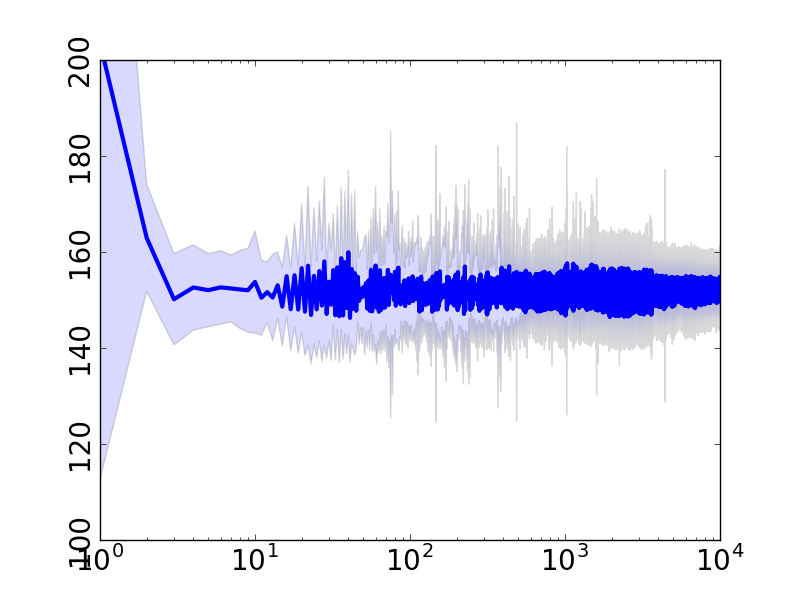}
         \label{fig:cost300}
     \end{subfigure}
    \vspace{-0.25cm}
    \caption*{Source: Author.}
    \label{cost_pre}
     
 \end{figure}

For the sake of measuring the algorithm's sensibility to parameter and hyperparameter initialization the three datasets were evaluated 30 times, with different initial guesses for centers $\mu_i$ and associated vectors $V_i$ and three different learning rates $\{0.03, 0.30, 3.00\}$. Regularization was not employed in this case ($\lambda = 0$). Training is made with the entire dataset in a full-batch fashion, hence, there is no validation/test set. Results are presented for the  cost function throughout 10000 epochs in the \textit{circles} dataset (Figure \ref{cost_pre}).

\begin{figure}[!htb]
    \centering
    \caption{Original space; vector field; and transformed space, respectively. }
    \begin{subfigure}[b]{0.3\textwidth}
        \includegraphics[width=\textwidth]{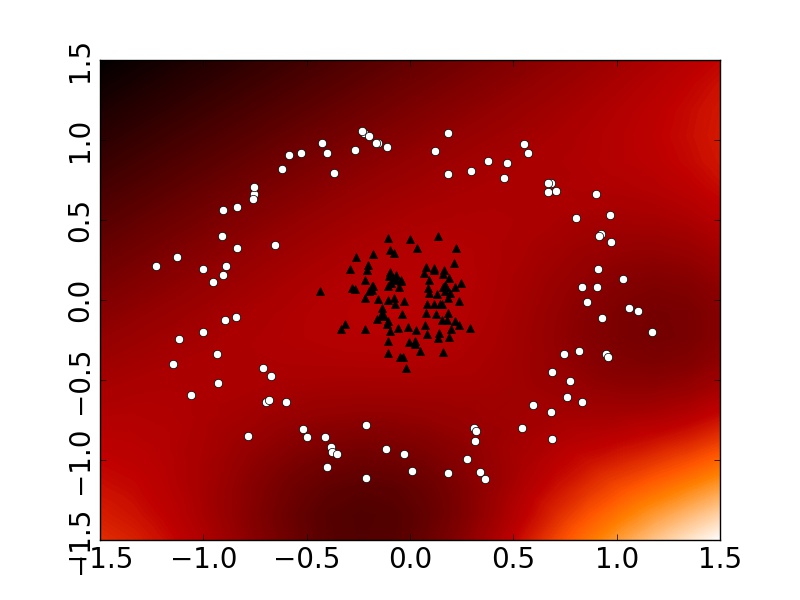}
        \label{fig:boundo030}
    \end{subfigure}
    ~ 
    \begin{subfigure}[b]{0.3\textwidth}
        \includegraphics[width=\textwidth]{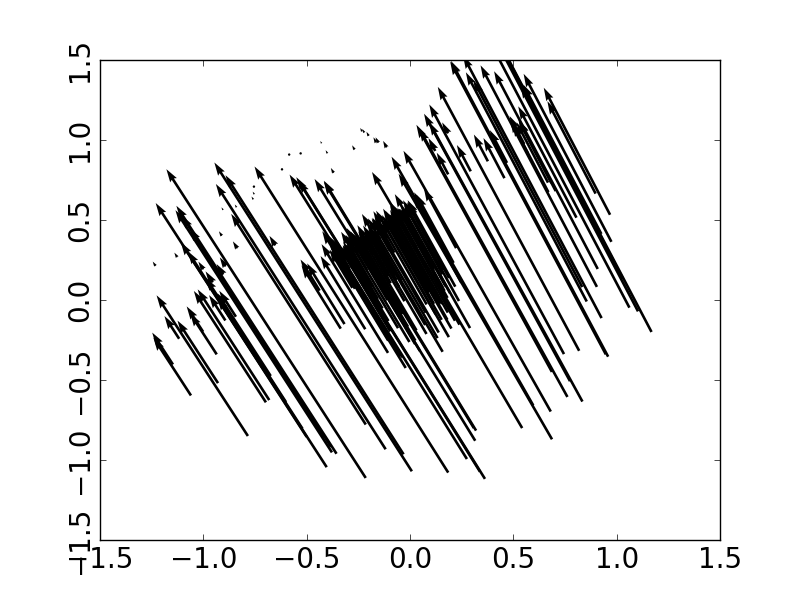}
        \label{fig:vec030}
    \end{subfigure}
    ~ 
    \begin{subfigure}[b]{0.3\textwidth}
        \includegraphics[width=\textwidth]{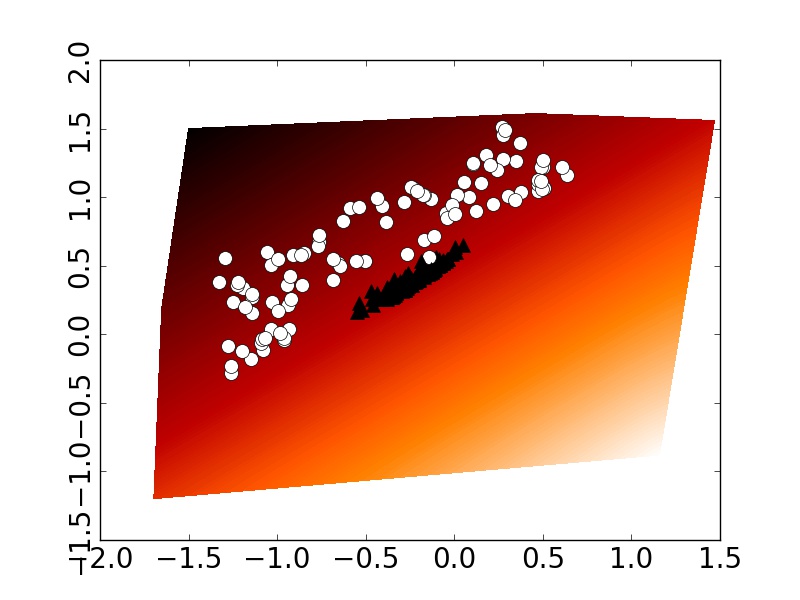}
        \label{fig:boundt030}
    \end{subfigure}
    \vspace{-0.25cm}
    \caption*{Source: Author.}
    \label{fig:bound}
\end{figure}

\begin{figure}[!htb]
    \centering\caption{Regularization over \textit{sin} dataset (5000 epochs, $\eta = 3.0$ and $\lambda=0.0005$).}
    \begin{subfigure}[b]{0.34\textwidth}
        \includegraphics[width=\textwidth]{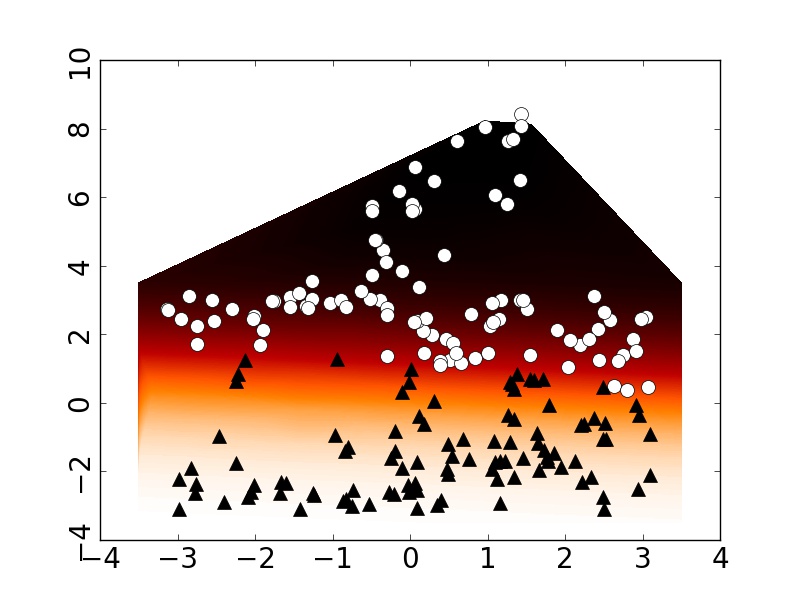}
        \label{fig:noreg300}
    \end{subfigure}
    ~ 
    \begin{subfigure}[b]{0.34\textwidth}
        \includegraphics[width=\textwidth]{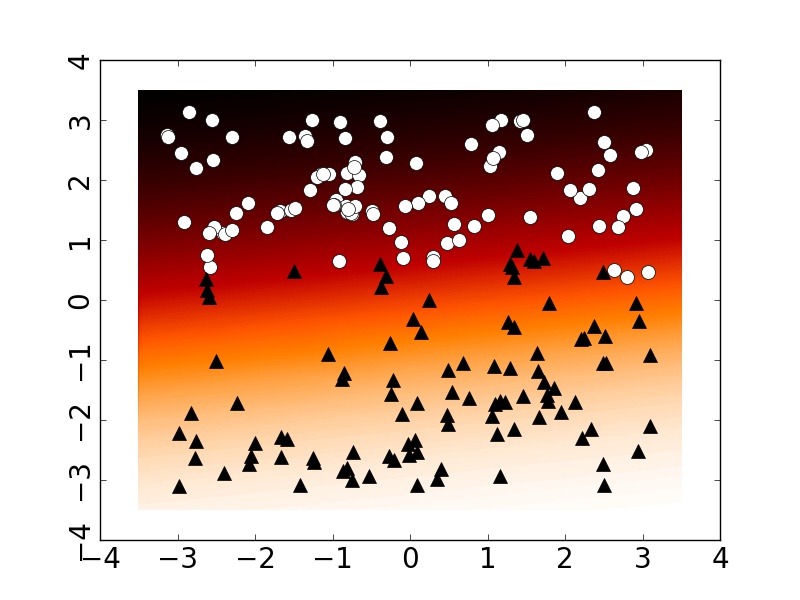}
        \label{fig:reg300}
    \end{subfigure}
    \vspace{-0.25cm}
    \caption*{Source: Author.}
    \label{fig:reg}
\end{figure}

Next, to better comprehend the algorithm's work, the boundary layer is calculated as a color map, with the original dataset points plotted over it. This way it is easier to see where the classifier is more certain of its decision and it is possible to analyze the relationship between the boundary layer in the original and transformed space (Figure \ref{fig:bound}). Finally, the effects of regularization can be visualized and compared in a color map of the \textit{sin} dataset (Figure \ref{fig:reg}).

Only the results for the \textit{circles} dataset are been shown here because both \textit{circles} and \textit{moons} have similar behaviour with it. Analyzing the cost function along the epochs for different learning rates shows the reduction of cost through epochs and an interesting pattern appears: as the learning rate increases the cost function (blue) and its standard deviation (light blue) becomes less smooth and the standard deviation increases as well.

In Figure \ref{fig:bound}, it is possible to see that the original boundary layer turns into a hyperplane on the transformed space. Although the algorithm achieved good classification by bending the space and extracting the center of the circle to the outside, it generates a superposition of different points in the original space. Thus, in the region where this happens, misclassification occurs which should be avoided.

One way to diminish the model's flexibility to create such extreme movements is by the use of regularization. Figure \ref{fig:reg} shows the regularization acting as a damper, smoothing the movements happening on the original space and preventing the overlapping of different points in the same region of the transformed space. Regularization has been used in Neural Networks for a long time now, reaching outstanding results in controlling overfitting, however, most of its motivation came from empirical results and insights taken from other problems, such as polynomial fitting. This result provides an example on how regularization acts to prevent the problem of overfitting.

Besides, because of how we chose the parameters that compose the vector field, there is a geometric meaning to the penalization caused by the regularization process. As we penalize the size of associated vectors, the vector field magnitude reduces, since the vectors give the direction and size of the steps taken by the data points.

The choice of $N =1$ and $h=1$ may facilitate overfitting, since the steps taken by data points are far greater and the restriction on how many movements can be made is at its maximum. It's possible to draw an analogy of regularization being automatically made when we choose good values of $h$ and $N$, as only small steps are taken and streamlines are followed, therefore avoiding the superposition of different classes.

The preliminary results show a promising future for this idea. The architecture was capable of learning the vector field as expected. Moreover, the final result can be visualized to be linearly separable by the last layer and it is possible to use regularization to control how strong the movements can be. It is important to move towards an evaluation of harder problems, with dimensions larger than two and complicated patterns. For this reason, the next chapter will provide the experiment design for such problems, which will be presented and discussed in Chapter 6.

\chapter{Design of Experiments}
\label{chapterDoe}

The preliminary results shown in the previous chapter provide us with a good visualization of the model behavior, functioning, and interpretation power, as well as some problems. However, we still need to analyze the model in a more robust manner, evaluating the results on harder problems, where the data dimension is greater than two (used in preliminary results). Furthermore, it will be shown that the model is capable of generalizing and how the behavior of hyperparameters influence computation time, model complexity and classification performance.

This chapter intends to detail the methodological design which satisfy our needs. The research method chosen is the computational experiment, and now we are going to explore the metrics, datasets, and structure of the problems to be analyzed.

\section{Datasets}

The choice of datasets also exerts an elevated importance. It tests the model against different cases of classes balancing (amount of data that belongs to each class), data dimension, and number of instances.

\begin{table}[!ht]
\begin{center}
\caption{Datasets chosen for experiment.}
\begin{tabular}{ |c|cccc| } 
\hline
Dataset & dimension & \# of instances & \% class 0 & \% class 1 \\
\hline 
Banknote & 4 & 1372 & 55.5 & 44.5\\ 
Ionosphere & 34 & 351& 64 & 36\\ 
Cryotherapy& 8 & 90 & 46.7 & 53.3 \\ 
Immunotherapy& 7 & 90 &  21.1 & 78.9\\
Pima Diabetes& 8 & 768 & 65.1 & 34.9 \\
\hline
\end{tabular}
\label{datasettable}
\caption*{ Source: Author.}
\end{center}
\end{table}

 The 5 datasets chosen \cite{Dua:2017}\cite{KHOZEIMEH2017167}\cite{KHOZEIMEH20171672} can be thorough analyzed by looking at Table \ref{datasettable}, where we can note the presence of an unbanlaced dataset ( Immunotherapy ). Moreover, the dimension varies from 4 to 34 and the number of available instances of training varies from 90 to 1372.
 
 The choice took into consideration 5 aspects: availability, cost, problem type, input data and feasibility. For the availability, the datasets needed to be open and easily accessible for reproducibility. As such, most come from the UCI database\cite{Dua:2017}, with the exception of the Pima Indians Diabetes \cite{pimakaggle}. Finally, only freely available, binary classification problems with  numerical input data for all entries and dimensions and that we can solve with limited computing resources are considered.

\section{Metrics}

First we need to study how the hyperparameters affect the performance of classification. For that, 3 well-known metrics will be used: the accuracy, mean squared error (MSE), and area under the receiver operating characteristics curve (AUCROC). Accuracy measures the percentage of correctly predicted values and is a classical interpretable metric to evaluate the classification performance. The AUCROC allows us to evaluate the performance independently of class balance inside the dataset, something that can pass unnoticed by looking only at accuracy. It can be seen as the expectation that a uniformly drawn random positive is ranked higher than an uniformly drawn random negative. At last, the MSE gives us the  measurement of mean the distance from our choice of class with respect to the expected output, whether or not they have been correctly classified. It can be helpful to understand if our predictor is good at achieving the expected value, or only at differentiating classes within a certain threshold.

The hyperparameters that need to be analyzed are the following: learning rate, mini-batch size, and number of gaussians inside the model. The learning rate dictates the step size of the gradient descent function inside the model, used to minimize the cost function of the classification problem. The number of gaussians used changes the complexity of the model, adding more free parameters. Finally, the mini-batch size changes the functioning of the learning process by changing the input used by the gradient descent.

The evaluation of these hyperparameters will be done on aforementioned datasets and the experiment will be repeated 30 times to guarantee a large enough sample for the Central Limit Theorem to hold true and our mean and standard deviation reported values approximate a normally distributed function. This leads to a more realistic value to be expected, since there are random factors that come in place when training the model, such as the weights initialization and batches generated.

By the end of those experiments it will be possible to do a reasonable analysis to choose the value  of the hyperparameters that will be used on the further experiments.

The next step consists of measuring the computational time and model complexity as we increase the dimension problem and the number of gaussians being used. A true measure of complexity for standard Neural Networks, such as the VC-dimension \cite{vapnik2015uniform}, is an open problem and we will adopt whats is typically reported in Deep Learning research by taking into consideration the number of free parameters that the model has. This will be calculated by  Equation \ref{freeparameter}.

Just like in the previous case, the experiments will be run 30 times, and mean and standard deviation will be reported. The hyperparameters, with the exception of the number of gaussians, will be fixed and the datasets used are going to be randomly generated with different dimensions.

Finally we can evaluate the model in terms of classification performance, and compare it against other well-known models of Machine Learning used in classification problems. The models chosen were the following: Naive Bayes(NB), Support Vector Machines(SVM) and  Shallow Feed Forward Neural Networks(FFNN). The NB model will act as a baseline of comparison, in some sense as a lower bound of what we expect any model to achieve. The SVM and FFNN are more complex, and we generally expect better results. The test metrics are the same used previously when evaluating the hyperparameters, that is: accuracy, MSE, and AUCROC.

\section{Experimental Planning}

Now that we have the metrics and the data which we want to work on, it is possible to plan the computational experiments that will be carried out. Table \ref{hyperparametertable} contains the hyperparameters search used. The number of gaussians and mini-batch size will be computed in a grid, with different learning rates being reported for each plot. This is two-dimensional visualization of a three dimension grid search including each hyperparameter. Other experiment variables are stated below:

\begin{itemize}
    \item  Epochs: 1000.
    \item Training-Test split: 80\%/20\%.
    \item No regularization ($\lambda = 0$).
    \item  Datasets : Banknote, Ionosphere, Cryotherapy, Immunotherapy and Pima Diabetes.
    \item  Metrics: accuracy (ACC), mean squared error (MSE), and area under the receiver operating curve (AUCROC).
\end{itemize}

\begin{table}[!ht]
\begin{center}
\caption{Hyperparameters grid for evaluation. The normalized values of hyperparameters are multiplied by the constant to suit different datasets.}
\begin{tabular}{ |c|cccc| } 
\hline
Hyperpameters & start & stop & constant & grid size \\
\hline 
number of gaussians & 0.50 & 5.00 & dimension (n) & 10\\ 
batch size  & 0.10 & 1.00 & training set size & 10\\ 
learning rate & 0.01 & 1.00 & 1& 5\\ 
\hline
\end{tabular}
\label{hyperparametertable}
\caption*{ Source: Author.}
\end{center}
\end{table}

After the completion of the first round of experiments and posterior analysis, a choice for mini-batch size, learning rate, and number of gaussians has been made and will be used on the last experiments. Table \ref{timecomplexitytable} presents the computational experiments to evaluate the computational time and complexity of the model in a grid search. Notice that learning rate does not matter, since we are not interested in prediction performance. Furthermore, even though it affects computing time, mini-batch size does not affect complexity, and will remain fixed as well.

\begin{table}[!ht]
\begin{center}
\caption{Computational experiments for evaluation of time and complexity of the proposed model.}
\begin{tabular}{ |c|cc| } 
\hline
metrics & dataset dim & \# gaussians \\
\hline 
Time & 1-100 & 1-100  \\ 
Complexity  & 1-100 & 1-100 \\ 
\hline
\end{tabular}
\label{timecomplexitytable}
\caption*{ Source: Author.}
\end{center}
\end{table}

Finally, the last round of experiments is designed to compare the classification accuracy of different models against the Vector Field Neural Networks (VFNN) model proposed in this dissertation. In order to maintain the comparison as fair as possible, the same number of epochs will be applied to all iterative models, and the number of neurons inside the FFNN hidden layers will be chosen to have a similar number of free parameters as VFNN. This is possible by computing the number of free parameters in a FFNN:

\begin{equation}\label{neural_p}
p = nl + 2l + 1,
\end{equation}
where $l$ is the number of neurons inside the hidden layer.

Combining Equations \ref{neural_p} and \ref{freeparameter} it is possible to calculate $l$ based on the choice of gaussians:

\begin{equation}\label{neural_free_p}
l = \frac{2Sn+n}{n+2}.
\end{equation}

For implementation purposes the number found in equation \ref{neural_free_p} will be rounded to the closest integer.

Table \ref{comparisontable} we have the hyperparameters used at the performance evaluation by the VFNN, after extracting the optimum from the hyperparameters grid evaluation. The remaining experiment variables are the following:

\begin{itemize}
    \item  Epochs: 5000.
    \item Training-Test split: 80\%/20\%.
    \item No regularization ($\lambda = 0$).
    \item  Datasets : Banknote, Ionosphere, Cryotherapy, Immunotherapy and Pima Diabetes.
    \item  Metrics: accuracy (ACC), mean squared error (MSE), and area under the receiver operating curve (AUCROC).
    \item Models: Naive Bayes (NB), Shallow Feed Forward Neural Networks (FFNN), Support Vector Machines, and Vector Field Neural Networks (VFNN).
\end{itemize}

\begin{table}[!ht]
\begin{center}
\caption{Hyperparameters used by the VFNN model for performance evaluation against other models.}
\begin{tabular}{ |c|ccc| } 
\hline
Dataset & number of gaussians & batch size  &learning rate \\
\hline 
Banknote & 20 & 987 & 0.2575\\ 
Ionosphere  & 280 & 33 & 0.7525\\ 
Cryotherapy & 64 & 21 & 0.2575\\ 
Immunotherapy  & 21 & 3 & 0.010\\ 
Pima Diabetes & 368 & 40 & 0.2575\\ 
\hline
\end{tabular}
\label{comparisontable}
\caption*{ Source: Author.}
\end{center}
\end{table}

\chapter{Results and Discussion}
\label{chapterResults}

\section{Hyperparameters}

Because a large amount of data that was generated by the experiments the complete results can be found at Appendices \ref{appendicebanknote}, \ref{appendiceionosphere}, \ref{appendicecryotherapy}, \ref{appendicepimadiabetes} and \ref{appendiceimmunotherapy}. Here, our attention will be directed at displaying and examining the behavior found, with examples extracted from the appendices.

\begin{figure} [!ht]
    \centering
    \caption{\vspace{-0.1cm} Accuracy mean and standard deviation measurements for the VFNN on the Banknote dataset with $\eta$ = 0.2575. }
    \begin{subfigure}[b]{0.45\textwidth}
        \includegraphics[width=\textwidth]{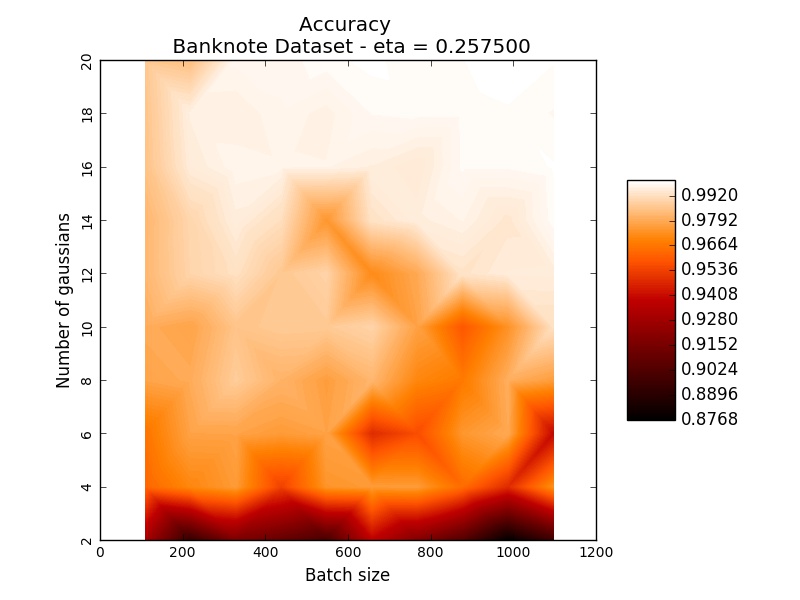}
        \label{fig:accbank025mean}
    \end{subfigure}
    \begin{subfigure}[b]{0.45\textwidth}
        \includegraphics[width=\textwidth]{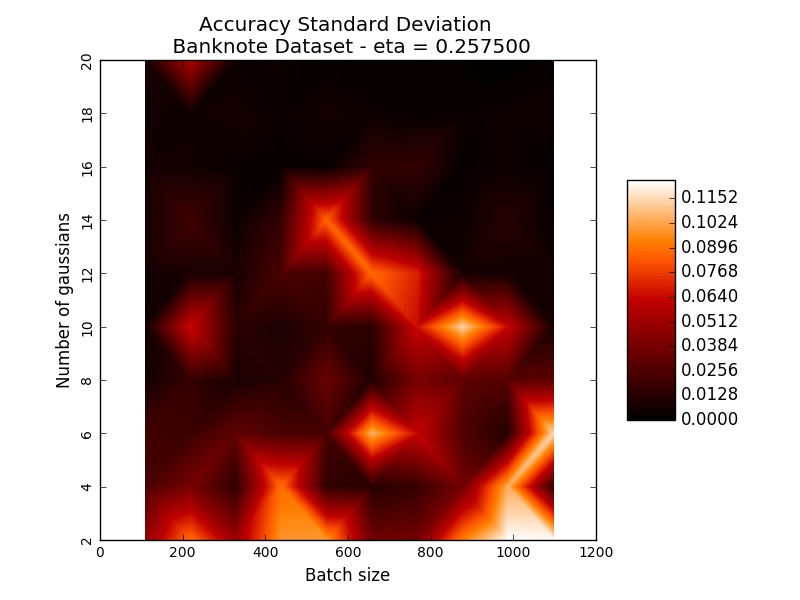}
        \label{fig:accbank025std}
    \end{subfigure}
    \vspace{-0.25cm}
    \caption*{\vspace{-0.1cm} Source: Author. }
    \label{fig:accbank025}
\end{figure}

The decision of visualizing the metrics results on two dimensional color plots was intentionally taken in order to display certain behaviors that one dimensional plots could not. At the Banknote dataset, Figure \ref{fig:accbank025} is specially worth of notice. Here, it is possible to see a bidimensional pattern that, as both batch size and number of gaussians increase, the accuracy is better.

\begin{figure} [!ht]
    \centering
    \caption{\vspace{-0.1cm} AUCROC mean for the VFNN on the Pima Diabetes dataset with $\eta$ = \{0.2575, 0.505\}. }
    \begin{subfigure}[b]{0.45\textwidth}
        \includegraphics[width=\textwidth]{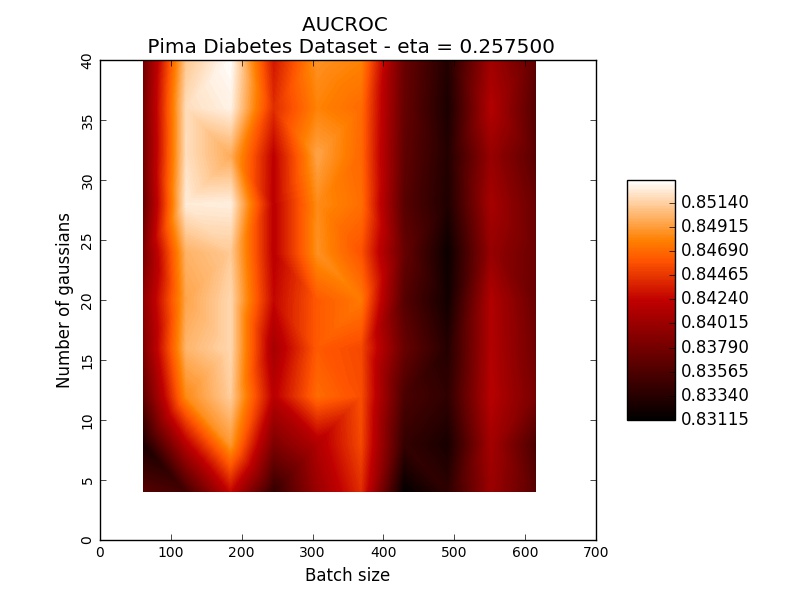}
        \label{fig:rocpima025mean}
    \end{subfigure}
    \begin{subfigure}[b]{0.45\textwidth}
        \includegraphics[width=\textwidth]{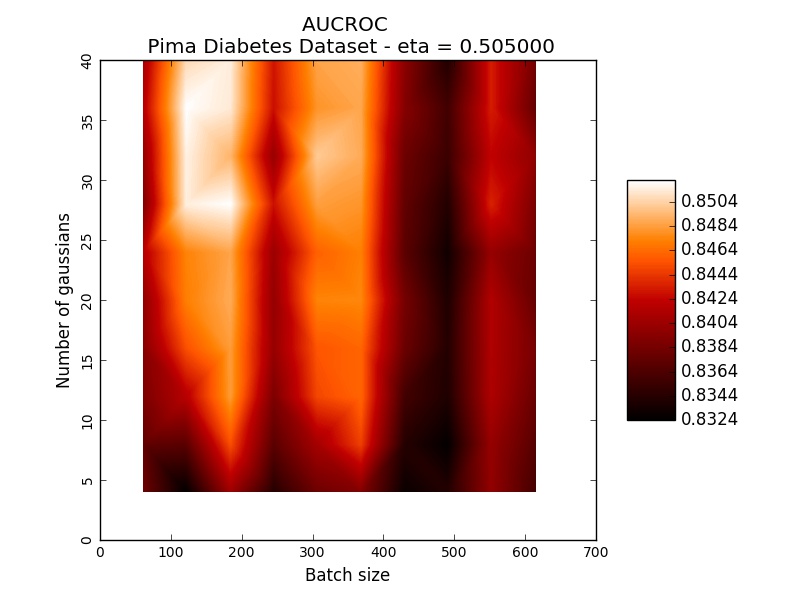}
        \label{fig:rocpima0505mean}
    \end{subfigure}
    \vspace{-0.25cm}
    \caption*{\vspace{-0.1cm} Source: Author. }
    \label{fig:aucrocpima}
\end{figure}

\begin{figure} [!ht]
    \centering
    \caption{\vspace{-0.1cm} Accuracy, AUCROC and MSE mean for the VFNN on the Cryotherapy dataset with $\eta$ = 0.2575. }
    \begin{subfigure}[b]{0.45\textwidth}
        \includegraphics[width=\textwidth]{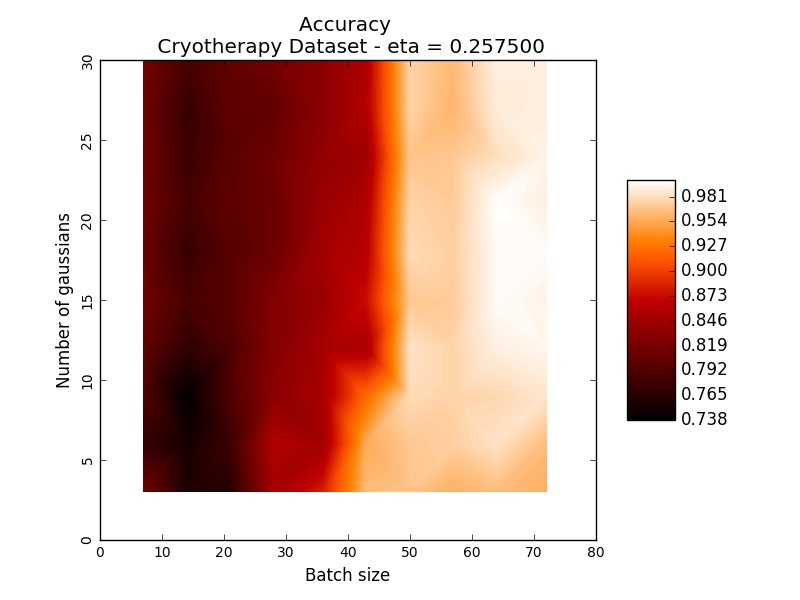}
        \label{fig:acccryo025mean}
    \end{subfigure}
    \begin{subfigure}[b]{0.45\textwidth}
        \includegraphics[width=\textwidth]{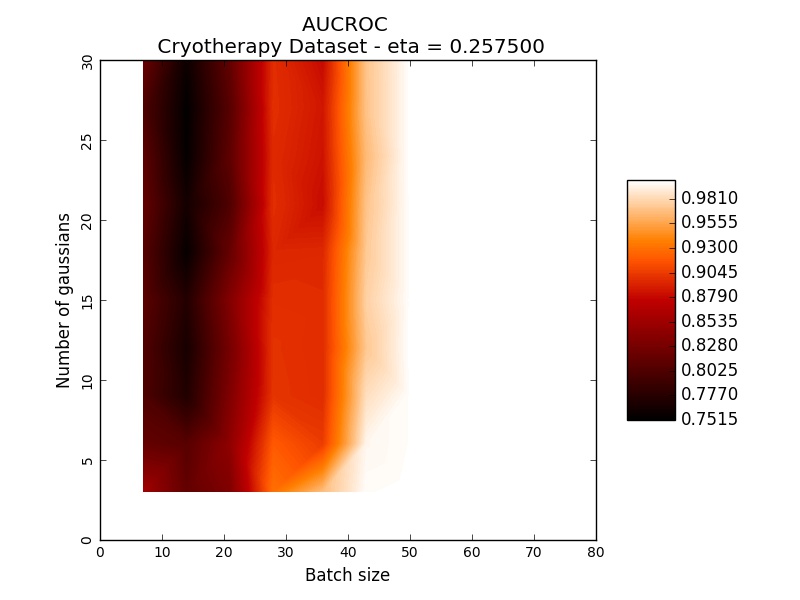}
        \label{fig:roccryo025mean}
    \end{subfigure}
    \begin{subfigure}[b]{0.45\textwidth}
        \includegraphics[width=\textwidth]{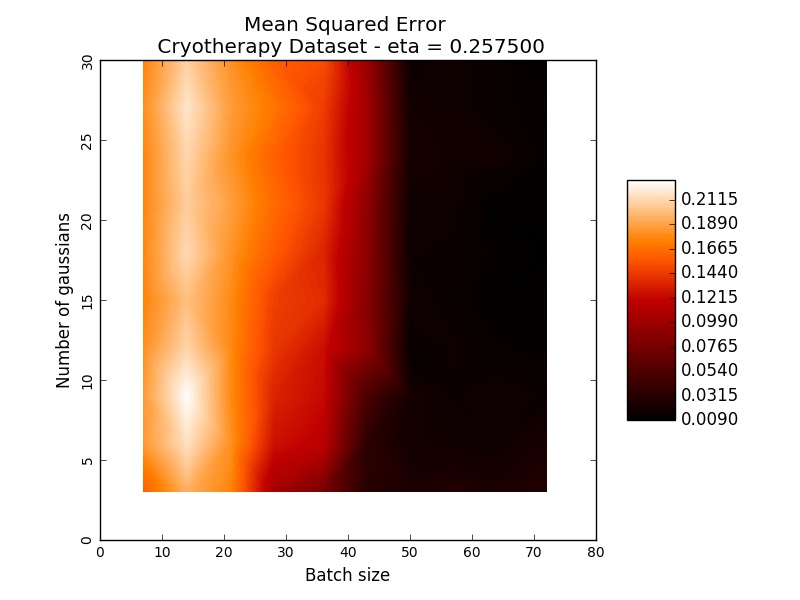}
        \label{fig:msecryo025mean}
    \end{subfigure}
    \vspace{-0.25cm}
    \caption*{\vspace{-0.1cm} Source: Author. }
    \label{fig:cryo025}
\end{figure}

\begin{figure} [!ht]
    \centering
    \caption{\vspace{-0.1cm} Accuracy, AUCROC and MSE mean for the VFNN on the Ionosphere dataset with $\eta$ = 0.7525. }
    \begin{subfigure}[b]{0.45\textwidth}
        \includegraphics[width=\textwidth]{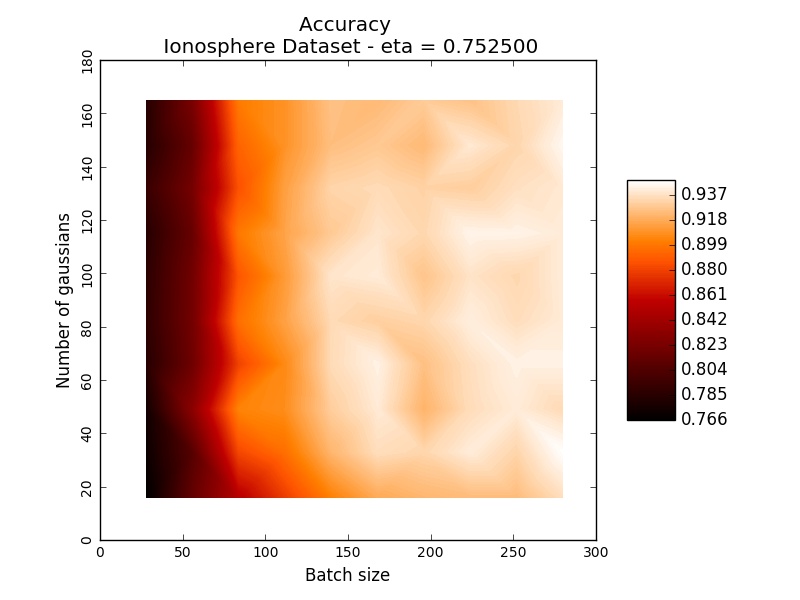}
        \label{fig:acciono025mean}
    \end{subfigure}
    \begin{subfigure}[b]{0.45\textwidth}
        \includegraphics[width=\textwidth]{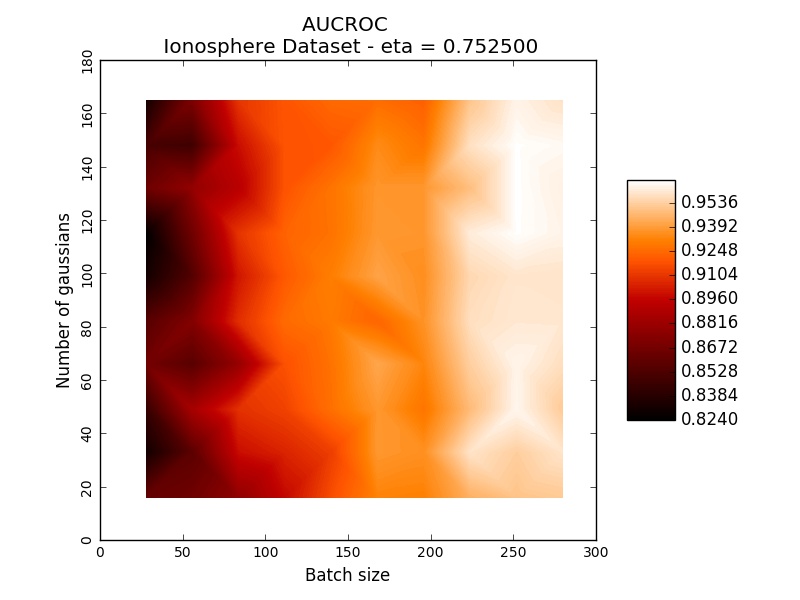}
        \label{fig:rociono025mean}
    \end{subfigure}
    \begin{subfigure}[b]{0.45\textwidth}
        \includegraphics[width=\textwidth]{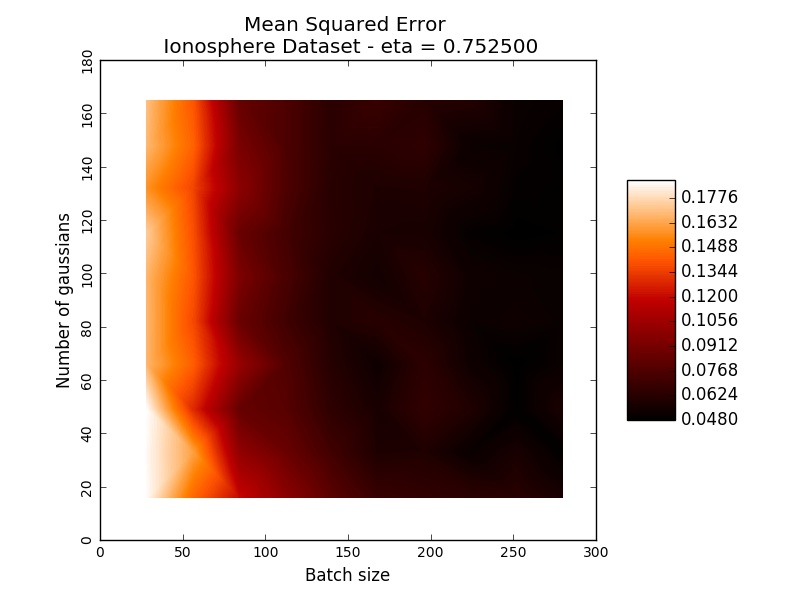}
        \label{fig:mseiono025mean}
    \end{subfigure}
    \vspace{-0.25cm}
    \caption*{\vspace{-0.1cm} Source: Author. }
    \label{fig:iono075}
\end{figure}

On the other hand a portion of plots in three datasets ( Ionosphere, Cryotherapy, and Pima Diabetes) have shown almost one dimensional behavior for a fixed learning rate. In Pima Diabetes case this happened at the AUCROC metric (Figure \ref{fig:aucrocpima}). At the Cryotherapy and Ionosphere datasets this was more evident as it happened in all metrics in at least one case (Figure \ref{fig:cryo025} and \ref{fig:iono075}). 

This type of pattern is troublesome since the change in the number of gaussians would not lead to any gain in terms of performance. Later, on the next section, it will be shown that the number of gaussians have a severe impact on training time, and, as such, to avoid plateaus like the ones exhibited here is important.

\begin{figure} [!ht]
    \centering
    \caption{\vspace{-0.1cm} Accuracy mean and standard deviation measurements for the VFNN on the Banknote dataset with $\eta$ = 0.2575. }
    \begin{subfigure}[b]{0.45\textwidth}
        \includegraphics[width=\textwidth]{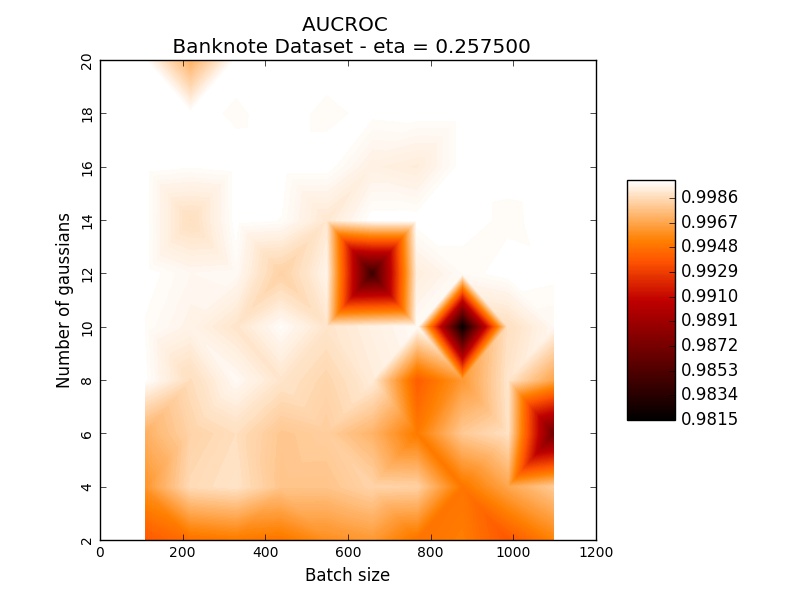}
        \label{fig:rocbank025mean}
    \end{subfigure}
    \begin{subfigure}[b]{0.45\textwidth}
        \includegraphics[width=\textwidth]{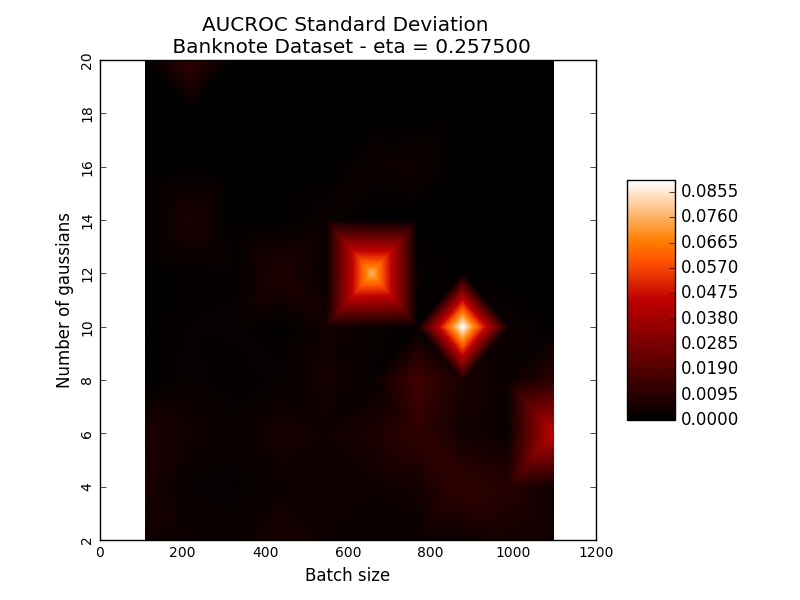}
        \label{fig:rocbank025std}
    \end{subfigure}
    \vspace{-0.25cm}
    \caption*{\vspace{-0.1cm} Source: Author. }
    \label{fig:rocbank025}
\end{figure}

Another pattern that can be noticed in some cases happens between the metric mean and standard deviation. Figure \ref{fig:rocbank025} is an example of this. The standard deviation is inversely proportional to the metric, that is, as the accuracy, or AUCROC, gets higher as the standard deviation decreases. This an interesting property since it would mean that hyperparameters which shown good performance also do so in an consistent fashion, that is, the higher the performance, more confindence

The change in learning rate also also exerts some influence, as the plot patterns and values change with different learning rates. It is also useful to take into consideration that the behavior and performance of $\eta = 0.01$ usually has a larger difference from the others. This may happens due to the fact that it is a smaller magnitude, and requires a large number of epochs to achieve similar results. Nevertheless, the best performance at the Immunotherapy was reached when we used $\eta = 0.01$, which shows that this hyperparameter also strongly depends on the dataset.

\begin{figure} [!ht]
    \centering
    \caption{\vspace{-0.1cm} Accuracy, AUCROC and MSE mean for the VFNN on the Banknote dataset with  = 1.0. }
    \begin{subfigure}[b]{0.45\textwidth}
        \includegraphics[width=\textwidth]{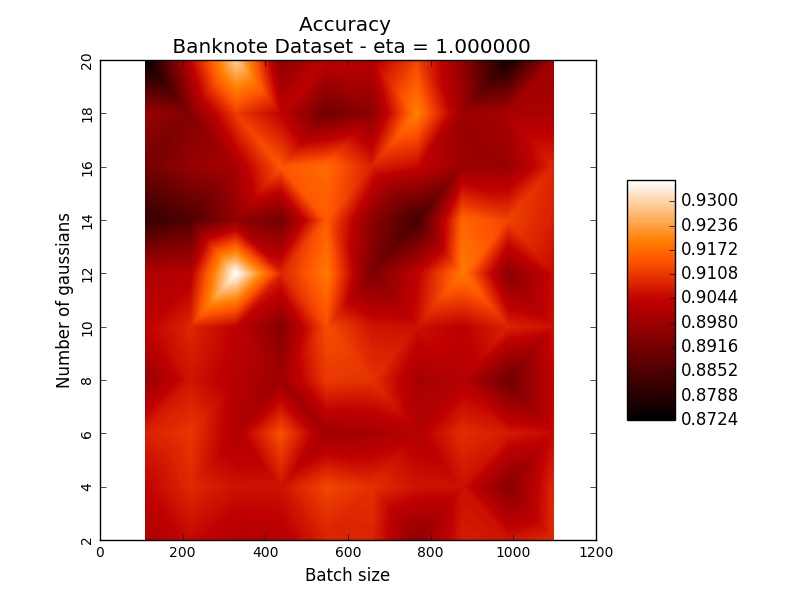}
        \label{fig:accbank100mean}
    \end{subfigure}
    \begin{subfigure}[b]{0.45\textwidth}
        \includegraphics[width=\textwidth]{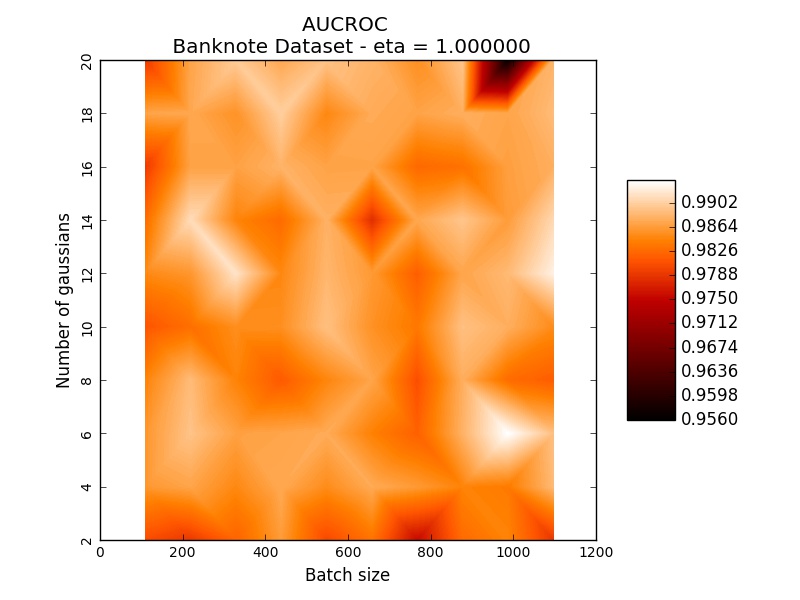}
        \label{fig:rocbank100mean}
    \end{subfigure}
    \begin{subfigure}[b]{0.45\textwidth}
        \includegraphics[width=\textwidth]{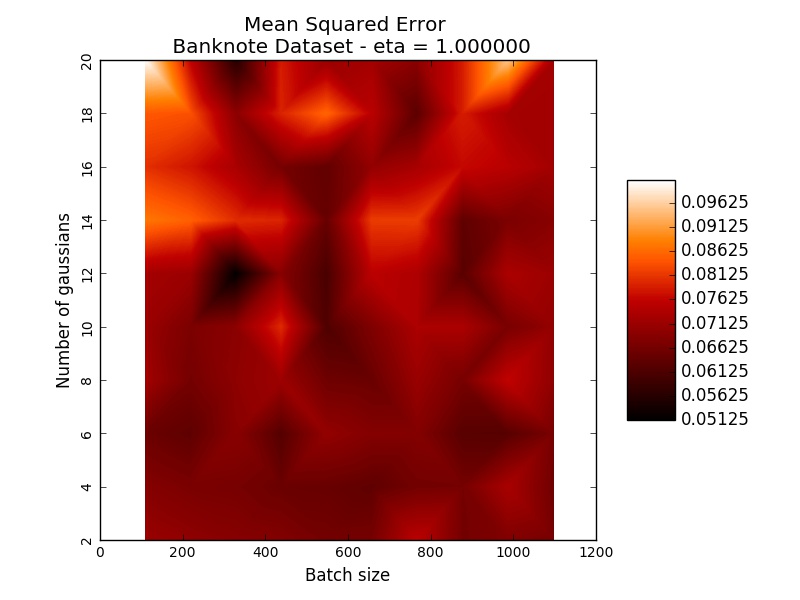}
        \label{fig:msebank100mean}
    \end{subfigure}
    \vspace{-0.25cm}
    \caption*{\vspace{-0.1cm} Source: Author. }
    \label{fig:bank100}
\end{figure}

\begin{figure} [!ht]
    \centering
    \caption{\vspace{-0.1cm} Accuracy, AUCROC and MSE mean for the VFNN on the Immunotherapy dataset with $\eta$ = 0.2575. }
    \begin{subfigure}[b]{0.45\textwidth}
        \includegraphics[width=\textwidth]{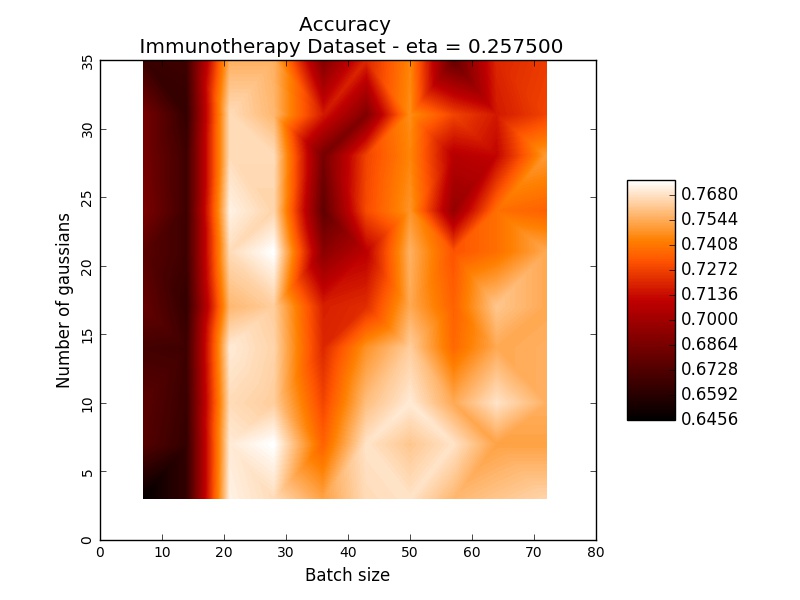}
        \label{fig:accimmu025mean}
    \end{subfigure}
    \begin{subfigure}[b]{0.45\textwidth}
        \includegraphics[width=\textwidth]{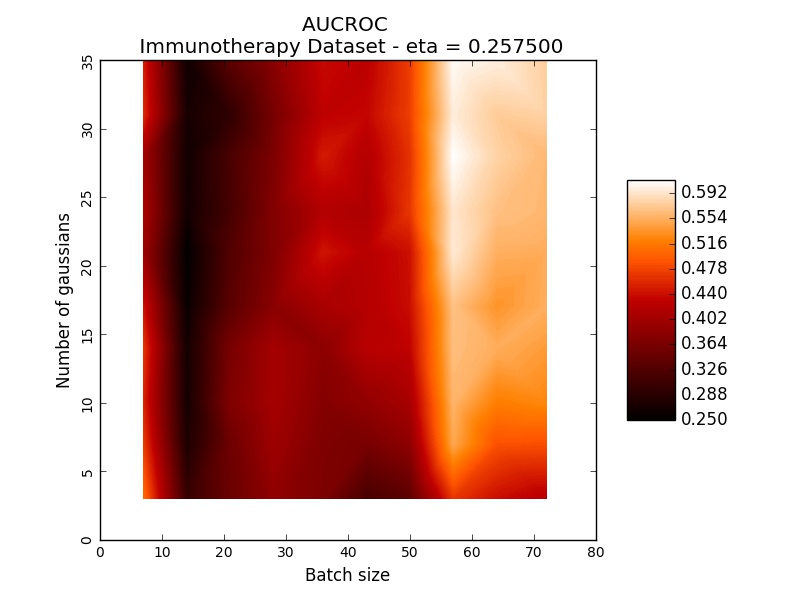}
        \label{fig:rocimmu025mean}
    \end{subfigure}
    \begin{subfigure}[b]{0.45\textwidth}
        \includegraphics[width=\textwidth]{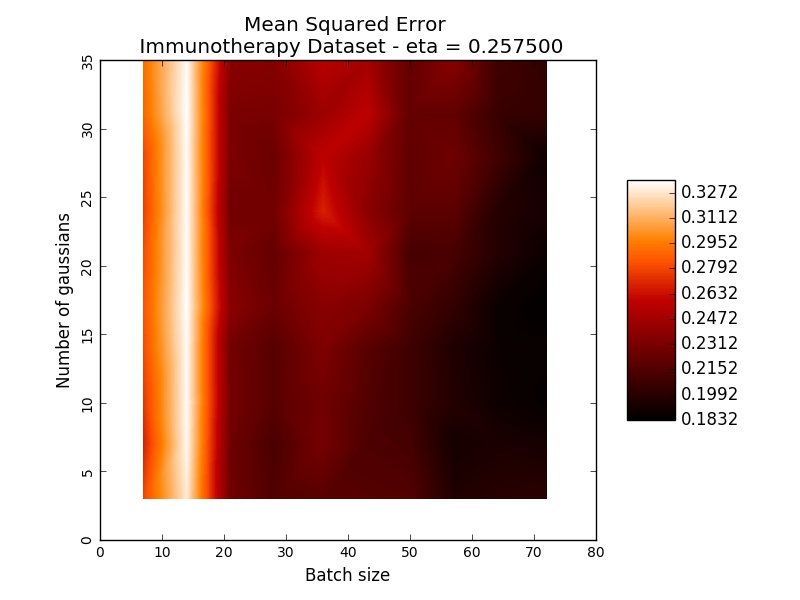}
        \label{fig:mseimmu025mean}
    \end{subfigure}
    \vspace{-0.25cm}
    \caption*{\vspace{-0.1cm} Source: Author. }
    \label{fig:immuno025}
\end{figure}

In some cases there is no clear pattern and there is a need to further investigate (Figure \ref{fig:bank100}). This happens in at least one scenario for all datasets. However, in the Immunotherapy dataset this behavior is found in all instances (Figures \ref{fig:immuno025}). Maybe this is due to the dataset having few examples and a pattern not easily learnable (sustained by poor performance by the models).

The analyses of hyperparameters influence on performance shows that the optimization of hyperparameters is highly non-convex, having many local minima and maxima. This exemplifies why it is so hard to correctly optimize and find the best hyperparameters when using a Machine Learning model.

In this section what does standout is the importance of the data when we are working with Machine Learning models, and more specifically, this model. We saw how the change in data can produce completely different patterns in the optimization of hyperparameters. Furthermore, not a single meaningful statement can be made concerning the entirety of experiments without taking the data into consideration. Data, more than hyperparameters optimization, initialization heuristics, and newer models, is the key behind Machine Learning.

\section{Time and Complexity}

Now that we have been able to analyze the hyperparameters, it is interesting to understand how complexity and computational time vary depending on the problem and on our choices of model.

\begin{figure}[!ht]
    \centering
    \caption{\vspace{-0.1cm} Train time and standard deviation for the VFNN. }
    \begin{subfigure}[b]{0.45\textwidth}
        \includegraphics[width=\textwidth]{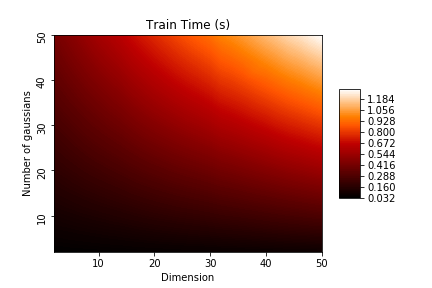}
        \label{fig:train_t}
    \end{subfigure}
    \begin{subfigure}[b]{0.45\textwidth}
        \includegraphics[width=\textwidth]{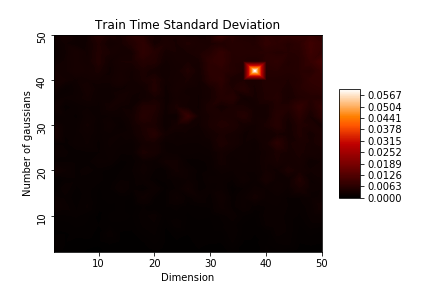}
        \label{fig:train_tstd}
    \end{subfigure}
    \vspace{-0.25cm}
    \caption*{\vspace{-0.1cm} Source: Author. }
    \label{fig:train_ts}
\end{figure}

\begin{figure}[!ht]
    \centering
    \caption{\vspace{-0.1cm} Prediction time and standard deviation for the VFNN. }
    \begin{subfigure}[b]{0.45\textwidth}
        \includegraphics[width=\textwidth]{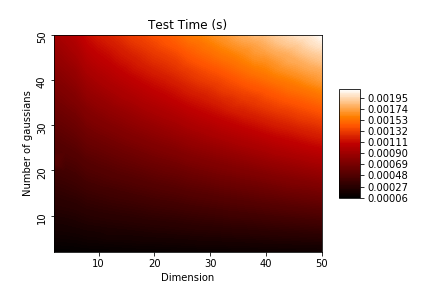}
        \label{fig:test_t}
    \end{subfigure}
    \begin{subfigure}[b]{0.45\textwidth}
        \includegraphics[width=\textwidth]{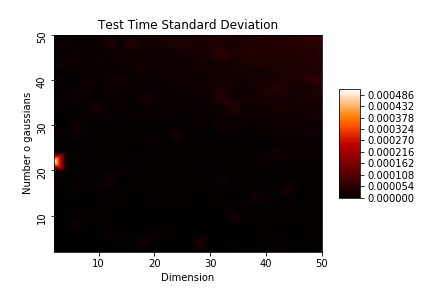}
        \label{fig:test_tstd}
    \end{subfigure}
    \vspace{-0.25cm}
    \caption*{\vspace{-0.1cm} Source: Author. }
    \label{fig:test_ts}
\end{figure}

Delving into  Figure \ref{fig:train_ts} and Figure \ref{fig:test_ts} for train and test time evaluation it is possible to see that the train time is three orders of magnitude larger than the test time. Furthermore, the prediction time is small even in high dimensions (maximum of 0.00195\emph{s} to predict the entire test set), which enables good response time after training. The standard deviation in both cases, as expected, is almost zero, and probably the variation exhibited are due to the processor's job scheduler.

\begin{figure}[!ht]
    \centering
    \caption{\vspace{-0.1cm} Complexity measurements for the VFNN. }
    \begin{subfigure}[b]{0.6\textwidth}
        \includegraphics[width=\textwidth]{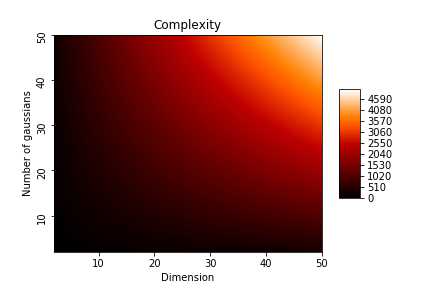}
        \label{fig:complex}
    \end{subfigure}
    \vspace{-0.25cm}
    \caption*{\vspace{-0.1cm} Source: Author. }
    \label{fig:complexity}
\end{figure}

Complexity (Figure \ref{fig:complexity}) has a similar shape and coloring to time plots. This was expected, since, when complexity increases the model gets larger, and more calculations are needed to evaluate the function. However, it is interesting to note that, while complexity increases linearly in both dimensions (from Equation \ref{freeparameter}) the time spent on computing grows larger with the number of gaussians than it does with data input dimension (darker regions concentrate on the bottom part). As such, it is important to take care when picking the number of gaussians, as it answers for a large portion of the computational effort.

\section{Performance Comparison}

Finally, we are now able to compare the proposed architecture against known models. It is important to remember that our model uses the optimum found in the hyperparameter analysis. Meanwhile, the hyperparameters for SVM and FFNN have been chosen to have a similar number of parameters as the VFNN for complexity purposes.

\begin{figure}[!ht]
    \centering
    \caption{Performance results for the Banknote dataset.}
    \begin{subfigure}[b]{0.48\linewidth}
         \includegraphics[width=\linewidth]{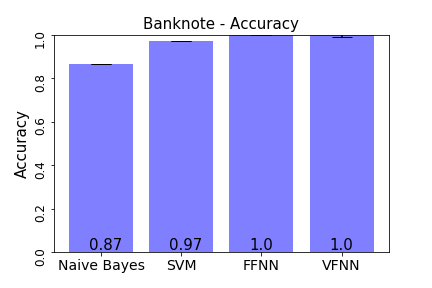}
         \label{fig:p_acc_banknote}
    \end{subfigure}
    \begin{subfigure}[b]{0.48\linewidth}
         \includegraphics[width=\linewidth]{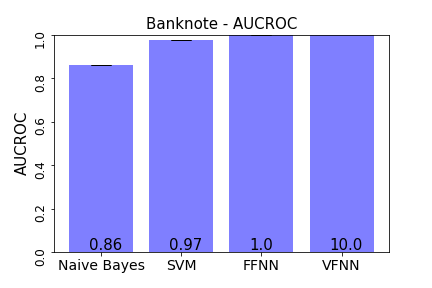}
         \label{fig:p_roc_banknote}
    \end{subfigure}
     ~ 
    \begin{subfigure}[b]{0.48\linewidth}
         \includegraphics[width=\linewidth]{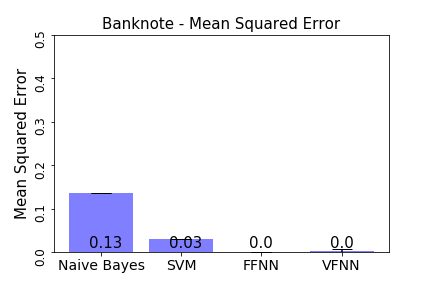}
         \label{fig:p_mse_banknote}
     \end{subfigure}
     \vspace{-0.25cm}
     \caption*{Source: Author.}
     \label{p_p_banknote}
 \end{figure}

The results for the Banknote dataset are presented in Figure \ref{p_p_banknote}. The VFNN model shows similar accuracy to the FFNN model, both being greater than SVM and Naive Bayes. The same behavior can be seen for both AUCROC and MSE. It is important to notice the larger standard deviation at the VFNN model, specially at the MSE metric. This indicates that, even though the mean is close, the performance of FFNN is more consistent, and better than the VFNN.

The best result on all three metrics by MLP and VFNN may happen due to the dataset conditions. Remember, the Banknote has only four dimensions and over 1372 examples available. Neural Networks are widely known to perform well in situations where there is a large amount of data available, in comparison to data dimension.

\begin{figure}[!ht]
    \centering
    \caption{Performance results for the Ionosphere dataset.}
    \begin{subfigure}[b]{0.48\linewidth}
         \includegraphics[width=\linewidth]{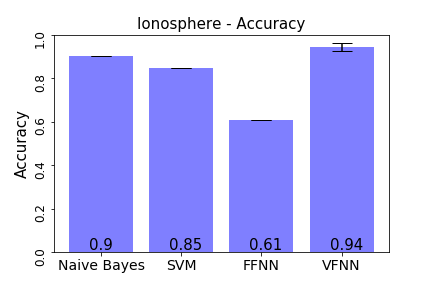}
         \label{fig:p_acc_iono}
    \end{subfigure}
    \begin{subfigure}[b]{0.48\linewidth}
         \includegraphics[width=\linewidth]{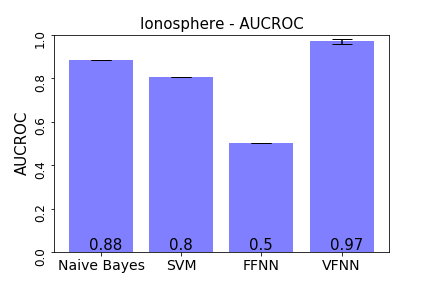}
         \label{fig:p_roc_iono}
    \end{subfigure}
     ~ 
    \begin{subfigure}[b]{0.48\linewidth}
         \includegraphics[width=\linewidth]{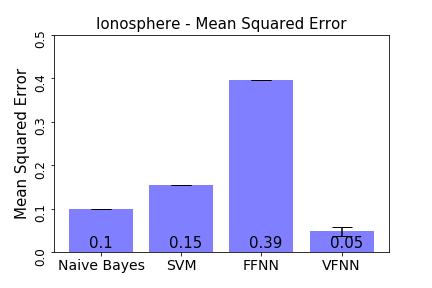}
         \label{fig:p_mse_iono}
     \end{subfigure}
     \vspace{-0.25cm}
     \caption*{Source: Author.}
     \label{p_p_iono}
 \end{figure}

The Ionosphere dataset (Figure \ref{p_p_iono}) has the most favorable result for the new model. Here, VFNN had the best performance on all metrics, and a standard deviation larger than other models, but low. It is followed by the Naive Bayes, SVM and finally FFNN in all metrics. The best performance in all metrics is probably due to dataset properties, although we can not pin down the specific reason.

\begin{figure}[!ht]
    \centering
    \caption{Performance results for the Cryotherapy dataset.}
    \begin{subfigure}[b]{0.48\linewidth}
         \includegraphics[width=\linewidth]{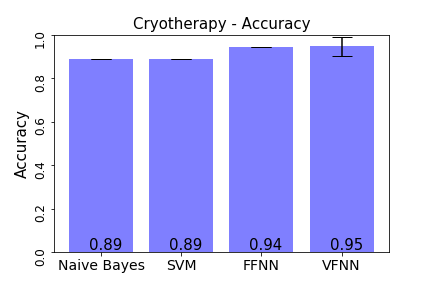}
         \label{fig:p_acc_cryo}
    \end{subfigure}
    \begin{subfigure}[b]{0.48\linewidth}
         \includegraphics[width=\linewidth]{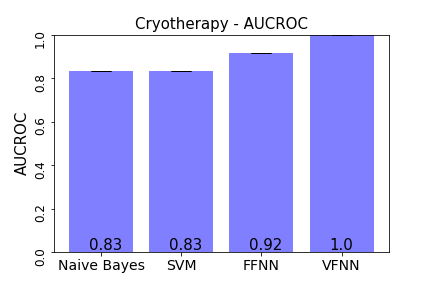}
         \label{fig:p_roc_cryo}
    \end{subfigure}
     ~ 
    \begin{subfigure}[b]{0.48\linewidth}
         \includegraphics[width=\linewidth]{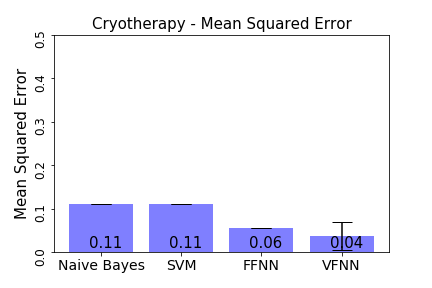}
         \label{fig:p_mse_cryo}
     \end{subfigure}
     \vspace{-0.25cm}
     \caption*{Source: Author.}
     \label{p_p_cryo}
 \end{figure}

Moving forward to Figure \ref{p_p_cryo} we have Cryotherapy results. The results are similar for accuracy, with FFNN and VFNN being higher than SVM and Naive Bayes. However, Naive Bayes closed the performance gap on SVM and the other. Also, the standard deviation for VFNN in accuracy is greater than on Banknote.

The AUCROC metric shows a different behavior, with the VFNN model performing better than FFNN and displaying a comparable standard deviation to the other models. FFNN maintain a better performance than SVM and Naive Bayes.

Finally, the MSE also goes in the VFNN favor, with the FFNN behind it, and Naive Bayes and SVM tied again with the worst performance. However, it is important to notice that the VFNN model has a better mean but a large standard deviation. This means that it may be consistently outperformed by the FFNN model depending on certain conditions. This large standard deviation on VFNN model is partially due to the randomness in initializing vectors $V_i$ and gaussian centers $\mu_i$ and is a problem that needs to be further addressed.

\begin{figure}[!ht]
    \centering
    \caption{Performance results for the Pima Diabetes dataset.}
    \begin{subfigure}[b]{0.48\linewidth}
         \includegraphics[width=\linewidth]{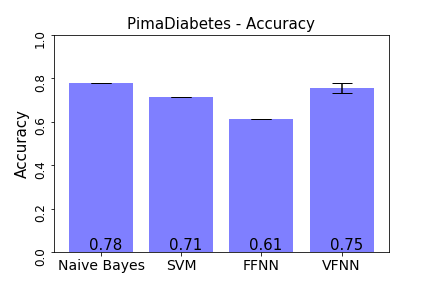}
         \label{fig:p_acc_pima}
    \end{subfigure}
    \begin{subfigure}[b]{0.48\linewidth}
         \includegraphics[width=\linewidth]{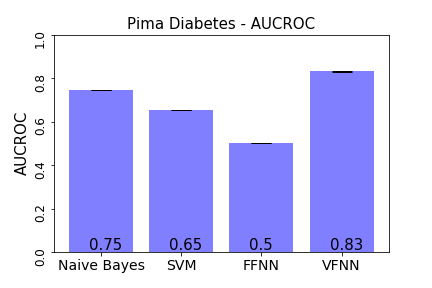}
         \label{fig:p_roc_pima}
    \end{subfigure}
     ~ 
    \begin{subfigure}[b]{0.48\linewidth}
         \includegraphics[width=\linewidth]{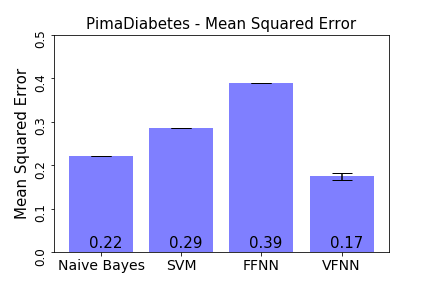}
         \label{fig:p_mse_pima}
     \end{subfigure}
     \vspace{-0.25cm}
     \caption*{Source: Author.}
     \label{p_p_pima}
 \end{figure}

The next dataset is the Pima Diabetes, with results shown in Figure \ref{p_p_pima}. Surprisingly, the simplest model, Naive Bayes, has the better overall accuracy, followed closely by the VFNN and SVM models, with FFNN now being at last. The standard deviation for accuracy for the VFNN  is still higher than other models, probably for the same reasons that have already being stated.

Nevertheless, the Pima Diabetes is an unbalanced dataset, with near 65\% being in one class. As such, the best metric to evaluate performance would be AUCROC, which is not affected by class balance. In this case, the VFNN has the better performance, followed by Naive Bayes, SVM, and FFNN. The standard deviation for VFNN is low compared to the accuracy counterpart, but still higher than for other classes. At the MSE metric the result that follows is the same as the one for AUCROC.

\begin{figure}[!ht]
    \centering
    \caption{Performance results for the Immunotherapy dataset.}
    \begin{subfigure}[b]{0.48\linewidth}
         \includegraphics[width=\linewidth]{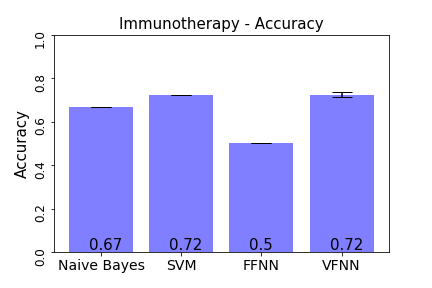}
         \label{fig:p_acc_immuno}
    \end{subfigure}
    \begin{subfigure}[b]{0.48\linewidth}
         \includegraphics[width=\linewidth]{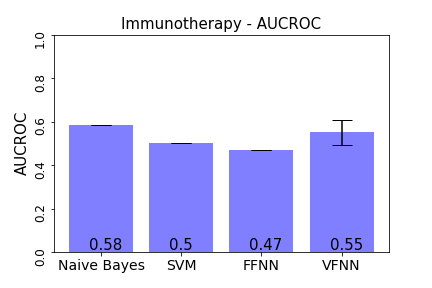}
         \label{fig:p_roc_immuno}
    \end{subfigure}
     ~ 
    \begin{subfigure}[b]{0.48\linewidth}
         \includegraphics[width=\linewidth]{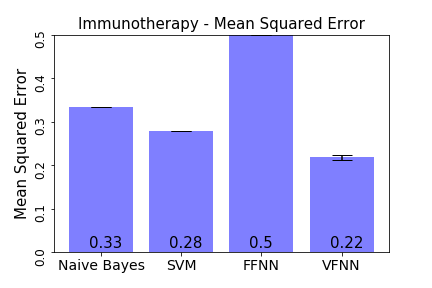}
         \label{fig:p_mse_immuno}
     \end{subfigure}
     \vspace{-0.25cm}
     \caption*{Source: Author.}
     \label{p_p_immuno}
 \end{figure}

At last, in the Immunotherapy dataset results (\ref{p_p_immuno}) the accuracy goes in favor of the SVM and VFNN models, followed by Naive Bayes and FFNN. The standard deviation for the VFNN continues to be the highest amonst all datasets.

Like the Pima Diabetes dataset, Immunotherapy is highly unbalanced, with one class having almost 80\% of examples. Besides that, Immunotherapy has few examples, with only 90 available examples (the same as Cryotherapy). With this two characteristics in mind, it is understandable that its performance results were the poorest among all datasets. Also, it means that the accuracy is not a proper metric, and the AUCROC and MSE should have a higher importance.

Analyzing the AUCROC tells us that the Naive Bayes has the best result, followed by VFNN, SVM and FFNN. The VFNN standard deviation is visually high, something that has not happened for other AUCROC results, and may have its source on the dataset aforementioned problems.

The MSE results mostly agrees with those of accuracy, with the difference of VFNN having a better performance than SVM.

Broadly speaking, the performance results show that the proposed VFNN model has learning capabilities near or better than those of established models such as Naive Bayes, SVM and FFNN, having the best or second best result at all datasets.

It is important, however, to note that it does not mean that the other models could not outperform the VFNN model. As was stated at the beginning of the section, the hyperparameters have been slightly optimized for the VFNN model, with others which rely heavily on hyperparameters (such as FFNN and SVM) having their hyperparameters set to similar values (epochs and number of neurons) or chosen by the library internal algorithm (learning algorithm, learning step, etc.). Moreover, the standard deviation for the VFNN is large for a few cases, even when compared to similar models like FFNN.

Also, even though widely known, these models are not the state-of-the-art models that one would find running in modern Machine Learning systems. As such, being able to perform better than basic models in most cases, specially given the circumstances, is not a completely unexpected result. Our objective is to see if the VFNN is a viable alternative that has certain benefits and it is worth exploring more not as a solution to all Machine Learning problems. And, with the results explored here, this objective was achieved.

\chapter{Conclusion}
\label{capituloConclusao}

After going through the computational experiments, analyses, mathematical formulation and so on we can draw a few conclusions. First, remembering our primary objectives, we were able to state several geometrical interpretations for Neural Networks and as a first contribution we established a mathematical relation between them. Not only that, but a new interpretation was also explored. Its limitations have been discussed briefly, as well as the benefit provided by each of the interpretations, be it mathematical, learning usefulness, visualization,  or mindset to approach the problem.

Beyond that, it was also possible to create an entirely new architecture of Neural Networks, which took the proposed interpretation as a starting point. 

This architecture initially proved to be able to learn in a feasible manner. More so, the regularization of model parameters has a direct geometrical meaning and interpretation, further explored to deal with some of the issues in learning problems.

After that, results were analyzed to provide further evidence that the model is capable of learning when compared against test samples in a variety of widely available datasets, hyperparameter choice, and randomly generated parameters. Furthermore, the results have shown patterns in a few cases, which can help the researcher to decide on hyperparameters choices. 

As it is common for Neural Networks, no hyperparameter proved to be the best answer for all cases, as it was initially expected. The hyperparameters metrics pattern changed between datasets. This once more shows that the most important part in a Machine Learning problem is the data. Data with different patterns, dimensions, behavior, etc. are capable of showing performance completely different at same initial conditions. The best course of action in a production setting would be to use an optimization technique.

Complexity and time were evaluated for varying dimensions and number of gaussians. The standard deviation was almost null, as expected. The results also shown that the complexity is linearly dependent on both, which was also expected from the formula from which it derived. Nonetheless, the number of gaussians showed to be more important to computation time than dataset dimension. This is a important consideration for future users of the architecture, as it is important to control the necessary computational time. 

At last, the model was compared against basic known classifiers such as SVM, NB and FFNN. The model reached good results in most datasets and metrics, being the best or second best model. It is important to remember that this happened with only a simple case of the methodology described in Chapter 4. Nevertheless, we should also note that the standard deviation was usually greater than all other models, and it used hyperparameters slightly optimized.

\section{Future Works}

This work was also important in drawing future directions. Changing the classifier to something like the softmax function \cite{michaelnielsen} would enable the architecture to deal with problems with more than two classes, providing viability to test the model against harder and larger problems.

The use of multiple steps with a small interval in Euler's method is one of the most important future extensions, as this would completely implement the idea of a flow moving particles in the original space. Furthermore, it is possible to try different numerical methods, as it was already mentioned, such as Crank Nicholson\cite{Butcher1987}. Those would need to be formulated into different architectures but may provide better and more stable results.

Moving from the ODE to the vector field, something important should be noticed: the vector field parameters could be changed during the iterative process. This the equivalent of a transient (time dependent) vector field. Its architecture may resemble that of a traditional Feed Forward Neural Network, where a new function (in this case, a new vector field), with new parameters act on the dataset in each subsequent hidden layer. The positive side of choosing a transient solution is the greater vector field flexibility which could lead to solving problems in fewer steps and with a better solution than the one from a stationary field. It would be possible to ensure that we are dealing with one vector field changing and not several disassociated fields if, for instance, we add hard constraints to how much a field can change since the last iteration.

\begin{figure}[!ht]
     \centering     
     \caption{Example of datasets,  with one class in blue and another in red, where rotation is not recommended (first case) and where it is necessary(second case).}
     \begin{subfigure}[b]{0.28\linewidth}
         \includegraphics[width=\linewidth]{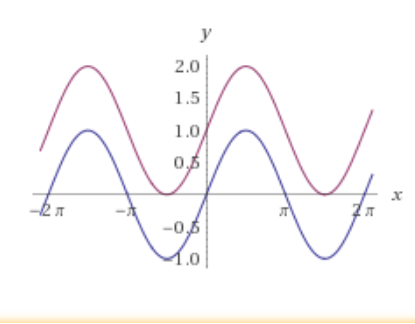}
         \label{fig:nearsin}
     \end{subfigure}
     ~ 
     \begin{subfigure}[b]{0.4\linewidth}
         \includegraphics[width=\linewidth]{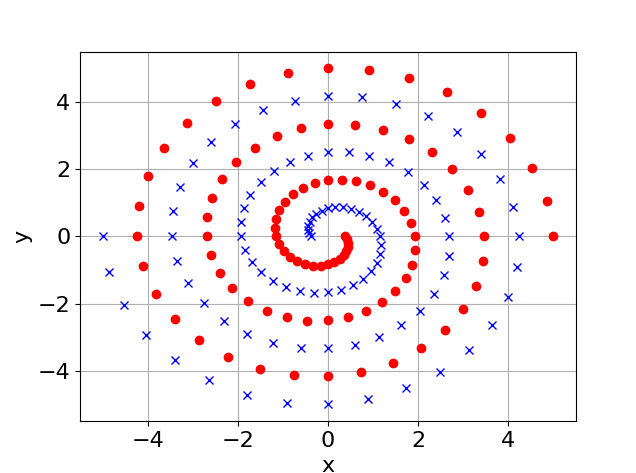}
         \label{fig:twospiral}
     \end{subfigure}
     \vspace{-0.25cm}
     \caption*{Source: \cite{wolfram_sin} and \cite{peter_spirals}.}
     \label{intro2}
 \end{figure}

There are several properties such as curl \cite{marsden2011vector} (which can be thought as rotation of the flow) that can be interesting to have or not, when one knows a bit about the structure of the problem at hand ,such as, Figure \ref{intro2}. 

\begin{figure}[!ht]
    \centering
    \caption{\vspace{-0.1cm} A region $W$ flowing along the streamlines of a vector field.}
    \begin{subfigure}[b]{0.45\textwidth}
        \includegraphics[width=\textwidth]{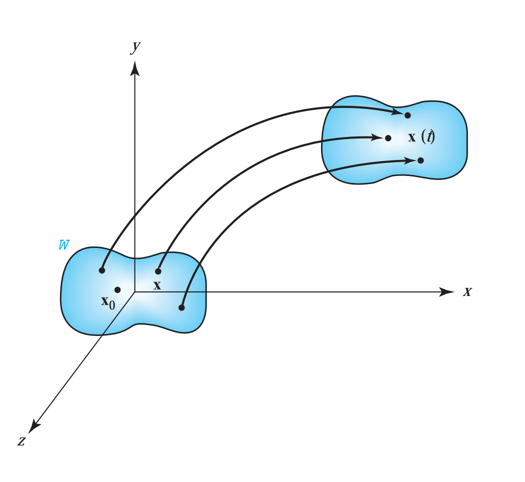}
        \label{fig:divergence}
    \end{subfigure}
    \vspace{-0.25cm}
    \caption*{\vspace{-0.1cm} Source: \cite{marsden2011vector}.}\label{fig:divergences}
\end{figure}

The user might want to add more mathematical elements to the field definition, or even change it completely to provide the desired characteristics. For instance, imagine that we need our data to expand in some sense. It is known that the divergent of the vector field, in physics, is directly connected to the expansion of a volume of gas (Figure \ref{fig:divergences}), after displacement \cite{marsden2011vector}. Also, other elements frequently seen and used by fluid dynamicists such as sinks and sources have all mathematical definitions, and could contribute positively to the desired movement. 

There are also many optimization methods and techniques used in Neural Networks training such as Momentum, Dropout, Adam, Adaboost, etc.  \cite{walia_opt} that provide better results than a simple gradient descent approach. Those could be implemented instead of the gradient descent, further increasing the performance of the model.
 
Finally, the dissertation was able to successfully achieve the objectives initially proposed, establishing a different geometrical and physical way to look at an important problem in Machine Learning, raised many new questions and provided a clear path forward.

\postextual
\bibliography{Postext/Referencias}

\providecommand{\abntreprintinfo}[1]{%
 \citeonline{#1}}
\setlength{\labelsep}{0pt}\begin{thebibliography}{}
\providecommand{\abntrefinfo}[3]{}
\providecommand{\abntbstabout}[1]{}
\abntbstabout{v<VERSION> }

\bibitem[Abu-Mostafa, Magdon-Ismail and Lin 2012]{abu2012learning}
\abntrefinfo{Abu-Mostafa, Magdon-Ismail and Lin}{ABU-MOSTAFA; MAGDON-ISMAIL;
  LIN}{2012}
{ABU-MOSTAFA, Y.~S.; MAGDON-ISMAIL, M.; LIN, H.-T. \textbf{Learning from data}.
  [S.l.]: AMLBook New York, NY, USA:, 2012.}

\bibitem[Alpha 2018]{wolfram_sin}
\abntrefinfo{Alpha}{ALPHA}{2018}
{ALPHA, W. \textbf{WolframAlpha}. [S.l.]: Wolfram Alpha LLC, 2018.
Https://www.wolframalpha.com/input/?i=sin(x)+and+sin(x)\%2B1.}

\bibitem[Barakat and Bradley 2006]{barakat2006rule}
\abntrefinfo{Barakat and Bradley}{BARAKAT; BRADLEY}{2006}
{BARAKAT, N.; BRADLEY, A.~P. Rule extraction from support vector machines:
  Measuring the explanation capability using the area under the roc curve. In:
  IEEE. \textbf{Pattern Recognition, 2006. ICPR 2006. 18th International
  Conference on}. [S.l.], 2006. v.~2, p. 812--815.}

\bibitem[Benenson 2016]{bennenson_records}
\abntrefinfo{Benenson}{BENENSON}{2016}
{BENENSON, R. \textbf{Classification datasets results}. 2016.
Http://rodrigob.github.io/are\_we\_there\_yet/build/classification\_datasets\_results.html.}

\bibitem[Bhande 2018]{anup}
\abntrefinfo{Bhande}{BHANDE}{2018}
{BHANDE, A. \textbf{What is underfitting and overfitting in machine learning
  and how to deal with it}. 2018.
Https://medium.com/greyatom/what-is-underfitting-and-overfitting-in-machine-learning-and-how-to-deal-with-it-6803a989c76.}

\bibitem[Brynjolfsson 2018]{brynjolfsson}
\abntrefinfo{Brynjolfsson}{BRYNJOLFSSON}{2018}
{BRYNJOLFSSON, E. \textbf{Machine learning will be the engine of global
  growth}. 2018.
Https://www.ft.com/content/133dc9c8-90ac-11e8-9609-3d3b945e78cf.}

\bibitem[Butcher 1987]{Butcher1987}
\abntrefinfo{Butcher}{BUTCHER}{1987}
{BUTCHER, J.~C. \textbf{The Numerical Analysis of Ordinary Differential
  Equations: Runge-Kutta and General Linear Methods}. New York, NY, USA:
  Wiley-Interscience, 1987.
ISBN 0-471-91046-5.}

\bibitem[Butler Keith T.;~Davies 2018]{Butler2018}
\abntrefinfo{Butler Keith T.;~Davies}{BUTLER KEITH T.;~DAVIES}{2018}
{BUTLER KEITH T.;~DAVIES, D. W. C. H. I. O. W.~A. Machine learning for
  molecular and materials science.
\textbf{Nature}, v.~559, p. 547--555, 2018.
Available at: \url{https://www.nature.com/articles/s41586-018-0337-2}.}

\bibitem[Chang et al. 2017]{chang2017multi}
\abntrefinfo{Chang et al.}{CHANG et al.}{2017a}
{CHANG, B. et al. Multi-level residual networks from dynamical systems view.
\textbf{arXiv preprint arXiv:1710.10348}, 2017.}

\bibitem[Chang et al. 2017]{chang2017reversible}
\abntrefinfo{Chang et al.}{CHANG et al.}{2017b}
{CHANG, B. et al. Reversible architectures for arbitrarily deep residual neural
  networks.
\textbf{arXiv preprint arXiv:1709.03698}, 2017.}

\bibitem[Columbus 2018]{columbus}
\abntrefinfo{Columbus}{COLUMBUS}{2018}
{COLUMBUS, L. \textbf{Roundup Of Machine Learning Forecasts And Market
  Estimates, 2018}. 2018.
Https://www.forbes.com/sites/louiscolumbus/2018/02/18/roundup-of-machine-learning-forecasts-and-market-estimates-2018.}

\bibitem[Cybenko 1989]{Cybenko1989}
\abntrefinfo{Cybenko}{CYBENKO}{1989}
{CYBENKO, G. Approximation by superpositions of a sigmoidal function.
\textbf{Mathematics of Control, Signals and Systems}, v.~2, n.~4, p. 303--314,
  Dec 1989.
ISSN 1435-568X.
Available at: \url{https://doi.org/10.1007/BF02551274}.}

\bibitem[DeGregory et al. 2018]{degregory}
\abntrefinfo{DeGregory et al.}{DEGREGORY et al.}{2018}
{DEGREGORY, K.~W. et al. A review of machine learning in obesity.
\textbf{Obesity Reviews}, v.~19, n.~5, p. 668--685, 2018.
Available at: \url{https://onlinelibrary.wiley.com/doi/abs/10.1111/obr.12667}.}

\bibitem[Dheeru and Taniskidou 2017]{Dua:2017}
\abntrefinfo{Dheeru and Taniskidou}{DHEERU; TANISKIDOU}{2017}
{DHEERU, D.; TANISKIDOU, E. K. \textbf{{UCI} Machine Learning Repository}.
  2017.
Available at: \url{http://archive.ics.uci.edu/ml}.}

\bibitem[E 2017]{E2017}
\abntrefinfo{E}{E}{2017}
{E, W. A proposal on machine learning via dynamical systems.
\textbf{Communications in Mathematics and Statistics}, v.~5, n.~1, p. 1--11,
  Mar 2017.
ISSN 2194-671X.
Available at: \url{https://doi.org/10.1007/s40304-017-0103-z}.}

\bibitem[Fox, McDonald and Pritchard 1985]{fox1985introduction}
\abntrefinfo{Fox, McDonald and Pritchard}{FOX; MCDONALD; PRITCHARD}{1985}
{FOX, R.~W.; MCDONALD, A.~T.; PRITCHARD, P.~J. \textbf{Introduction to fluid
  mechanics}. [S.l.]: John Wiley \& Sons, New York, 1985.}

\bibitem[Freitas, Wieser and Apweiler 2010]{freitas2010importance}
\abntrefinfo{Freitas, Wieser and Apweiler}{FREITAS; WIESER; APWEILER}{2010}
{FREITAS, A.~A.; WIESER, D.~C.; APWEILER, R. On the importance of
  comprehensible classification models for protein function prediction.
\textbf{IEEE/ACM Transactions on Computational Biology and Bioinformatics
  (TCBB)}, IEEE Computer Society Press, v.~7, n.~1, p. 172--182, 2010.}

\bibitem[Gergel 2017]{peter_spirals}
\abntrefinfo{Gergel}{GERGEL}{2017}
{GERGEL, P. Improving the performance of impulse neuro–glial network. In:
  VYDAVATEľSTVO UNIVERZITY KOMENSKéHO. \textbf{In Kognícia a umelý život}.
  [S.l.], 2017. v.~2, p. 60--63.}

\bibitem[Goodfellow, Bengio and Courville 2016]{goodfellow2016deep}
\abntrefinfo{Goodfellow, Bengio and Courville}{GOODFELLOW; BENGIO;
  COURVILLE}{2016}
{GOODFELLOW, I.; BENGIO, Y.; COURVILLE, A. \textbf{Deep Learning}. [S.l.]: MIT
  Press, 2016.}

\bibitem[Goyal et al. 2016]{goyal2016interpreting}
\abntrefinfo{Goyal et al.}{GOYAL et al.}{2016}
{GOYAL, Y. et al. Interpreting visual question answering models. In:
  \textbf{ICML Workshop on Visualization for Deep Learning}. [S.l.: s.n.],
  2016. v.~2.}

\bibitem[Haber and Ruthotto 2017]{DBLP:journals/corr/HaberR17}
\abntrefinfo{Haber and Ruthotto}{HABER; RUTHOTTO}{2017}
{HABER, E.; RUTHOTTO, L. Stable architectures for deep neural networks.
\textbf{CoRR}, abs/1705.03341, 2017.
Available at: \url{http://arxiv.org/abs/1705.03341}.}

\bibitem[Hariharan 2018]{freecodecamp}
\abntrefinfo{Hariharan}{HARIHARAN}{2018}
{HARIHARAN, A. \textbf{How to Use Machine Learning to Predict the Quality of
  Wines}. 2018.
Https://medium.freecodecamp.org/using-machine-learning-to-predict-the-quality-of-wines-9e2e13d7480d.}

\bibitem[He et al. 2015]{DBLP:journals/corr/HeZRS15}
\abntrefinfo{He et al.}{HE et al.}{2015}
{HE, K. et al. Deep residual learning for image recognition.
\textbf{CoRR}, abs/1512.03385, 2015.
Available at: \url{http://arxiv.org/abs/1512.03385}.}

\bibitem[Hornik 1991]{hornik1991approximation}
\abntrefinfo{Hornik}{HORNIK}{1991}
{HORNIK, K. Approximation capabilities of multilayer feedforward networks.
\textbf{Neural networks}, Elsevier, v.~4, n.~2, p. 251--257, 1991.}

\bibitem[Hornik, Stinchcombe and White 1989]{HORNIK1989359}
\abntrefinfo{Hornik, Stinchcombe and White}{HORNIK; STINCHCOMBE; WHITE}{1989}
{HORNIK, K.; STINCHCOMBE, M.; WHITE, H. Multilayer feedforward networks are
  universal approximators.
\textbf{Neural Networks}, v.~2, n.~5, p. 359 -- 366, 1989.
ISSN 0893-6080.
Available at:
  \url{http://www.sciencedirect.com/science/article/pii/0893608089900208}.}

\bibitem[Kaggle 2018]{pimakaggle}
\abntrefinfo{Kaggle}{KAGGLE}{2018}
{KAGGLE. \textbf{Pima Indians Diabetes Database}. 2018.
Https://www.kaggle.com/uciml/pima-indians-diabetes-database.}

\bibitem[Kaparthy 2015]{kaparthy_rnn}
\abntrefinfo{Kaparthy}{KAPARTHY}{2015}
{KAPARTHY, A. \textbf{The Unreasonable Effectiveness of Recurrent Neural
  Networks}. 2015.
Http://karpathy.github.io/2015/05/21/rnn-effectiveness/.}

\bibitem[Kaster 2018]{stackplay}
\abntrefinfo{Kaster}{KASTER}{2018}
{KASTER. \textbf{Looking for a 3D smooth step function}. 2018.
Https://math.stackexchange.com/questions/417717/looking\-for\-a\-3d\-smooth\-step\-function.}

\bibitem[Khozeimeh et al. 2017]{KHOZEIMEH2017167}
\abntrefinfo{Khozeimeh et al.}{KHOZEIMEH et al.}{2017a}
{KHOZEIMEH, F. et al. An expert system for selecting wart treatment method.
\textbf{Computers in Biology and Medicine}, v.~81, p. 167 -- 175, 2017.
ISSN 0010-4825.
Available at:
  \url{http://www.sciencedirect.com/science/article/pii/S001048251730001X}.}

\bibitem[Khozeimeh et al. 2017]{KHOZEIMEH20171672}
\abntrefinfo{Khozeimeh et al.}{KHOZEIMEH et al.}{2017b}
{KHOZEIMEH, F. et al. Intralesional immunotherapy compared to cryotherapy in
  the treatment of warts.
\textbf{International Journal of Dermatology}, v.~56, n.~4, p. 474--478, 2017.
Available at: \url{https://onlinelibrary.wiley.com/doi/abs/10.1111/ijd.13535}.}

\bibitem[Kohani 2017]{alireza}
\abntrefinfo{Kohani}{KOHANI}{2017}
{KOHANI, A.~R. \textbf{Regression vs Classification}. 2017.
Https://medium.com/@ali\_88273/regression-vs-classification-87c224350d69.}

\bibitem[Laidlaw et al. 2001]{vecfield2d}
\abntrefinfo{Laidlaw et al.}{LAIDLAW et al.}{2001}
{LAIDLAW, D.~H. et al. Quantitative comparative evaluation of 2d vector field
  visualization methods. In:  \textbf{Proceedings Visualization, 2001. VIS
  '01.} [S.l.: s.n.], 2001. p. 143--150.}

\bibitem[Li et al. 2017]{DBLP:journals/corr/abs-1710-09513}
\abntrefinfo{Li et al.}{LI et al.}{2017}
{LI, Q. et al. Maximum principle based algorithms for deep learning.
\textbf{CoRR}, abs/1710.09513, 2017.
Available at: \url{http://arxiv.org/abs/1710.09513}.}

\bibitem[Liu and Salinas 2015]{LIU20151636}
\abntrefinfo{Liu and Salinas}{LIU; SALINAS}{2015}
{LIU, N.~T.; SALINAS, J. Machine learning in burn care and research: A
  systematic review of the literature.
\textbf{Burns}, v.~41, n.~8, p. 1636 -- 1641, 2015.
ISSN 0305-4179.
Available at:
  \url{http://www.sciencedirect.com/science/article/pii/S0305417915002004}.}

\bibitem[Lu et al. 2017]{DBLP:journals/corr/abs-1710-10121}
\abntrefinfo{Lu et al.}{LU et al.}{2017}
{LU, Y. et al. Beyond finite layer neural networks: Bridging deep architectures
  and numerical differential equations.
\textbf{CoRR}, abs/1710.10121, 2017.
Available at: \url{http://arxiv.org/abs/1710.10121}.}

\bibitem[Marsden and Tromba 2011]{marsden2011vector}
\abntrefinfo{Marsden and Tromba}{MARSDEN; TROMBA}{2011}
{MARSDEN, J.; TROMBA, A. \textbf{Vector Calculus}. W. H. Freeman, 2011.
ISBN 9781429215084. Available at:
  \url{https://books.google.com.br/books?id=b3oVDAEACAAJ}.}

\bibitem[McCulloch and Pitts 1943]{mcculloch1943logical}
\abntrefinfo{McCulloch and Pitts}{MCCULLOCH; PITTS}{1943}
{MCCULLOCH, W.~S.; PITTS, W. A logical calculus of the ideas immanent in
  nervous activity.
\textbf{The bulletin of mathematical biophysics}, Springer, v.~5, n.~4, p.
  115--133, 1943.}

\bibitem[Mescheder, Nowozin and Geiger 2017]{mescheder2017numerics}
\abntrefinfo{Mescheder, Nowozin and Geiger}{MESCHEDER; NOWOZIN; GEIGER}{2017}
{MESCHEDER, L.; NOWOZIN, S.; GEIGER, A. The numerics of gans.
\textbf{arXiv preprint arXiv:1705.10461}, 2017.}

\bibitem[Nielsen 2015]{michaelnielsen}
\abntrefinfo{Nielsen}{NIELSEN}{2015}
{NIELSEN, M.~A. \textbf{Neural Networks and Deep Learning}. [S.l.]:
  Determination Press, 2015.
Http://neuralnetworksanddeeplearning.com.}

\bibitem[Olah 2014]{colah}
\abntrefinfo{Olah}{OLAH}{2014}
{OLAH, C. \textbf{Neural Networks, Manifolds and Topology}. 2014.
Http://colah.github.io/posts/2014-03-NN-Manifolds-Topology/.}

\bibitem[Olah 2015]{colahfunc}
\abntrefinfo{Olah}{OLAH}{2015a}
{OLAH, C. \textbf{Neural Networks, Types, and Functional Programming}. 2015.
Http://colah.github.io/posts/2015-09-NN-Types-FP/.}

\bibitem[Olah 2015]{colah_lstm}
\abntrefinfo{Olah}{OLAH}{2015b}
{OLAH, C. \textbf{Understanding LSTM Networks}. 2015.
Http://colah.github.io/posts/2015-08-Understanding-LSTMs/.}

\bibitem[PARLIAMENT and UNION 2016]{eu-679-2016}
\abntrefinfo{PARLIAMENT and UNION}{PARLIAMENT; UNION}{2016}
{PARLIAMENT, E.; UNION, T. C. O. T.~E. \textbf{REGULATION (EU) 2016/679 OF THE
  EUROPEAN PARLIAMENT AND OF THE COUNCIL of 27 April 2016}. 2016.
Http://ec.europa.eu/justice/data-protection/reform/files/regulation\_oj\_en.pdf.}

\bibitem[Pedregosa et al. 2011]{scikit-learn}
\abntrefinfo{Pedregosa et al.}{PEDREGOSA et al.}{2011}
{PEDREGOSA, F. et al. Scikit-learn: Machine learning in {P}ython.
\textbf{Journal of Machine Learning Research}, v.~12, p. 2825--2830, 2011.}

\bibitem[Rosenblatt 1962]{rosenblatt1961principles}
\abntrefinfo{Rosenblatt}{ROSENBLATT}{1962}
{ROSENBLATT, F. \textbf{Principles of Neurodynamics: Perceptrons and the Theory
  of Brain Mechanisms}. [S.l.]: Spartan Books, 1962.}

\bibitem[ScikitLearn 2018]{scikitunder}
\abntrefinfo{ScikitLearn}{SCIKITLEARN}{2018}
{SCIKITLEARN. \textbf{Underfitting vs. Overfitting}. 2018.
Http://scikit-learn.org/stable/auto\_examples/model\_selection/plot\_underfitting\_overfitting.html.}

\bibitem[Sharma 2017]{towardsdata}
\abntrefinfo{Sharma}{SHARMA}{2017}
{SHARMA, S. \textbf{Activation Functions: Neural Networks}. 2017.
Https://towardsdatascience.com/activation-functions-neural-networks-1cbd9f8d91d6.}

\bibitem[Silver et al. 2016]{silver2016mastering}
\abntrefinfo{Silver et al.}{SILVER et al.}{2016}
{SILVER, D. et al. Mastering the game of go with deep neural networks and tree
  search.
\textbf{nature}, Nature Publishing Group, v.~529, n.~7587, p.~484, 2016.}

\bibitem[Smilkov and Carter 2018]{tensorplay}
\abntrefinfo{Smilkov and Carter}{SMILKOV; CARTER}{2018}
{SMILKOV, D.; CARTER, S. \textbf{Tinker With a Neural Network}. 2018.
Http://playground.tensorflow.org.}

\bibitem[Vapnik and Chervonenkis 2015]{vapnik2015uniform}
\abntrefinfo{Vapnik and Chervonenkis}{VAPNIK; CHERVONENKIS}{2015}
{VAPNIK, V.~N.; CHERVONENKIS, A.~Y. On the uniform convergence of relative
  frequencies of events to their probabilities. In:  \textbf{Measures of
  complexity}. [S.l.]: Springer, 2015. p. 11--30.}

\bibitem[Vieira et al. 2018]{vieira}
\abntrefinfo{Vieira et al.}{VIEIRA et al.}{2018}
{VIEIRA, D. et al. Vector field based neural networks. In:  \textbf{Proceedings
  of the European Symposium on Artificial Neural Networks, Computational
  Intelligence and Machine Learning}. [s.n.], 2018. p. 579--584. Available at:
  \url{https://www.elen.ucl.ac.be/Proceedings/esann/esannpdf/es2018-192.pdf}.}

\bibitem[Walia 2017]{walia_opt}
\abntrefinfo{Walia}{WALIA}{2017}
{WALIA, A.~S. \textbf{Types of Optimization Algorithms used in Neural Networks
  and Ways to Optimize Gradient Descent}. 2017.
Https://towardsdatascience.com/types-of-optimization-algorithms-used-in-neural-networks-and-ways-to-optimize-gradient-95ae5d39529f.}

\end{thebibliography}
%
%
\begin{apendicesenv}
\chapter{Banknote Hyperparameter Results}
\label{appendicebanknote}

\begin{figure} [!ht]
    \centering
    \caption{\vspace{-0.1cm} Accuracy mean and standard deviation measurements for the VFNN on the Banknote dataset with eta = 0.01. }
    \begin{subfigure}[b]{0.45\textwidth}
        \includegraphics[width=\textwidth]{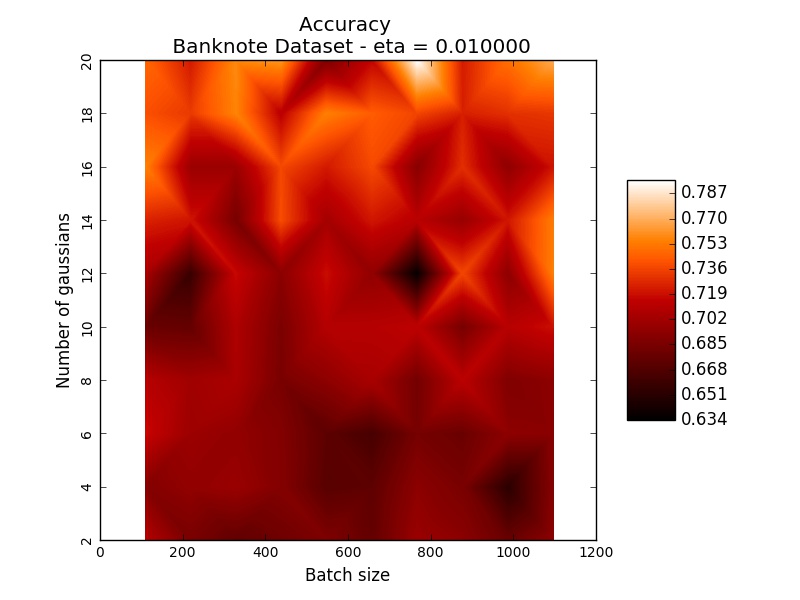}
        \label{fig:apaccbank001mean}
    \end{subfigure}
    \begin{subfigure}[b]{0.45\textwidth}
        \includegraphics[width=\textwidth]{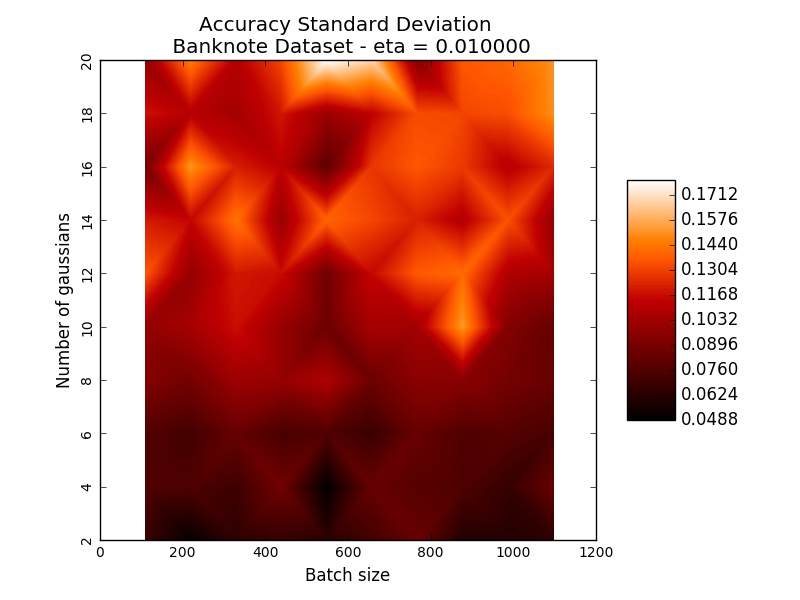}
        \label{fig:apaccbank001std}
    \end{subfigure}
    \vspace{-0.25cm}
    \caption*{\vspace{-0.1cm} Source: Author. }
    \label{fig:apaccbank001}
\end{figure}

\begin{figure} [!ht]
    \centering
    \caption{\vspace{-0.1cm} Accuracy mean and standard deviation measurements for the VFNN on the Banknote dataset with eta = 0.2575. }
    \begin{subfigure}[b]{0.45\textwidth}
        \includegraphics[width=\textwidth]{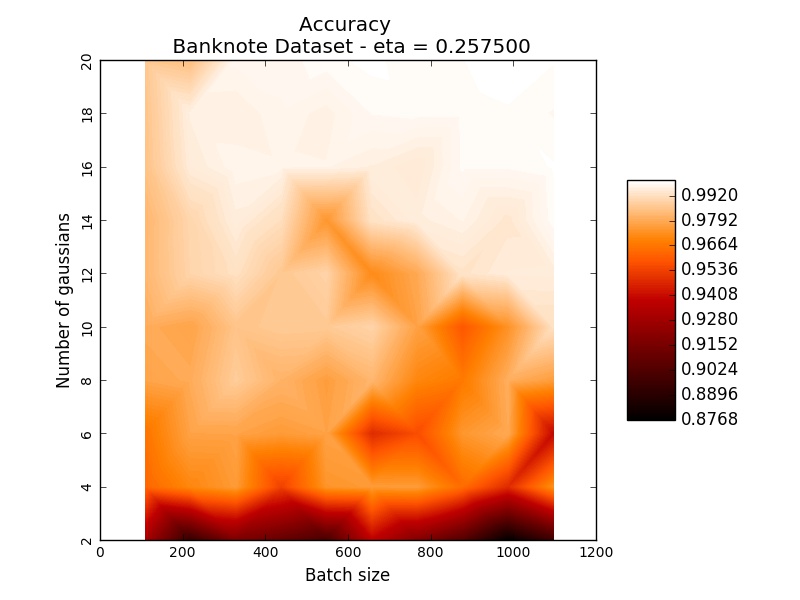}
        \label{fig:apaccbank025mean}
    \end{subfigure}
    \begin{subfigure}[b]{0.45\textwidth}
        \includegraphics[width=\textwidth]{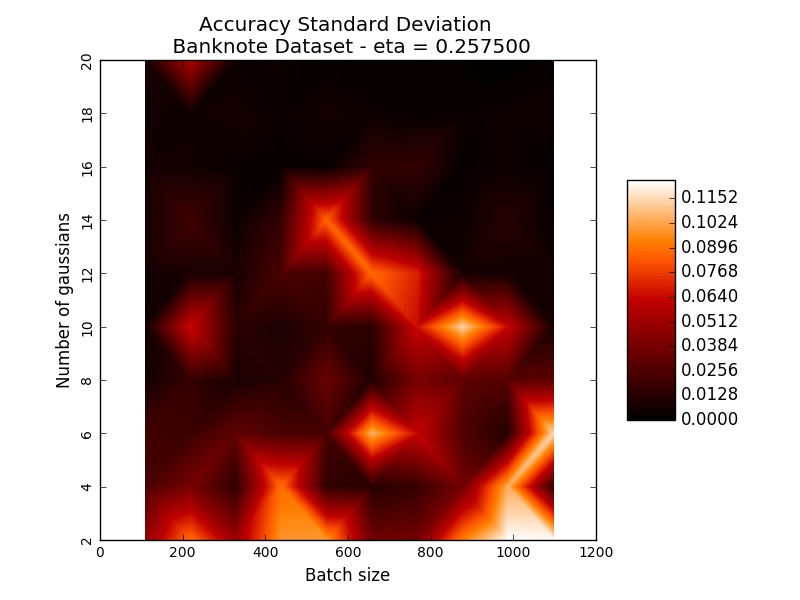}
        \label{fig:apaccbank025std}
    \end{subfigure}
    \vspace{-0.25cm}
    \caption*{\vspace{-0.1cm} Source: Author. }
    \label{fig:apaccbank025}
\end{figure}

\begin{figure} [!ht]
    \centering
    \caption{\vspace{-0.1cm} Accuracy mean and standard deviation measurements for the VFNN on the Banknote dataset with eta = 0.505. }
    \begin{subfigure}[b]{0.45\textwidth}
        \includegraphics[width=\textwidth]{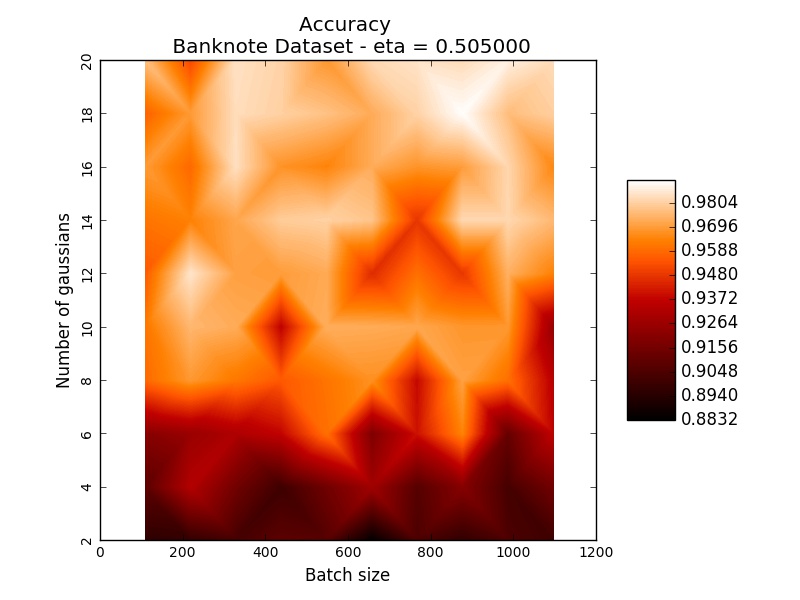}
        \label{fig:apaccbank0505mean}
    \end{subfigure}
    \begin{subfigure}[b]{0.45\textwidth}
        \includegraphics[width=\textwidth]{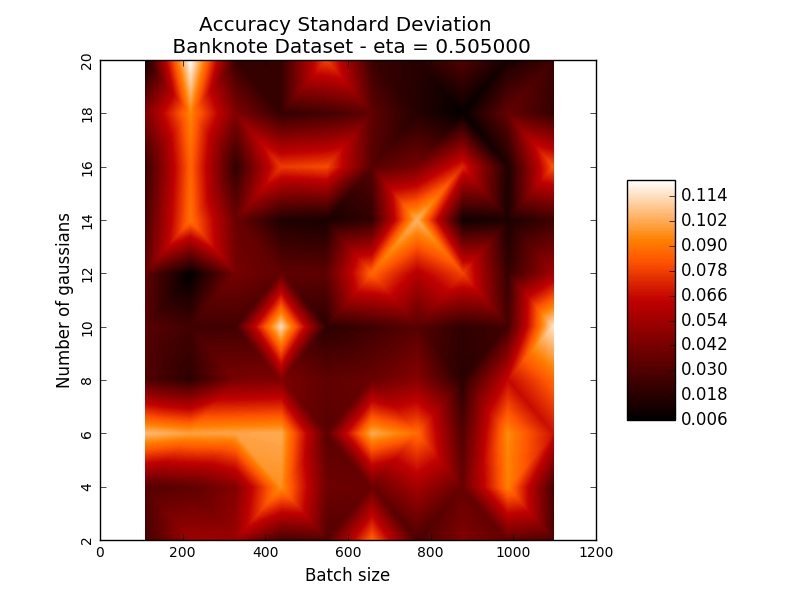}
        \label{fig:apaccbank0505std}
    \end{subfigure}
    \vspace{-0.25cm}
    \caption*{\vspace{-0.1cm} Source: Author. }
    \label{fig:apaccbank505}
\end{figure}

\begin{figure} [!ht]
    \centering
    \caption{\vspace{-0.1cm} Accuracy mean and standard deviation measurements for the VFNN on the Banknote dataset with eta = 0.7525. }
    \begin{subfigure}[b]{0.45\textwidth}
        \includegraphics[width=\textwidth]{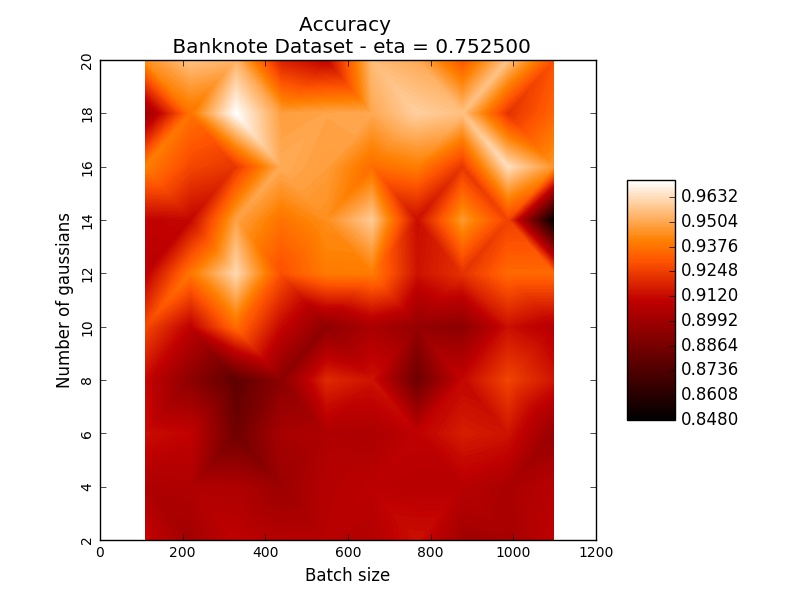}
        \label{fig:apaccbank075mean}
    \end{subfigure}
    \begin{subfigure}[b]{0.45\textwidth}
        \includegraphics[width=\textwidth]{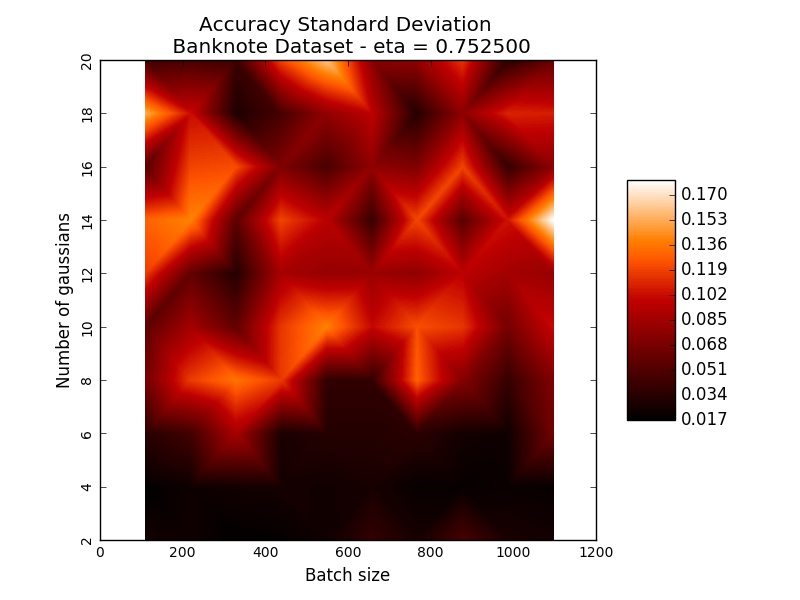}
        \label{fig:apaccbank075std}
    \end{subfigure}
    \vspace{-0.25cm}
    \caption*{\vspace{-0.1cm} Source: Author. }
    \label{fig:apaccbank75}
\end{figure}

\begin{figure} [!ht]
    \centering
    \caption{\vspace{-0.1cm} Accuracy mean and standard deviation measurements for the VFNN on the Banknote dataset with eta = 1.0. }
    \begin{subfigure}[b]{0.45\textwidth}
        \includegraphics[width=\textwidth]{Figures/HyperparametersResults/AccuracyBanknote1000000.jpg}
        \label{fig:apaccbank100mean}
    \end{subfigure}
    \begin{subfigure}[b]{0.45\textwidth}
        \includegraphics[width=\textwidth]{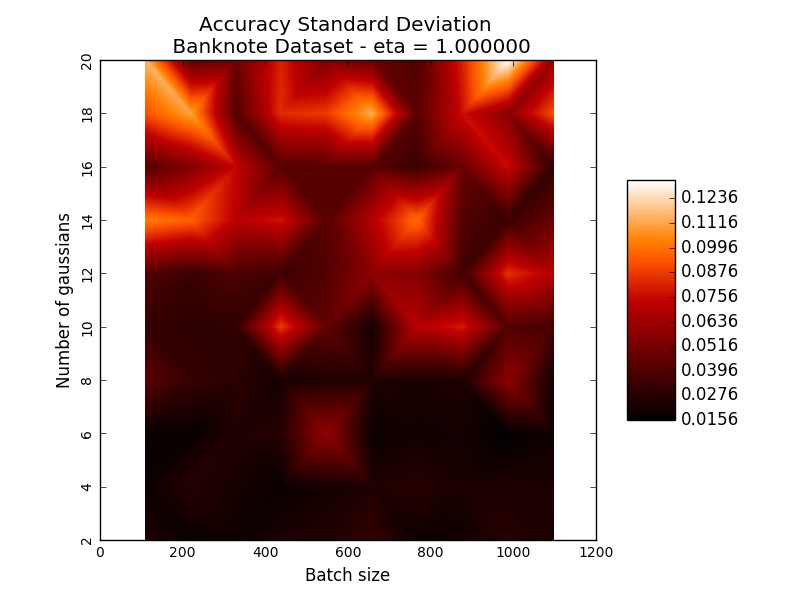}
        \label{fig:apaccbank100std}
    \end{subfigure}
    \vspace{-0.25cm}
    \caption*{\vspace{-0.1cm} Source: Author. }
    \label{fig:apaccbank100}
\end{figure}

\begin{figure} [!ht]
    \centering
    \caption{\vspace{-0.1cm} AUCROC mean and standard deviation measurements for the VFNN on the Banknote dataset with eta = 0.01. }
    \begin{subfigure}[b]{0.45\textwidth}
        \includegraphics[width=\textwidth]{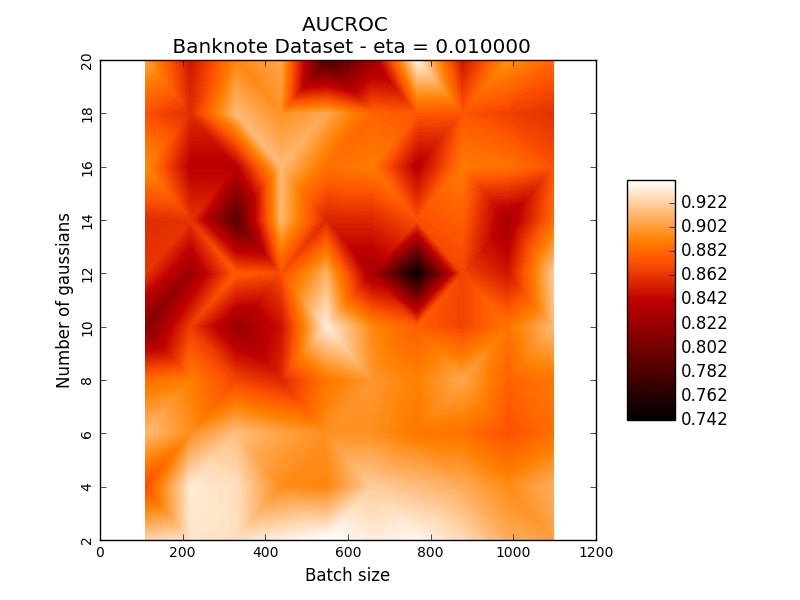}
        \label{fig:apaucbank001mean}
    \end{subfigure}
    \begin{subfigure}[b]{0.45\textwidth}
        \includegraphics[width=\textwidth]{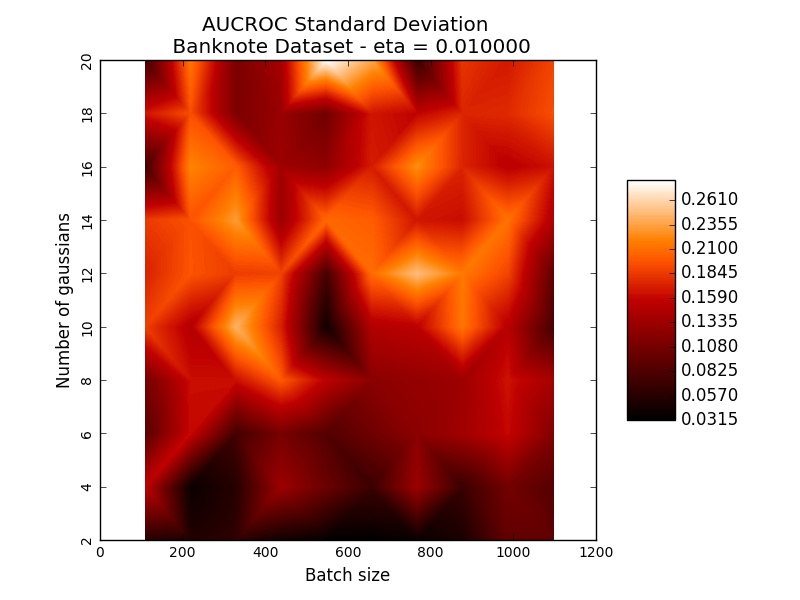}
        \label{fig:apaucbank001std}
    \end{subfigure}
    \vspace{-0.25cm}
    \caption*{\vspace{-0.1cm} Source: Author. }
    \label{fig:apaucbank001}
\end{figure}

\begin{figure} [!ht]
    \centering
    \caption{\vspace{-0.1cm} AUCROC mean and standard deviation measurements for the VFNN on the Banknote dataset with eta = 0.2575. }
    \begin{subfigure}[b]{0.45\textwidth}
        \includegraphics[width=\textwidth]{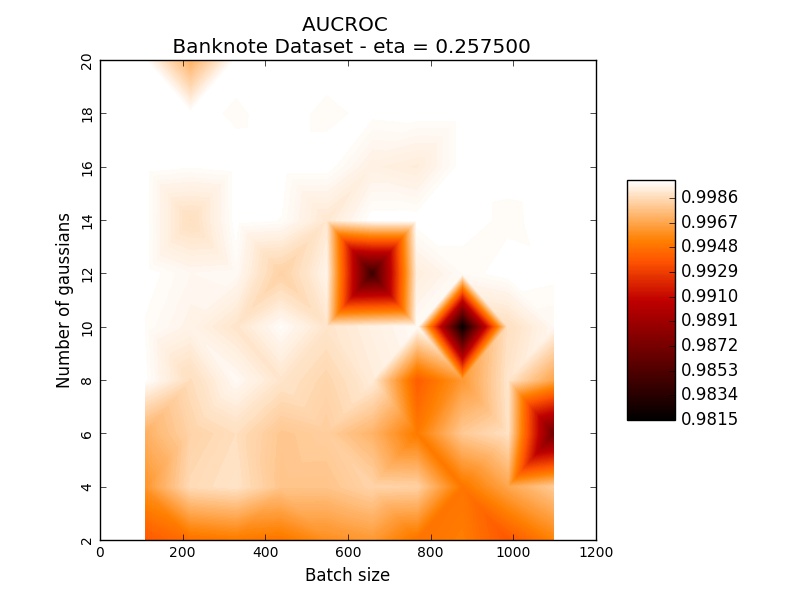}
        \label{fig:apaucbank025mean}
    \end{subfigure}
    \begin{subfigure}[b]{0.45\textwidth}
        \includegraphics[width=\textwidth]{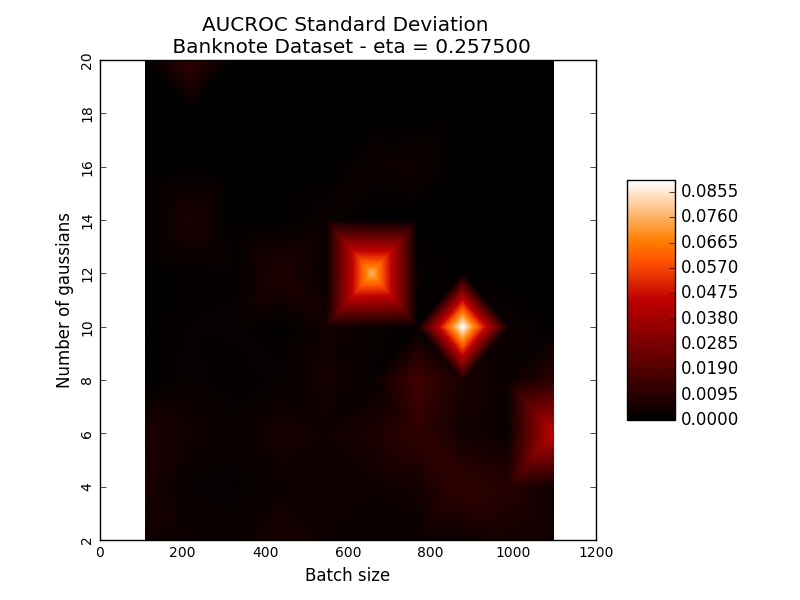}
        \label{fig:apaucbank025std}
    \end{subfigure}
    \vspace{-0.25cm}
    \caption*{\vspace{-0.1cm} Source: Author. }
    \label{fig:apaucbank025}
\end{figure}

\begin{figure} [!ht]
    \centering
    \caption{\vspace{-0.1cm} AUCROC mean and standard deviation measurements for the VFNN on the Banknote dataset with eta = 0.505. }
    \begin{subfigure}[b]{0.45\textwidth}
        \includegraphics[width=\textwidth]{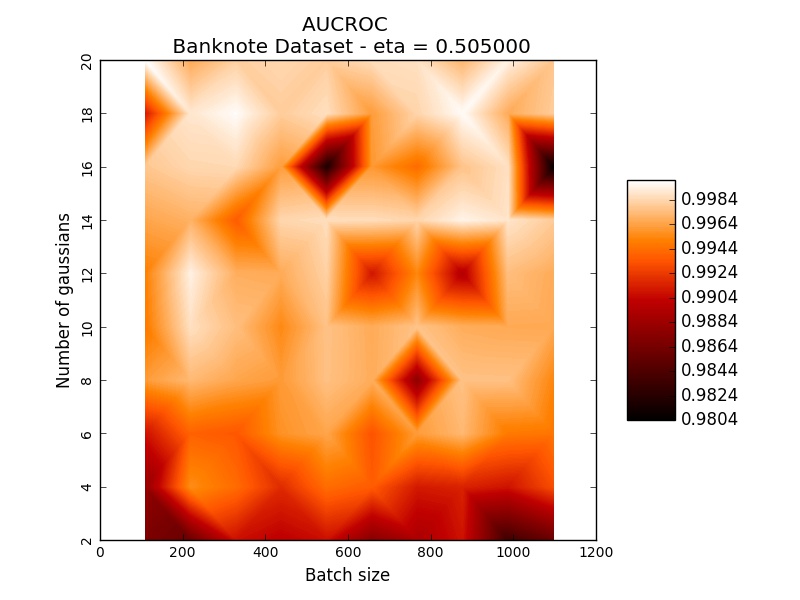}
        \label{fig:apaucbank0505mean}
    \end{subfigure}
    \begin{subfigure}[b]{0.45\textwidth}
        \includegraphics[width=\textwidth]{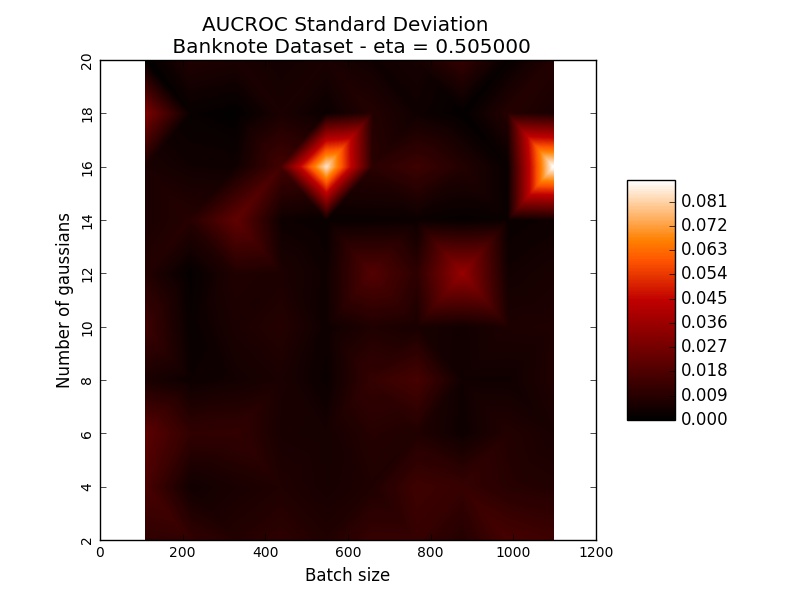}
        \label{fig:apaucbank0505std}
    \end{subfigure}
    \vspace{-0.25cm}
    \caption*{\vspace{-0.1cm} Source: Author. }
    \label{fig:apaucbank0505}
\end{figure}

\begin{figure} [!ht]
    \centering
    \caption{\vspace{-0.1cm} AUCROC mean and standard deviation measurements for the VFNN on the Banknote dataset with eta = 0.7525. }
    \begin{subfigure}[b]{0.45\textwidth}
        \includegraphics[width=\textwidth]{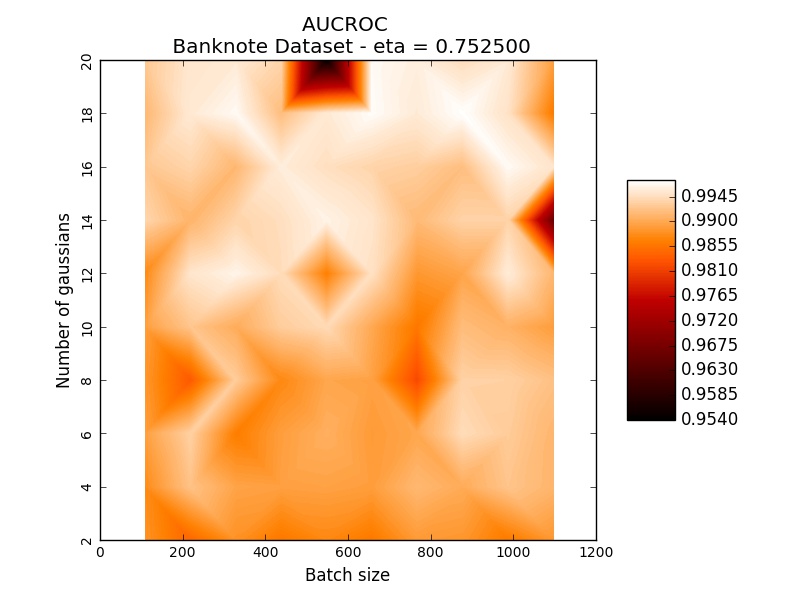}
        \label{fig:apaucbank075mean}
    \end{subfigure}
    \begin{subfigure}[b]{0.45\textwidth}
        \includegraphics[width=\textwidth]{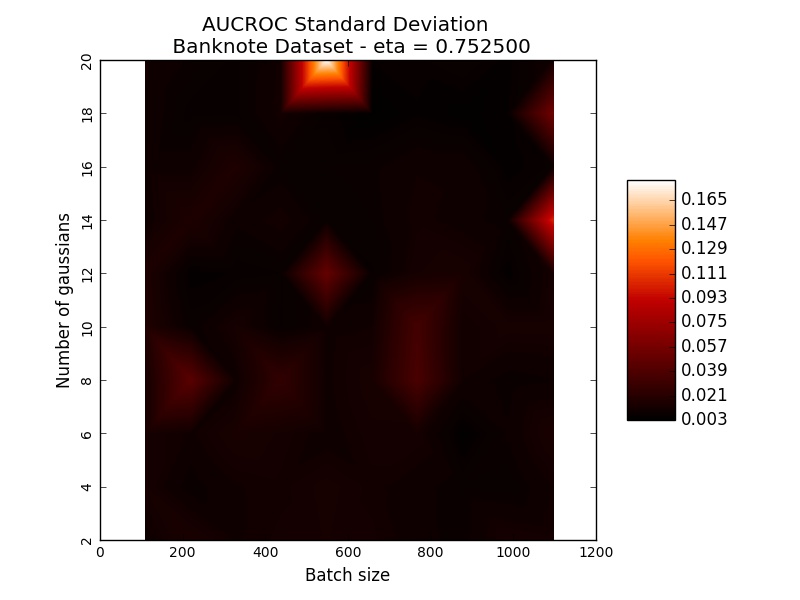}
        \label{fig:apaucbank075std}
    \end{subfigure}
    \vspace{-0.25cm}
    \caption*{\vspace{-0.1cm} Source: Author. }
    \label{fig:apaucbank075}
\end{figure}

\begin{figure} [!ht]
    \centering
    \caption{\vspace{-0.1cm} AUCROC mean and standard deviation measurements for the VFNN on the Banknote dataset with eta = 1.0. }
    \begin{subfigure}[b]{0.45\textwidth}
        \includegraphics[width=\textwidth]{Figures/HyperparametersResults/AUCROCBanknote1000000.jpg}
        \label{fig:apaucbank100mean}
    \end{subfigure}
    \begin{subfigure}[b]{0.45\textwidth}
        \includegraphics[width=\textwidth]{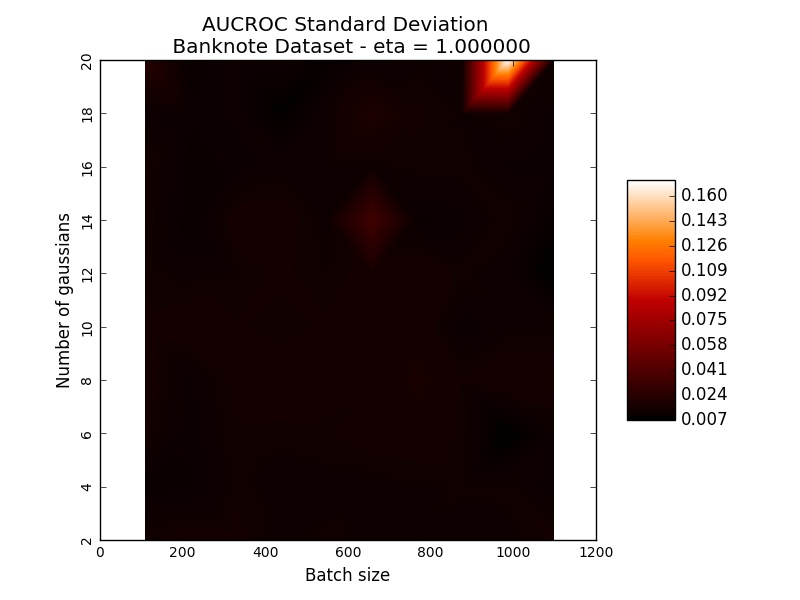}
        \label{fig:apaucbank100std}
    \end{subfigure}
    \vspace{-0.25cm}
    \caption*{\vspace{-0.1cm} Source: Author. }
    \label{fig:apaucbank100}
\end{figure}

\begin{figure} [!ht]
    \centering
    \caption{\vspace{-0.1cm} Mean Squared Error and standard deviation measurements for the VFNN on the Banknote dataset with eta = 0.01. }
    \begin{subfigure}[b]{0.45\textwidth}
        \includegraphics[width=\textwidth]{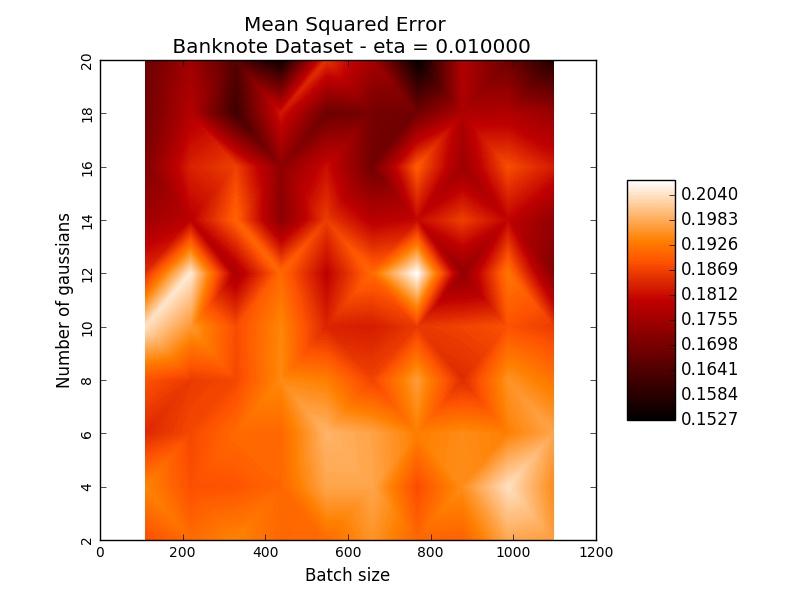}
        \label{fig:apmsebank001mean}
    \end{subfigure}
    \begin{subfigure}[b]{0.45\textwidth}
        \includegraphics[width=\textwidth]{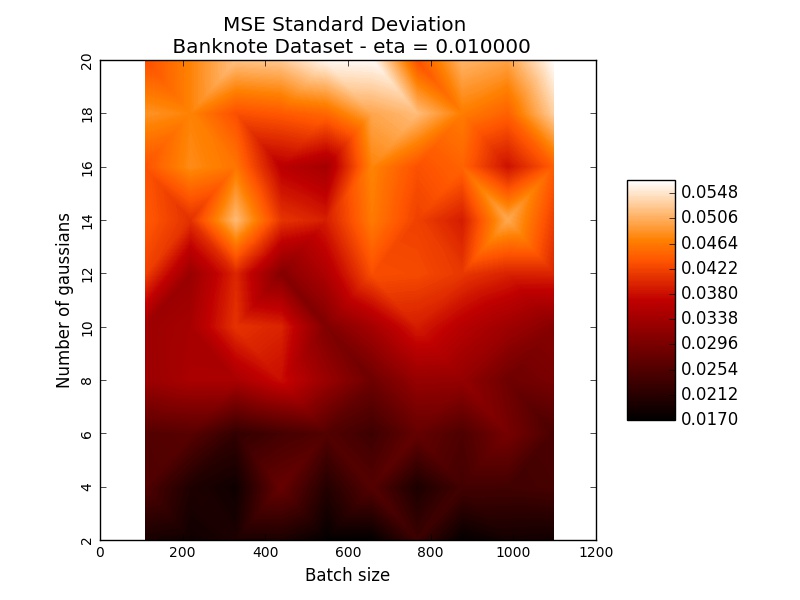}
        \label{fig:apmsebank001std}
    \end{subfigure}
    \vspace{-0.25cm}
    \caption*{\vspace{-0.1cm} Source: Author. }
    \label{fig:apmsebank001}
\end{figure}

\begin{figure} [!ht]
    \centering
    \caption{\vspace{-0.1cm} Mean Squared Error and standard deviation measurements for the VFNN on the Banknote dataset with eta = 0.2575. }
    \begin{subfigure}[b]{0.45\textwidth}
        \includegraphics[width=\textwidth]{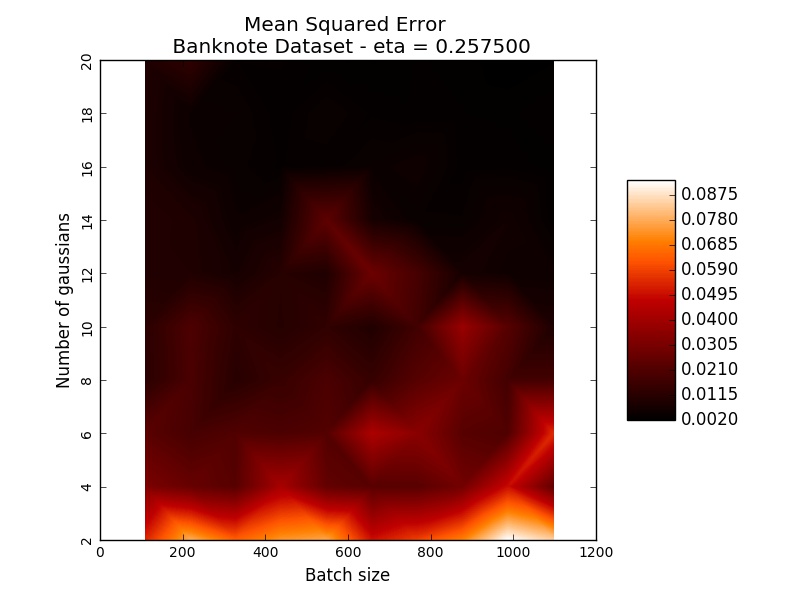}
        \label{fig:apmsebank025mean}
    \end{subfigure}
    \begin{subfigure}[b]{0.45\textwidth}
        \includegraphics[width=\textwidth]{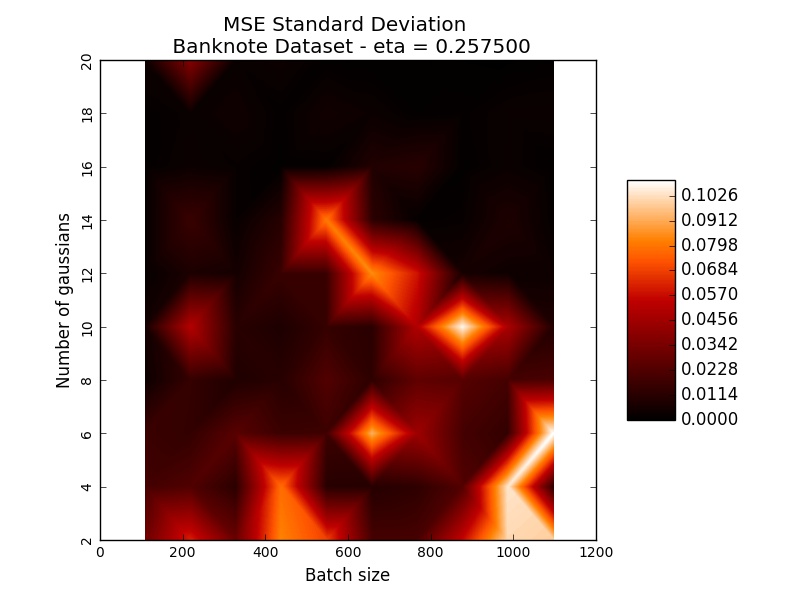}
        \label{fig:apmsebank025std}
    \end{subfigure}
    \vspace{-0.25cm}
    \caption*{\vspace{-0.1cm} Source: Author. }
    \label{fig:apmsebank025}
\end{figure}

\begin{figure} [!ht]
    \centering
    \caption{\vspace{-0.1cm} Mean Squared Error and standard deviation measurements for the VFNN on the Banknote dataset with eta = 0.505. }
    \begin{subfigure}[b]{0.45\textwidth}
        \includegraphics[width=\textwidth]{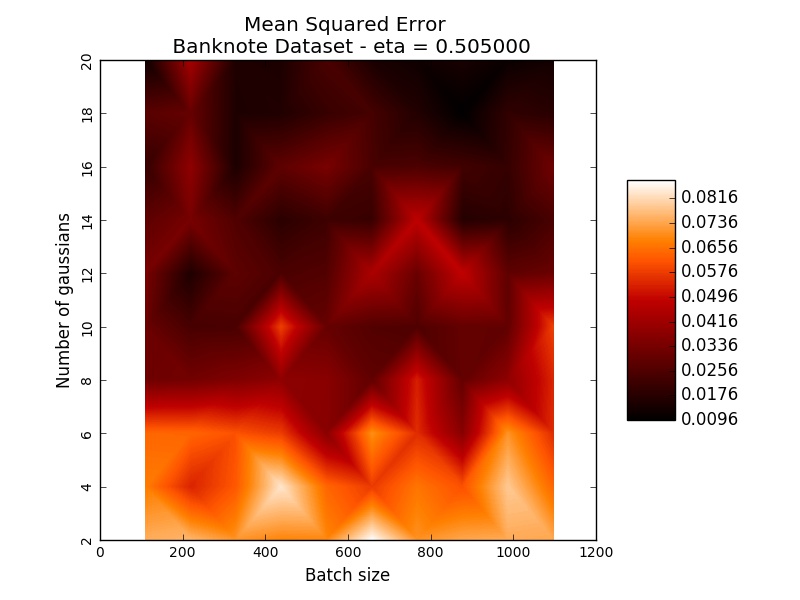}
        \label{fig:apmsebank0505mean}
    \end{subfigure}
    \begin{subfigure}[b]{0.45\textwidth}
        \includegraphics[width=\textwidth]{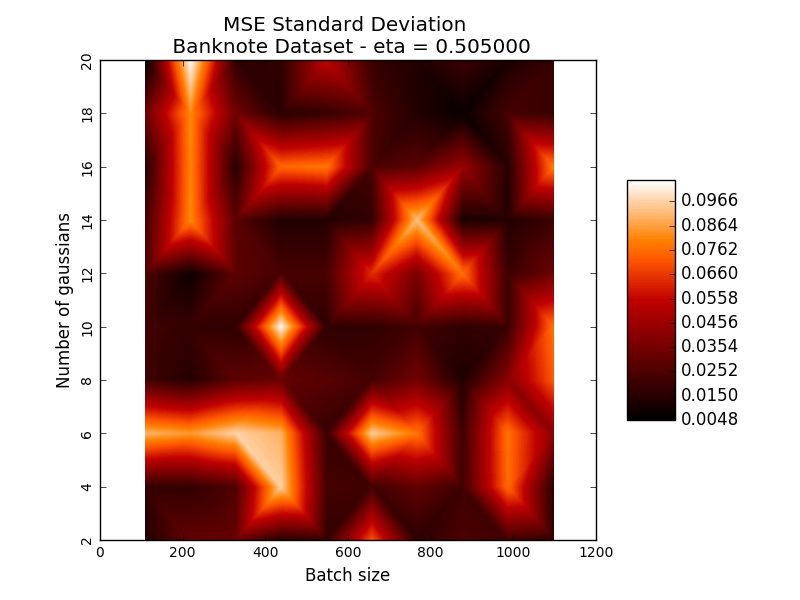}
        \label{fig:apmsebank0505std}
    \end{subfigure}
    \vspace{-0.25cm}
    \caption*{\vspace{-0.1cm} Source: Author. }
    \label{fig:apmsebank0505}
\end{figure}

\begin{figure} [!ht]
    \centering
    \caption{\vspace{-0.1cm} Mean Squared Error and standard deviation measurements for the VFNN on the Banknote dataset with eta = 0.7525. }
    \begin{subfigure}[b]{0.45\textwidth}
        \includegraphics[width=\textwidth]{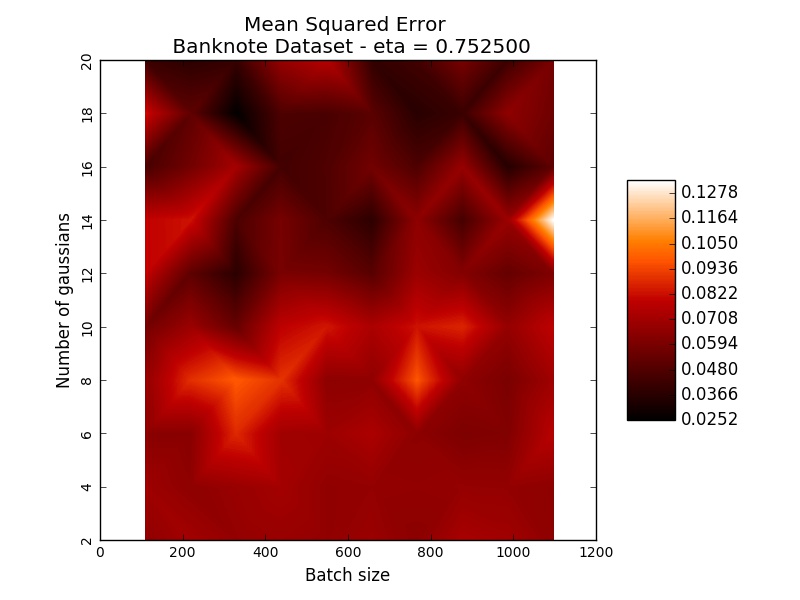}
        \label{fig:apmsebank075mean}
    \end{subfigure}
    \begin{subfigure}[b]{0.45\textwidth}
        \includegraphics[width=\textwidth]{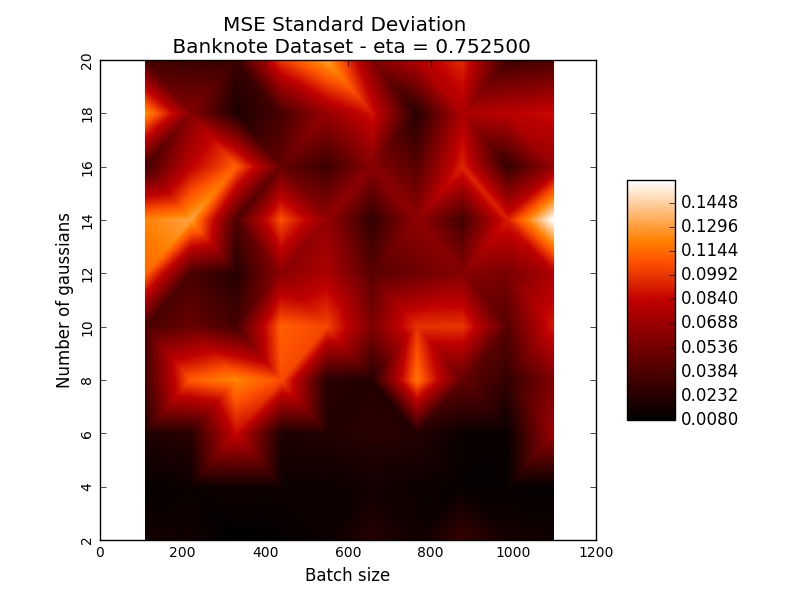}
        \label{fig:apmsebank075std}
    \end{subfigure}
    \vspace{-0.25cm}
    \caption*{\vspace{-0.1cm} Source: Author. }
    \label{fig:apmsebank075}
\end{figure}

\begin{figure} [!ht]
    \centering
    \caption{\vspace{-0.1cm} Mean Squared Error and standard deviation measurements for the VFNN on the Banknote dataset with eta = 1.0. }
    \begin{subfigure}[b]{0.45\textwidth}
        \includegraphics[width=\textwidth]{Figures/HyperparametersResults/MeanSquaredErrorBanknote1000000.jpg}
        \label{fig:apmsebank100mean}
    \end{subfigure}
    \begin{subfigure}[b]{0.45\textwidth}
        \includegraphics[width=\textwidth]{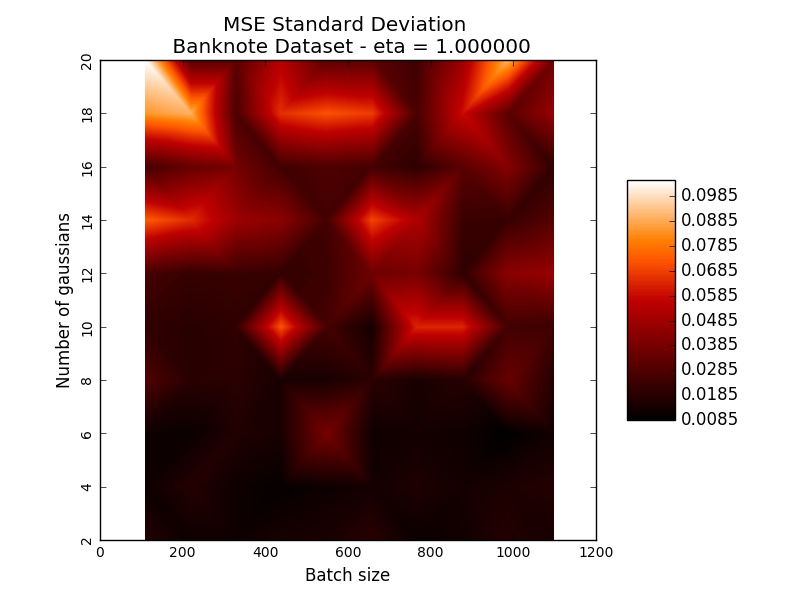}
        \label{fig:apmsebank100std}
    \end{subfigure}
    \vspace{-0.25cm}
    \caption*{\vspace{-0.1cm} Source: Author. }
    \label{fig:apmsebank0100}
\end{figure}

\chapter{Ionosphere Hyperparameter Results}
\label{appendiceionosphere}

\begin{figure} [!ht]
    \centering
    \caption{\vspace{-0.1cm} Accuracy mean and standard deviation measurements for the VFNN on the Ionosphere dataset with eta = 0.01. }
    \begin{subfigure}[b]{0.45\textwidth}
        \includegraphics[width=\textwidth]{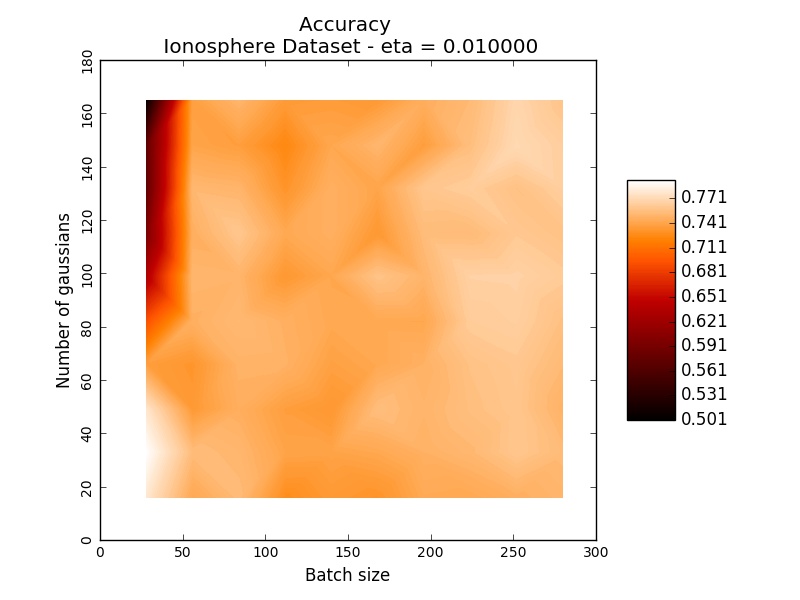}
        \label{fig:apacciono001mean}
    \end{subfigure}
    \begin{subfigure}[b]{0.45\textwidth}
        \includegraphics[width=\textwidth]{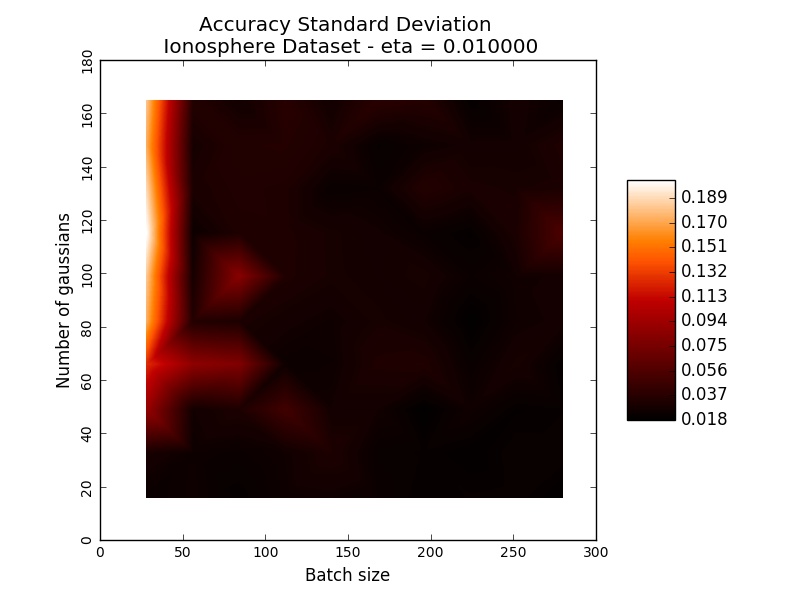}
        \label{fig:apacciono001std}
    \end{subfigure}
    \vspace{-0.25cm}
    \caption*{\vspace{-0.1cm} Source: Author. }
    \label{fig:apacciono001}
\end{figure}

\begin{figure} [!ht]
    \centering
    \caption{\vspace{-0.1cm} Accuracy mean and standard deviation measurements for the VFNN on the Ionosphere dataset with eta = 0.2575. }
    \begin{subfigure}[b]{0.45\textwidth}
        \includegraphics[width=\textwidth]{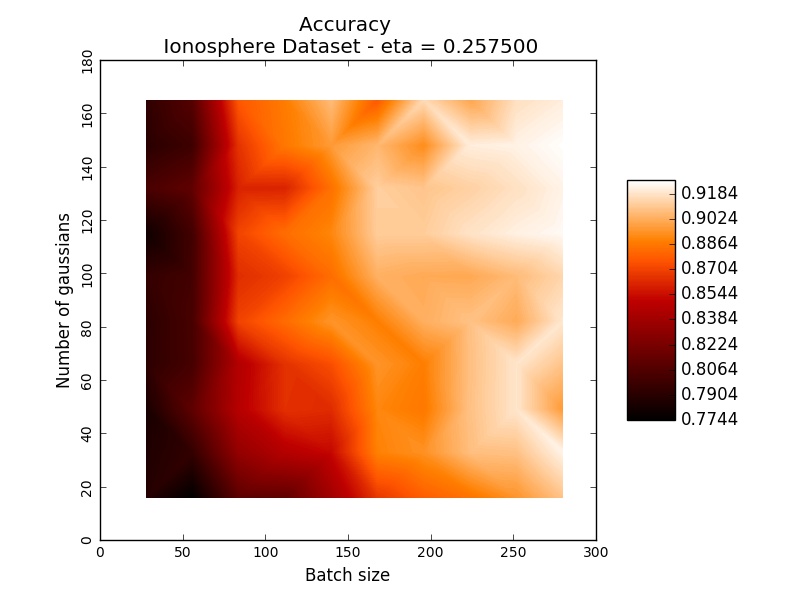}
        \label{fig:apacciono025mean}
    \end{subfigure}
    \begin{subfigure}[b]{0.45\textwidth}
        \includegraphics[width=\textwidth]{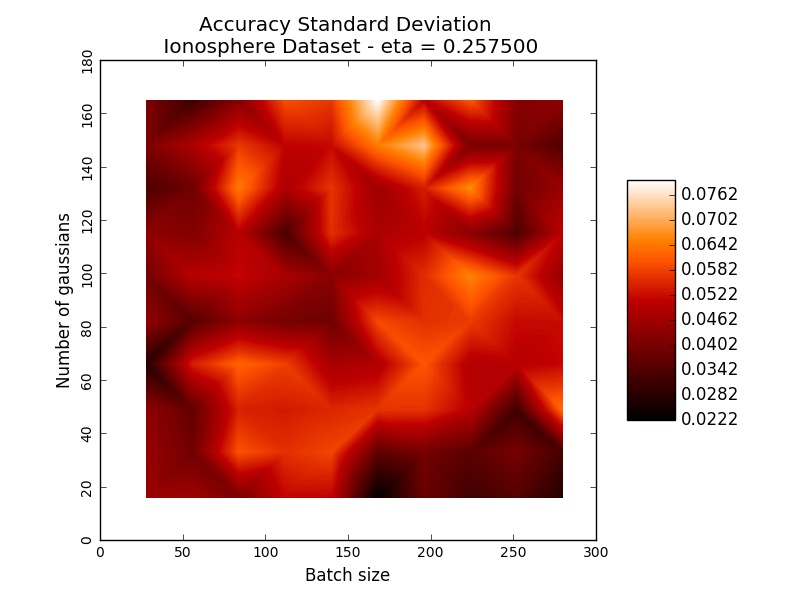}
        \label{fig:apacciono025std}
    \end{subfigure}
    \vspace{-0.25cm}
    \caption*{\vspace{-0.1cm} Source: Author. }
    \label{fig:apacciono025}
\end{figure}

\begin{figure} [!ht]
    \centering
    \caption{\vspace{-0.1cm} Accuracy mean and standard deviation measurements for the VFNN on the Ionosphere dataset with eta = 0.505. }
    \begin{subfigure}[b]{0.45\textwidth}
        \includegraphics[width=\textwidth]{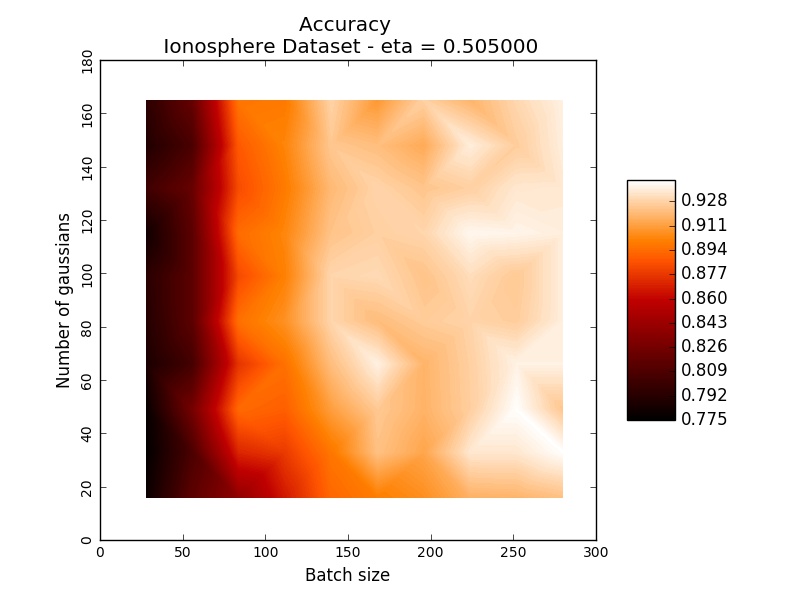}
        \label{fig:apacciono0505mean}
    \end{subfigure}
    \begin{subfigure}[b]{0.45\textwidth}
        \includegraphics[width=\textwidth]{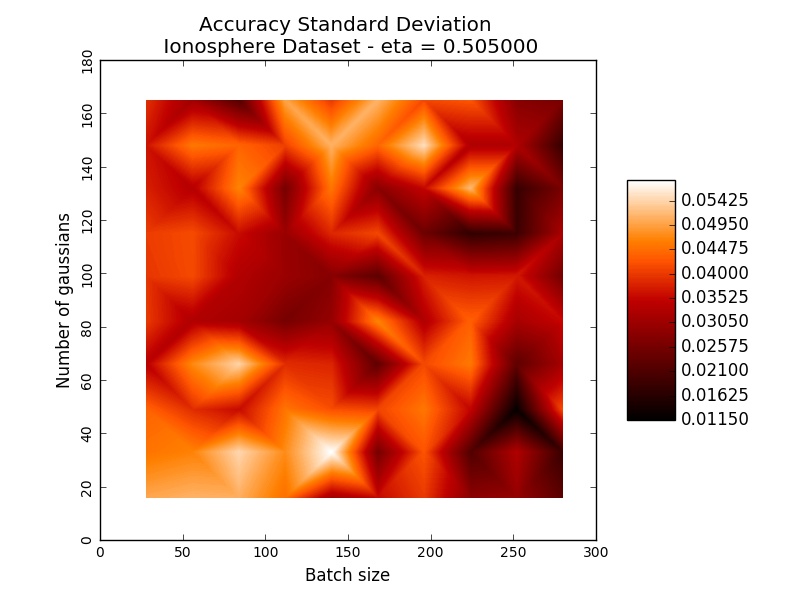}
        \label{fig:apacciono0505std}
    \end{subfigure}
    \vspace{-0.25cm}
    \caption*{\vspace{-0.1cm} Source: Author. }
    \label{fig:apacciono505}
\end{figure}

\begin{figure} [!ht]
    \centering
    \caption{\vspace{-0.1cm} Accuracy mean and standard deviation measurements for the VFNN on the Ionosphere dataset with eta = 0.7525. }
    \begin{subfigure}[b]{0.45\textwidth}
        \includegraphics[width=\textwidth]{Figures/HyperparametersResults/AccuracyIonosphere0752500.jpg}
        \label{fig:apacciono075mean}
    \end{subfigure}
    \begin{subfigure}[b]{0.45\textwidth}
        \includegraphics[width=\textwidth]{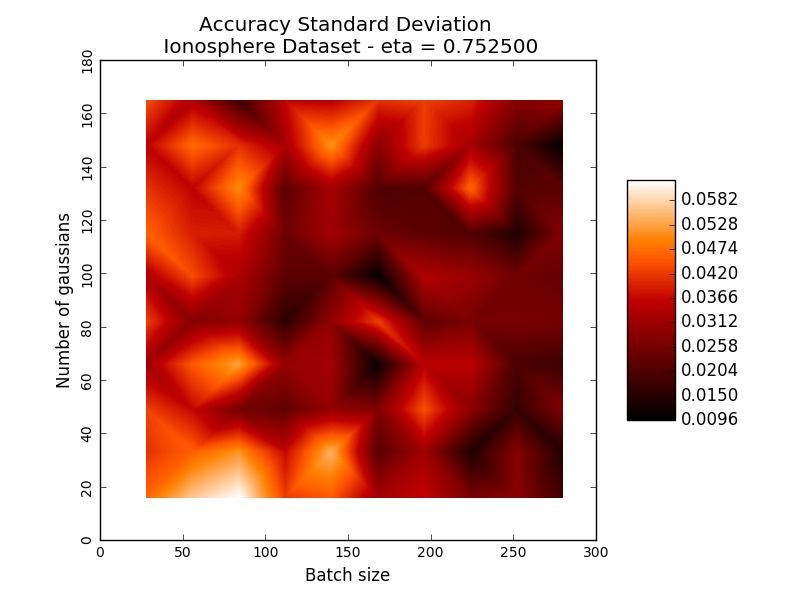}
        \label{fig:apacciono075std}
    \end{subfigure}
    \vspace{-0.25cm}
    \caption*{\vspace{-0.1cm} Source: Author. }
    \label{fig:apacciono75}
\end{figure}

\begin{figure} [!ht]
    \centering
    \caption{\vspace{-0.1cm} Accuracy mean and standard deviation measurements for the VFNN on the Ionosphere dataset with eta = 1.0. }
    \begin{subfigure}[b]{0.45\textwidth}
        \includegraphics[width=\textwidth]{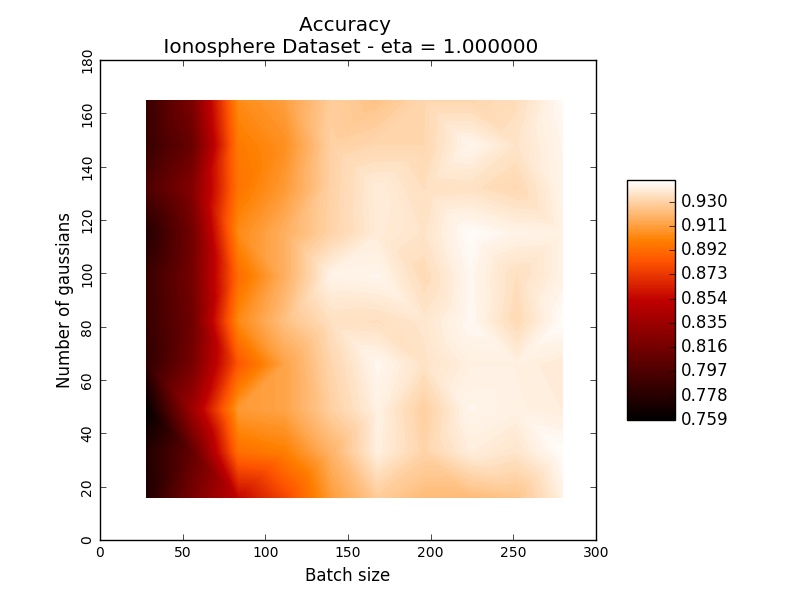}
        \label{fig:apacciono100mean}
    \end{subfigure}
    \begin{subfigure}[b]{0.45\textwidth}
        \includegraphics[width=\textwidth]{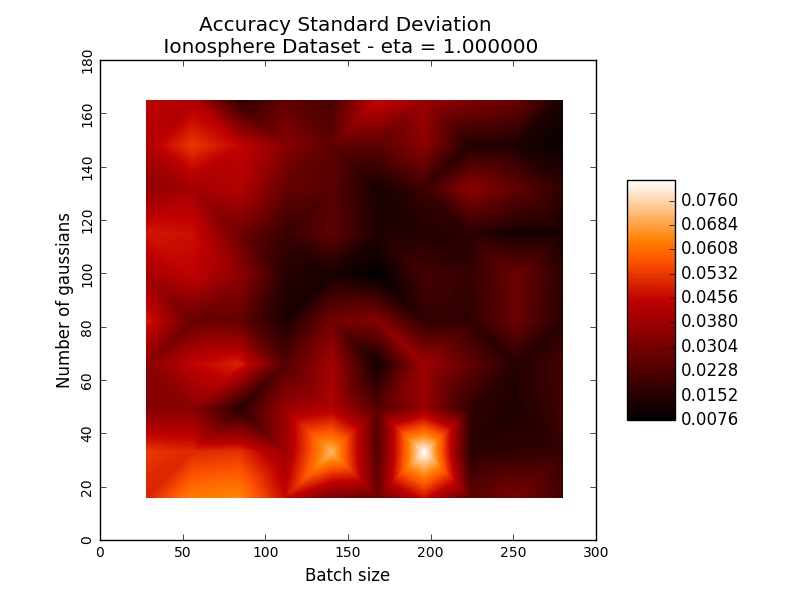}
        \label{fig:apacciono100std}
    \end{subfigure}
    \vspace{-0.25cm}
    \caption*{\vspace{-0.1cm} Source: Author. }
    \label{fig:apacciono100}
\end{figure}

\begin{figure} [!ht]
    \centering
    \caption{\vspace{-0.1cm} AUCROC mean and standard deviation measurements for the VFNN on the Ionosphere dataset with eta = 0.01. }
    \begin{subfigure}[b]{0.45\textwidth}
        \includegraphics[width=\textwidth]{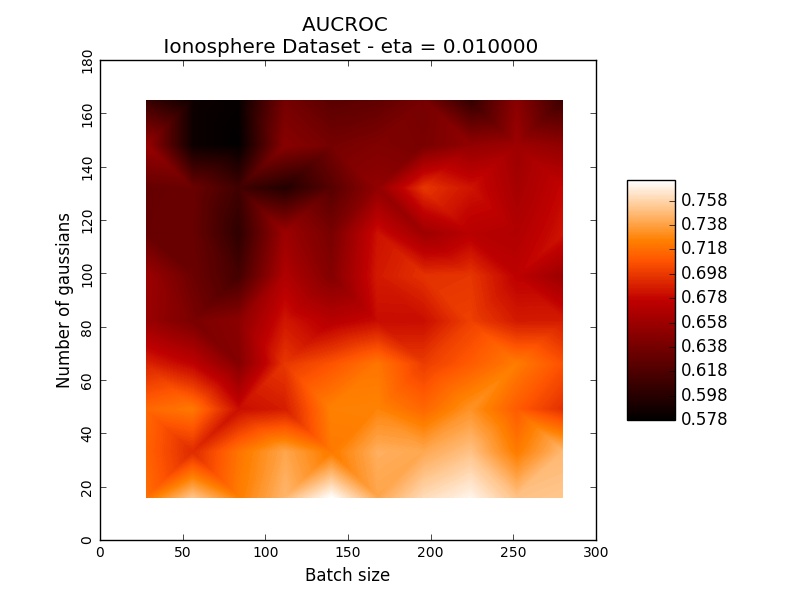}
        \label{fig:apauciono001mean}
    \end{subfigure}
    \begin{subfigure}[b]{0.45\textwidth}
        \includegraphics[width=\textwidth]{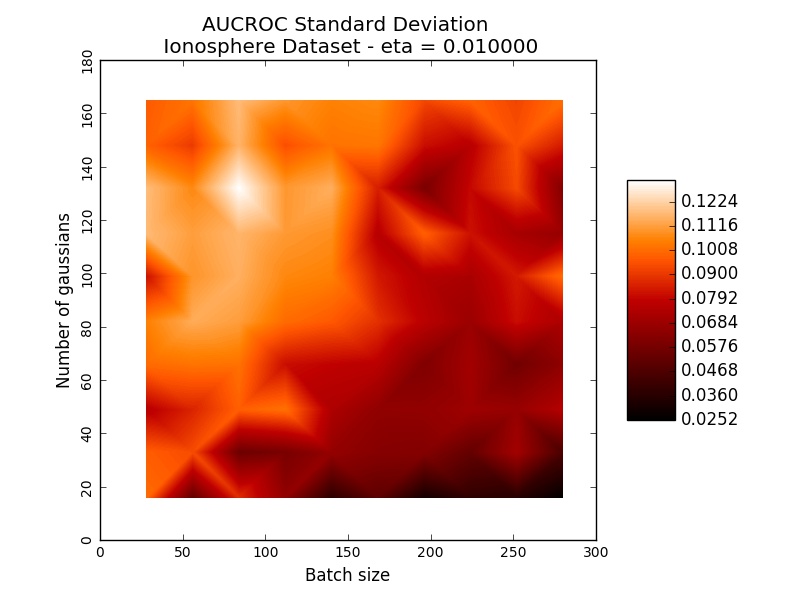}
        \label{fig:apauciono001std}
    \end{subfigure}
    \vspace{-0.25cm}
    \caption*{\vspace{-0.1cm} Source: Author. }
    \label{fig:apauciono001}
\end{figure}

\begin{figure} [!ht]
    \centering
    \caption{\vspace{-0.1cm} AUCROC mean and standard deviation measurements for the VFNN on the Ionosphere dataset with eta = 0.2575. }
    \begin{subfigure}[b]{0.45\textwidth}
        \includegraphics[width=\textwidth]{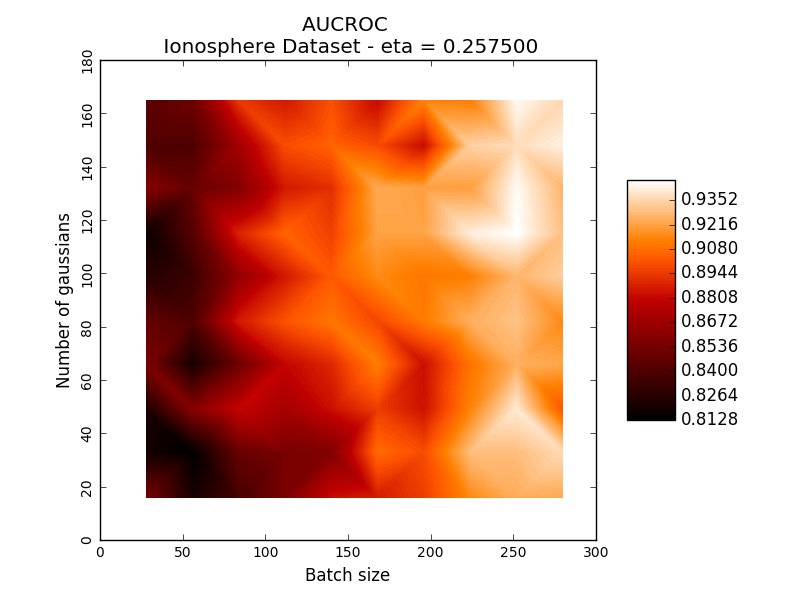}
        \label{fig:apauciono025mean}
    \end{subfigure}
    \begin{subfigure}[b]{0.45\textwidth}
        \includegraphics[width=\textwidth]{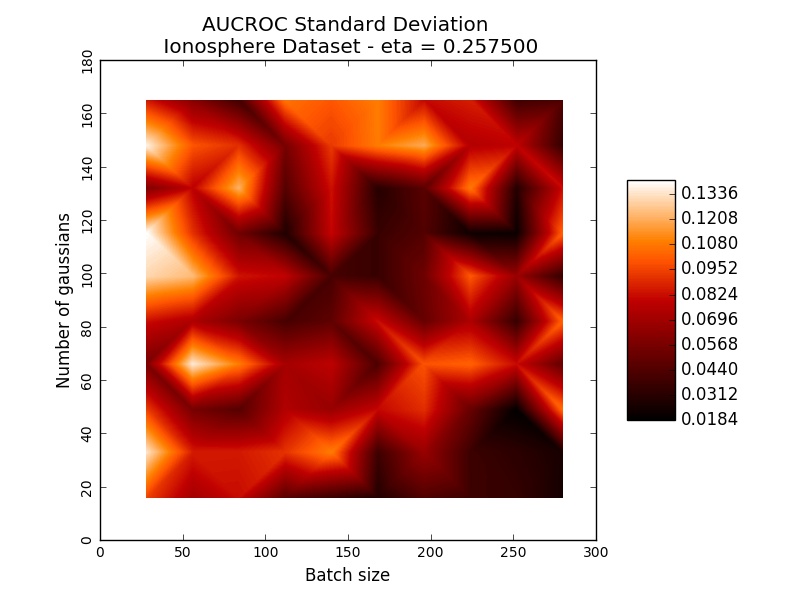}
        \label{fig:apauciono025std}
    \end{subfigure}
    \vspace{-0.25cm}
    \caption*{\vspace{-0.1cm} Source: Author. }
    \label{fig:apauciono025}
\end{figure}

\begin{figure} [!ht]
    \centering
    \caption{\vspace{-0.1cm} AUCROC mean and standard deviation measurements for the VFNN on the Ionosphere dataset with eta = 0.505. }
    \begin{subfigure}[b]{0.45\textwidth}
        \includegraphics[width=\textwidth]{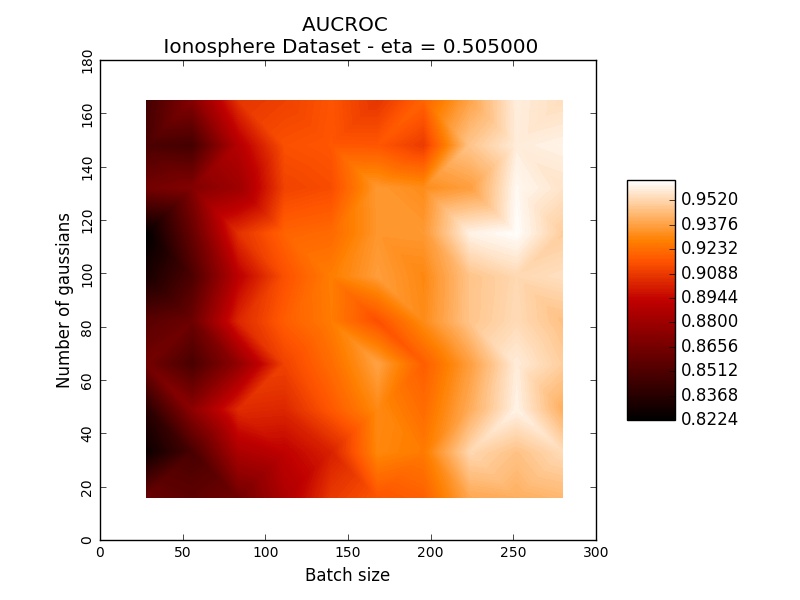}
        \label{fig:apauciono0505mean}
    \end{subfigure}
    \begin{subfigure}[b]{0.45\textwidth}
        \includegraphics[width=\textwidth]{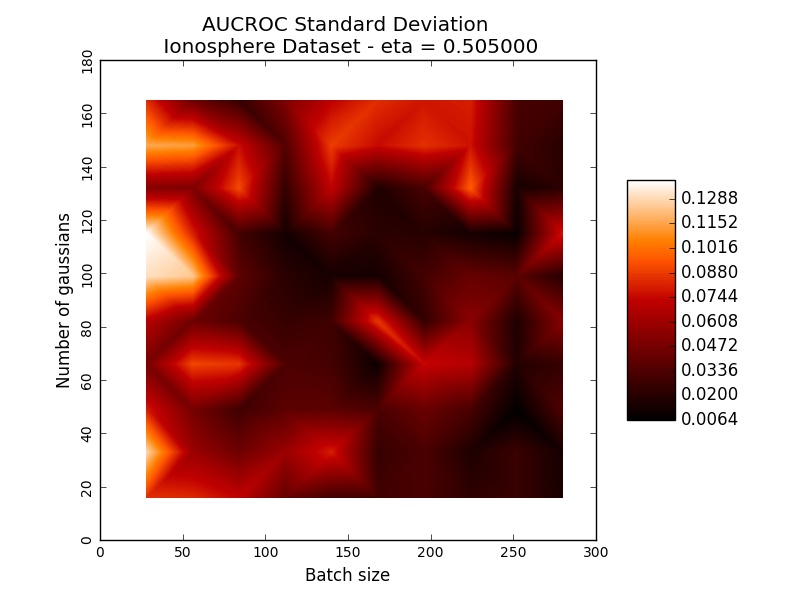}
        \label{fig:apauciono0505std}
    \end{subfigure}
    \vspace{-0.25cm}
    \caption*{\vspace{-0.1cm} Source: Author. }
    \label{fig:apauciono0505}
\end{figure}

\begin{figure} [!ht]
    \centering
    \caption{\vspace{-0.1cm} AUCROC mean and standard deviation measurements for the VFNN on the Ionosphere dataset with eta = 0.7525. }
    \begin{subfigure}[b]{0.45\textwidth}
        \includegraphics[width=\textwidth]{Figures/HyperparametersResults/AUCROCIonosphere0752500.jpg}
        \label{fig:apauciono075mean}
    \end{subfigure}
    \begin{subfigure}[b]{0.45\textwidth}
        \includegraphics[width=\textwidth]{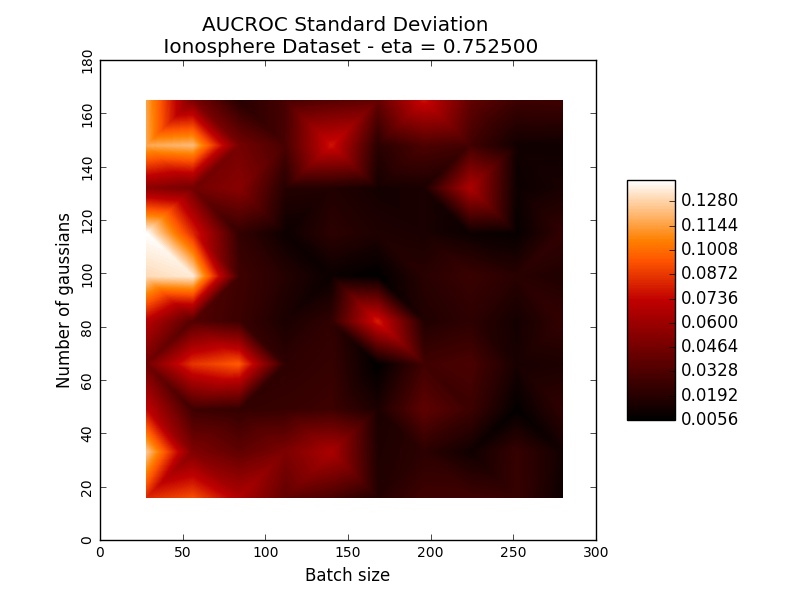}
        \label{fig:apauciono075std}
    \end{subfigure}
    \vspace{-0.25cm}
    \caption*{\vspace{-0.1cm} Source: Author. }
    \label{fig:apauciono075}
\end{figure}

\begin{figure} [!ht]
    \centering
    \caption{\vspace{-0.1cm} AUCROC mean and standard deviation measurements for the VFNN on the Ionosphere dataset with eta = 1.0. }
    \begin{subfigure}[b]{0.45\textwidth}
        \includegraphics[width=\textwidth]{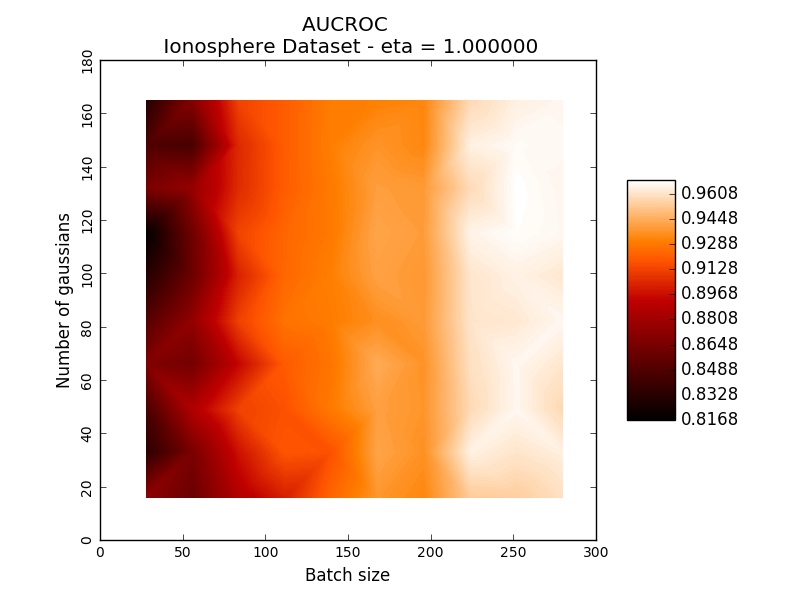}
        \label{fig:apauciono100mean}
    \end{subfigure}
    \begin{subfigure}[b]{0.45\textwidth}
        \includegraphics[width=\textwidth]{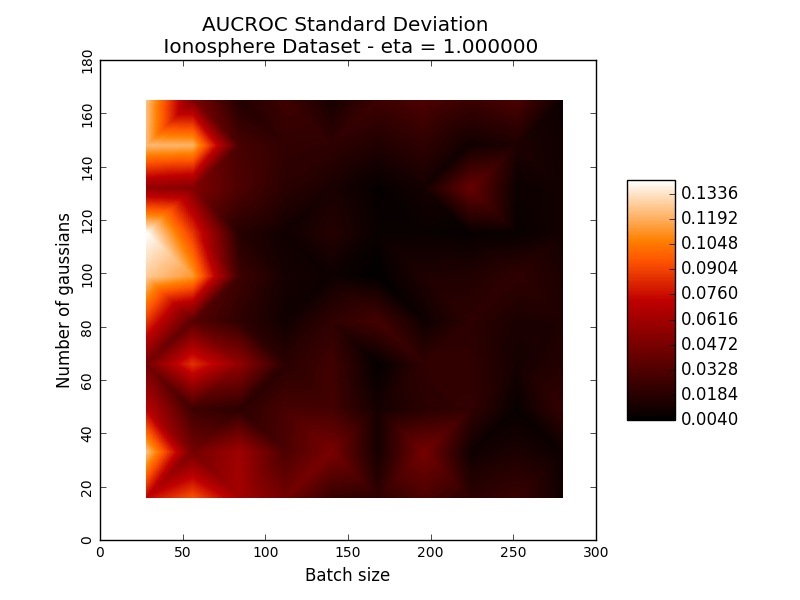}
        \label{fig:apauciono100std}
    \end{subfigure}
    \vspace{-0.25cm}
    \caption*{\vspace{-0.1cm} Source: Author. }
    \label{fig:apauciono100}
\end{figure}

\begin{figure} [!ht]
    \centering
    \caption{\vspace{-0.1cm} Mean Squared Error and standard deviation measurements for the VFNN on the Ionosphere dataset with eta = 0.01. }
    \begin{subfigure}[b]{0.45\textwidth}
        \includegraphics[width=\textwidth]{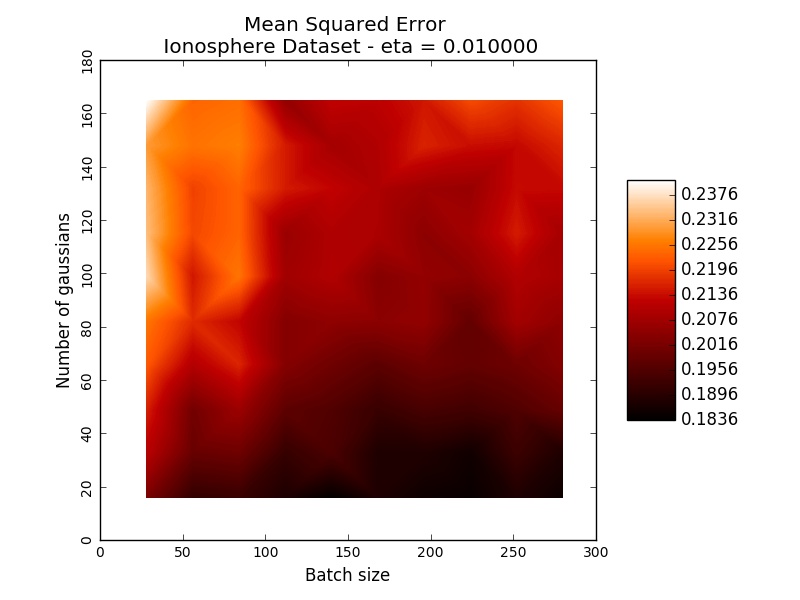}
        \label{fig:apmseiono001mean}
    \end{subfigure}
    \begin{subfigure}[b]{0.45\textwidth}
        \includegraphics[width=\textwidth]{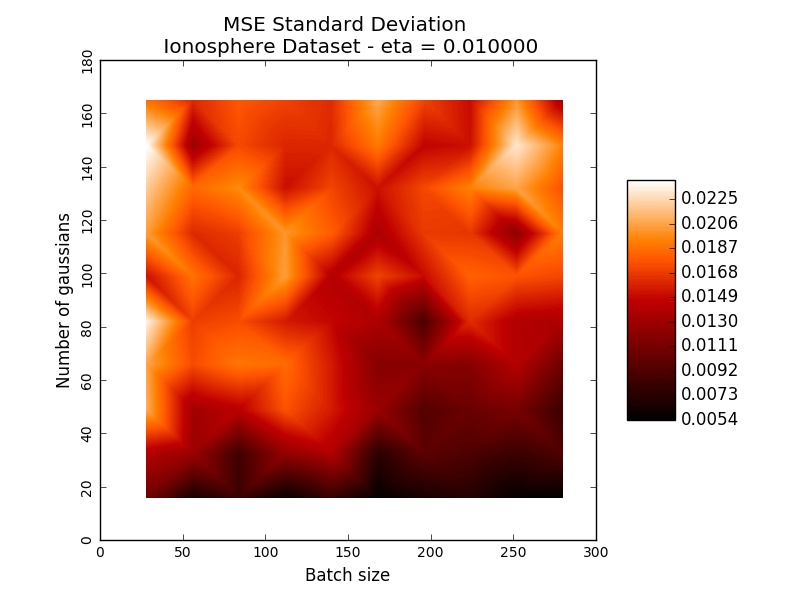}
        \label{fig:apmseiono001std}
    \end{subfigure}
    \vspace{-0.25cm}
    \caption*{\vspace{-0.1cm} Source: Author. }
    \label{fig:apmseiono001}
\end{figure}

\begin{figure} [!ht]
    \centering
    \caption{\vspace{-0.1cm} Mean Squared Error and standard deviation measurements for the VFNN on the Ionosphere dataset with eta = 0.2575. }
    \begin{subfigure}[b]{0.45\textwidth}
        \includegraphics[width=\textwidth]{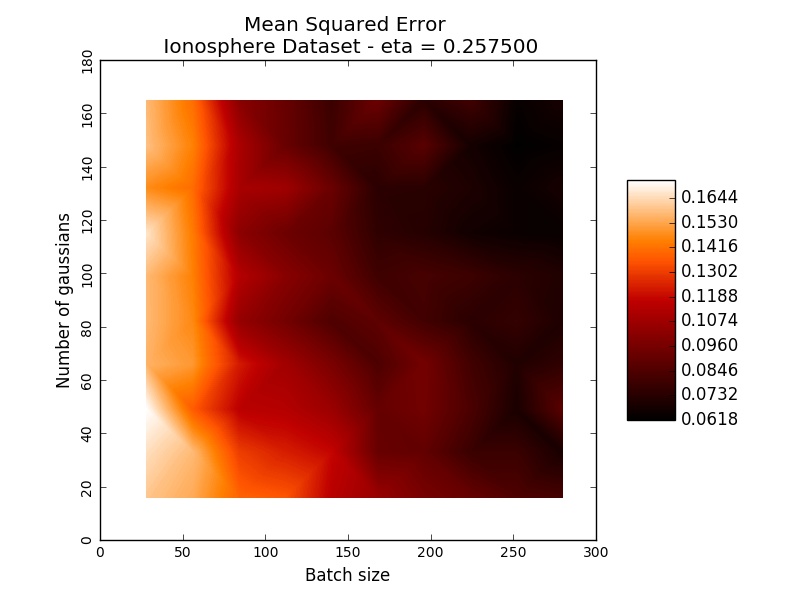}
        \label{fig:apmseiono025mean}
    \end{subfigure}
    \begin{subfigure}[b]{0.45\textwidth}
        \includegraphics[width=\textwidth]{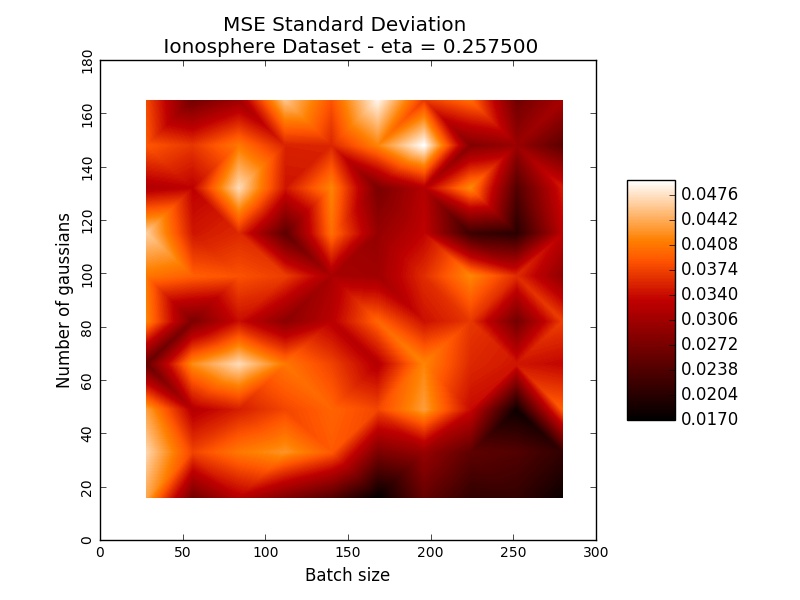}
        \label{fig:apmseiono025std}
    \end{subfigure}
    \vspace{-0.25cm}
    \caption*{\vspace{-0.1cm} Source: Author. }
    \label{fig:apmseiono025}
\end{figure}

\begin{figure} [!ht]
    \centering
    \caption{\vspace{-0.1cm} Mean Squared Error and standard deviation measurements for the VFNN on the Ionosphere dataset with eta = 0.505. }
    \begin{subfigure}[b]{0.45\textwidth}
        \includegraphics[width=\textwidth]{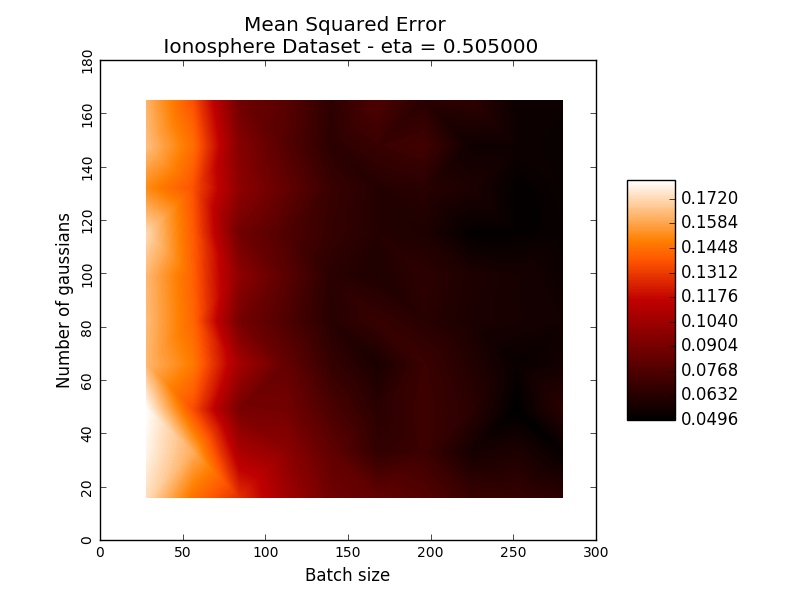}
        \label{fig:apmseiono0505mean}
    \end{subfigure}
    \begin{subfigure}[b]{0.45\textwidth}
        \includegraphics[width=\textwidth]{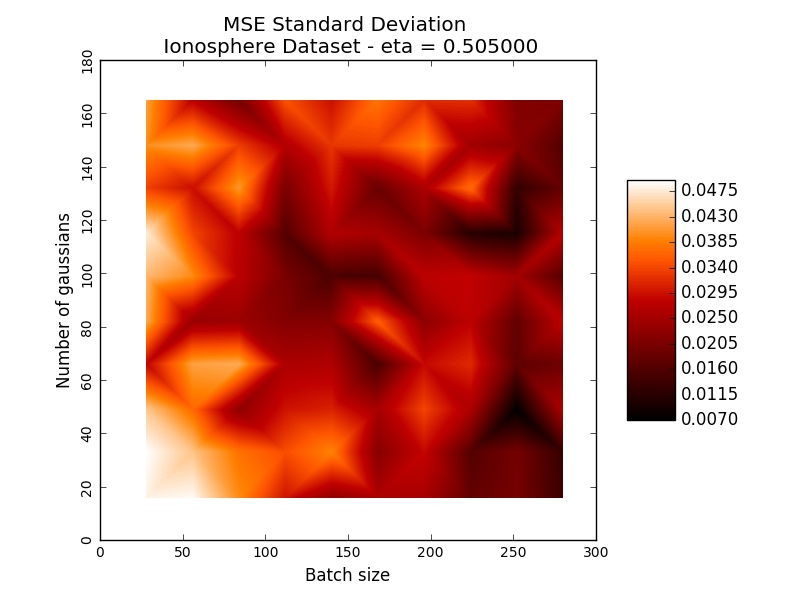}
        \label{fig:apmseiono0505std}
    \end{subfigure}
    \vspace{-0.25cm}
    \caption*{\vspace{-0.1cm} Source: Author. }
    \label{fig:apmseiono0505}
\end{figure}

\begin{figure} [!ht]
    \centering
    \caption{\vspace{-0.1cm} Mean Squared Error and standard deviation measurements for the VFNN on the Ionosphere dataset with eta = 0.7525. }
    \begin{subfigure}[b]{0.45\textwidth}
        \includegraphics[width=\textwidth]{Figures/HyperparametersResults/MeanSquaredErrorIonosphere0752500.jpg}
        \label{fig:apmseiono075mean}
    \end{subfigure}
    \begin{subfigure}[b]{0.45\textwidth}
        \includegraphics[width=\textwidth]{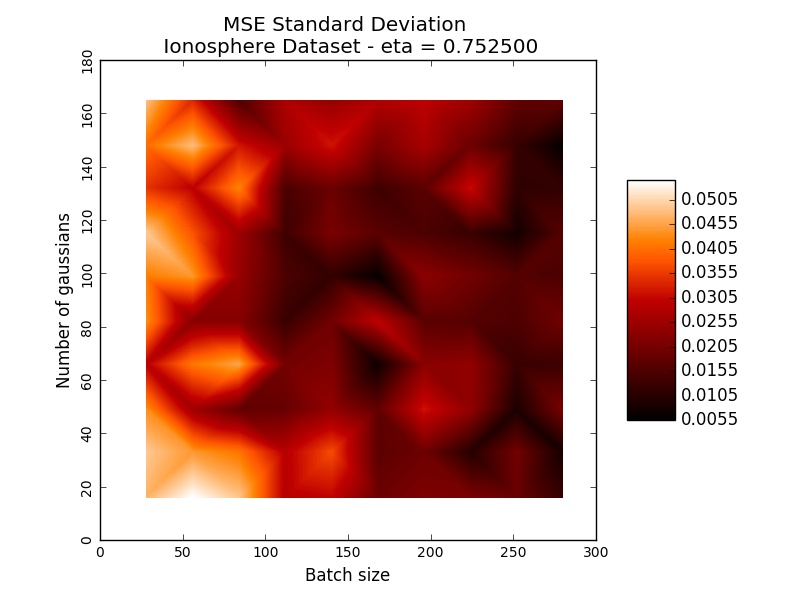}
        \label{fig:apmseiono075std}
    \end{subfigure}
    \vspace{-0.25cm}
    \caption*{\vspace{-0.1cm} Source: Author. }
    \label{fig:apmseiono075}
\end{figure}

\begin{figure} [!ht]
    \centering
    \caption{\vspace{-0.1cm} Mean Squared Error and standard deviation measurements for the VFNN on the Ionosphere dataset with eta = 1.0. }
    \begin{subfigure}[b]{0.45\textwidth}
        \includegraphics[width=\textwidth]{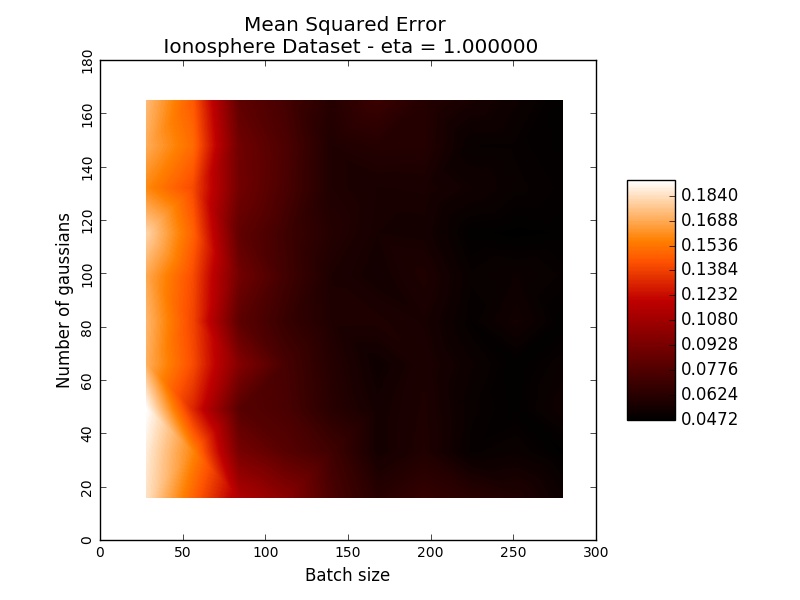}
        \label{fig:apmseiono100mean}
    \end{subfigure}
    \begin{subfigure}[b]{0.45\textwidth}
        \includegraphics[width=\textwidth]{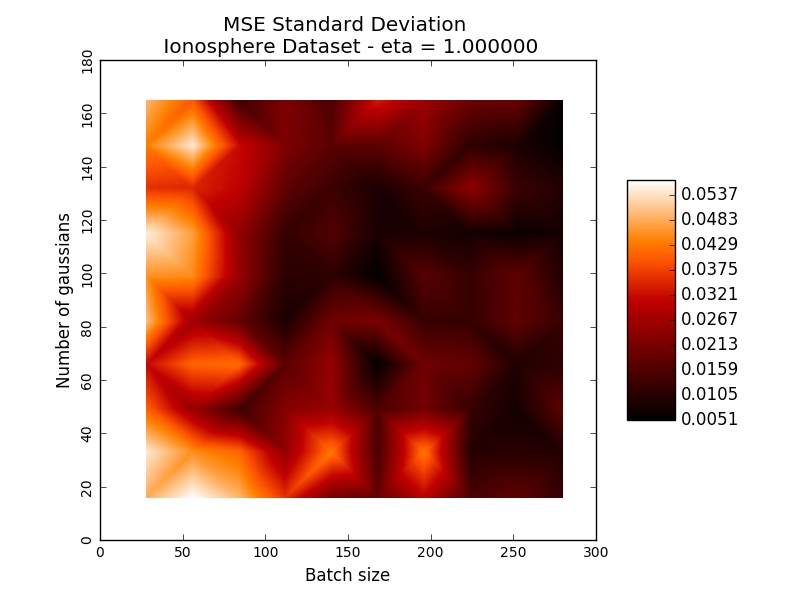}
        \label{fig:apmseiono100std}
    \end{subfigure}
    \vspace{-0.25cm}
    \caption*{\vspace{-0.1cm} Source: Author. }
    \label{fig:apmseiono0100}
\end{figure}

\chapter{Cryotherapy Hyperparameter Results}
\label{appendicecryotherapy}

\begin{figure} [!ht]
    \centering
    \caption{\vspace{-0.1cm} Accuracy mean and standard deviation measurements for the VFNN on the Cryotherapy dataset with eta = 0.01. }
    \begin{subfigure}[b]{0.45\textwidth}
        \includegraphics[width=\textwidth]{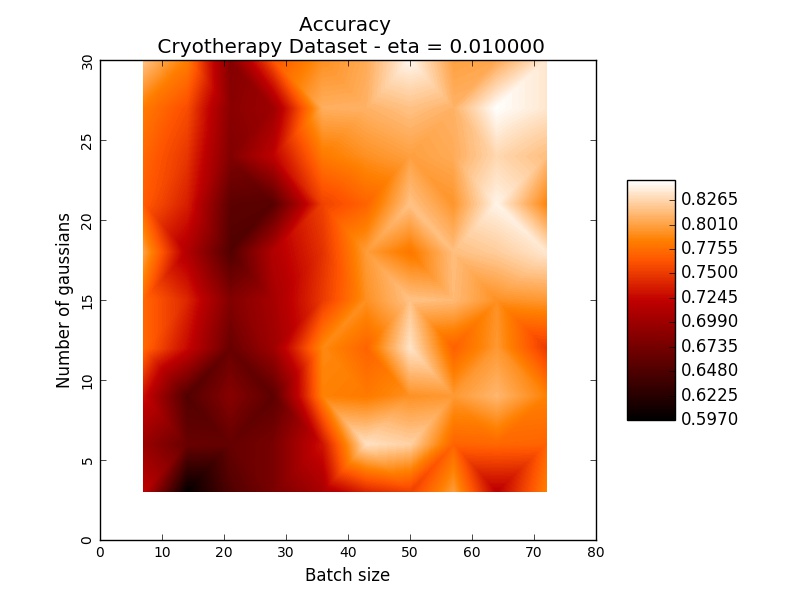}
        \label{fig:apacccryo001mean}
    \end{subfigure}
    \begin{subfigure}[b]{0.45\textwidth}
        \includegraphics[width=\textwidth]{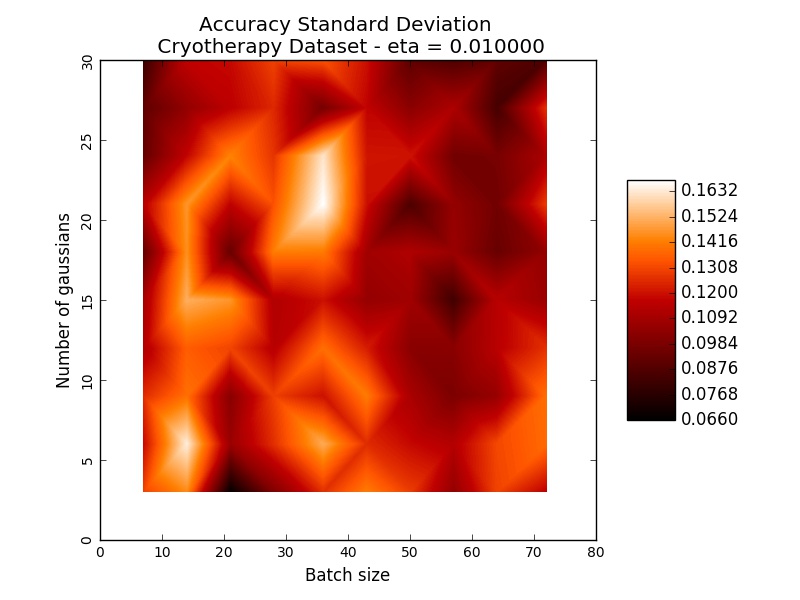}
        \label{fig:apacccryo001std}
    \end{subfigure}
    \vspace{-0.25cm}
    \caption*{\vspace{-0.1cm} Source: Author. }
    \label{fig:apacccryo001}
\end{figure}

\begin{figure} [!ht]
    \centering
    \caption{\vspace{-0.1cm} Accuracy mean and standard deviation measurements for the VFNN on the Cryotherapy dataset with eta = 0.2575. }
    \begin{subfigure}[b]{0.45\textwidth}
        \includegraphics[width=\textwidth]{Figures/HyperparametersResults/AccuracyCryotherapy0257500.jpg}
        \label{fig:apacccryo025mean}
    \end{subfigure}
    \begin{subfigure}[b]{0.45\textwidth}
        \includegraphics[width=\textwidth]{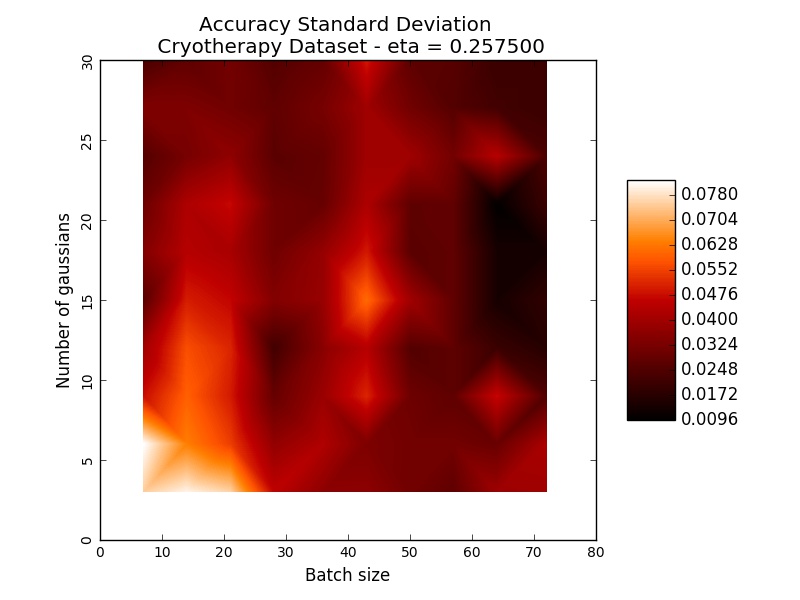}
        \label{fig:apacccryo025std}
    \end{subfigure}
    \vspace{-0.25cm}
    \caption*{\vspace{-0.1cm} Source: Author. }
    \label{fig:apacccryo025}
\end{figure}

\begin{figure} [!ht]
    \centering
    \caption{\vspace{-0.1cm} Accuracy mean and standard deviation measurements for the VFNN on the Cryotherapy dataset with eta = 0.505. }
    \begin{subfigure}[b]{0.45\textwidth}
        \includegraphics[width=\textwidth]{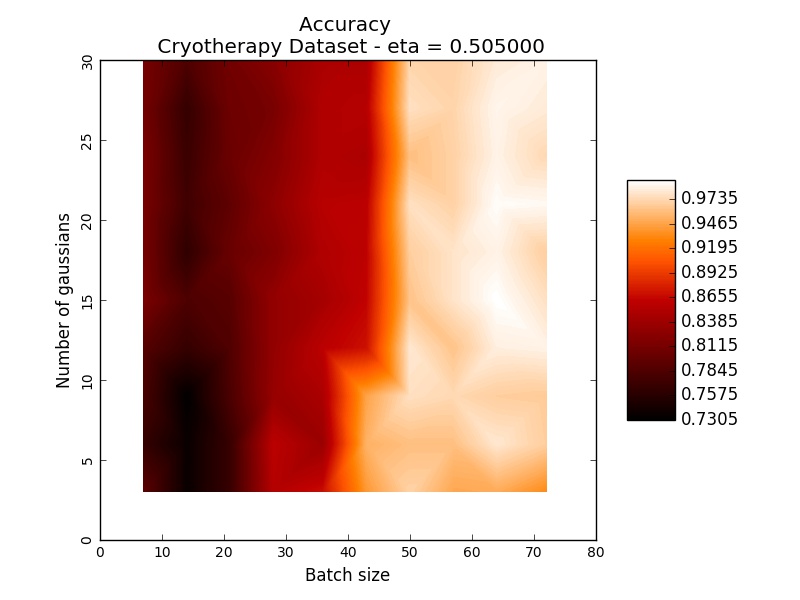}
        \label{fig:apacccryo0505mean}
    \end{subfigure}
    \begin{subfigure}[b]{0.45\textwidth}
        \includegraphics[width=\textwidth]{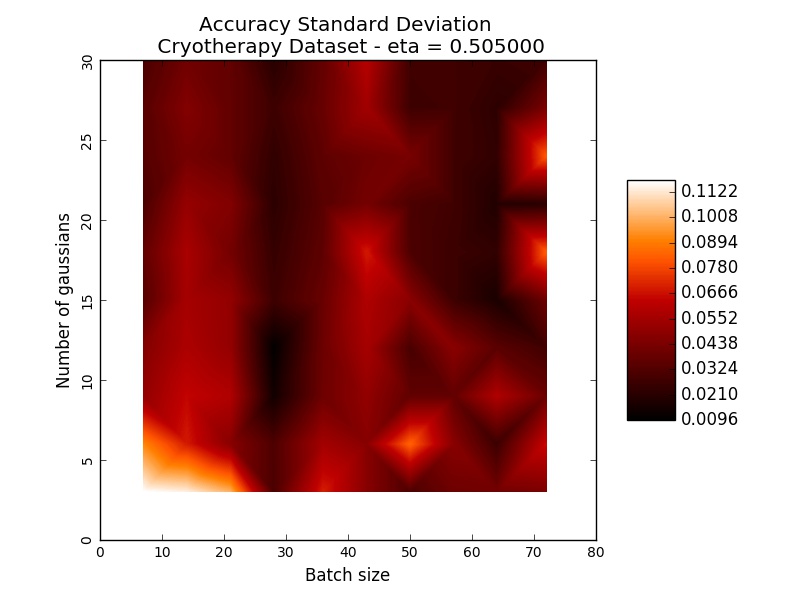}
        \label{fig:apacccryo0505std}
    \end{subfigure}
    \vspace{-0.25cm}
    \caption*{\vspace{-0.1cm} Source: Author. }
    \label{fig:apacccryo505}
\end{figure}

\begin{figure} [!ht]
    \centering
    \caption{\vspace{-0.1cm} Accuracy mean and standard deviation measurements for the VFNN on the Cryotherapy dataset with eta = 0.7525. }
    \begin{subfigure}[b]{0.45\textwidth}
        \includegraphics[width=\textwidth]{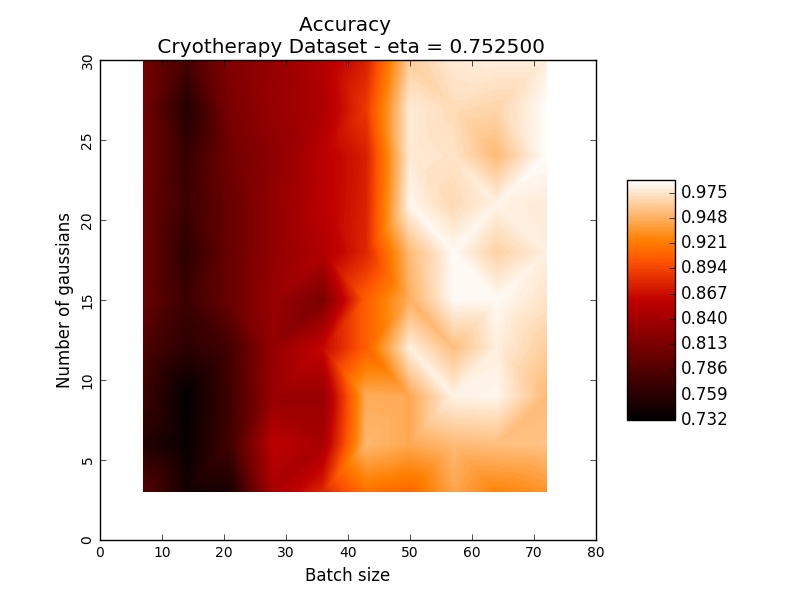}
        \label{fig:apacccryo075mean}
    \end{subfigure}
    \begin{subfigure}[b]{0.45\textwidth}
        \includegraphics[width=\textwidth]{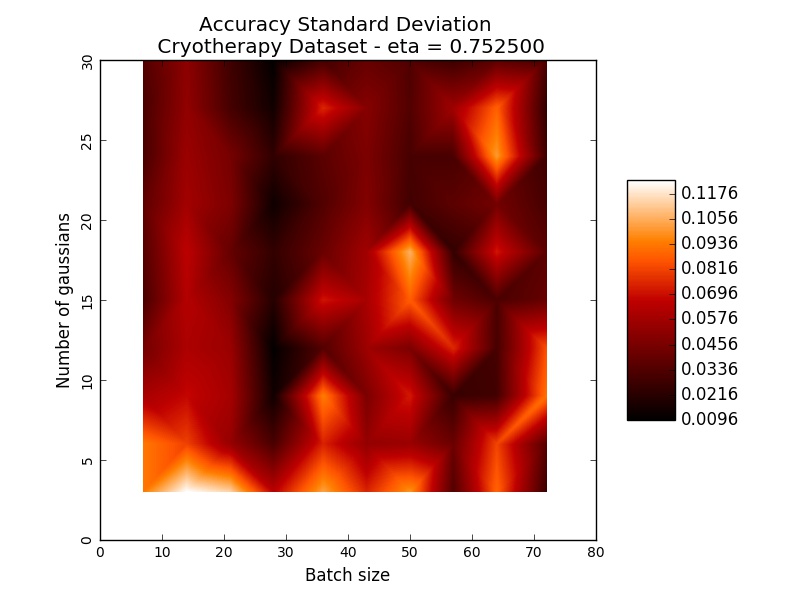}
        \label{fig:apacccryo075std}
    \end{subfigure}
    \vspace{-0.25cm}
    \caption*{\vspace{-0.1cm} Source: Author. }
    \label{fig:apacccryo75}
\end{figure}

\begin{figure} [!ht]
    \centering
    \caption{\vspace{-0.1cm} Accuracy mean and standard deviation measurements for the VFNN on the Cryotherapy dataset with eta = 1.0. }
    \begin{subfigure}[b]{0.45\textwidth}
        \includegraphics[width=\textwidth]{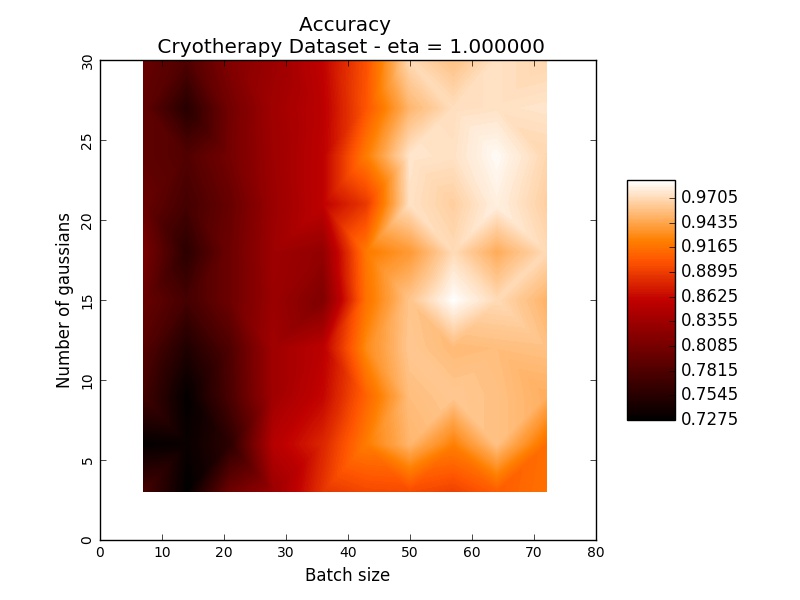}
        \label{fig:apacccryo100mean}
    \end{subfigure}
    \begin{subfigure}[b]{0.45\textwidth}
        \includegraphics[width=\textwidth]{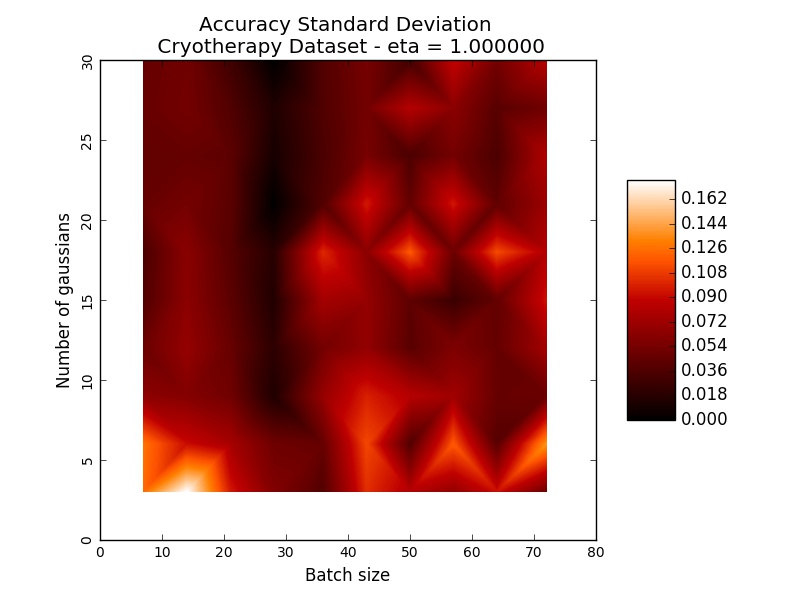}
        \label{fig:apacccryo100std}
    \end{subfigure}
    \vspace{-0.25cm}
    \caption*{\vspace{-0.1cm} Source: Author. }
    \label{fig:apacccryo100}
\end{figure}

\begin{figure} [!ht]
    \centering
    \caption{\vspace{-0.1cm} AUCROC mean and standard deviation measurements for the VFNN on the Cryotherapy dataset with eta = 0.01. }
    \begin{subfigure}[b]{0.45\textwidth}
        \includegraphics[width=\textwidth]{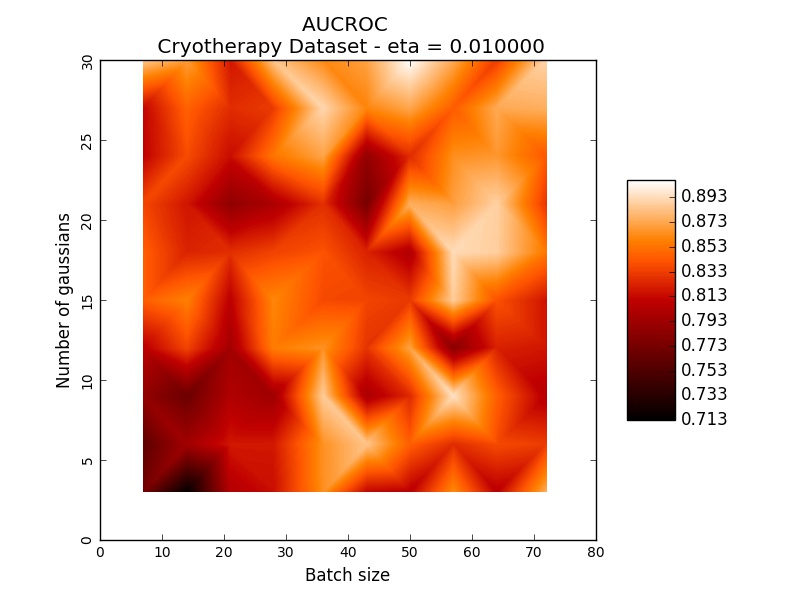}
        \label{fig:apauccryo001mean}
    \end{subfigure}
    \begin{subfigure}[b]{0.45\textwidth}
        \includegraphics[width=\textwidth]{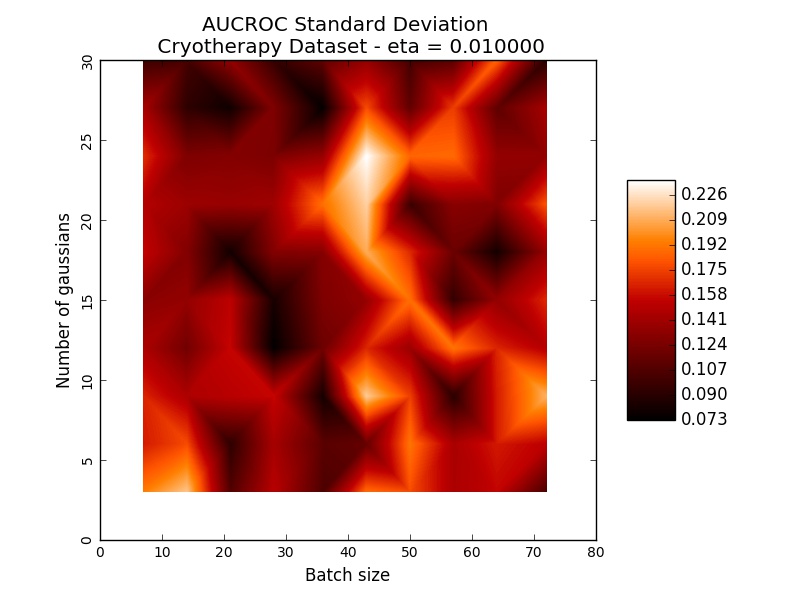}
        \label{fig:apauccryo001std}
    \end{subfigure}
    \vspace{-0.25cm}
    \caption*{\vspace{-0.1cm} Source: Author. }
    \label{fig:apauccryo001}
\end{figure}

\begin{figure} [!ht]
    \centering
    \caption{\vspace{-0.1cm} AUCROC mean and standard deviation measurements for the VFNN on the Cryotherapy dataset with eta = 0.2575. }
    \begin{subfigure}[b]{0.45\textwidth}
        \includegraphics[width=\textwidth]{Figures/HyperparametersResults/AUCROCCryotherapy0257500.jpg}
        \label{fig:apauccryo025mean}
    \end{subfigure}
    \begin{subfigure}[b]{0.45\textwidth}
        \includegraphics[width=\textwidth]{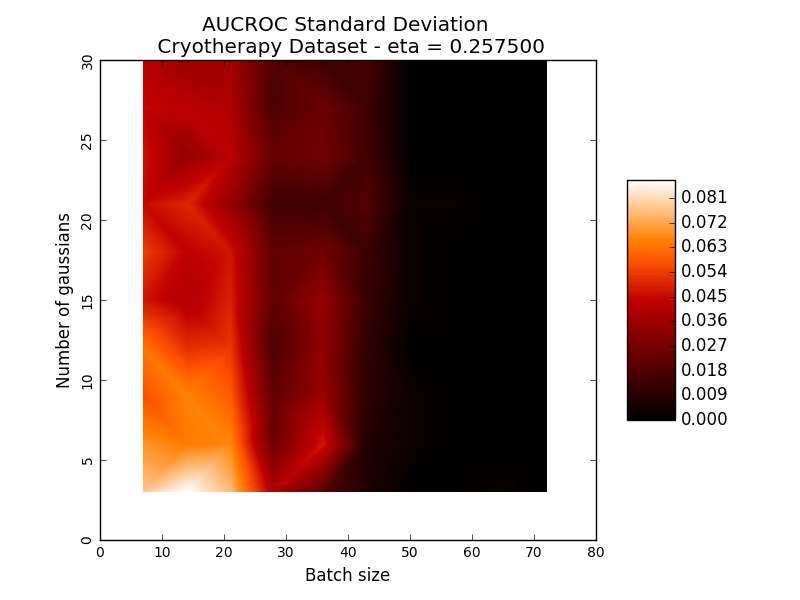}
        \label{fig:apauccryo025std}
    \end{subfigure}
    \vspace{-0.25cm}
    \caption*{\vspace{-0.1cm} Source: Author. }
    \label{fig:apauccryo025}
\end{figure}

\begin{figure} [!ht]
    \centering
    \caption{\vspace{-0.1cm} AUCROC mean and standard deviation measurements for the VFNN on the Cryotherapy dataset with eta = 0.505. }
    \begin{subfigure}[b]{0.45\textwidth}
        \includegraphics[width=\textwidth]{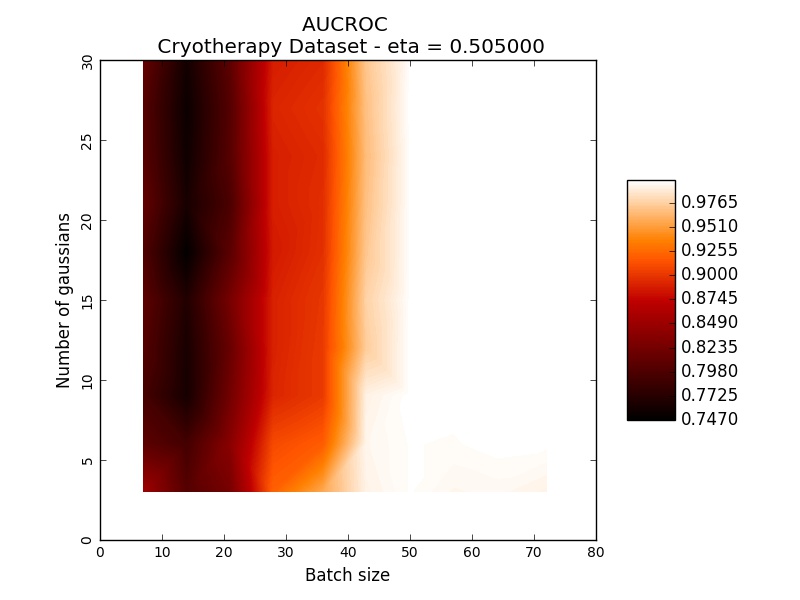}
        \label{fig:apauccryo0505mean}
    \end{subfigure}
    \begin{subfigure}[b]{0.45\textwidth}
        \includegraphics[width=\textwidth]{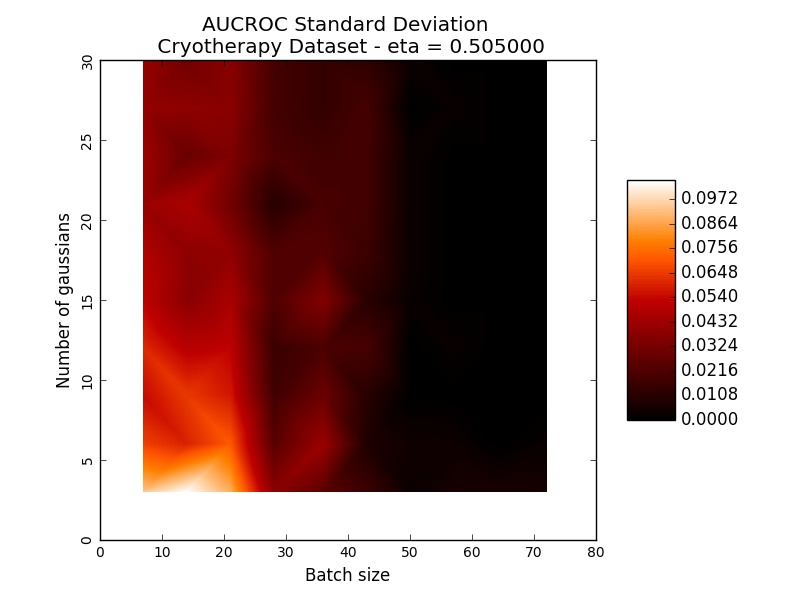}
        \label{fig:apauccryo0505std}
    \end{subfigure}
    \vspace{-0.25cm}
    \caption*{\vspace{-0.1cm} Source: Author. }
    \label{fig:apauccryo0505}
\end{figure}

\begin{figure} [!ht]
    \centering
    \caption{\vspace{-0.1cm} AUCROC mean and standard deviation measurements for the VFNN on the Cryotherapy dataset with eta = 0.7525. }
    \begin{subfigure}[b]{0.45\textwidth}
        \includegraphics[width=\textwidth]{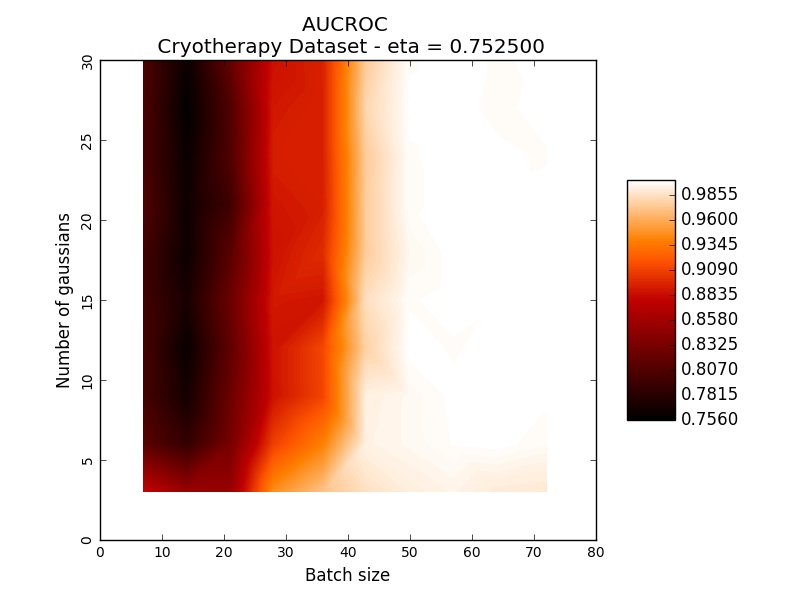}
        \label{fig:apauccryo075mean}
    \end{subfigure}
    \begin{subfigure}[b]{0.45\textwidth}
        \includegraphics[width=\textwidth]{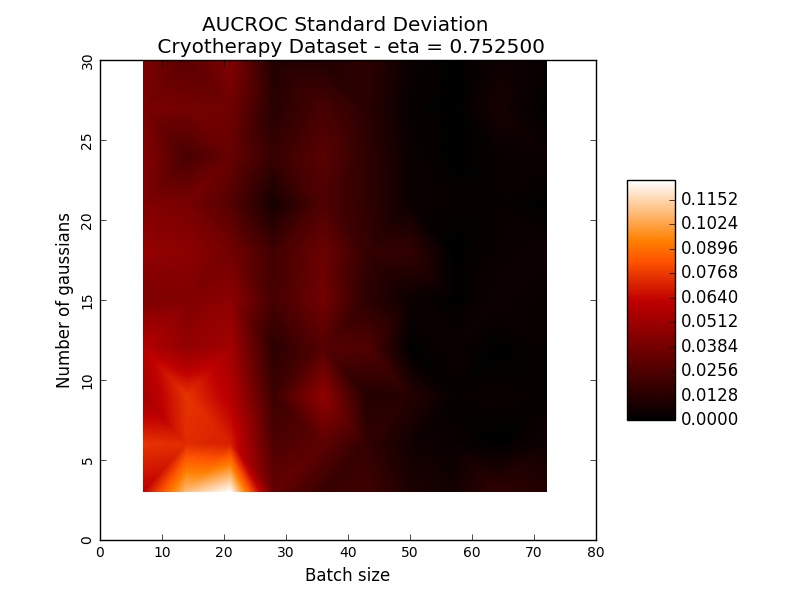}
        \label{fig:apauccryo075std}
    \end{subfigure}
    \vspace{-0.25cm}
    \caption*{\vspace{-0.1cm} Source: Author. }
    \label{fig:apauccryo075}
\end{figure}

\begin{figure} [!ht]
    \centering
    \caption{\vspace{-0.1cm} AUCROC mean and standard deviation measurements for the VFNN on the Cryotherapy dataset with eta = 1.0. }
    \begin{subfigure}[b]{0.45\textwidth}
        \includegraphics[width=\textwidth]{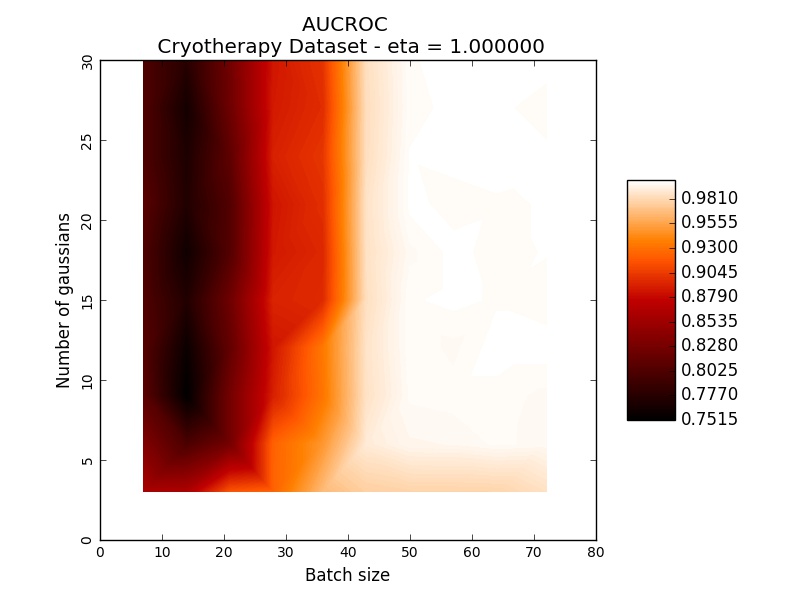}
        \label{fig:apauccryo100mean}
    \end{subfigure}
    \begin{subfigure}[b]{0.45\textwidth}
        \includegraphics[width=\textwidth]{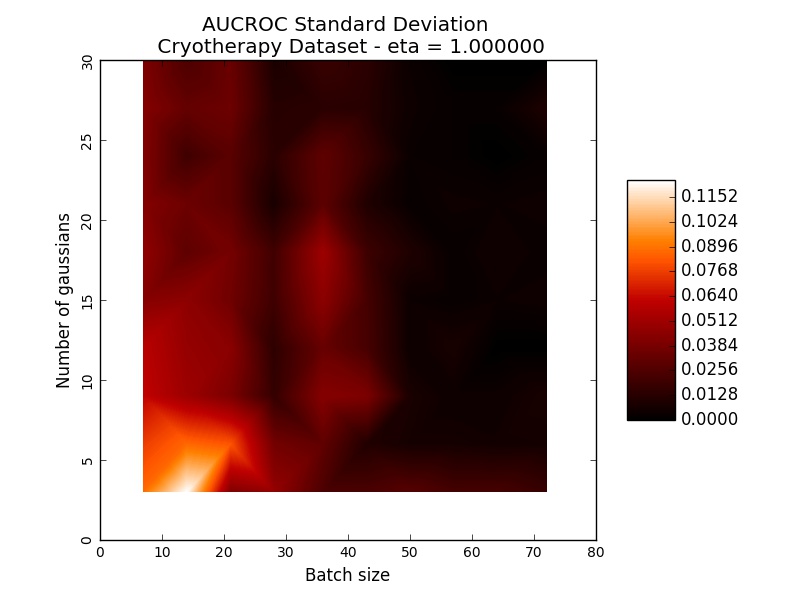}
        \label{fig:apauccryo100std}
    \end{subfigure}
    \vspace{-0.25cm}
    \caption*{\vspace{-0.1cm} Source: Author. }
    \label{fig:apauccryo100}
\end{figure}

\begin{figure} [!ht]
    \centering
    \caption{\vspace{-0.1cm} Mean Squared Error and standard deviation measurements for the VFNN on the Cryotherapy dataset with eta = 0.01. }
    \begin{subfigure}[b]{0.45\textwidth}
        \includegraphics[width=\textwidth]{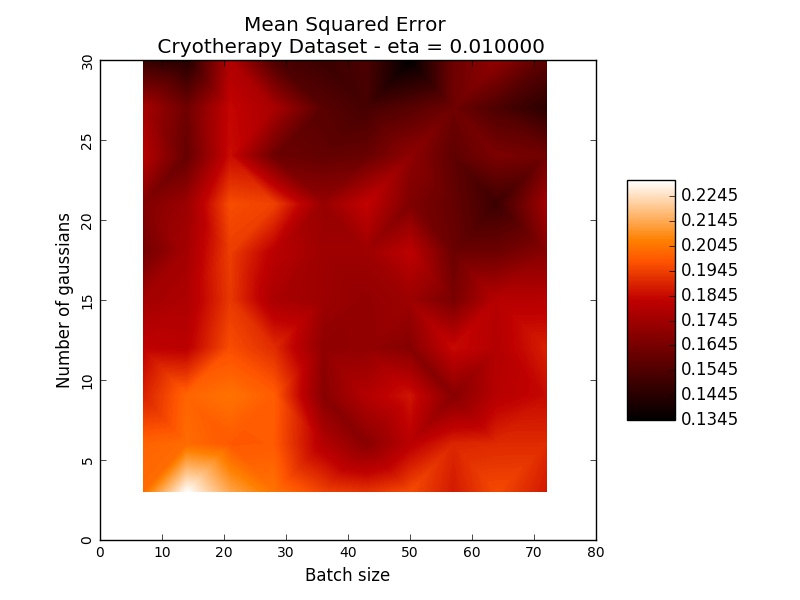}
        \label{fig:apmsecryo001mean}
    \end{subfigure}
    \begin{subfigure}[b]{0.45\textwidth}
        \includegraphics[width=\textwidth]{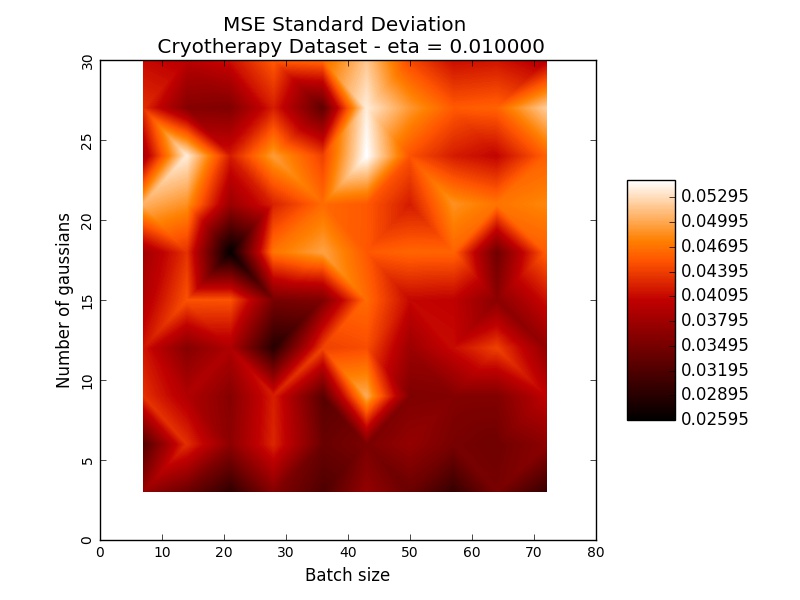}
        \label{fig:apmsecryo001std}
    \end{subfigure}
    \vspace{-0.25cm}
    \caption*{\vspace{-0.1cm} Source: Author. }
    \label{fig:apmsecryo001}
\end{figure}

\begin{figure} [!ht]
    \centering
    \caption{\vspace{-0.1cm} Mean Squared Error and standard deviation measurements for the VFNN on the Cryotherapy dataset with eta = 0.2575. }
    \begin{subfigure}[b]{0.45\textwidth}
        \includegraphics[width=\textwidth]{Figures/HyperparametersResults/MeanSquaredErrorCryotherapy0257500.jpg}
        \label{fig:apmsecryo025mean}
    \end{subfigure}
    \begin{subfigure}[b]{0.45\textwidth}
        \includegraphics[width=\textwidth]{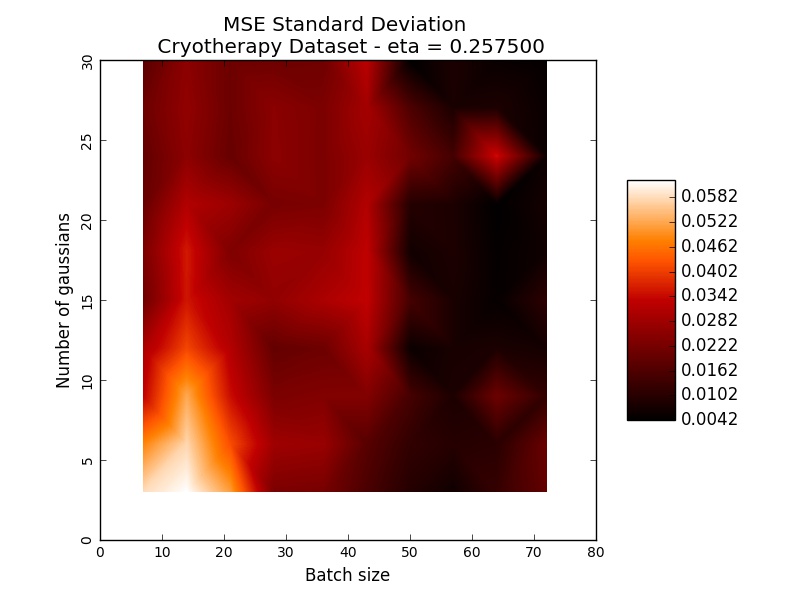}
        \label{fig:apmsecryo025std}
    \end{subfigure}
    \vspace{-0.25cm}
    \caption*{\vspace{-0.1cm} Source: Author. }
    \label{fig:apmsecryo025}
\end{figure}

\begin{figure} [!ht]
    \centering
    \caption{\vspace{-0.1cm} Mean Squared Error and standard deviation measurements for the VFNN on the Cryotherapy dataset with eta = 0.505. }
    \begin{subfigure}[b]{0.45\textwidth}
        \includegraphics[width=\textwidth]{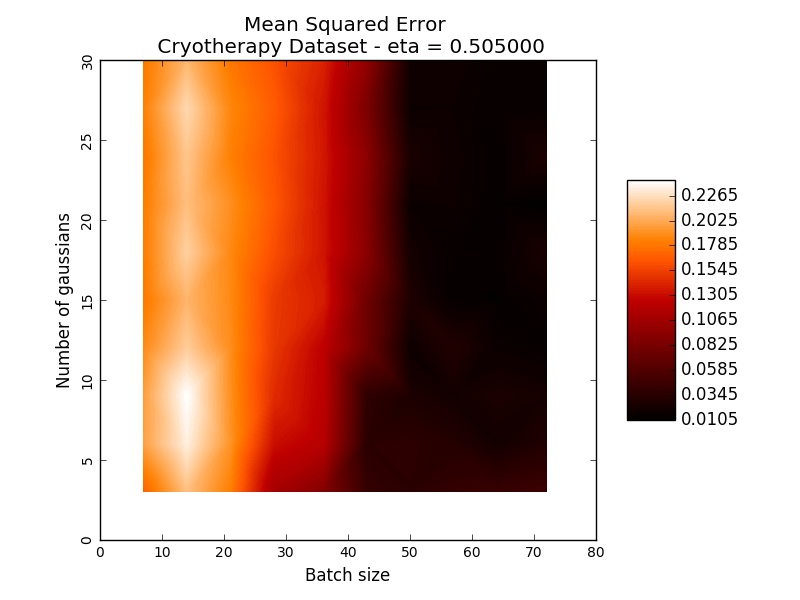}
        \label{fig:apmsecryo0505mean}
    \end{subfigure}
    \begin{subfigure}[b]{0.45\textwidth}
        \includegraphics[width=\textwidth]{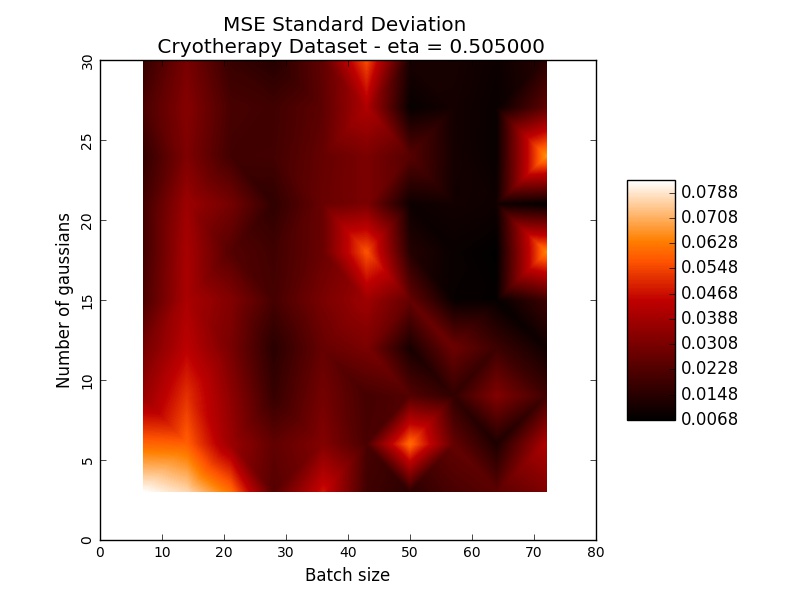}
        \label{fig:apmsecryo0505std}
    \end{subfigure}
    \vspace{-0.25cm}
    \caption*{\vspace{-0.1cm} Source: Author. }
    \label{fig:apmsecryo0505}
\end{figure}

\begin{figure} [!ht]
    \centering
    \caption{\vspace{-0.1cm} Mean Squared Error and standard deviation measurements for the VFNN on the Cryotherapy dataset with eta = 0.7525. }
    \begin{subfigure}[b]{0.45\textwidth}
        \includegraphics[width=\textwidth]{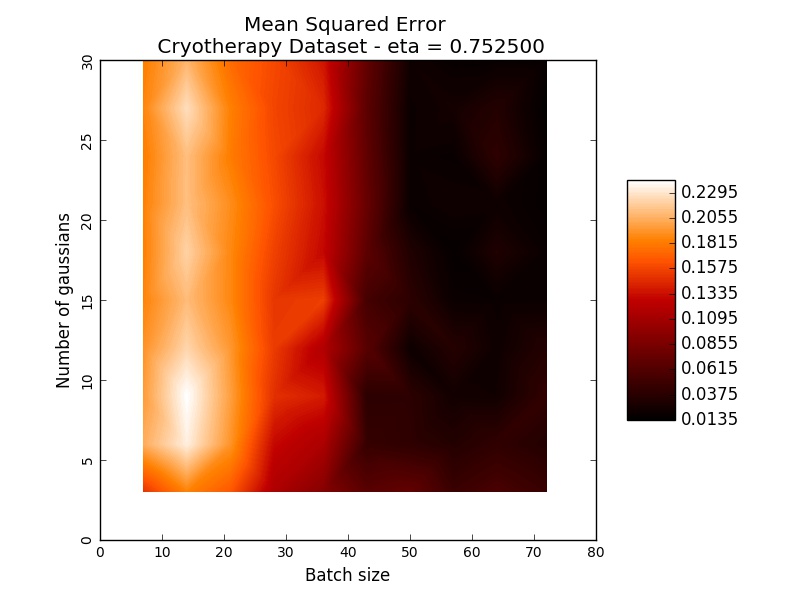}
        \label{fig:apmsecryo075mean}
    \end{subfigure}
    \begin{subfigure}[b]{0.45\textwidth}
        \includegraphics[width=\textwidth]{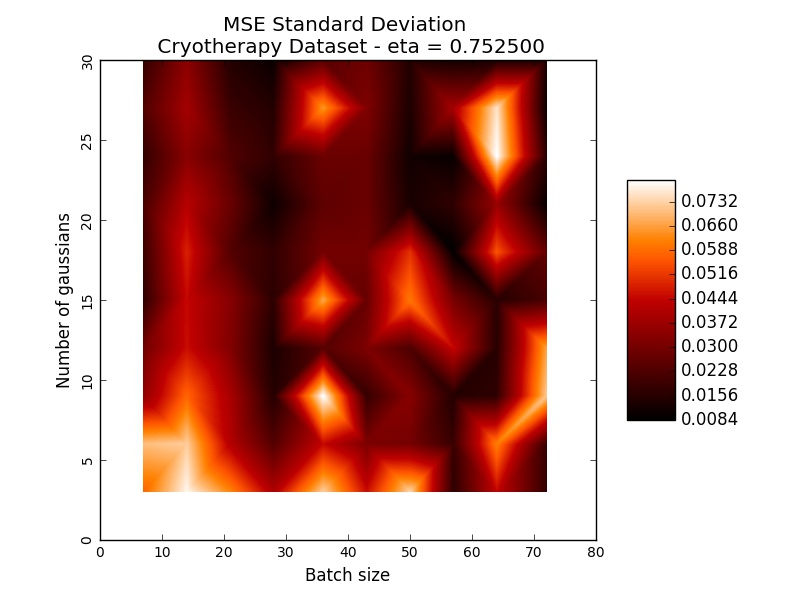}
        \label{fig:apmsecryo075std}
    \end{subfigure}
    \vspace{-0.25cm}
    \caption*{\vspace{-0.1cm} Source: Author. }
    \label{fig:apmsecryo075}
\end{figure}

\begin{figure} [!ht]
    \centering
    \caption{\vspace{-0.1cm} Mean Squared Error and standard deviation measurements for the VFNN on the Cryotherapy dataset with eta = 1.0. }
    \begin{subfigure}[b]{0.45\textwidth}
        \includegraphics[width=\textwidth]{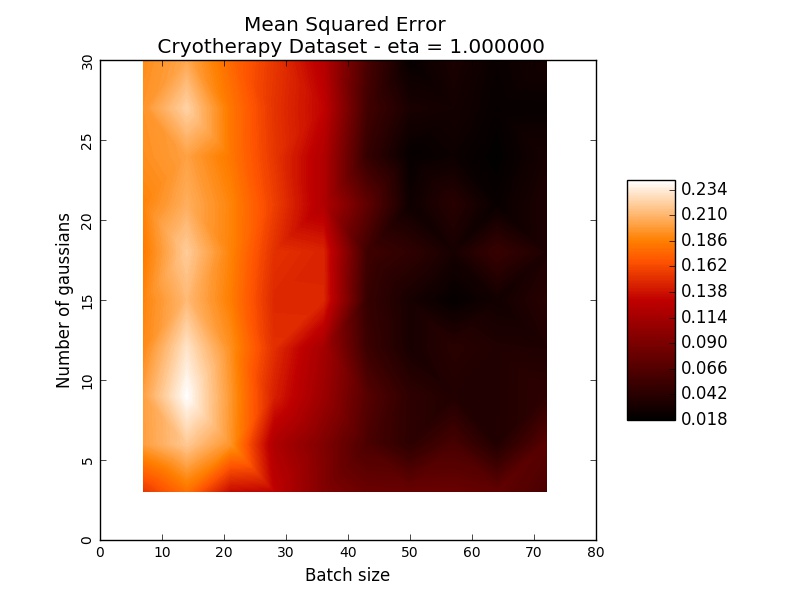}
        \label{fig:apmsecryo100mean}
    \end{subfigure}
    \begin{subfigure}[b]{0.45\textwidth}
        \includegraphics[width=\textwidth]{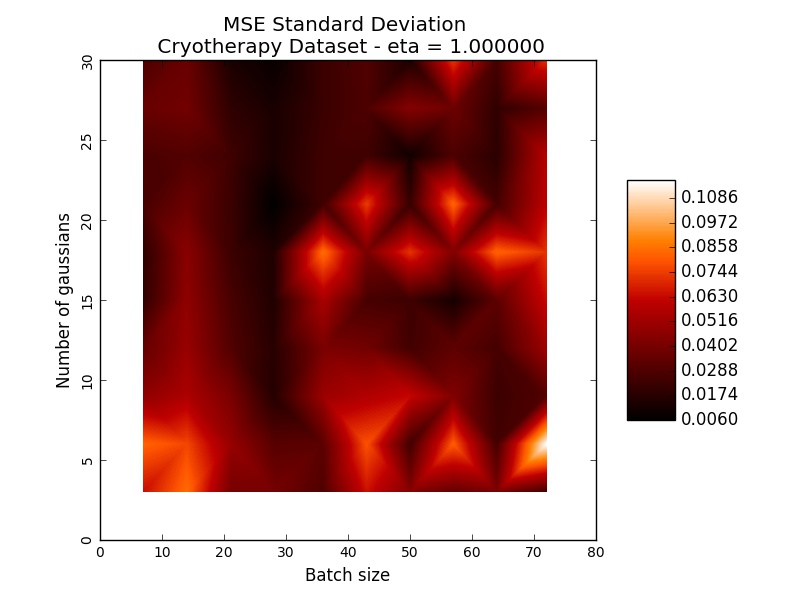}
        \label{fig:apmsecryo100std}
    \end{subfigure}
    \vspace{-0.25cm}
    \caption*{\vspace{-0.1cm} Source: Author. }
    \label{fig:apmsecryo0100}
\end{figure}

\chapter{Pima Diabetes Hyperparameter Results}
\label{appendicepimadiabetes}

\begin{figure} [!ht]
    \centering
    \caption{\vspace{-0.1cm} Accuracy mean and standard deviation measurements for the VFNN on the Pima Diabetes dataset with eta = 0.01. }
    \begin{subfigure}[b]{0.45\textwidth}
        \includegraphics[width=\textwidth]{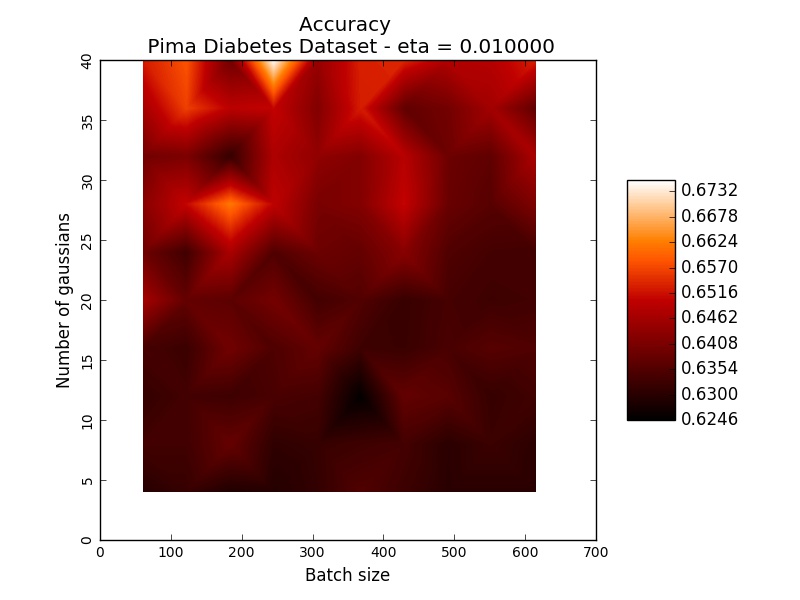}
        \label{fig:apaccpima001mean}
    \end{subfigure}
    \begin{subfigure}[b]{0.45\textwidth}
        \includegraphics[width=\textwidth]{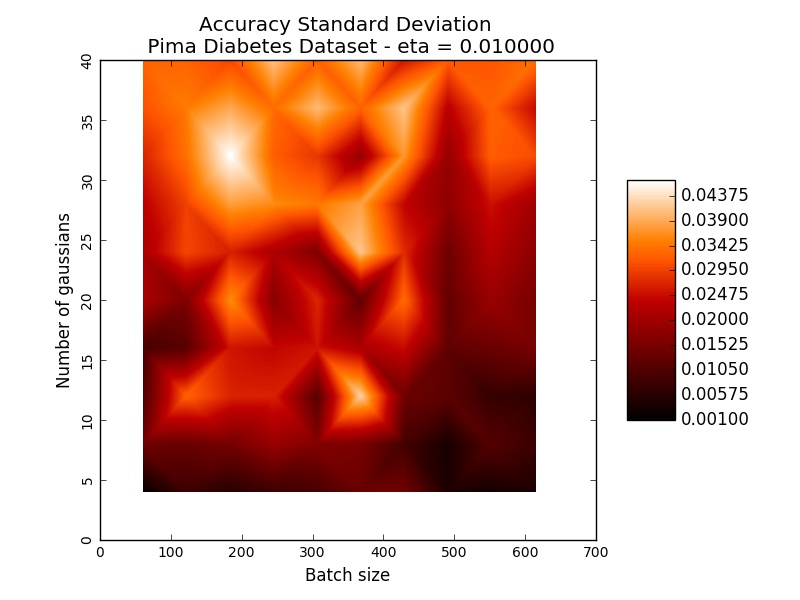}
        \label{fig:apaccpima001std}
    \end{subfigure}
    \vspace{-0.25cm}
    \caption*{\vspace{-0.1cm} Source: Author. }
    \label{fig:apaccpima001}
\end{figure}

\begin{figure} [!ht]
    \centering
    \caption{\vspace{-0.1cm} Accuracy mean and standard deviation measurements for the VFNN on the Pima Diabetes dataset with eta = 0.2575. }
    \begin{subfigure}[b]{0.45\textwidth}
        \includegraphics[width=\textwidth]{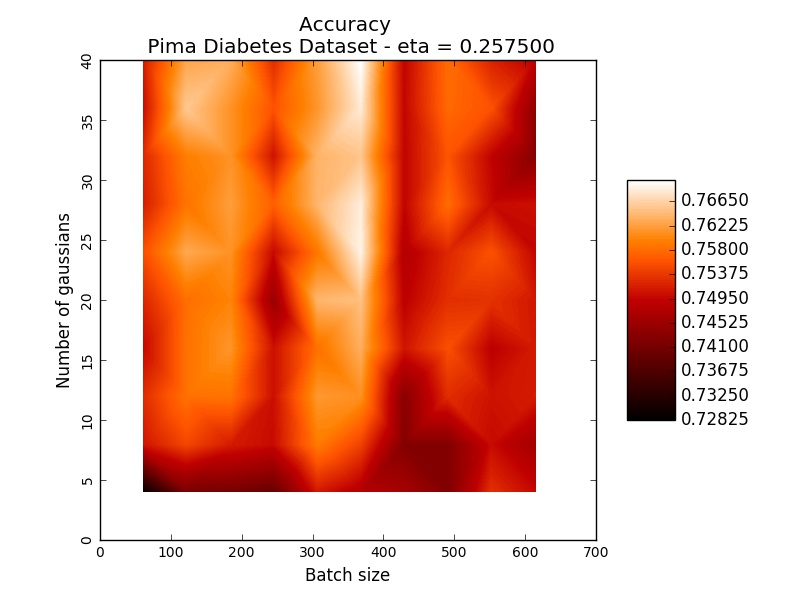}
        \label{fig:apaccpima025mean}
    \end{subfigure}
    \begin{subfigure}[b]{0.45\textwidth}
        \includegraphics[width=\textwidth]{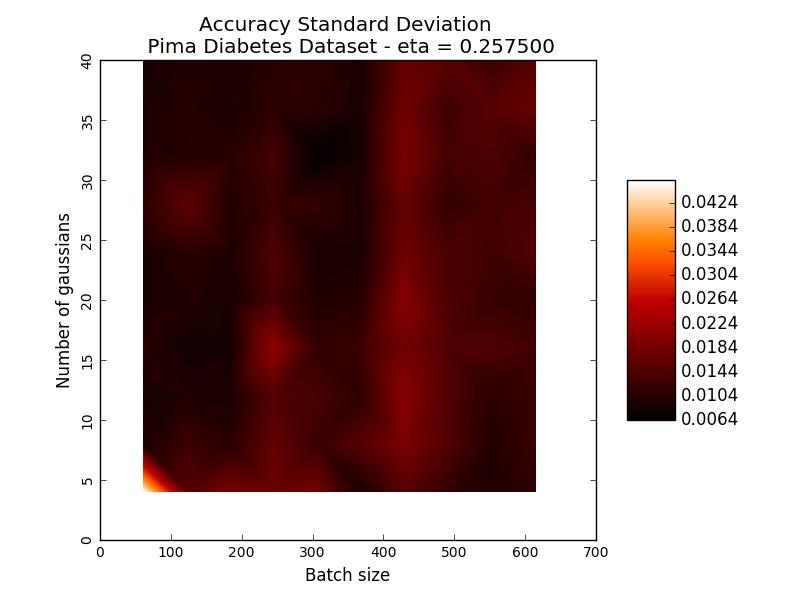}
        \label{fig:apaccpima025std}
    \end{subfigure}
    \vspace{-0.25cm}
    \caption*{\vspace{-0.1cm} Source: Author. }
    \label{fig:apaccpima025}
\end{figure}

\begin{figure} [!ht]
    \centering
    \caption{\vspace{-0.1cm} Accuracy mean and standard deviation measurements for the VFNN on the Pima Diabetes dataset with eta = 0.505. }
    \begin{subfigure}[b]{0.45\textwidth}
        \includegraphics[width=\textwidth]{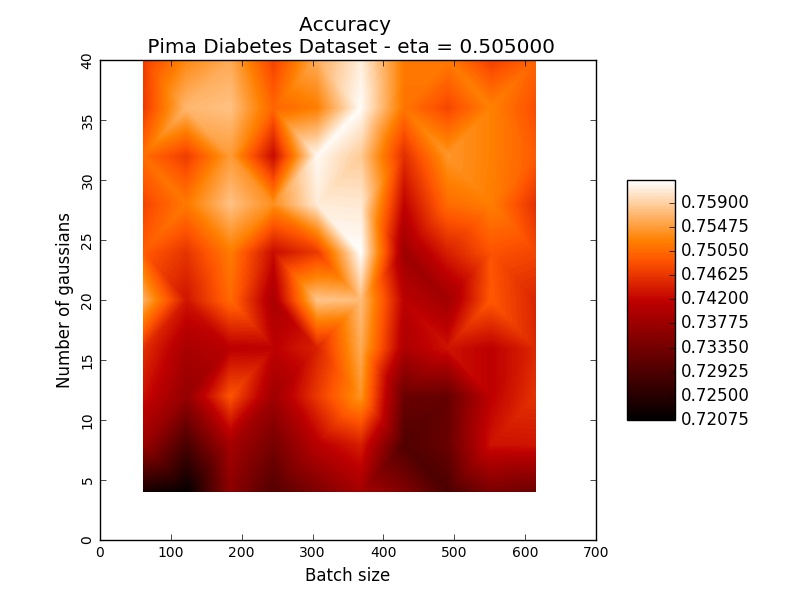}
        \label{fig:apaccpima0505mean}
    \end{subfigure}
    \begin{subfigure}[b]{0.45\textwidth}
        \includegraphics[width=\textwidth]{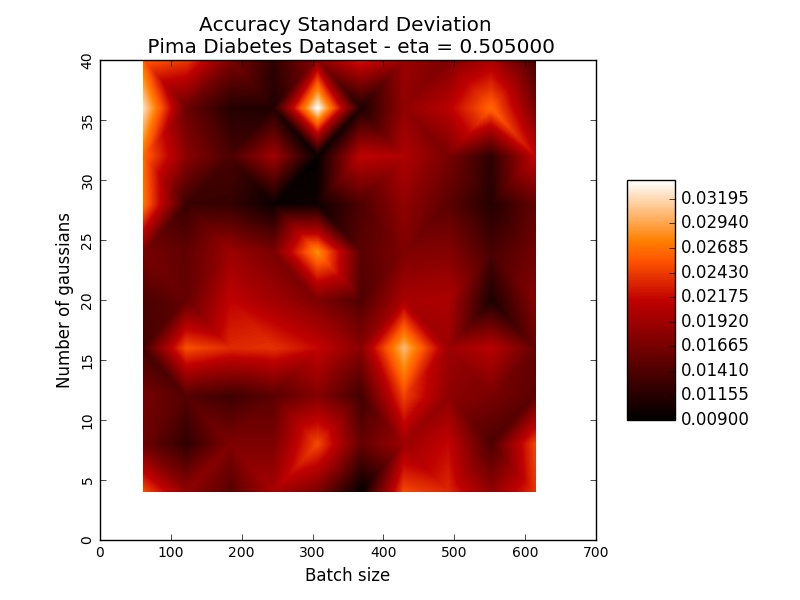}
        \label{fig:apaccpima0505std}
    \end{subfigure}
    \vspace{-0.25cm}
    \caption*{\vspace{-0.1cm} Source: Author. }
    \label{fig:apaccpima505}
\end{figure}

\begin{figure} [!ht]
    \centering
    \caption{\vspace{-0.1cm} Accuracy mean and standard deviation measurements for the VFNN on the Pima Diabetes dataset with eta = 0.7525. }
    \begin{subfigure}[b]{0.45\textwidth}
        \includegraphics[width=\textwidth]{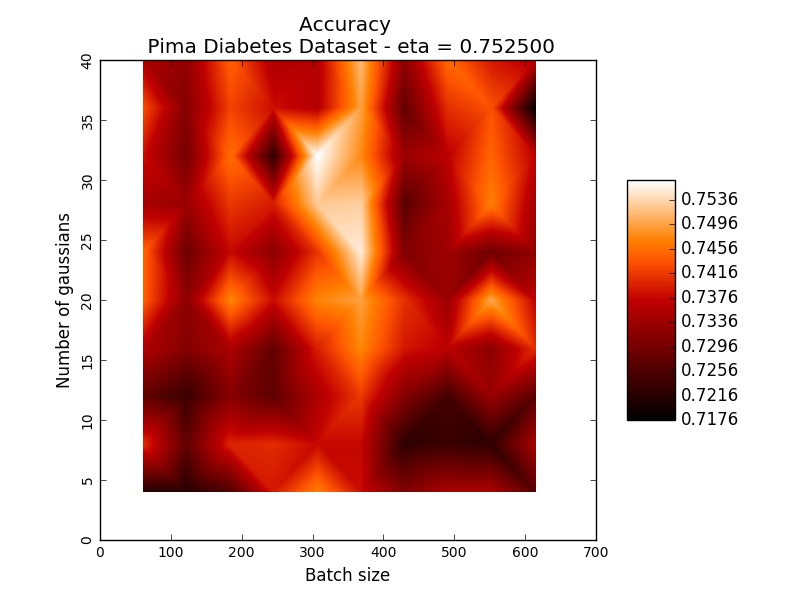}
        \label{fig:apaccpima075mean}
    \end{subfigure}
    \begin{subfigure}[b]{0.45\textwidth}
        \includegraphics[width=\textwidth]{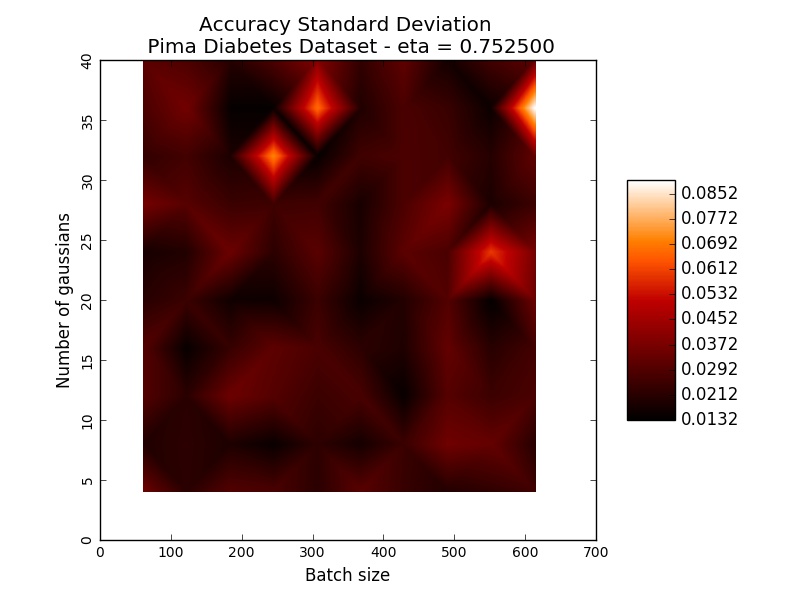}
        \label{fig:apaccpima075std}
    \end{subfigure}
    \vspace{-0.25cm}
    \caption*{\vspace{-0.1cm} Source: Author. }
    \label{fig:apaccpima75}
\end{figure}

\begin{figure} [!ht]
    \centering
    \caption{\vspace{-0.1cm} Accuracy mean and standard deviation measurements for the VFNN on the Pima Diabetes dataset with eta = 1.0. }
    \begin{subfigure}[b]{0.45\textwidth}
        \includegraphics[width=\textwidth]{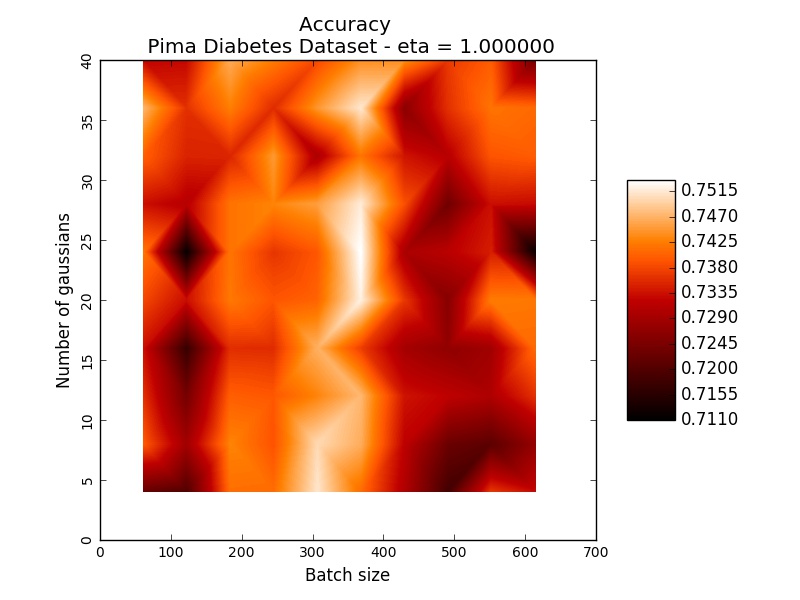}
        \label{fig:apaccpima100mean}
    \end{subfigure}
    \begin{subfigure}[b]{0.45\textwidth}
        \includegraphics[width=\textwidth]{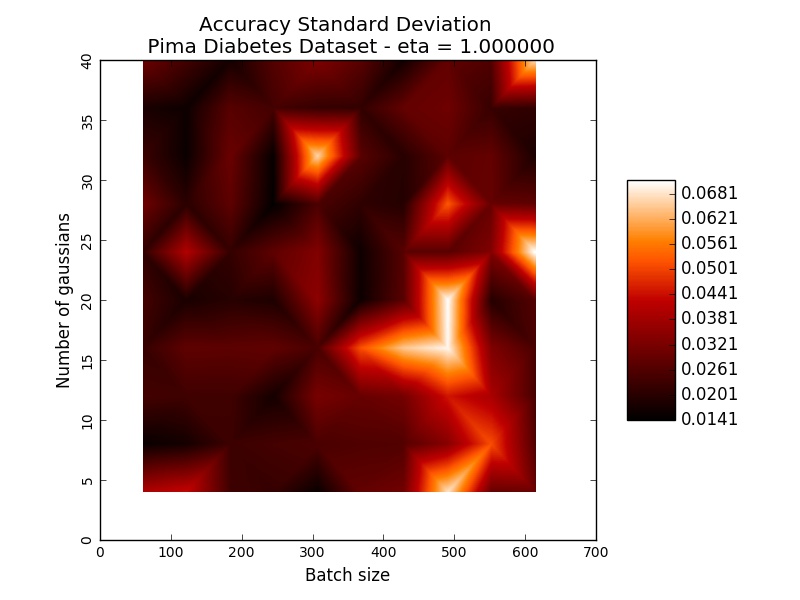}
        \label{fig:apaccpima100std}
    \end{subfigure}
    \vspace{-0.25cm}
    \caption*{\vspace{-0.1cm} Source: Author. }
    \label{fig:apaccpima100}
\end{figure}

\begin{figure} [!ht]
    \centering
    \caption{\vspace{-0.1cm} AUCROC mean and standard deviation measurements for the VFNN on the Pima Diabetes dataset with eta = 0.01. }
    \begin{subfigure}[b]{0.45\textwidth}
        \includegraphics[width=\textwidth]{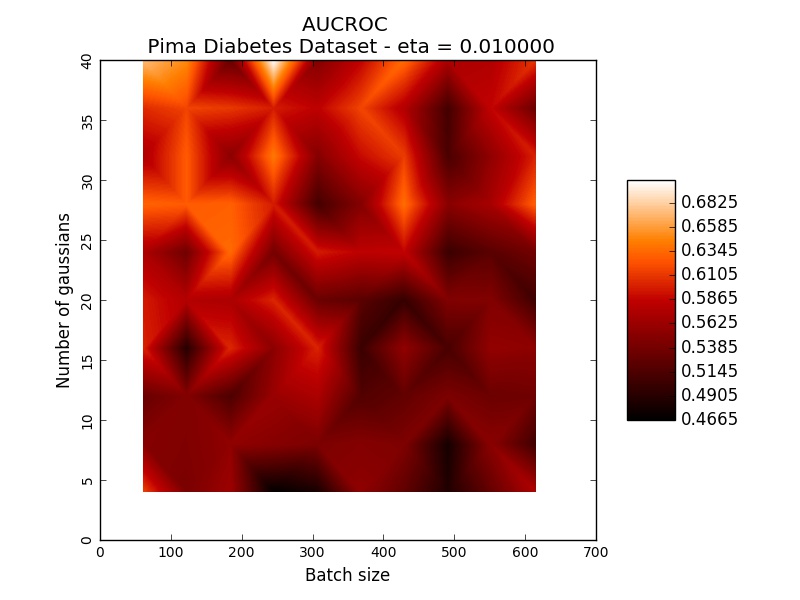}
        \label{fig:apaucpima001mean}
    \end{subfigure}
    \begin{subfigure}[b]{0.45\textwidth}
        \includegraphics[width=\textwidth]{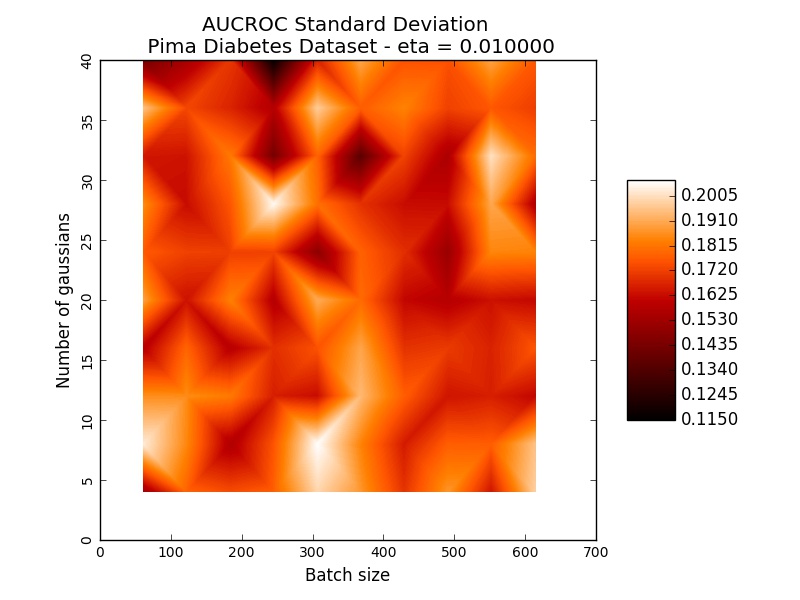}
        \label{fig:apaucpima001std}
    \end{subfigure}
    \vspace{-0.25cm}
    \caption*{\vspace{-0.1cm} Source: Author. }
    \label{fig:apaucpima001}
\end{figure}

\begin{figure} [!ht]
    \centering
    \caption{\vspace{-0.1cm} AUCROC mean and standard deviation measurements for the VFNN on the Pima Diabetes dataset with eta = 0.2575. }
    \begin{subfigure}[b]{0.45\textwidth}
        \includegraphics[width=\textwidth]{Figures/HyperparametersResults/AUCROCPimaDiabetes0257500.jpg}
        \label{fig:apaucpima025mean}
    \end{subfigure}
    \begin{subfigure}[b]{0.45\textwidth}
        \includegraphics[width=\textwidth]{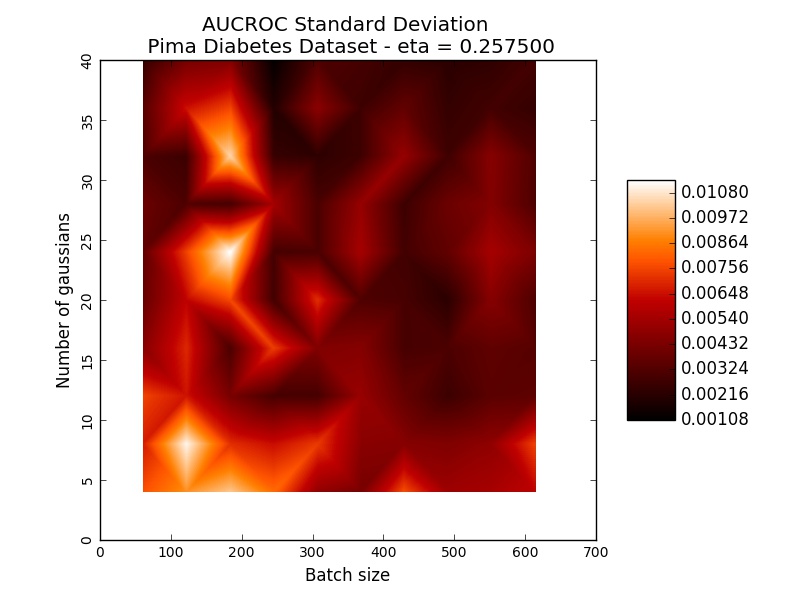}
        \label{fig:apaucpima025std}
    \end{subfigure}
    \vspace{-0.25cm}
    \caption*{\vspace{-0.1cm} Source: Author. }
    \label{fig:apaucpima025}
\end{figure}

\begin{figure} [!ht]
    \centering
    \caption{\vspace{-0.1cm} AUCROC mean and standard deviation measurements for the VFNN on the Pima Diabetes dataset with eta = 0.505. }
    \begin{subfigure}[b]{0.45\textwidth}
        \includegraphics[width=\textwidth]{Figures/HyperparametersResults/AUCROCPimaDiabetes0505000.jpg}
        \label{fig:apaucpima0505mean}
    \end{subfigure}
    \begin{subfigure}[b]{0.45\textwidth}
        \includegraphics[width=\textwidth]{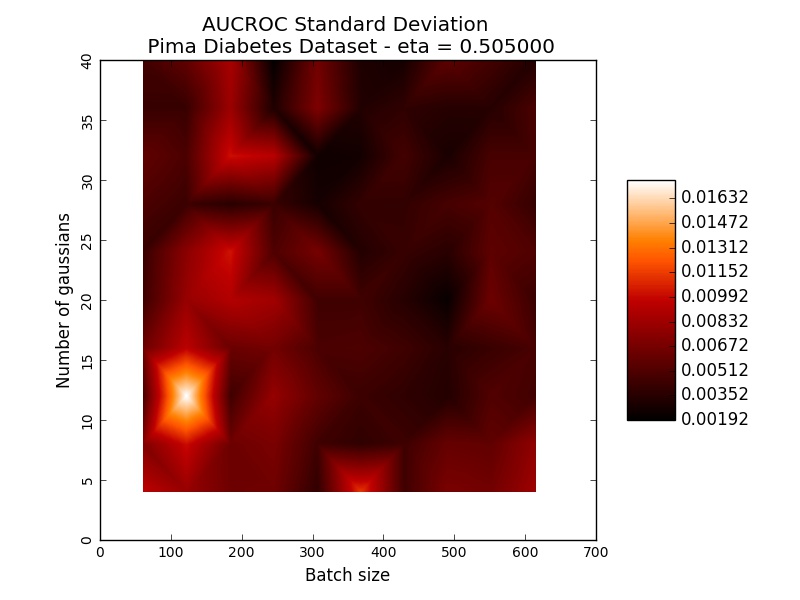}
        \label{fig:apaucpima0505std}
    \end{subfigure}
    \vspace{-0.25cm}
    \caption*{\vspace{-0.1cm} Source: Author. }
    \label{fig:apaucpima0505}
\end{figure}

\begin{figure} [!ht]
    \centering
    \caption{\vspace{-0.1cm} AUCROC mean and standard deviation measurements for the VFNN on the PimaDiabetes dataset with eta = 0.7525. }
    \begin{subfigure}[b]{0.45\textwidth}
        \includegraphics[width=\textwidth]{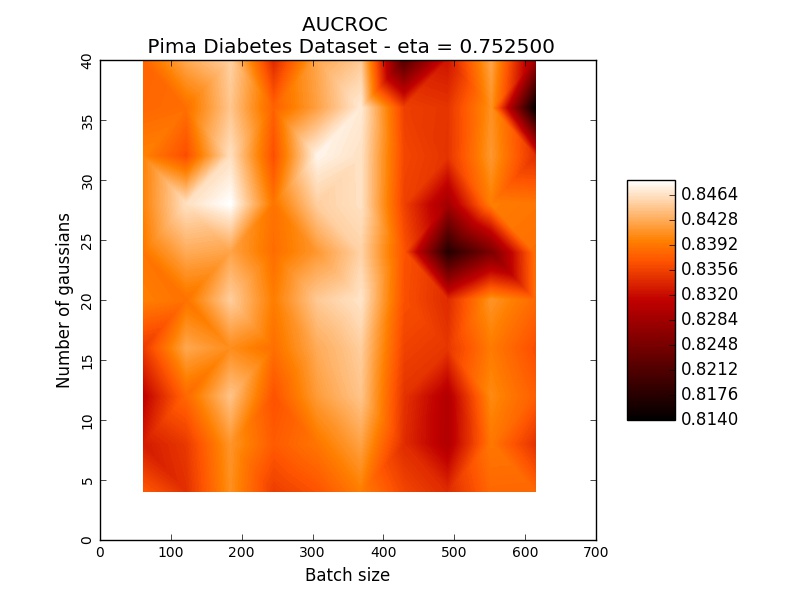}
        \label{fig:apaucpima075mean}
    \end{subfigure}
    \begin{subfigure}[b]{0.45\textwidth}
        \includegraphics[width=\textwidth]{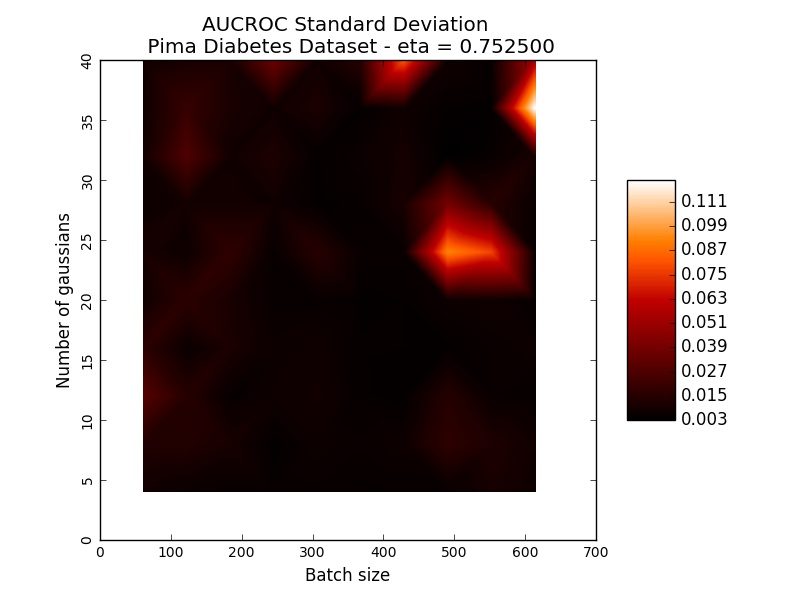}
        \label{fig:apaucpima075std}
    \end{subfigure}
    \vspace{-0.25cm}
    \caption*{\vspace{-0.1cm} Source: Author. }
    \label{fig:apaucpima075}
\end{figure}

\begin{figure} [!ht]
    \centering
    \caption{\vspace{-0.1cm} AUCROC mean and standard deviation measurements for the VFNN on the Pima Diabetes dataset with eta = 1.0. }
    \begin{subfigure}[b]{0.45\textwidth}
        \includegraphics[width=\textwidth]{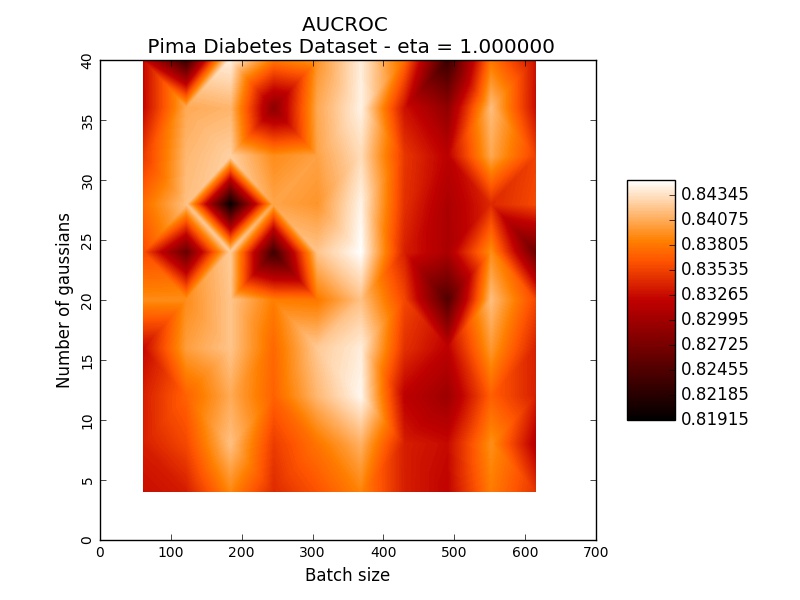}
        \label{fig:apaucpima100mean}
    \end{subfigure}
    \begin{subfigure}[b]{0.45\textwidth}
        \includegraphics[width=\textwidth]{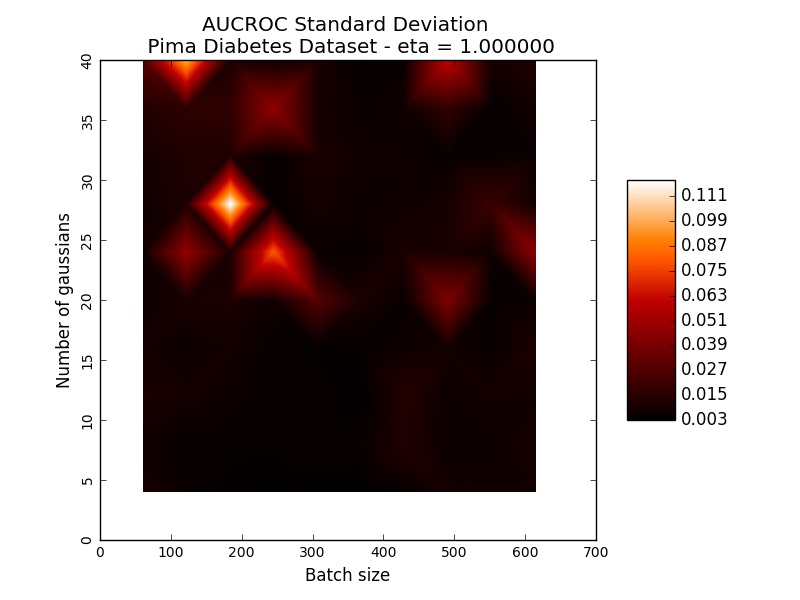}
        \label{fig:apaucpima100std}
    \end{subfigure}
    \vspace{-0.25cm}
    \caption*{\vspace{-0.1cm} Source: Author. }
    \label{fig:apaucpima100}
\end{figure}

\begin{figure} [!ht]
    \centering
    \caption{\vspace{-0.1cm} Mean Squared Error and standard deviation measurements for the VFNN on the Pima Diabetes dataset with eta = 0.01. }
    \begin{subfigure}[b]{0.45\textwidth}
        \includegraphics[width=\textwidth]{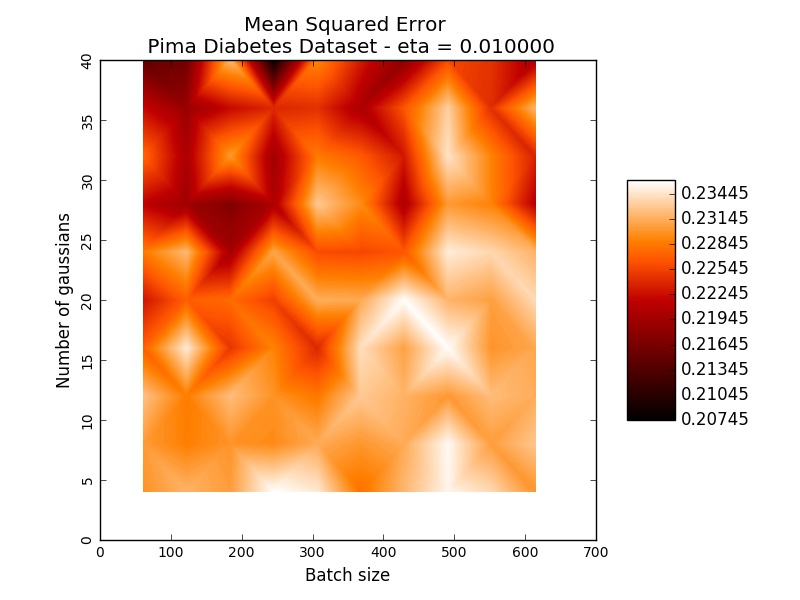}
        \label{fig:apmsepima001mean}
    \end{subfigure}
    \begin{subfigure}[b]{0.45\textwidth}
        \includegraphics[width=\textwidth]{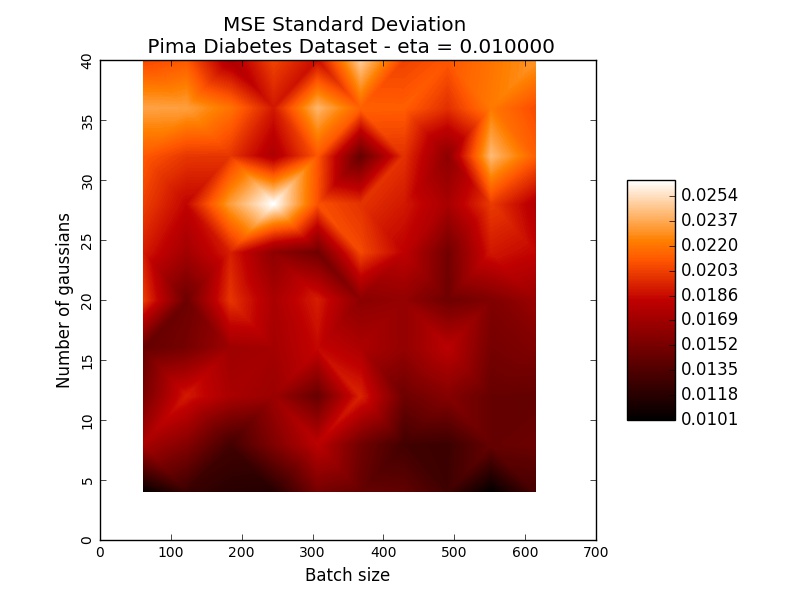}
        \label{fig:apmsepima001std}
    \end{subfigure}
    \vspace{-0.25cm}
    \caption*{\vspace{-0.1cm} Source: Author. }
    \label{fig:apmsepima001}
\end{figure}

\begin{figure} [!ht]
    \centering
    \caption{\vspace{-0.1cm} Mean Squared Error and standard deviation measurements for the VFNN on the Pima Diabetes dataset with eta = 0.2575. }
    \begin{subfigure}[b]{0.45\textwidth}
        \includegraphics[width=\textwidth]{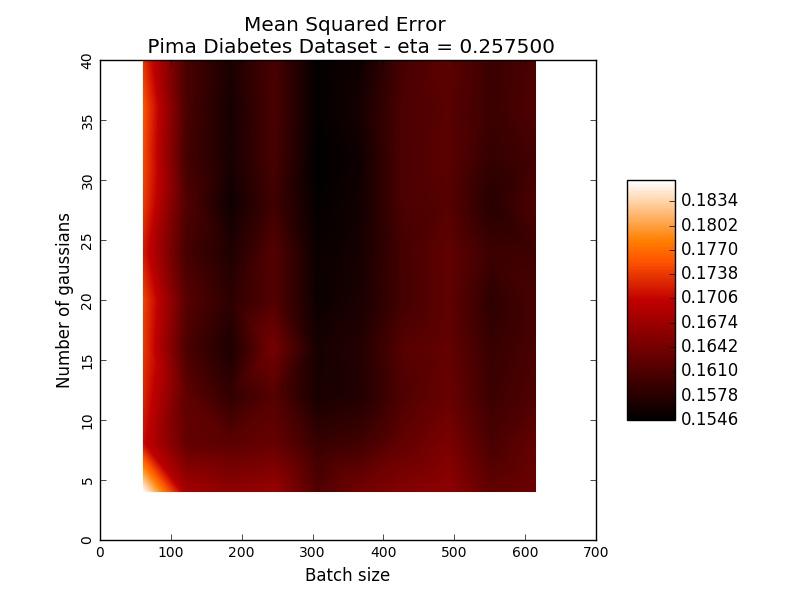}
        \label{fig:apmsepima025mean}
    \end{subfigure}
    \begin{subfigure}[b]{0.45\textwidth}
        \includegraphics[width=\textwidth]{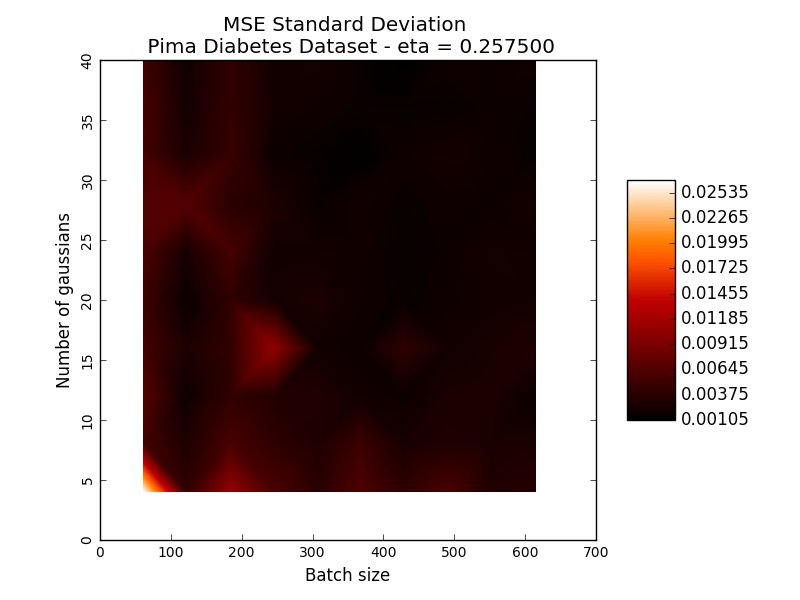}
        \label{fig:apmsepima025std}
    \end{subfigure}
    \vspace{-0.25cm}
    \caption*{\vspace{-0.1cm} Source: Author. }
    \label{fig:apmsepima025}
\end{figure}

\begin{figure} [!ht]
    \centering
    \caption{\vspace{-0.1cm} Mean Squared Error and standard deviation measurements for the VFNN on the PimaDiabetes dataset with eta = 0.505. }
    \begin{subfigure}[b]{0.45\textwidth}
        \includegraphics[width=\textwidth]{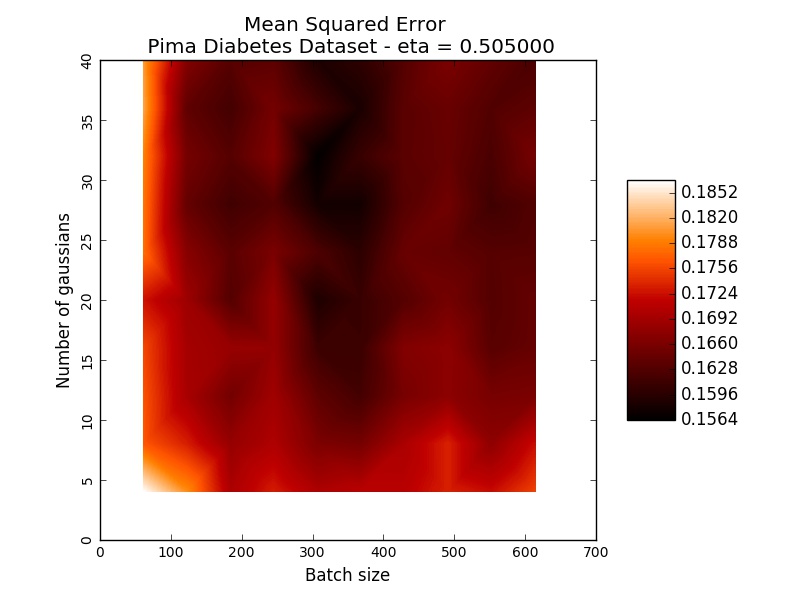}
        \label{fig:apmsepima0505mean}
    \end{subfigure}
    \begin{subfigure}[b]{0.45\textwidth}
        \includegraphics[width=\textwidth]{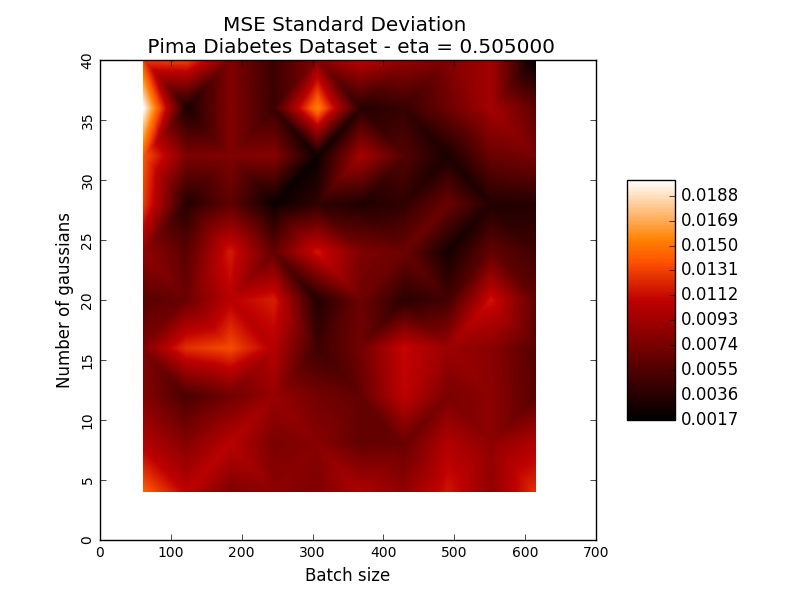}
        \label{fig:apmsepima0505std}
    \end{subfigure}
    \vspace{-0.25cm}
    \caption*{\vspace{-0.1cm} Source: Author. }
    \label{fig:apmsepima0505}
\end{figure}

\begin{figure} [!ht]
    \centering
    \caption{\vspace{-0.1cm} Mean Squared Error and standard deviation measurements for the VFNN on the Pima Diabetes dataset with eta = 0.7525. }
    \begin{subfigure}[b]{0.45\textwidth}
        \includegraphics[width=\textwidth]{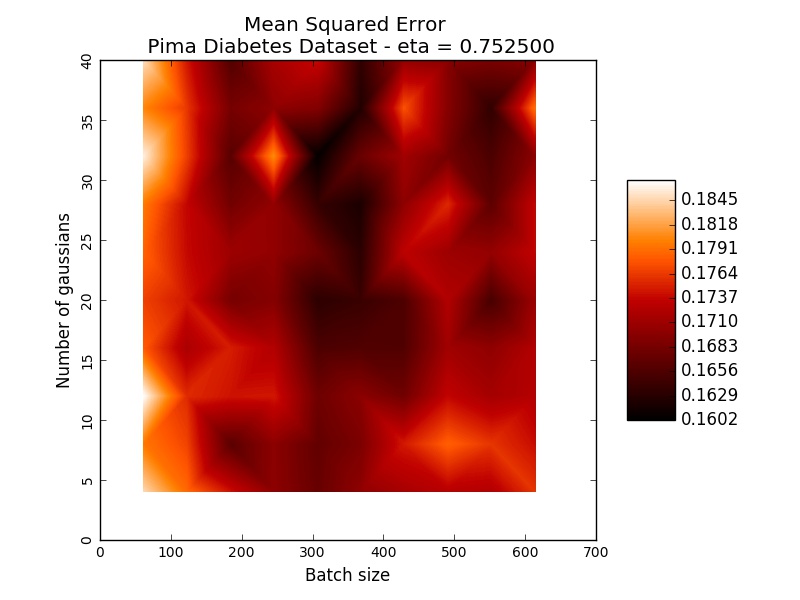}
        \label{fig:apmsepima075mean}
    \end{subfigure}
    \begin{subfigure}[b]{0.45\textwidth}
        \includegraphics[width=\textwidth]{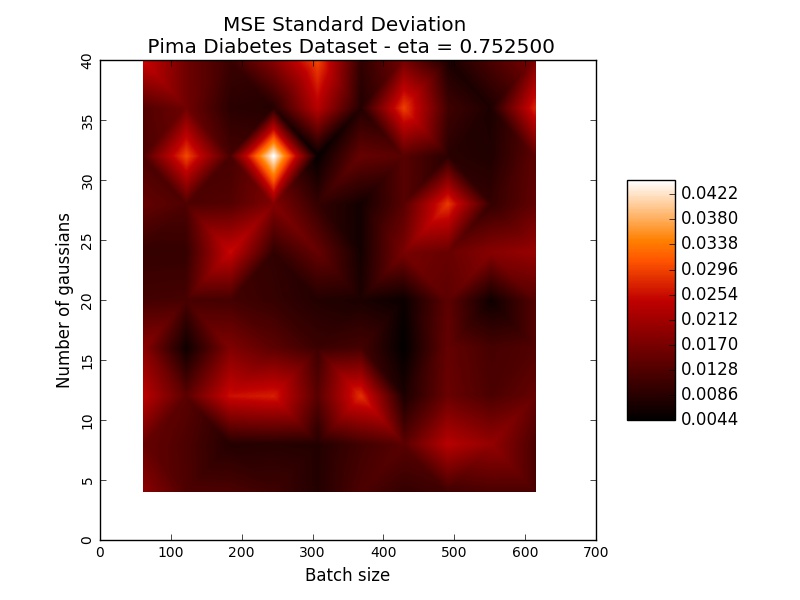}
        \label{fig:apmsepima075std}
    \end{subfigure}
    \vspace{-0.25cm}
    \caption*{\vspace{-0.1cm} Source: Author. }
    \label{fig:apmsepima075}
\end{figure}

\begin{figure} [!ht]
    \centering
    \caption{\vspace{-0.1cm} Mean Squared Error and standard deviation measurements for the VFNN on the Pima Diabetes dataset with eta = 1.0. }
    \begin{subfigure}[b]{0.45\textwidth}
        \includegraphics[width=\textwidth]{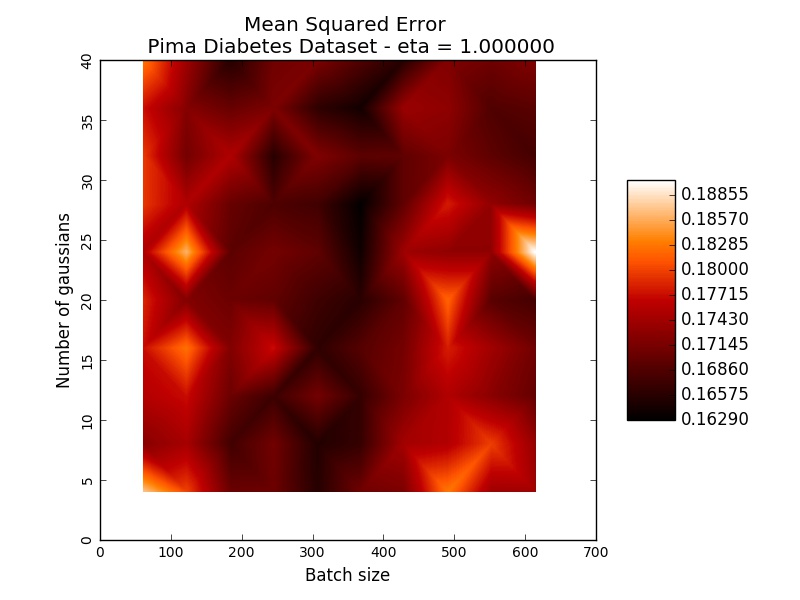}
        \label{fig:apmsepima100mean}
    \end{subfigure}
    \begin{subfigure}[b]{0.45\textwidth}
        \includegraphics[width=\textwidth]{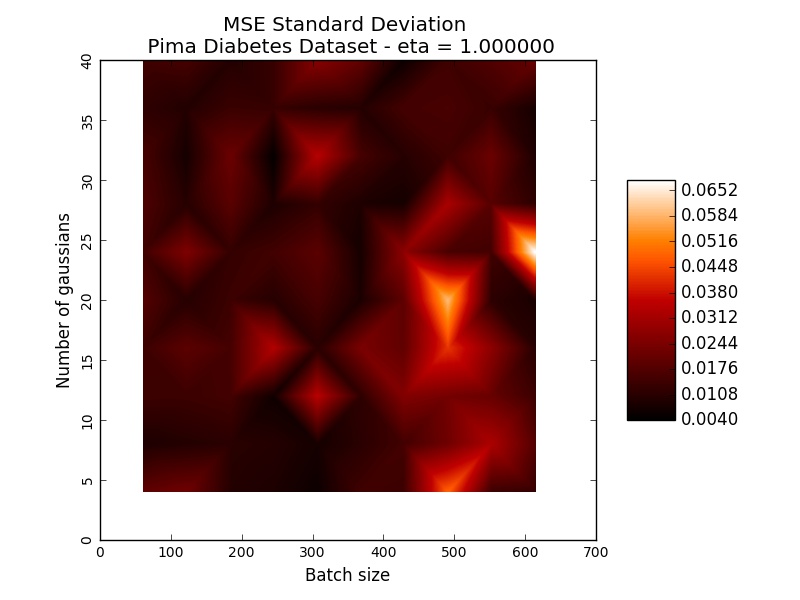}
        \label{fig:apmsepima100std}
    \end{subfigure}
    \vspace{-0.25cm}
    \caption*{\vspace{-0.1cm} Source: Author. }
    \label{fig:apmsepima0100}
\end{figure}

\chapter{Immunotherapy Hyperparameter Results}
\label{appendiceimmunotherapy}

\begin{figure} [!ht]
    \centering
    \caption{\vspace{-0.1cm} Accuracy mean and standard deviation measurements for the VFNN on the Immunotherapy dataset with eta = 0.01. }
    \begin{subfigure}[b]{0.4\textwidth}
        \includegraphics[width=\textwidth]{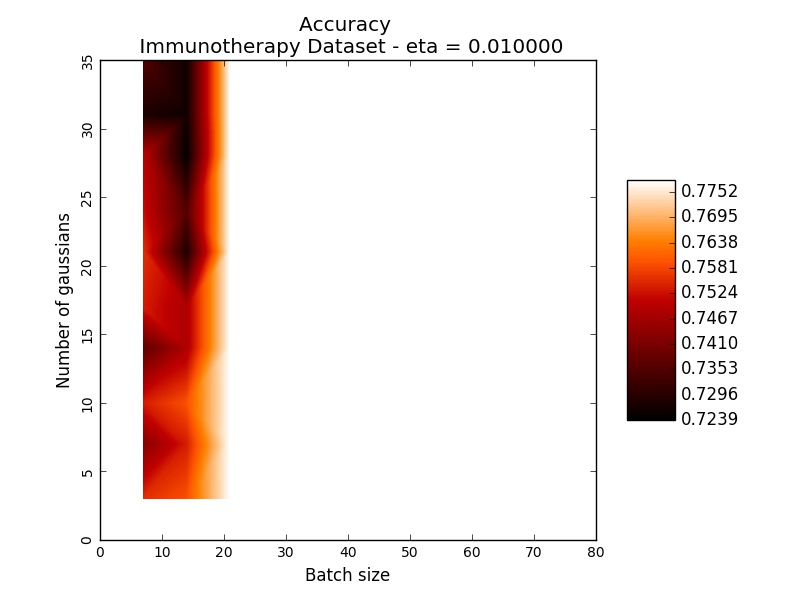}
        \label{fig:apaccimmu001mean}
    \end{subfigure}
    \begin{subfigure}[b]{0.4\textwidth}
        \includegraphics[width=\textwidth]{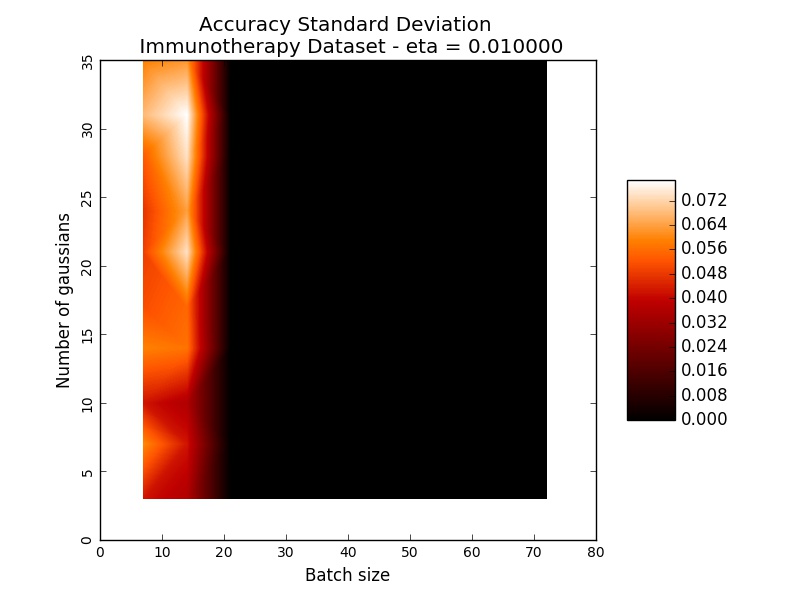}
        \label{fig:apaccimmu001std}
    \end{subfigure}
    \vspace{-0.25cm}
    \caption*{\vspace{-0.1cm} Source: Author. }
    \label{fig:apaccimmu001}
\end{figure}

\begin{figure} [!ht]
    \centering
    \caption{\vspace{-0.1cm} Accuracy mean and standard deviation measurements for the VFNN on the Immunotherapy dataset with eta = 0.2575. }
    \begin{subfigure}[b]{0.4\textwidth}
        \includegraphics[width=\textwidth]{Figures/HyperparametersResults/AccuracyImmunotherapy0257500.jpg}
        \label{fig:apaccimmu025mean}
    \end{subfigure}
    \begin{subfigure}[b]{0.4\textwidth}
        \includegraphics[width=\textwidth]{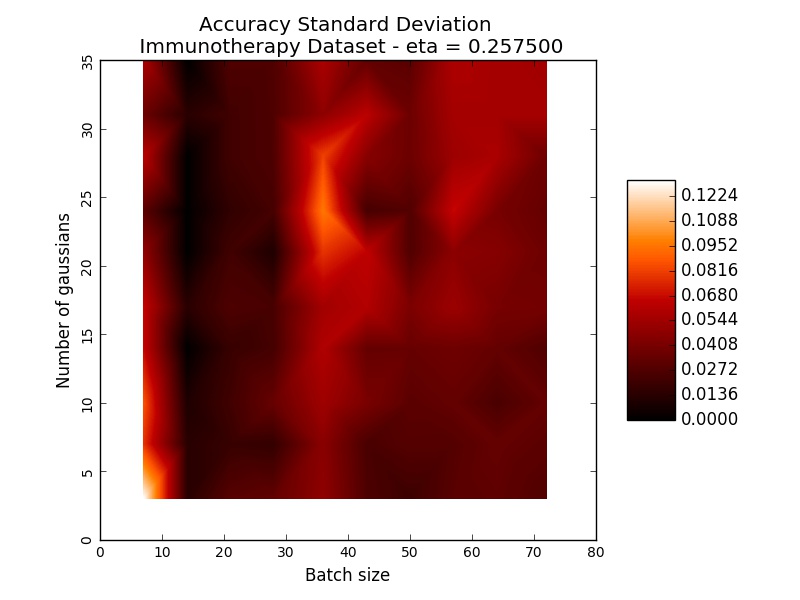}
        \label{fig:apaccimmu025std}
    \end{subfigure}
    \vspace{-0.25cm}
    \caption*{\vspace{-0.1cm} Source: Author. }
    \label{fig:apaccimmu025}
\end{figure}

\begin{figure} [!ht]
    \centering
    \caption{\vspace{-0.1cm} Accuracy mean and standard deviation measurements for the VFNN on the Immunotherapy dataset with eta = 0.505. }
    \begin{subfigure}[b]{0.4\textwidth}
        \includegraphics[width=\textwidth]{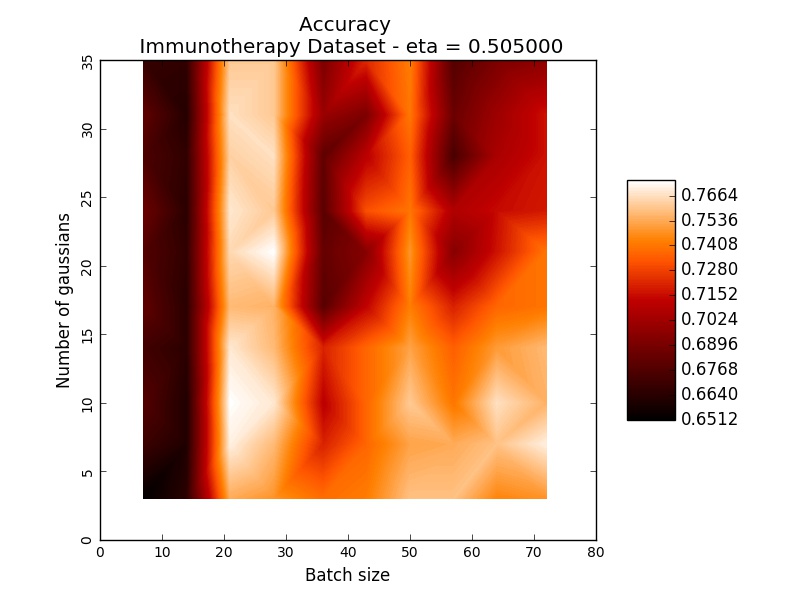}
        \label{fig:apaccimmu0505mean}
    \end{subfigure}
    \begin{subfigure}[b]{0.4\textwidth}
        \includegraphics[width=\textwidth]{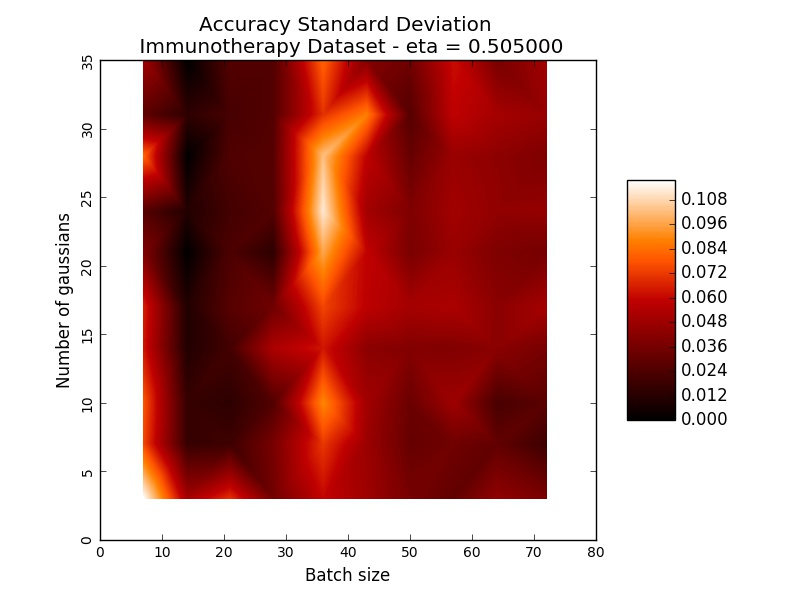}
        \label{fig:apaccimmu0505std}
    \end{subfigure}
    \vspace{-0.25cm}
    \caption*{\vspace{-0.1cm} Source: Author. }
    \label{fig:apaccimmu505}
\end{figure}

\begin{figure} [!ht]
    \centering
    \caption{\vspace{-0.1cm} Accuracy mean and standard deviation measurements for the VFNN on the Immunotherapy dataset with eta = 0.7525. }
    \begin{subfigure}[b]{0.4\textwidth}
        \includegraphics[width=\textwidth]{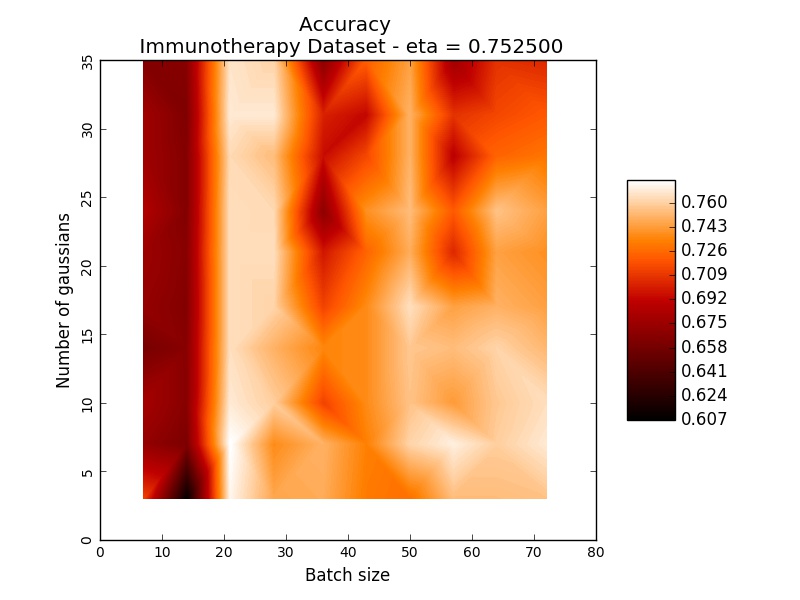}
        \label{fig:apaccimmu075mean}
    \end{subfigure}
    \begin{subfigure}[b]{0.4\textwidth}
        \includegraphics[width=\textwidth]{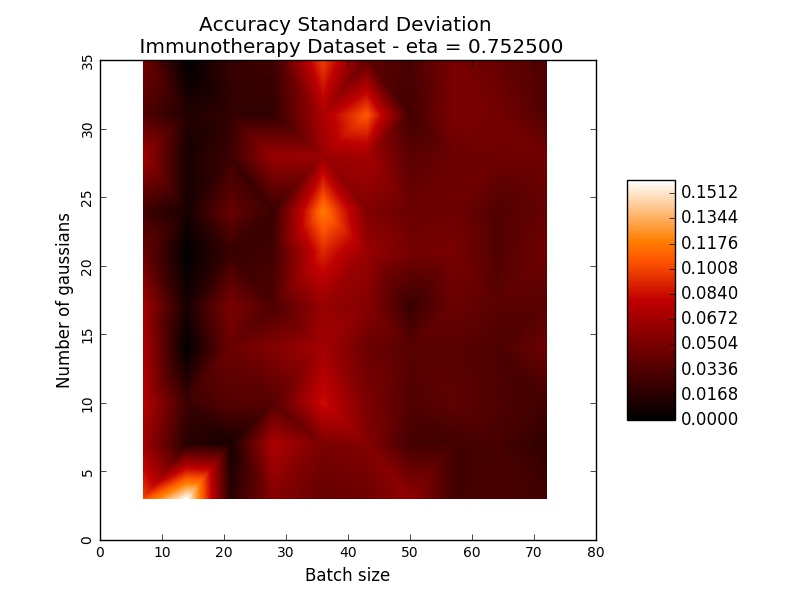}
        \label{fig:apaccimmu075std}
    \end{subfigure}
    \vspace{-0.25cm}
    \caption*{\vspace{-0.1cm} Source: Author. }
    \label{fig:apaccimmu75}
\end{figure}

\begin{figure} [!ht]
    \centering
    \caption{\vspace{-0.1cm} Accuracy mean and standard deviation measurements for the VFNN on the Immunotherapy dataset with eta = 1.0. }
    \begin{subfigure}[b]{0.4\textwidth}
        \includegraphics[width=\textwidth]{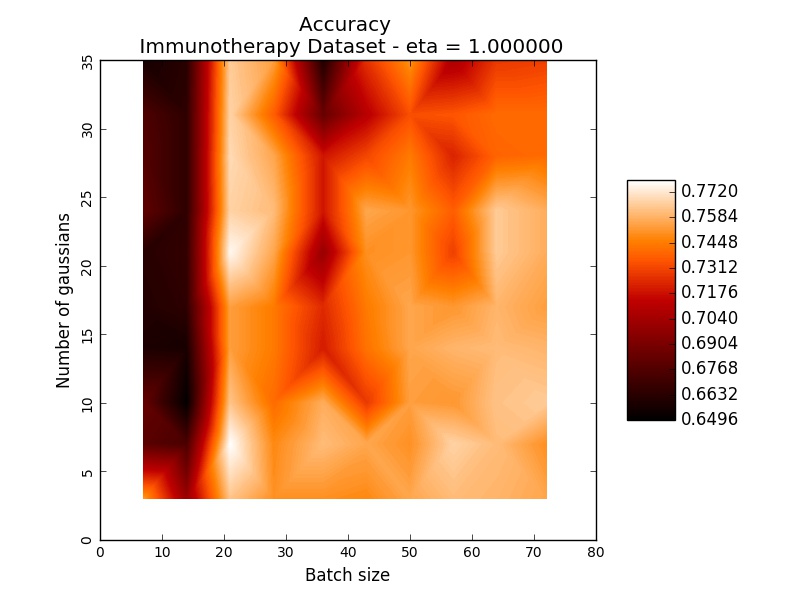}
        \label{fig:apaccimmu100mean}
    \end{subfigure}
    \begin{subfigure}[b]{0.4\textwidth}
        \includegraphics[width=\textwidth]{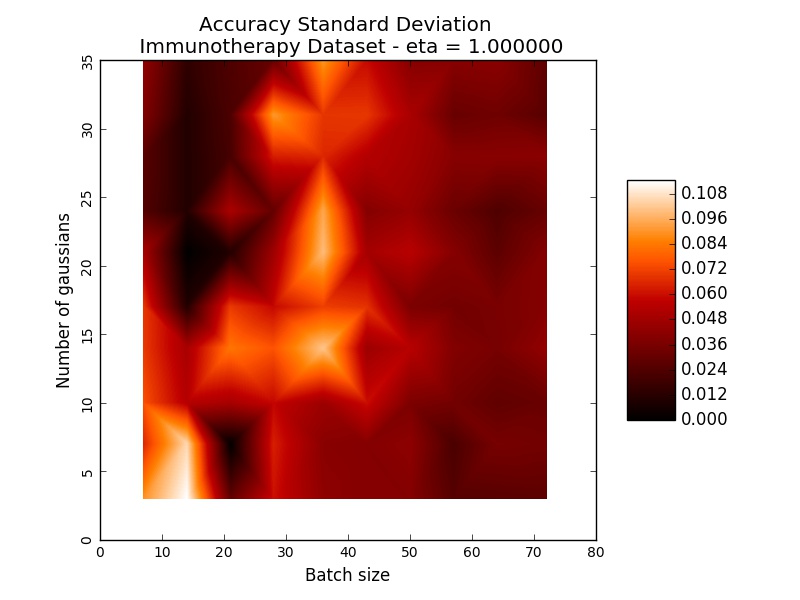}
        \label{fig:apaccimmu100std}
    \end{subfigure}
    \vspace{-0.25cm}
    \caption*{\vspace{-0.1cm} Source: Author. }
    \label{fig:apaccimmu100}
\end{figure}

\begin{figure} [!ht]
    \centering
    \caption{\vspace{-0.1cm} AUCROC mean and standard deviation measurements for the VFNN on the Immunotherapy dataset with eta = 0.01. }
    \begin{subfigure}[b]{0.4\textwidth}
        \includegraphics[width=\textwidth]{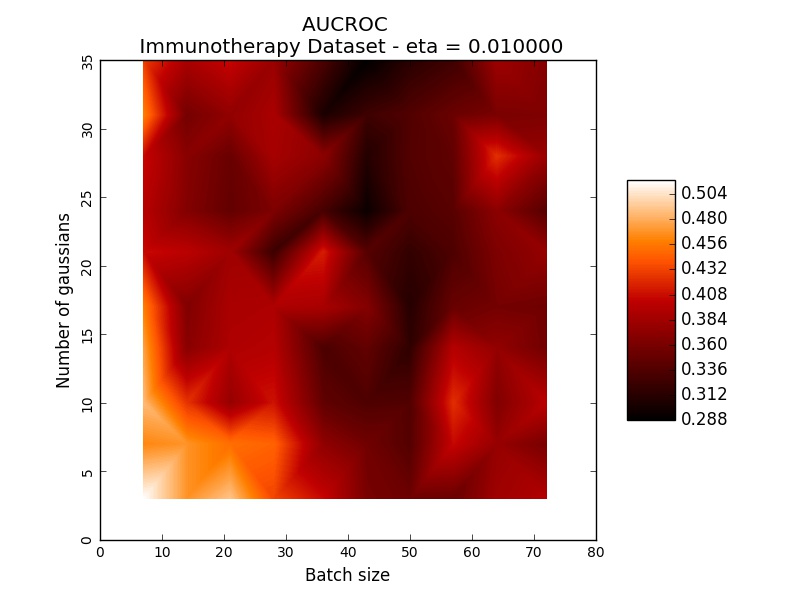}
        \label{fig:apaucimmu001mean}
    \end{subfigure}
    \begin{subfigure}[b]{0.4\textwidth}
        \includegraphics[width=\textwidth]{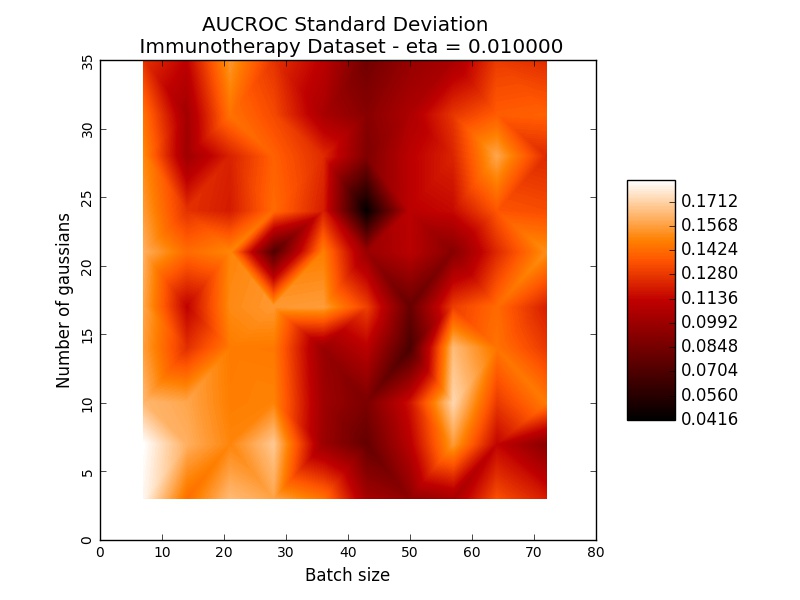}
        \label{fig:apaucimmu001std}
    \end{subfigure}
    \vspace{-0.25cm}
    \caption*{\vspace{-0.1cm} Source: Author. }
    \label{fig:apaucimmu001}
\end{figure}

\begin{figure} [!ht]
    \centering
    \caption{\vspace{-0.1cm} AUCROC mean and standard deviation measurements for the VFNN on the Immunotherapy dataset with eta = 0.2575. }
    \begin{subfigure}[b]{0.4\textwidth}
        \includegraphics[width=\textwidth]{Figures/HyperparametersResults/AUCROCImmunotherapy0257500.jpg}
        \label{fig:apaucimmu025mean}
    \end{subfigure}
    \begin{subfigure}[b]{0.4\textwidth}
        \includegraphics[width=\textwidth]{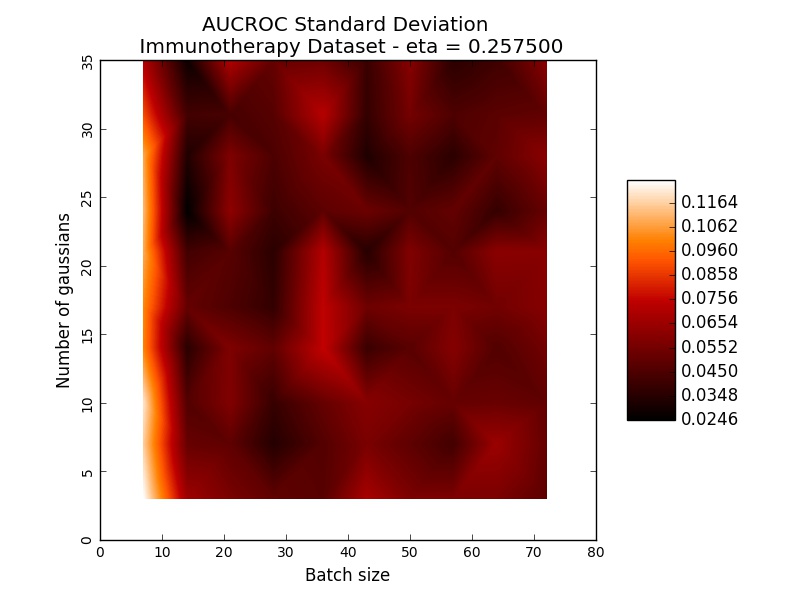}
        \label{fig:apaucimmu025std}
    \end{subfigure}
    \vspace{-0.25cm}
    \caption*{\vspace{-0.1cm} Source: Author. }
    \label{fig:apaucimmu025}
\end{figure}

\begin{figure} [!ht]
    \centering
    \caption{\vspace{-0.1cm} AUCROC mean and standard deviation measurements for the VFNN on the Immunotherapy dataset with eta = 0.505. }
    \begin{subfigure}[b]{0.4\textwidth}
        \includegraphics[width=\textwidth]{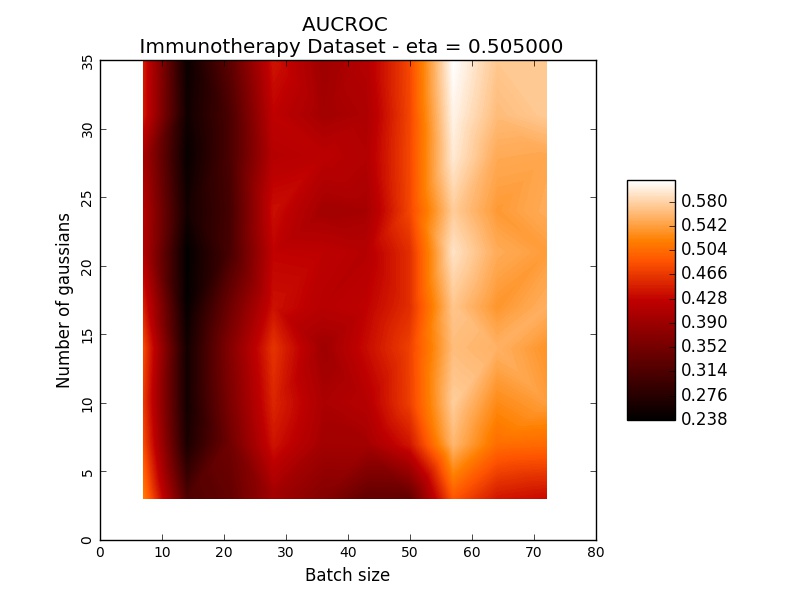}
        \label{fig:apaucimmu0505mean}
    \end{subfigure}
    \begin{subfigure}[b]{0.4\textwidth}
        \includegraphics[width=\textwidth]{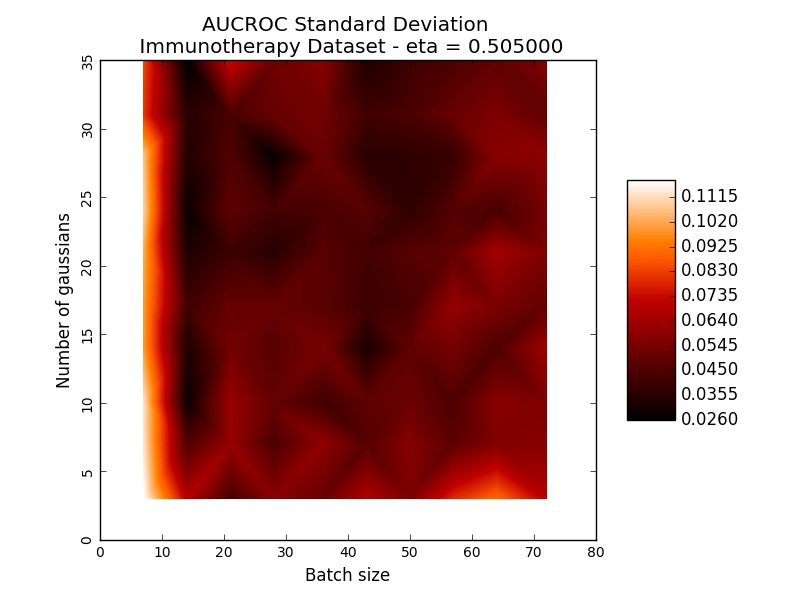}
        \label{fig:apaucimmu0505std}
    \end{subfigure}
    \vspace{-0.25cm}
    \caption*{\vspace{-0.1cm} Source: Author. }
    \label{fig:apaucimmu0505}
\end{figure}

\begin{figure} [!ht]
    \centering
    \caption{\vspace{-0.1cm} AUCROC mean and standard deviation measurements for the VFNN on the Immunotherapy dataset with eta = 0.7525. }
    \begin{subfigure}[b]{0.4\textwidth}
        \includegraphics[width=\textwidth]{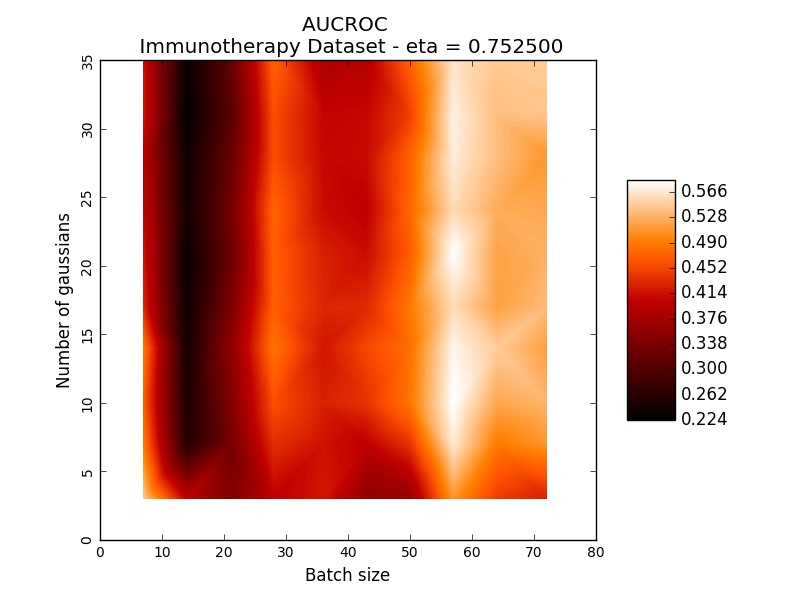}
        \label{fig:apaucimmu075mean}
    \end{subfigure}
    \begin{subfigure}[b]{0.4\textwidth}
        \includegraphics[width=\textwidth]{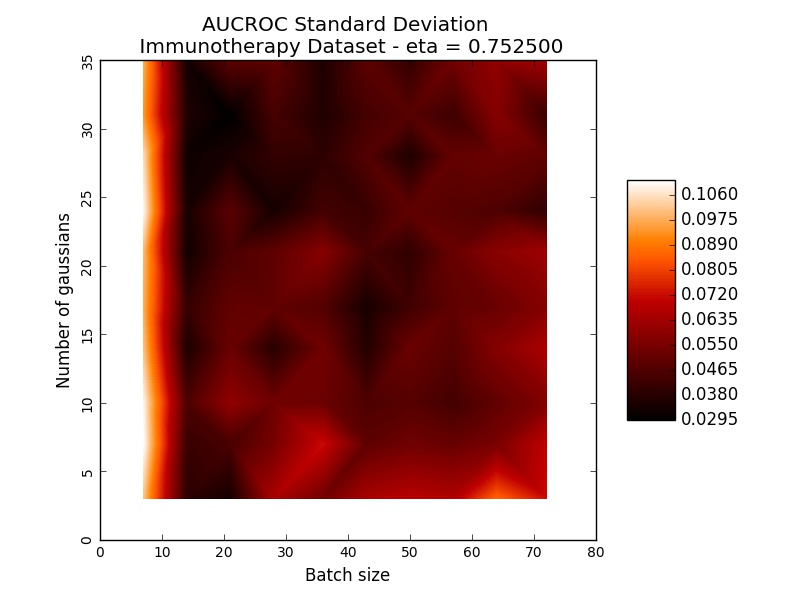}
        \label{fig:apaucimmu075std}
    \end{subfigure}
    \vspace{-0.25cm}
    \caption*{\vspace{-0.1cm} Source: Author. }
    \label{fig:apaucimmu075}
\end{figure}

\begin{figure} [!ht]
    \centering
    \caption{\vspace{-0.1cm} AUCROC mean and standard deviation measurements for the VFNN on the Immunotherapy dataset with eta = 1.0. }
    \begin{subfigure}[b]{0.4\textwidth}
        \includegraphics[width=\textwidth]{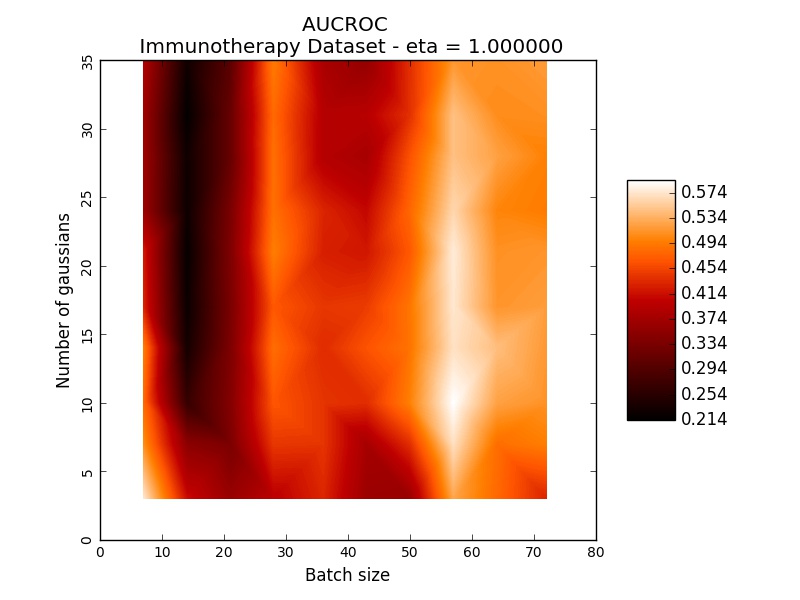}
        \label{fig:apaucimmu100mean}
    \end{subfigure}
    \begin{subfigure}[b]{0.4\textwidth}
        \includegraphics[width=\textwidth]{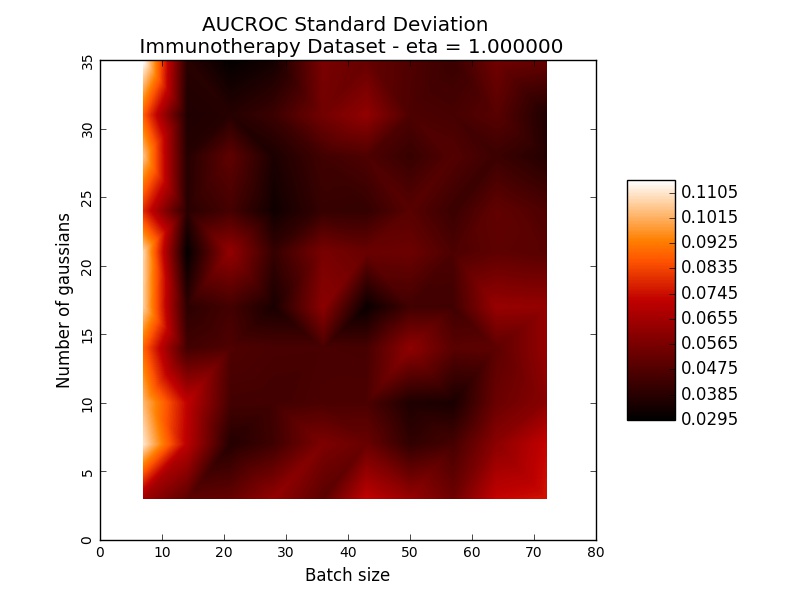}
        \label{fig:apaucimmu100std}
    \end{subfigure}
    \vspace{-0.25cm}
    \caption*{\vspace{-0.1cm} Source: Author. }
    \label{fig:apaucimmu100}
\end{figure}

\begin{figure} [!ht]
    \centering
    \caption{\vspace{-0.1cm} Mean Squared Error and standard deviation measurements for the VFNN on the Immunotherapy dataset with eta = 0.01. }
    \begin{subfigure}[b]{0.4\textwidth}
        \includegraphics[width=\textwidth]{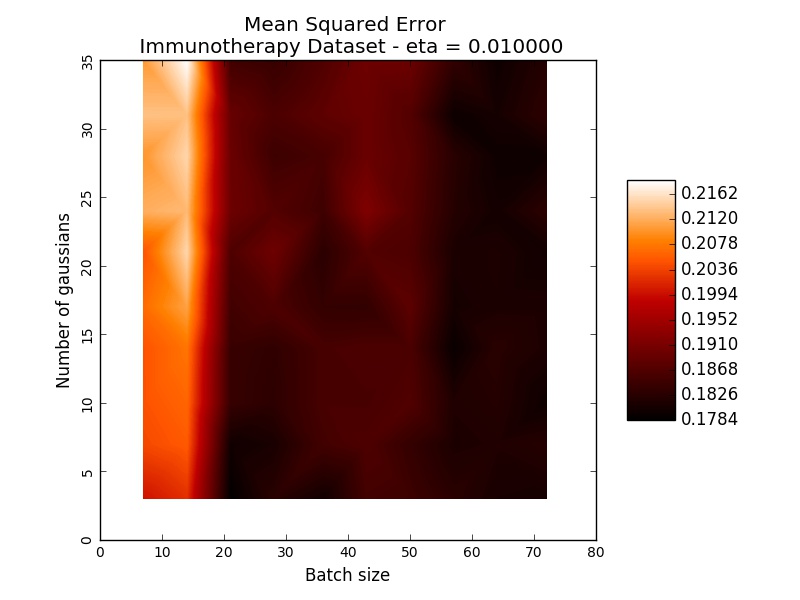}
        \label{fig:apmseimmu001mean}
    \end{subfigure}
    \begin{subfigure}[b]{0.4\textwidth}
        \includegraphics[width=\textwidth]{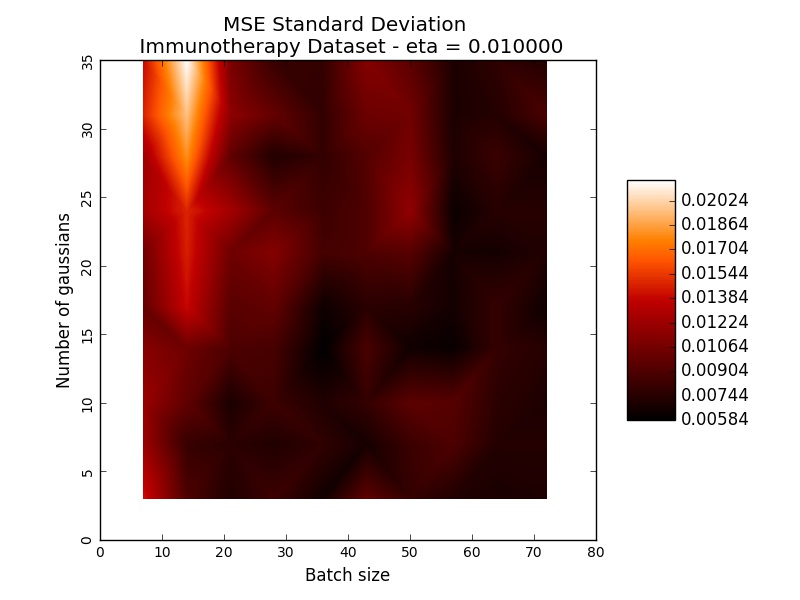}
        \label{fig:apmseimmu001std}
    \end{subfigure}
    \vspace{-0.25cm}
    \caption*{\vspace{-0.1cm} Source: Author. }
    \label{fig:apmseimmu001}
\end{figure}

\begin{figure} [!ht]
    \centering
    \caption{\vspace{-0.1cm} Mean Squared Error and standard deviation measurements for the VFNN on the Immunotherapy dataset with eta = 0.2575. }
    \begin{subfigure}[b]{0.4\textwidth}
        \includegraphics[width=\textwidth]{Figures/HyperparametersResults/MeanSquaredErrorImmunotherapy0257500.jpg}
        \label{fig:apmseimmu025mean}
    \end{subfigure}
    \begin{subfigure}[b]{0.4\textwidth}
        \includegraphics[width=\textwidth]{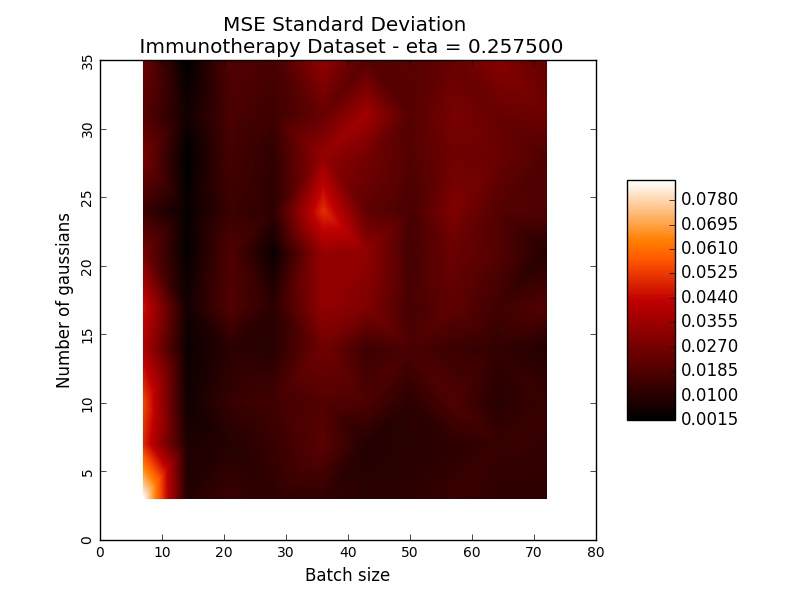}
        \label{fig:apmseimmu025std}
    \end{subfigure}
    \vspace{-0.25cm}
    \caption*{\vspace{-0.1cm} Source: Author. }
    \label{fig:apmseimmu025}
\end{figure}

\begin{figure} [!ht]
    \centering
    \caption{\vspace{-0.1cm} Mean Squared Error and standard deviation measurements for the VFNN on the Immunotherapy dataset with eta = 0.505. }
    \begin{subfigure}[b]{0.4\textwidth}
        \includegraphics[width=\textwidth]{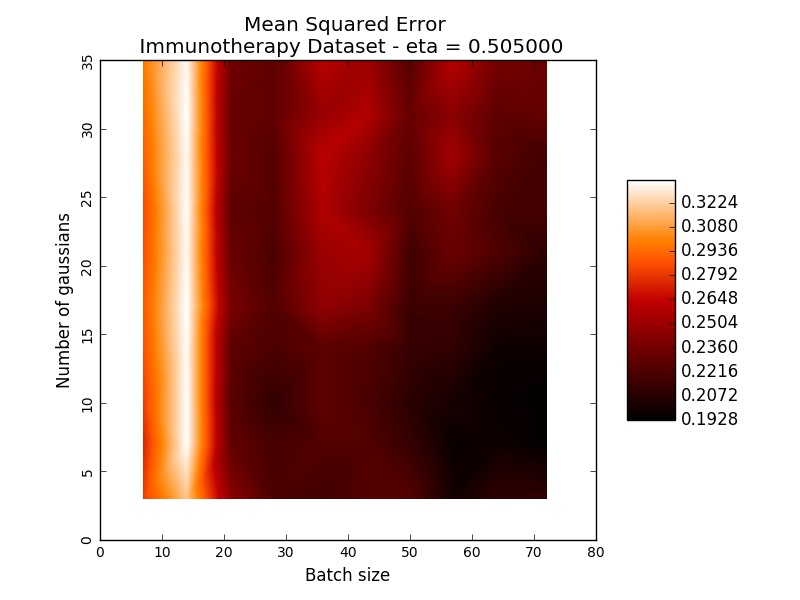}
        \label{fig:apmseimmu0505mean}
    \end{subfigure}
    \begin{subfigure}[b]{0.4\textwidth}
        \includegraphics[width=\textwidth]{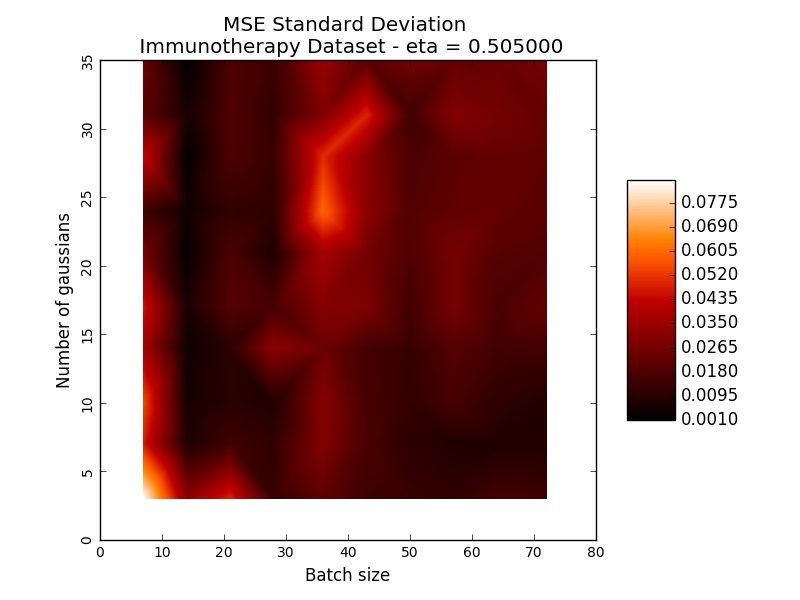}
        \label{fig:apmseimmu0505std}
    \end{subfigure}
    \vspace{-0.25cm}
    \caption*{\vspace{-0.1cm} Source: Author. }
    \label{fig:apmseimmu0505}
\end{figure}

\begin{figure} [!ht]
    \centering
    \caption{\vspace{-0.1cm} Mean Squared Error and standard deviation measurements for the VFNN on the Immunotherapy dataset with eta = 0.7525. }
    \begin{subfigure}[b]{0.4\textwidth}
        \includegraphics[width=\textwidth]{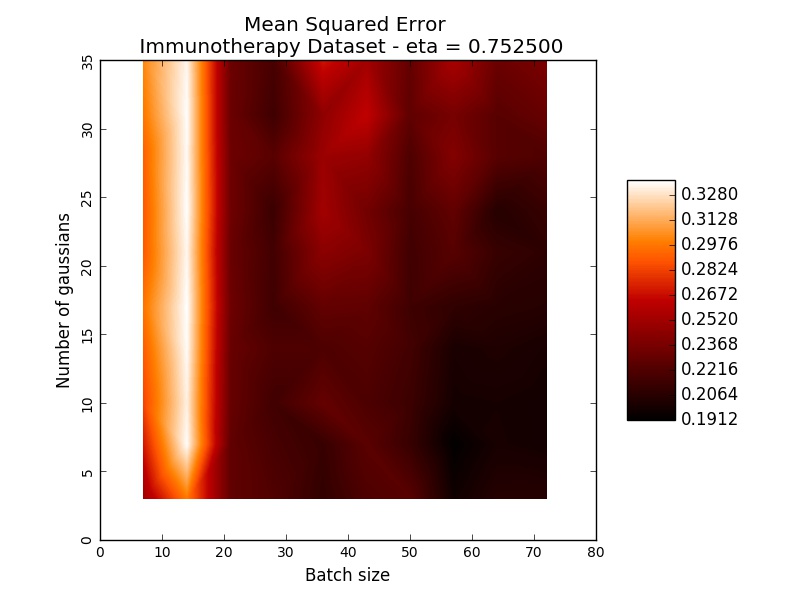}
        \label{fig:apmseimmu075mean}
    \end{subfigure}
    \begin{subfigure}[b]{0.4\textwidth}
        \includegraphics[width=\textwidth]{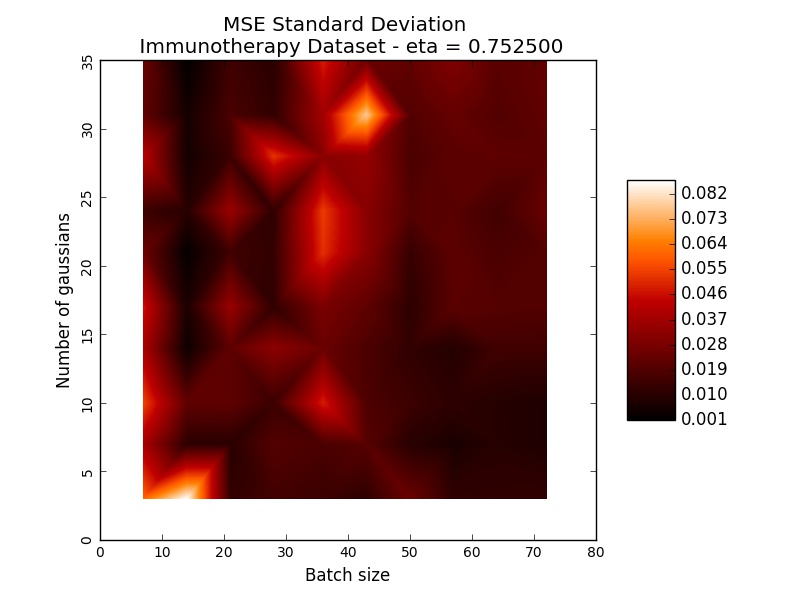}
        \label{fig:apmseimmu075std}
    \end{subfigure}
    \vspace{-0.25cm}
    \caption*{\vspace{-0.1cm} Source: Author. }
    \label{fig:apmseimmu075}
\end{figure}

\begin{figure} [!ht]
    \centering
    \caption{\vspace{-0.1cm} Mean Squared Error and standard deviation measurements for the VFNN on the Immunotherapy dataset with eta = 1.0. }
    \begin{subfigure}[b]{0.4\textwidth}
        \includegraphics[width=\textwidth]{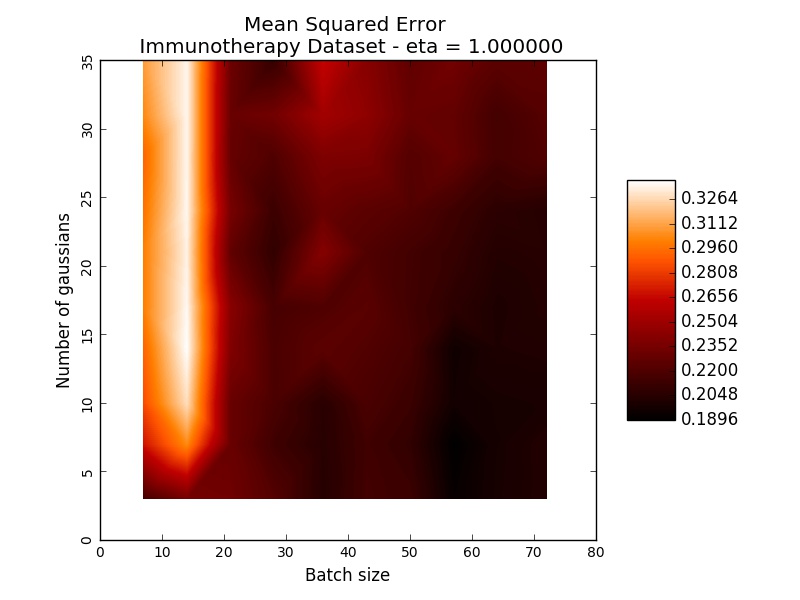}
        \label{fig:apmseimmu100mean}
    \end{subfigure}
    \begin{subfigure}[b]{0.4\textwidth}
        \includegraphics[width=\textwidth]{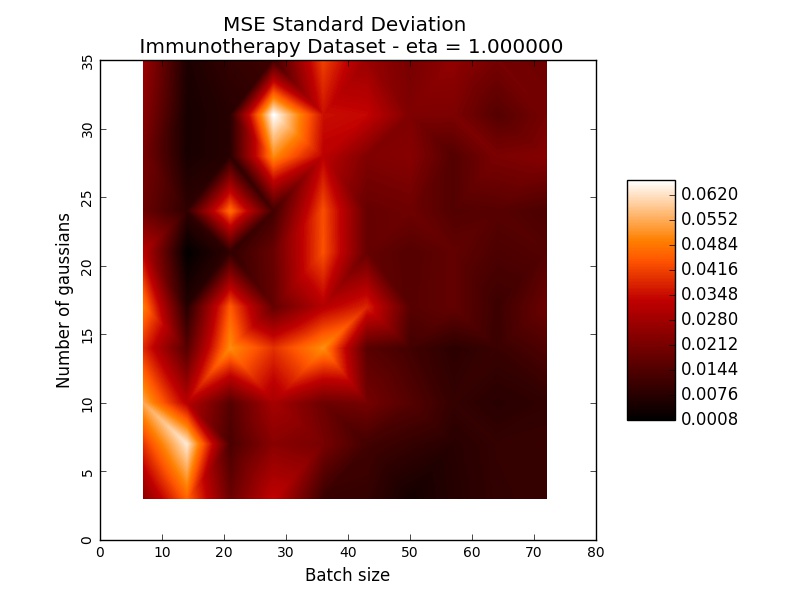}
        \label{fig:apmseimmu100std}
    \end{subfigure}
    \vspace{-0.25cm}
    \caption*{\vspace{-0.1cm} Source: Author. }
    \label{fig:apmseimmu0100}
\end{figure}

\chapter{Time and Complexity Results}
\label{appendicetimeandcomplexity}

\begin{figure} [!ht]
    \centering
    \caption{\vspace{-0.1cm} Train time and standard deviation for the VFNN. }
    \begin{subfigure}[b]{0.45\textwidth}
        \includegraphics[width=\textwidth]{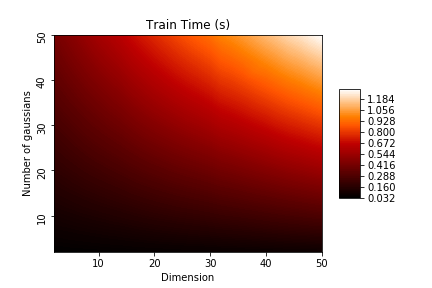}
        \label{fig:aptrain_time}
    \end{subfigure}
    \begin{subfigure}[b]{0.45\textwidth}
        \includegraphics[width=\textwidth]{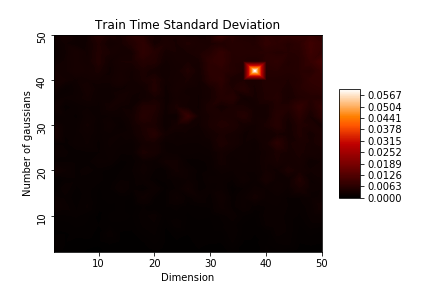}
        \label{fig:aptrain_timestd}
    \end{subfigure}
    \vspace{-0.25cm}
    \caption*{\vspace{-0.1cm} Source: Author. }
    \label{fig:aptrain_times}
\end{figure}

\begin{figure} [!ht]
    \centering
    \caption{\vspace{-0.1cm} Train time and standard deviation for the VFNN. }
    \begin{subfigure}[b]{0.45\textwidth}
        \includegraphics[width=\textwidth]{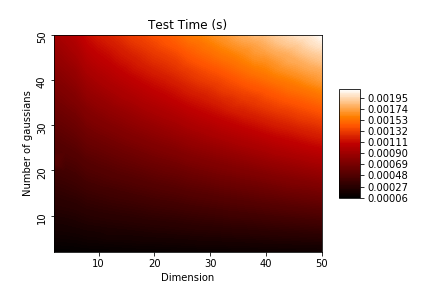}
        \label{fig:aptest_time}
    \end{subfigure}
    \begin{subfigure}[b]{0.45\textwidth}
        \includegraphics[width=\textwidth]{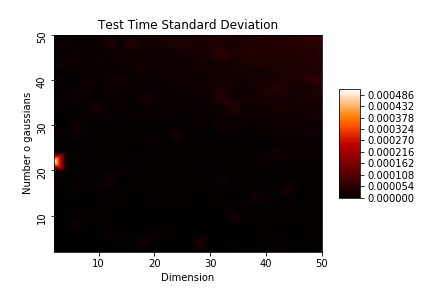}
        \label{fig:aptest_timestd}
    \end{subfigure}
    \vspace{-0.25cm}
    \caption*{\vspace{-0.1cm} Source: Author. }
    \label{fig:aptest_times}
\end{figure}

\begin{figure} [!ht]
    \centering
    \caption{\vspace{-0.1cm} Complexity measurements for the VFNN. }
    \begin{subfigure}[b]{0.45\textwidth}
        \includegraphics[width=\textwidth]{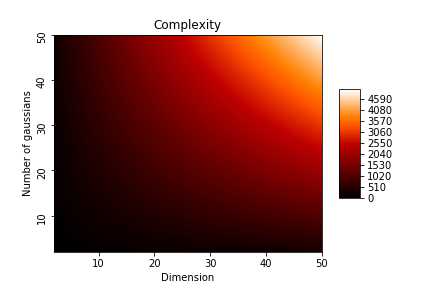}
        \label{fig:apcomp}
    \end{subfigure}
    \vspace{-0.25cm}
    \caption*{\vspace{-0.1cm} Source: Author. }
    \label{fig:apcomplexity}
\end{figure}

\chapter{Performance Results}

\begin{figure}[!ht]
    \centering
    \caption{Performance results for the Banknote dataset.}
    \begin{subfigure}[b]{0.48\linewidth}
         \includegraphics[width=\linewidth]{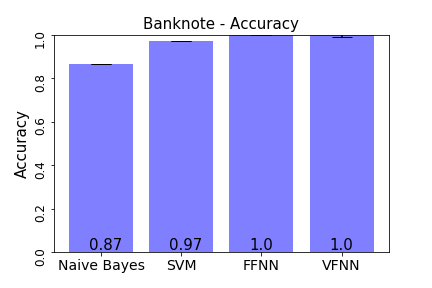}
         \label{fig:app_acc_banknote}
    \end{subfigure}
    \begin{subfigure}[b]{0.48\linewidth}
         \includegraphics[width=\linewidth]{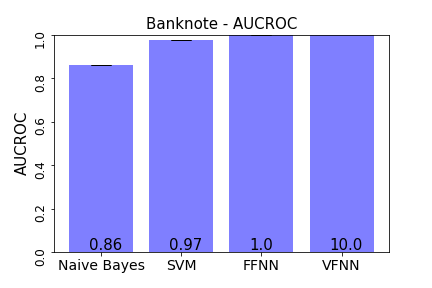}
         \label{fig:app_roc_banknote}
    \end{subfigure}
     ~ 
    \begin{subfigure}[b]{0.48\linewidth}
         \includegraphics[width=\linewidth]{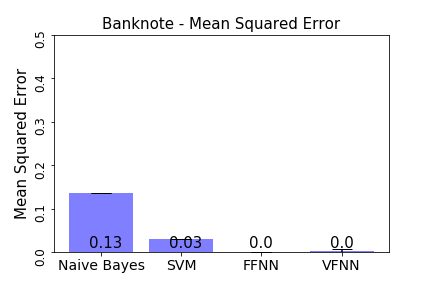}
         \label{fig:app_mse_banknote}
     \end{subfigure}
     \vspace{-0.25cm}
     \caption*{Source: Author.}
     \label{app_p_banknote}
 \end{figure}

\begin{figure}[!ht]
    \centering
    \caption{Performance results for the Ionosphere dataset.}
    \begin{subfigure}[b]{0.48\linewidth}
         \includegraphics[width=\linewidth]{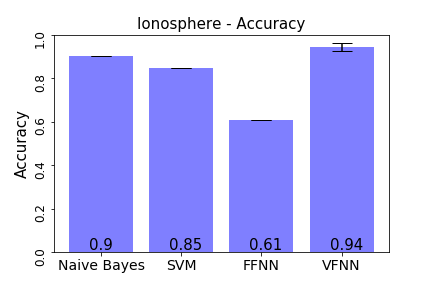}
         \label{fig:app_acc_iono}
    \end{subfigure}
    \begin{subfigure}[b]{0.48\linewidth}
         \includegraphics[width=\linewidth]{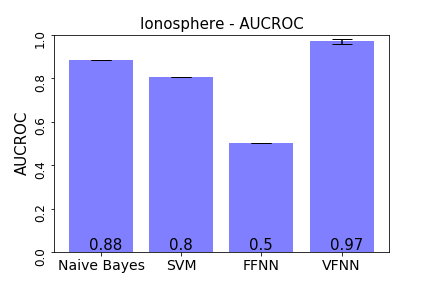}
         \label{fig:app_roc_iono}
    \end{subfigure}
     ~ 
    \begin{subfigure}[b]{0.48\linewidth}
         \includegraphics[width=\linewidth]{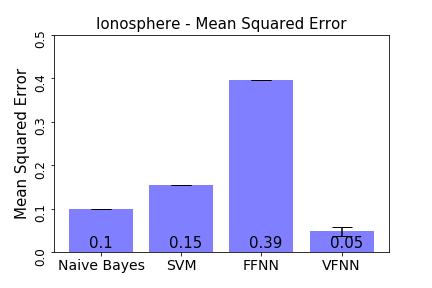}
         \label{fig:app_mse_iono}
     \end{subfigure}
     \vspace{-0.25cm}
     \caption*{Source: Author.}
     \label{app_p_iono}
 \end{figure}

\begin{figure}[!ht]
    \centering
    \caption{Performance results for the Cryotherapy dataset.}
    \begin{subfigure}[b]{0.48\linewidth}
         \includegraphics[width=\linewidth]{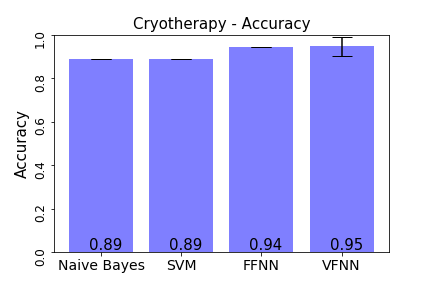}
         \label{fig:app_acc_cryo}
    \end{subfigure}
    \begin{subfigure}[b]{0.48\linewidth}
         \includegraphics[width=\linewidth]{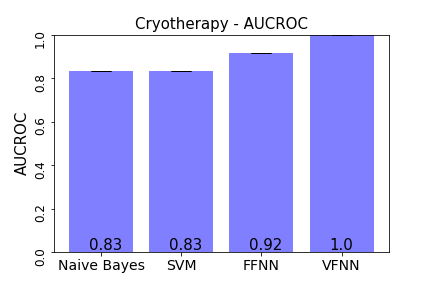}
         \label{fig:app_roc_cryo}
    \end{subfigure}
     ~ 
    \begin{subfigure}[b]{0.48\linewidth}
         \includegraphics[width=\linewidth]{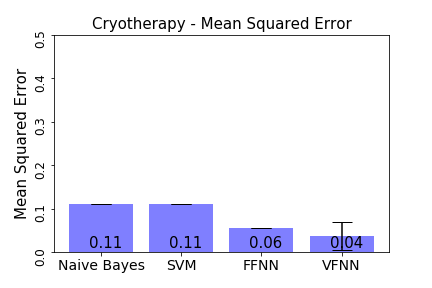}
         \label{fig:app_mse_cryo}
     \end{subfigure}
     \vspace{-0.25cm}
     \caption*{Source: Author.}
     \label{app_p_cryo}
 \end{figure}

\begin{figure}[!ht]
    \centering
    \caption{Performance results for the Pima Diabetes dataset.}
    \begin{subfigure}[b]{0.48\linewidth}
         \includegraphics[width=\linewidth]{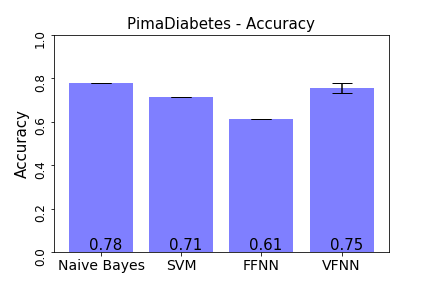}
         \label{fig:app_acc_pima}
    \end{subfigure}
    \begin{subfigure}[b]{0.48\linewidth}
         \includegraphics[width=\linewidth]{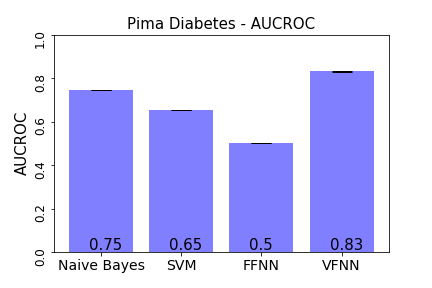}
         \label{fig:app_roc_pima}
    \end{subfigure}
     ~ 
    \begin{subfigure}[b]{0.48\linewidth}
         \includegraphics[width=\linewidth]{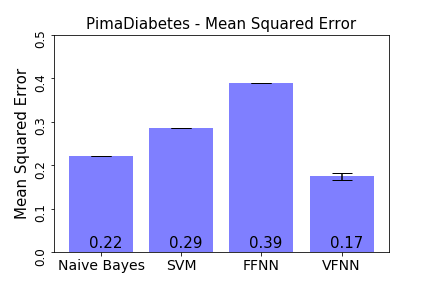}
         \label{fig:app_mse_pima}
     \end{subfigure}
     \vspace{-0.25cm}
     \caption*{Source: Author.}
     \label{app_p_pima}
 \end{figure}

\begin{figure}[!ht]
    \centering
    \caption{Performance results for the Immunotherapy dataset.}
    \begin{subfigure}[b]{0.48\linewidth}
         \includegraphics[width=\linewidth]{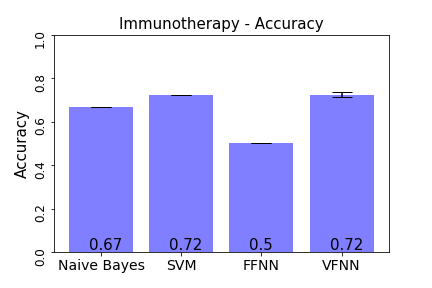}
         \label{fig:app_acc_immuno}
    \end{subfigure}
    \begin{subfigure}[b]{0.48\linewidth}
         \includegraphics[width=\linewidth]{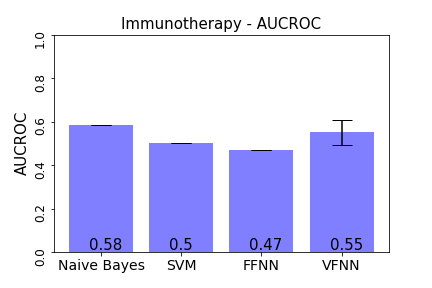}
         \label{fig:app_roc_immuno}
    \end{subfigure}
     ~ 
    \begin{subfigure}[b]{0.48\linewidth}
         \includegraphics[width=\linewidth]{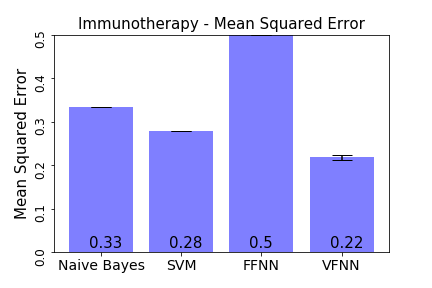}
         \label{fig:app_mse_immuno}
     \end{subfigure}
     \vspace{-0.25cm}
     \caption*{Source: Author.}
     \label{app_p_immuno}
 \end{figure}

\end{apendicesenv}

\end{document}